\PassOptionsToPackage{most,breakable}{tcolorbox}
\PassOptionsToPackage{dvipsnames,table}{xcolor}
\PassOptionsToPackage{numbers,compress}{natbib}
\documentclass[11pt]{article}

\usepackage[final]{arxiv}

\usepackage[T1]{fontenc}
\usepackage[utf8]{inputenc}
\usepackage{amsmath,amssymb}
\usepackage{amsfonts}
\usepackage{mathpazo}
\usepackage[xcharter,bigdelims,vvarbb]{newtxmath}
\usepackage[scaled]{helvet}
\usepackage[scaled=1.1]{zlmtt}
\usepackage{latexsym}

\usepackage{pifont}
\usepackage{placeins}
\usepackage{booktabs}
\usepackage{multirow}
\usepackage{tabularx}
\usepackage{subcaption}
\usepackage{arydshln}
\usepackage{adjustbox}
\usepackage{makecell}
\usepackage{enumitem}
\usepackage{rotating}
\usepackage{titletoc}
\usepackage{comment}
\usepackage{wrapfig}
\usepackage{needspace}
\usepackage{nicefrac}
\usepackage{microtype}

\usepackage{siunitx}
\usepackage{dblfloatfix}
\usepackage{graphicx}
\usepackage{cuted}
\usepackage{capt-of}
\usepackage{xspace}

\setlist[itemize]{leftmargin=2em}

\usepackage{tikz}
\usetikzlibrary{fadings}
\usepackage{array}
\usepackage{tcolorbox}
\usepackage{threeparttable}

\newcommand{\benchmark}{SpatialBench\xspace}
\newcommand{\dataset}{DA-Next-5M\xspace}
\newcommand{\ours}{DA-Next\xspace}

\newcommand{\cmark}{\textcolor{green!60!black}{\ding{51}}}
\newcommand{\xmark}{\textcolor{red}{\ding{55}}}
\definecolor{subcol}{RGB}{255,255,255}
\definecolor{bestred}{RGB}{120,170,220}
\definecolor{secondorange}{RGB}{175,205,235}
\definecolor{thirdyellow}{RGB}{215,232,248}
\definecolor{trainmark}{RGB}{0,180,0}
\definecolor{oomred}{RGB}{252,228,228}
\definecolor{catgray}{RGB}{230,230,230}
\definecolor{altcolblue}{RGB}{225,237,250}
\definecolor{pretrainmark}{RGB}{197,90,17}
\definecolor{gaingreen}{RGB}{30,140,60}
\definecolor{lossred}{RGB}{200,30,30}
\definecolor{navyblue}{HTML}{0071BC}

\DeclareRobustCommand{\observationbox}[2][blue!5]{%
\begin{tcolorbox}[
        breakable,
        left=0pt,
        right=0pt,
        top=0pt,
        bottom=0pt,
        colback=#1,
        colframe=#1,
        width=\columnwidth,
        arc=1pt,outer arc=1pt,
        ]
        #2
\end{tcolorbox}
}

\firstpagelogos{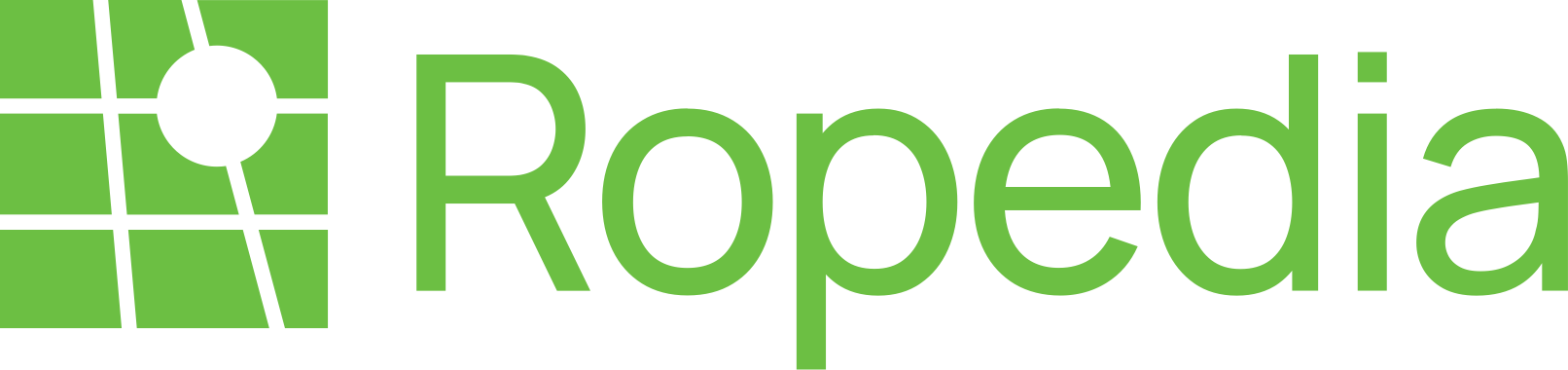}{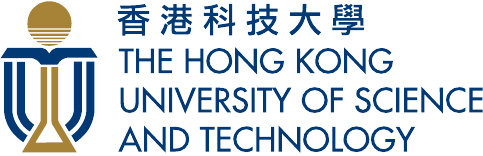}{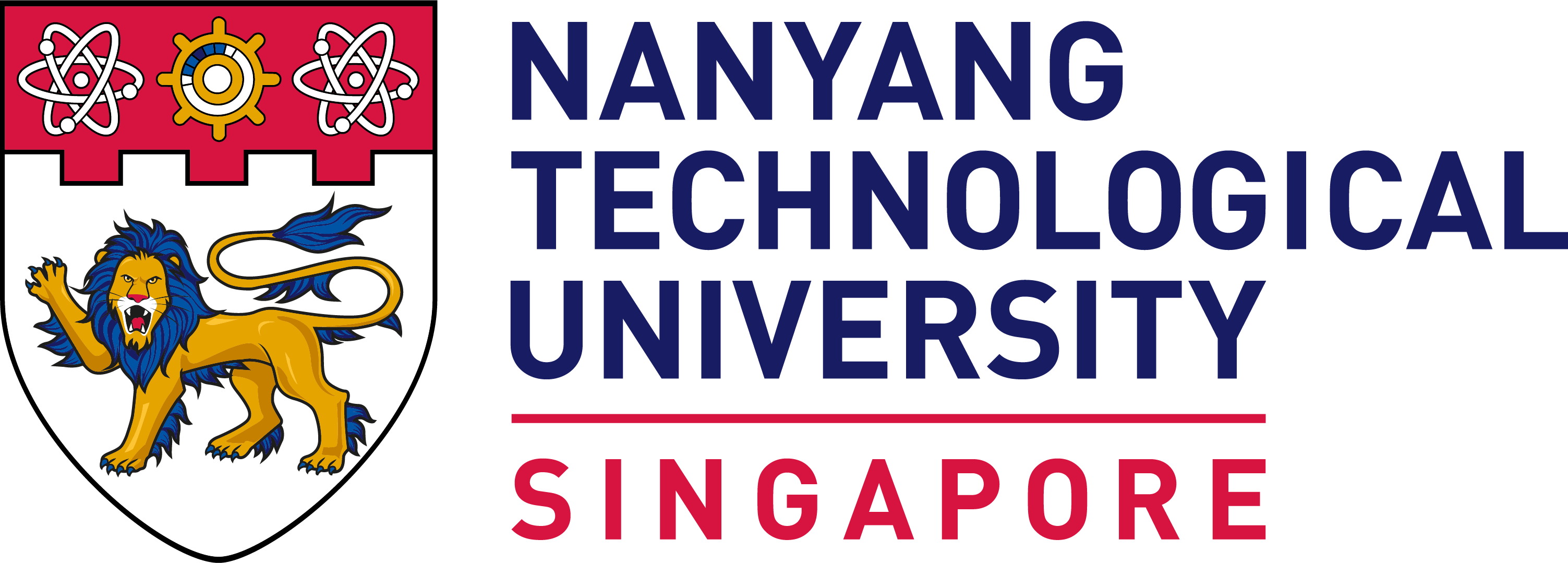}
\abstractlogo{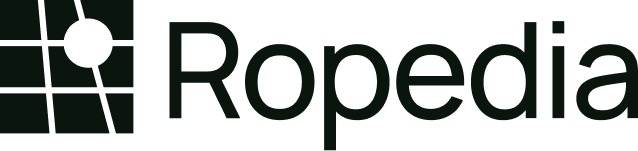}

\title{\benchmark}
\subtitle{Is Your Spatial Foundation Model an All-Round Player?}

\newcommand{\coremark}{\ensuremath{^{\star}}}
\newcommand{\leadmark}{\ensuremath{^{\ddagger}}}
\newcommand{\corrmark}{\ensuremath{^{\dagger}}}

\newcommand{\authorfootnotes}{%
  \begingroup
  \renewcommand{\thefootnote}{}%
  \footnotetext{$\star$ Core contributors. $\ddagger$ Project lead. $\dagger$ Corresponding author.}%
  \endgroup
  \addtocounter{footnote}{-1}%
}

\author{
  \AuthorName{Haosong Peng}{1\coremark}\quad
  \RopediaAuthorName{Hao Li}{2\,3\coremark\leadmark}\quad
  \AuthorName{Jiaqi Chen}{3\coremark}\quad
  \AuthorName{Yuhao Pan}{1\coremark}\quad
  \RopediaAuthorName{Runmao Yao}{2}\quad
  \AuthorName{Yalun Dai}{2}\\[0.55ex]
  \AuthorName{Fushuo Huo}{4}\quad
  \RopediaAuthorName{Fangzhou Hong}{2}\quad
  \RopediaAuthorName{Zhaoxi Chen}{2}\quad
  \AuthorName{Haozhao Wang}{5}\\[0.55ex]
  \AuthorName{Dingwen Zhang}{3}\quad
  \RopediaAuthorName{Ziwei Liu}{2}\quad
  \AuthorName{Wenchao Xu}{1\corrmark}\\[1.0ex]
  {\normalfont\footnotesize
  \mbox{\textsuperscript{1}Hong Kong University of Science and Technology}\quad
  \mbox{\textsuperscript{2}Nanyang Technological University}\quad
  \mbox{\textsuperscript{\ropediamark}Ropedia}\\
  \mbox{\textsuperscript{3}Northwestern Polytechnical University}\quad
  \mbox{\textsuperscript{4}Southeast University}\quad
  \mbox{\textsuperscript{5}Huazhong University of Science and Technology}}
}

\begin{document}
\maketitle
\authorfootnotes

\begin{paperresources}
\paperresourceicon{\resourceprojecticon}{Project Page}{https://ropedia.github.io/SpatialBench/}{ropedia.github.io/SpatialBench}
\quad
\paperresourceicon{\resourcehficon}{Dataset}{https://huggingface.co/datasets/ropedia-ai/DA-Next-5M}{ropedia-ai/DA-Next-5M}\\
\paperresourceicon{\resourcegithubicon}{Code}{https://github.com/Ropedia/SpatialBench}{github.com/Ropedia/SpatialBench}
\qquad \quad \quad
\paperresourceicon{\resourcehficon}{Model}{https://huggingface.co/ropedia-ai/DA-Next}{ropedia-ai/DA-Next}
\end{paperresources}

\vspace{7pt}
{\centering
\includegraphics[width=1\textwidth]{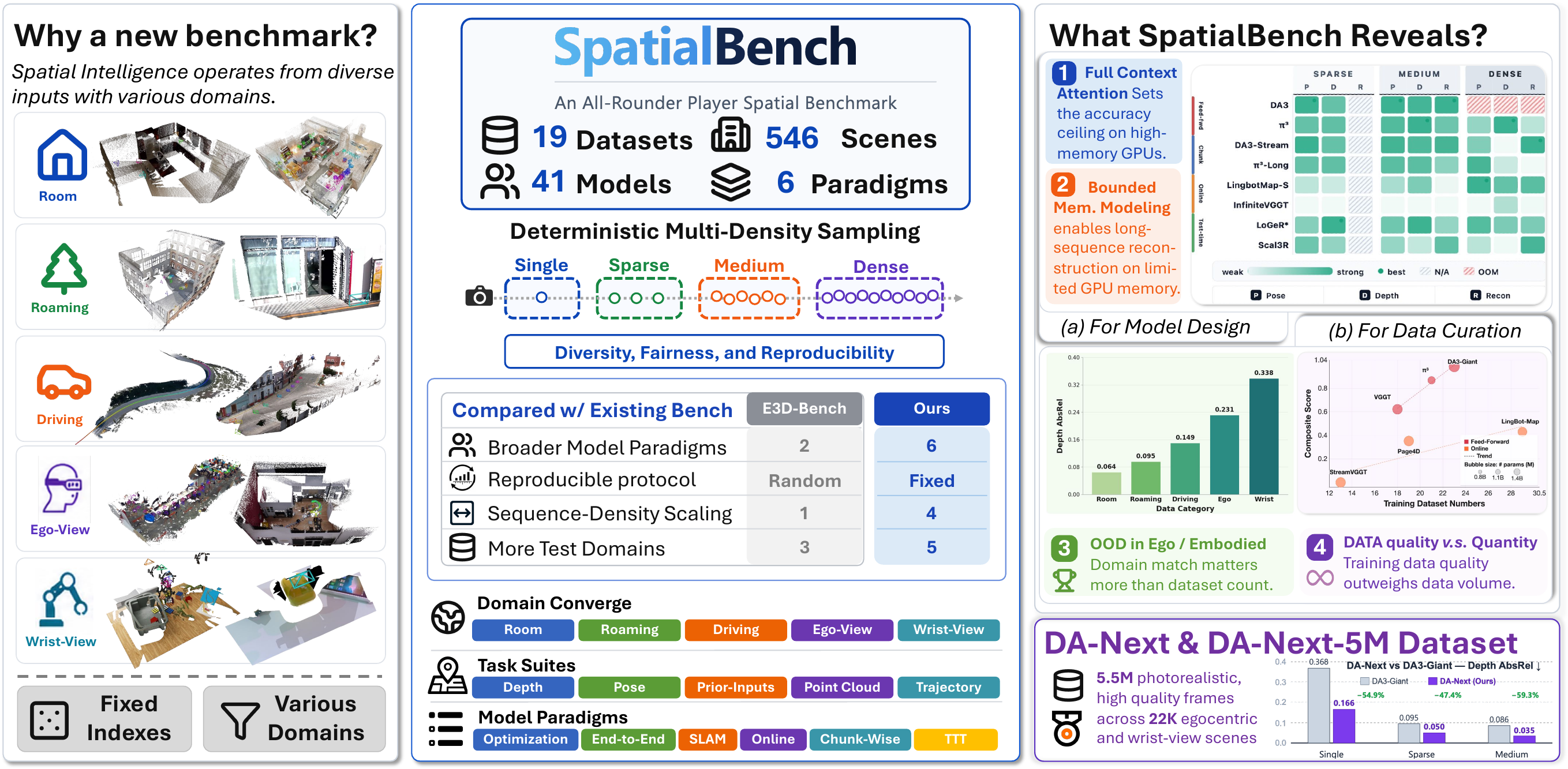}
\par\vspace{6pt}
\begin{minipage}{\textwidth}
\footnotesize
\refstepcounter{figure}\label{fig:teaser}
Figure 1 | \textbf{\benchmark} provides a reproducible, cross-paradigm benchmark spanning 19 datasets, 546 scenes, 41 models, and 6 paradigms under deterministic multi-density sampling.
Our analysis reveals insights on model design, domain generalization, data curation, and beyond, complemented by \ours and \dataset to address the embodied domain gap.
\end{minipage}
\par}

\vspace{8pt}
\begin{abstract}
While spatial foundation models have demonstrated impressive performance on standard datasets, a critical question remains: are they truly \textbf{\textit{all-round players}} capable of generalizing robustly across diverse downstream tasks, arbitrary viewpoints, shifting scene domains, varying input densities, and specific hardware constraints?
Answering this overarching question requires a holistic assessment, yet current models are mainly evaluated on specific domains for which they were specifically designed or trained.
Such evaluations are intrinsically limited by narrow paradigm coverage, limited scene domains, and arbitrary frame sampling, making it fundamentally difficult to assess their true generalization capabilities.
To address this gap, we present \textbf{\benchmark}, a cross-paradigm, domain-diverse benchmark for spatial foundation models with deterministic sampling.
\benchmark features unprecedented scale and rigorous deterministic design, comprising \textbf{19} datasets and \textbf{546} scenes across \textbf{5} diverse spatial domains. It comprehensively evaluates \textbf{41} models across \textbf{6} paradigms on \textbf{5} task suites under \textbf{4} different input density settings.
Our extensive evaluation reveals that current models are not yet all-round players, and uncovers crucial insights for future advancement. Specifically, we demonstrate that full-context attention maximizes accuracy while bounded-memory strategies unlock long-sequence scalability.
Moreover, our empirical evaluations in challenging embodied and egocentric tasks demonstrate that strict domain alignment and high data quality are far more critical to performance than simple dataset scaling.
Furthermore, to address the largest data gap identified in our analysis, we go beyond evaluation by introducing a large-scale dataset, \textbf{DA-Next-5M}, and a strong baseline model, \textbf{DA-Next}, pushing the boundaries of spatial representation learning.
\end{abstract}

\section{Introduction}

Spatial foundation models have already been widely deployed across robotics~\citep{yang2026robo3r,lin2025evo,zhang2025spatial}, AR/VR~\citep{yang2026neoverse}, autonomous driving~\citep{chen2025dggt}, and embodied AI~\citep{kareer2025egomimic,zheng2026egoscale}. 
This extensive adoption is driven by their remarkable ability to recover accurate 3D structures from mere images or videos, establishing them as general-purpose visual geometry backbones for spatial intelligence. 
However, operating in these real-world applications is inherently chaotic and far more demanding than standard reconstruction benchmarks. 
To truly support these downstream tasks, a robust model must maintain its reliability when confronting unpredictable scene domain shifts, highly variable sparse-to-dense input regimes, and strict hardware memory constraints. 
This raises the central question of this work: \emph{if spatial foundation models are expected to support general-purpose spatial intelligence, can they truly serve as robust all-round players across the diverse conditions of the 3D world?}

However, existing evaluations fall short in several critical aspects. 
First, they cover only a narrow slice of today's model paradigms. 
Spatial foundation models now span feed-forward~\citep{wang2025vggt,keetha2026mapanything,lin2025depth}, optimization-based~\citep{wang2024dust3r,leroy2024mast3r,zhang2024monst3r}, streaming~\citep{streamVGGT,stream3r2025,chen2026lingbotmap}, SLAM-based~\citep{maggio2025vggt-slam,murai2024_mast3rslam}, chunk-based~\citep{deng2025vggtlong}, and test-time training (TTT)~\citep{chen2025ttt3r,xie2026scal3r,zhang2026loger} approaches, yet most benchmarks evaluate only one or a few of them under separate protocols. 
Second, current comparisons are often not standardized. 
Even when papers report results on the same dataset, they may use different scene splits, private subsets, frame indices, temporal windows, or input densities, making direct comparison ambiguous. 
Third, existing protocols rarely expose how models scale with sequence density. 
For example, a model that works well on sparse image sets may fail on dense long videos because of memory growth, accumulated drift, or degraded global consistency; conversely, bounded-memory methods (e.g., online, chunk-wise, and TTT) may be undervalued when evaluation is restricted to short sequences. 
Finally, test domains remain too limited for assessing real-world spatial intelligence. 
Standard indoor or object-centric reconstruction datasets do not capture the diversity of robotics, autonomous driving, egocentric perception, and wrist-mounted manipulation settings. 
These gaps motivate a benchmark that is cross-paradigm, deterministic, density-aware, and domain-diverse: one that can fairly compare models, reveal how performance changes from sparse views to dense streams, and diagnose where current spatial foundation models succeed or fail.

To address the aforementioned challenges, we introduce \textbf{\benchmark}. 
By incorporating a deterministic density-aware protocol, broad domain diversity, and cross-paradigm comparisons, this comprehensive benchmark serves as a beacon to guide and verify spatial foundation models toward becoming true \textbf{all-round players}.
\benchmark is built around three core design principles:
\noindent\textbf{(1) Deterministic Multi-Density Evaluation Protocol.}
To systematically assess model robustness across varying input scales, \benchmark adopts a deterministic sampling strategy to precompute frame indices across \textbf{4} distinct density regimes: \textit{single-frame}, \textit{sparse}, \textit{medium}, and \textit{dense}. 
By evaluating each scene under these standardized configurations across several key metrics, our protocol ensures both a comprehensive understanding of model performance and full reproducibility across different paradigms.
\noindent\textbf{(2) Broad Domain Coverage Across 19 Datasets.}
\benchmark aggregates \textbf{19} datasets and \textbf{546} scenes in total, spanning a comprehensive range of conditions, including \textit{indoor} and \textit{outdoor} environments, \textit{static} and \textit{dynamic} scenes, \textit{real-world} and \textit{synthetic} data, and diverse viewpoint types.
Each scene is annotated with orthogonal tags along these axes, enabling fine-grained cross-domain filtering and aggregation, supporting over 100 distinct evaluation configurations that far exceed any existing benchmark.
\noindent\textbf{(3) Comprehensive and Cross-Paradigm Model Comparison.}
\benchmark provides unified adapters for \textbf{31} state-of-the-art models and \textbf{41} variants in total, spanning all six reconstruction paradigms: \textit{optimization-based}, \textit{end-to-end feed-forward}, \textit{online streaming}, \textit{chunk-wise}, \textit{TTT-based}, and \textit{SLAM-based} systems.
All methods are evaluated under a unified protocol, enabling fair and direct comparison across several geometric tasks, including \textit{depth} and \textit{camera pose estimation}, \textit{reconstruction}, \textit{prior-enhanced geometry prediction}, and \textit{trajectory estimation}.

We further conduct extensive analysis experiments on \benchmark, revealing several key insights:
\textbf{(1) Full-context attention defines the accuracy upper bound}, with globally coupled feed-forward models consistently outperforming bounded-memory approaches under the same input budget.
\textbf{(2) Bounded-memory models unlock long-horizon scalability}, enabling continuous reconstruction beyond the memory limits of full-context models, at the cost of geometry estimation accuracy.
\textbf{(3) Data quality outweighs data volume}, as carefully curated pseudo-GT supervision consistently outperforms larger but noisier training mixtures.
\textbf{(4) Egocentric and wrist-view domains remain the dominant OOD failure modes}, exposing a field-level gap that cannot be addressed by scaling existing training mixtures alone.

To further address the gap in egocentric and wrist-view domains, we curate \textbf{\dataset}, a dataset comprising \textbf{22K} scenes with \textbf{5.5M} frames of 3D data in total from egocentric and robot wrist-view sources. 
We train our proposed \textbf{Depth-Anything-Next} (\ours) on \dataset, establishing a strong domain-specific baseline for these underexplored viewpoints.
The key contributions of our work are summarized as follows.
\begin{itemize}
    \item \benchmark is the first standardized benchmark for comprehensive evaluation of 3D spatial foundation models on several geometry tasks, aggregating 19 diverse datasets and 546 scenes, and providing unified adapters for 32 methods and 41 variants across all six paradigms.
    \item Through extensive experiments on \benchmark, we conduct a comprehensive cross-paradigm analysis and derive key insights into model robustness, domain generalization, and input-density scaling behavior, highlighting promising directions for future research.
    \item Experiments show that \ours achieves substantial gains over DA3-Giant: $+47\%$/$+59\%$ in depth estimation and $+3.1\%$/$+5.5\%$ in pose estimation on sparse/medium inputs, demonstrating that targeted in-domain data curation effectively closes the embodied domain gap.
\end{itemize}

\section{Related Work}
\label{appendix:related}

\textbf{Spatial foundation models for visual geometry.}
Recent advances in visual geometry have shifted 3D reconstruction from optimization-heavy pipelines toward spatial foundation models that directly infer scene geometry, camera parameters, and point cloud from images. 
Early influential systems such as DUSt3R~\citep{wang2024dust3r} and MASt3R~\citep{leroy2024mast3r} reformulate geometric reconstruction as dense pointmap prediction, substantially simplifying pose-free 3D reconstruction and 3D-grounded image matching. 
Although these methods still rely on global alignment or optimization-based post-processing, they establish a strong foundation for subsequent feed-forward reconstruction models. 
Building on this direction, end-to-end feed-forward methods aim to recover visual geometry in a single network pass. 
VGGT~\citep{wang2025vggt} predicts camera parameters, depth maps, pointmaps, and point tracks in a unified transformer framework, while Fast3R~\citep{yang2025fast3r} scales feed-forward reconstruction to large unordered image collections and FastVGGT~\citep{shen2025fastvggt} accelerates VGGT-style inference without retraining. 
MUSt3R~\citep{cabon2025must3r} extends stereo-style reconstruction to multi-view settings, and MapAnything~\citep{keetha2026mapanything} supports flexible geometric inputs such as poses, depths, intrinsics, and partial reconstructions for universal metric 3D reconstruction. 
More model families further expand this paradigm: OmniVGGT~\citep{peng2025omnivggt} incorporates omni-modality prior for reconstruction; $\pi^3$ removes the dependence on reference frames through a fully permutation-equivariant architecture, predicting affine-invariant camera poses, and scale-invariant local pointmaps; AMB3R~\citep{wang2025amb3r} introduces a backend module for more accurate metric-scale reconstruction; DA3 and its variants~\citep{lin2025depth} recover consistent 3D geometry from arbitrary visual inputs across multiple model scales; and WorldMirror~\citep{liu2025worldmirror} explores any-prior prompting to unify diverse 3D representations. 
Together, these feed-forward spatial foundation models demonstrate strong reconstruction capability on bounded image sets, but their performance can degrade when applied to long videos, streaming inputs, or large-scale scenes where memory, consistency, and drift become critical.
Moreover, processing long sequences with these models incurs prohibitive GPU memory consumption and increased inference latency.

\textbf{Long-sequence, online, and test-time training models.}
To handle realistic video streams, recent work has extended spatial foundation models from bounded image sets to online, chunk-wise, SLAM-based, and test-time adaptive settings. 
Online and streaming methods maintain temporal or spatial memory, recurrent states, or compact historical context as new frames arrive. 
Spann3R~\citep{wang2024spann3r} introduces spatial memory for incremental 3D reconstruction, CUT3R~\citep{wang2025cut3r} uses a persistent recurrent state for continuous 3D perception, and MonST3R~\citep{zhang2024monst3r} extends DUSt3R-style reconstruction to dynamic scenes with motion. 
Point3R~\citep{point3r} employs explicit spatial pointer memory for streaming reconstruction, while Stream3R~\citep{stream3r2025}, StreamVGGT~\citep{streamVGGT}, Page4D~\citep{zhou2025page4d}, InfiniteVGGT~\citep{yuan2026infinitevggt}, WinT3R~\citep{li2025wint3r}, LongStream~\citep{cheng2026longstream}, and LingBot-Map~\citep{chen2026lingbotmap} investigate different memory mechanisms, window designs, causal attention strategies, and long-horizon update rules for scalable online geometry estimation. 
Another line processes long videos in chunks and then aligns local reconstructions into a global scene. VGGT-Long~\citep{deng2025vggtlong}, $\pi^3$-Long~\citep{deng2025vggtlong}, and DA3-Streaming~\citep{deng2025vggtlong} follow this chunk-wise strategy to extend powerful feed-forward backbones or model variants to kilometer-scale or long-sequence reconstruction. 
In parallel, SLAM-based systems such as MASt3R-SLAM~\citep{murai2024_mast3rslam} and VGGT-SLAM~\citep{maggio2025vggt-slam} combine learned 3D priors with classical mapping and tracking components to improve real-time dense reconstruction. 
Finally, test-time training methods, including TTT3R~\citep{chen2025ttt3r}, Scal3R~\citep{xie2026scal3r}, ZipMap~\citep{jin2026zipmap} and LoGeR~\citep{zhang2026loger}, adapt the model or scene representation during inference to improve large-scale consistency and reduce drift. 
These methods reveal an emerging trend: the central challenge of spatial foundation models is no longer only accurate single-shot reconstruction, but also scalable memory management, temporal consistency, dynamic-scene robustness, and long-range geometric alignment under realistic visual streams.

\textbf{Related benchmarks for visual geometry.}
Several recent efforts have introduced systematic benchmarks for 3D reconstruction and visual geometry. Robust MVD~\citep{schroeppel2022rmvd} focuses on cross-dataset generalization for multi-view depth estimation, while E3D-Bench~\citep{cong2025e3d} provides a broader evaluation covering depth, reconstruction, pose estimation, and novel-view synthesis.
In addition, several model works construct their own evaluation protocols, including DA3~\citep{lin2025depth}, $\pi^3$~\citep{wang2025pi3}, and MapAnything~\citep{keetha2026mapanything}.
Among these, E3D-Bench is the most comprehensive standalone effort, supporting cross-method comparison across multiple tasks. 
However, it does not provide comparisons across various domains and paradigms.
The remaining model-specific suites are largely tied to individual model studies and lack a unified protocol for controlled cross-paradigm comparison.
In contrast, \benchmark provides a standalone, deterministic, and tag-aware benchmark that enables systematic analysis across diverse input densities, viewpoint types, scene dynamics, and foundation model paradigms.
\section{\benchmark Design}
\begin{figure*}
    \centering
    \includegraphics[width=1\linewidth]{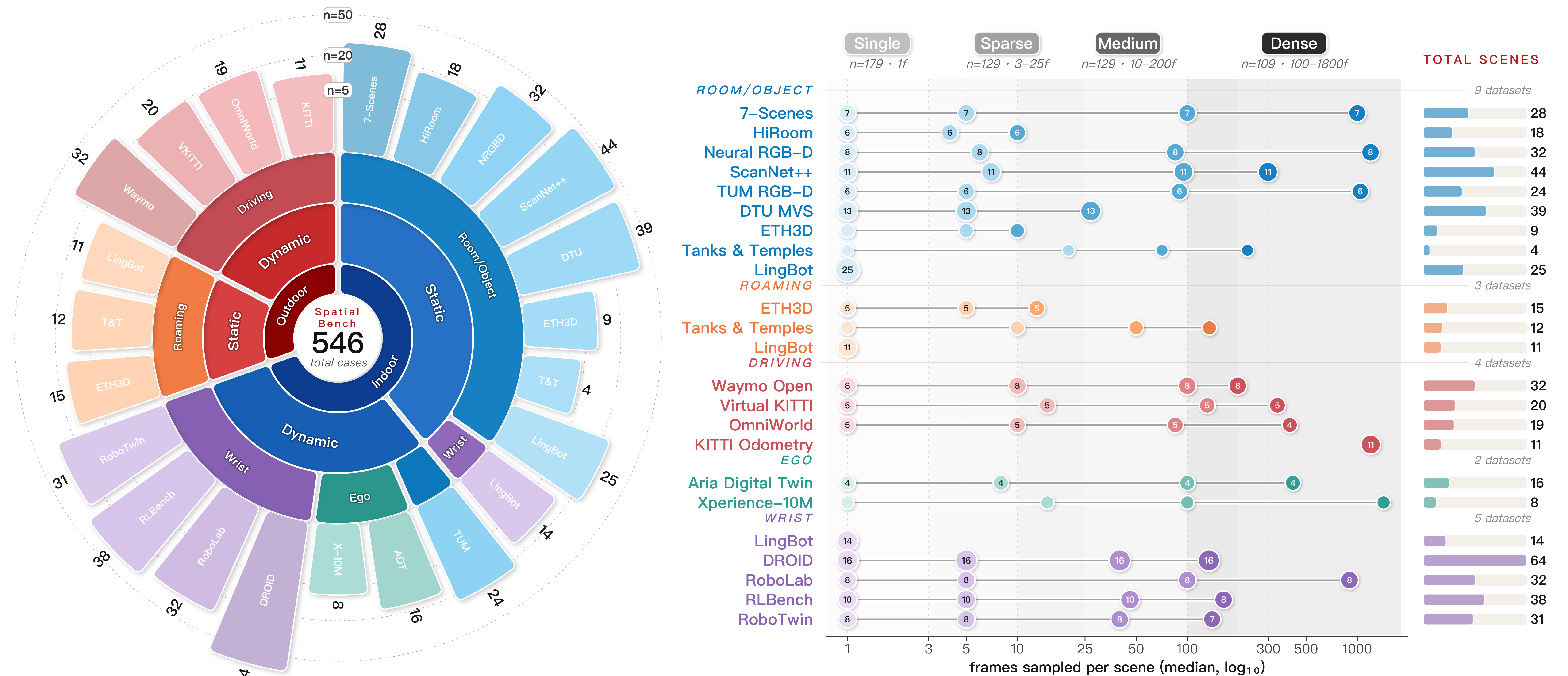}
    \caption{\textbf{Overview of \benchmark.} (Left) Distribution of all scene categories and their corresponding counts. (Right) Data sources and the median number of frames per scene under sparse, medium, and dense input settings.
    The number on each circle indicates the number of scenes.
}
    \label{fig:statistics}
\end{figure*}

\benchmark is built upon a large-scale collection of heterogeneous 3D vision datasets, covering a diverse spectrum of scene categories, capture conditions, and viewpoint configurations.
Fig.~\ref{fig:statistics} provides an overview of \benchmark: the left panel shows the breakdown of scene categories and their corresponding counts, and the right panel reports the data sources alongside the median number of frames per scene across different settings.
This multi-dimensional design allows \benchmark to evaluate model capabilities 
across a wide range of conditions in a principled and systematic way.

\subsection{Data Collection and Curation}
\label{sec:data}

\benchmark unifies heterogeneous 3D vision datasets under a common, deterministic evaluation protocol.
Raw datasets are first normalized into a shared per-scene representation comprising \textit{RGB frames}, \textit{metric depth maps}, \textit{camera-to-world poses}, and \textit{camera intrinsics}, and are subsequently curated into a fixed set of evaluation scene indices.
Each scene index is stored as a JSON record that specifies, for every (scene, view-density) pair, the exact frame indices to be consumed by a method.
By decoupling data ingestion from evaluation, this design ensures that all methods are assessed on identical inputs and that results remain fully reproducible across repeated runs.


We aggregate 19 publicly available real-world and synthetic datasets, spanning the principal axes relevant to modern 3D perception: \emph{environment} (indoor/outdoor), \emph{dynamics} (static/dynamic), \emph{viewpoint} (normal/egocentric/wrist), and \emph{data type} (real/synthetic).
For example, the whole dataset can be classified into four distinct subsets according to the \emph{dynamics} and \emph{data type} axes: static-real, static-synthetic, dynamic-real, and dynamic-synthetic:
(1) \textbf{Static-real.} 7-Scenes~\citep{shotton20137scenes}, DTU~\citep{jensen2014dtu}, NRGBD~\citep{azinovic2022Neuralrgbd}, Scannet++~\citep{yeshwanth2023scannet++}, Tanks~\&~Temples~\citep{Knapitsch2017tanks}, and ETH3D~\citep{schops2017eth3d} provide high-quality ground-truth geometry under static conditions, covering settings from close-range tabletop scans to large-scale outdoor architecture.
(2) \textbf{Static-synthetic.} Hiroom~\citep{lin2025depth}.
(3) \textbf{Dynamic-real.} TUM-Dynamic~\citep{sturm12tumdynamic}, DROID~\citep{khazatsky2024droid}, Xperience~\citep{xperience_10m}, Waymo~\citep{sun2020waymo}, and KITTI-Odometry~\citep{geiger2012kittiod} capture dynamic indoor activities as well as street-scale driving scenarios.
(4) \textbf{Dynamic-synthetic.} ADT~\citep{pan2023aria}, RLBench~\citep{james2020rlbench} with Colosseum~\citep{pumacay2024colosseum},  RoboTwin~\citep{chen2025robotwin}, Robolab~\citep{yang2026robolab}, Virtual KITTI~2~\citep{cabon2020vkitti}, and OmniWorld-Game~\citep{zhou2025omniworld} provide dense photorealistic sequences for dynamic and robotic settings that are otherwise costly to acquire in the real world.
We also collect a \textbf{Single-frame Mixture} including Lingbot-Depth~\citep{lingbot-depth2026} and all the above datasets, which contributes one-shot \texttt{rgb/depth/intrinsic} triplets, used exclusively in the monocular depth evaluation.
We refer the reader to Tab.~\ref{tab:datasets_summary} in Appendix~\ref{appendix:full_benchmark} for a complete overview of all datasets included in \benchmark.

\begin{figure}[t]
  \centering
  \includegraphics[width=0.86\textwidth]{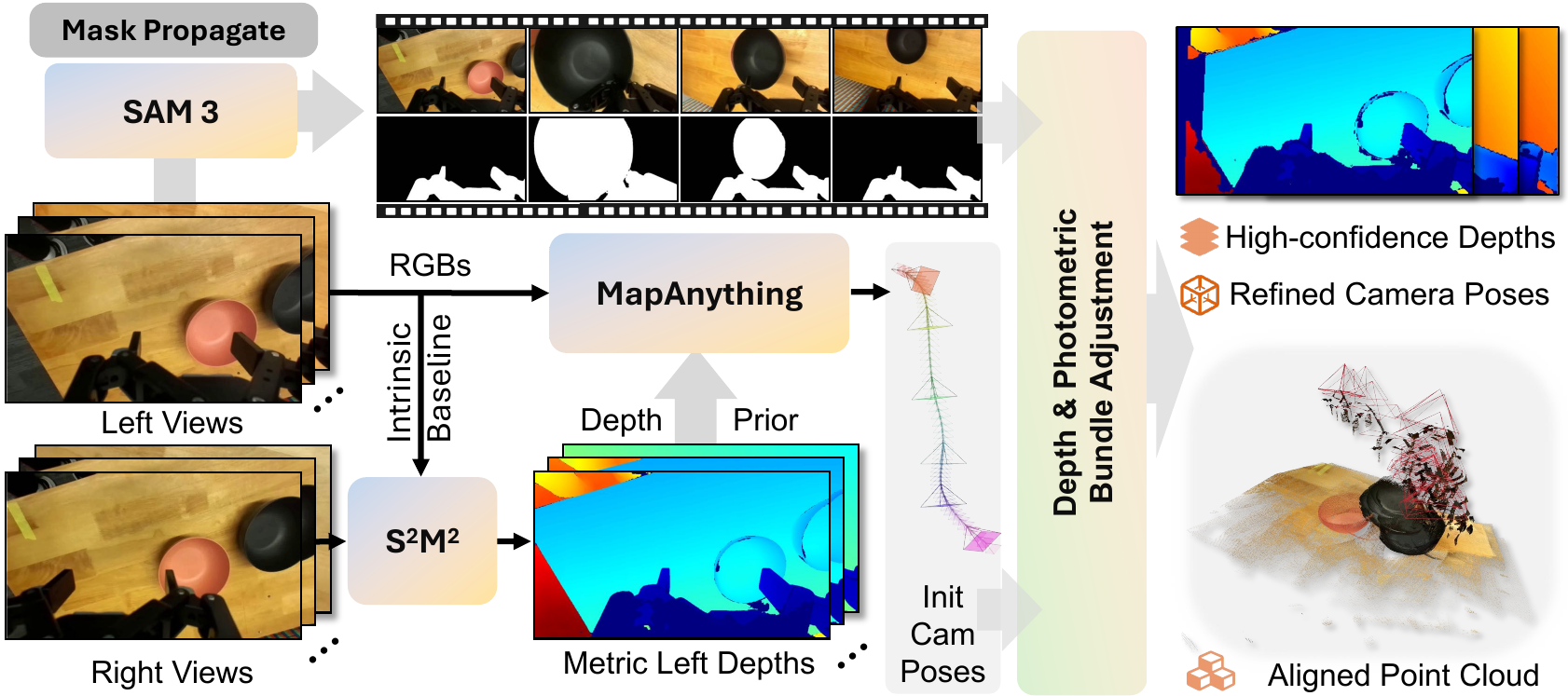}
  \vspace{-8pt}
  \caption{\textbf{DROID data curation pipeline.} We obtain metric depth from stereo sequences via S$^2$M$^2$~\protect\citep{min2025s2m2}, with initial camera poses estimated by MapAnything~\protect\citep{keetha2026mapanything}. 
  Gripper and contact affordance masks are segmented using SAM3. 
  Then the camera poses are refined via bundle adjustment using the RGBs, initial camera poses, and masks.}
  \label{fig:droid}
\end{figure}

To obtain high-quality, depth-consistent real-world wrist-view sequences, we design a dedicated data curation pipeline for the DROID~\citep{khazatsky2024droid} dataset, as illustrated in Fig.~\ref{fig:droid}.
We feed stereo video sequences into the S$^2$M$^2$~\citep{min2025s2m2} stereo depth estimation model to obtain per-frame metric depth for the left image, with unreliable points filtered out via confidence thresholding.
The resulting image sequence and metric depth maps are then passed to MapAnything~\citep{keetha2026mapanything} to obtain initial camera poses.
In parallel, we apply SAM3~\citep{carion2025sam3} to segment dynamic regions, including the gripper and objects it interacts with, on a set of keyframes, and propagate the masks to the full sequence. 
These masks exclude dynamic foreground regions from the Bundle Adjustment optimization, which assumes a static scene background.
Finally, leveraging the initial camera poses along with RGB images and the obtained masks, we perform depth \& photometric bundle adjustment to refine the camera poses, yielding globally aligned point clouds.
Other data pipeline and implementation details are provided in Appendix~\ref{appendix:pipeline}.

\subsection{Multi-density Evaluation Regimes}
\label{sec:data:regimes}

A central principle of our benchmark is that each method is evaluated across multiple temporal resolutions on the same scene, rather than on arbitrarily truncated clips. 
From each curated scene, we generate four parallel entries corresponding to distinct view-density regimes: \textbf{\textsc{Single}}, \textbf{\textsc{Sparse}}, \textbf{\textsc{Medium}}, and \textbf{\textsc{Dense}}. 
These regimes are designed to probe complementary failure modes: \textsc{Single} isolates monocular depth priors; \textsc{Sparse} stresses wide-baseline reconstruction from unordered views; \textsc{Medium} reflects the moderate-overlap inputs typical of SfM and SLAM; and \textsc{Dense} evaluates long-horizon online estimation. 
Because all four regimes are derived from the same underlying scene, scene difficulty and density-related difficulty can be disentangled.

\textbf{\textsc{Single}}.
For each scene, we fix a single deterministic frame index that is consistent across all evaluations.
This ensures that frame selection is reproducible across machines and independent of wall-clock time.

\textbf{\textsc{Sparse}}.
Sparse-view selection is formulated as a \textit{weighted set-cover problem} over the scene's 3D voxel support.
Let $\mathcal{V}$ denote the set of all voxels in the scene, and let $\mathcal{F}$ denote the candidate frames.
Each frame $f \in \mathcal{F}$ covers a subset of voxels $V_f \subseteq \mathcal{V}$.
We greedily select frames to maximize cumulative voxel coverage until a small frame budget $K$ is reached.
This deterministic procedure promotes viewpoint diversity and is robust to variations in trajectory speed, producing a compact set of views that jointly covers the scene rather than merely temporally distant frames.
The full selection objective is given in Appendix~\ref{app:regime-details}.

\textbf{\textsc{Medium}}.
The medium regime retains the set-cover formulation of the sparse regime but favors view overlap over diversity.
Let $\mathcal{V}$ be the coarsened voxel set and $\mathcal{F}$ the candidate frames.
The selected frame set $\mathcal{S}$ is obtained by greedily maximizing voxel coverage, subject to a length-adaptive frame budget.
This procedure encourages mid-range overlap among views rather than extreme viewpoint diversity.
The corresponding budgeted objective is detailed in Appendix~\ref{app:regime-details}.

\textbf{\textsc{Dense}}.
The dense regime targets the opposite end of the spectrum: an online, long-horizon setting in which a method must ingest essentially every frame of a trajectory and reconstruct temporally coherent geometry. 
Accordingly, the goal is not to \emph{select} views in the set-cover sense, but to preserve temporal continuity while avoiding the trivial inflation of evaluation cost caused by near-duplicate consecutive frames. 
In practice, however, processing arbitrarily long trajectories can exceed the memory limits of some methods; therefore, we impose a maximum frame budget to ensure that the dense evaluation remains feasible across all methods.


\subsection{Evaluated Models} 
In this work, we evaluate 31 methods with 41 model variants in total. 
We categorize the evaluated methods into six paradigms: \textit{1) Optimization-based} methods, including DUSt3R~\citep{wang2024dust3r} and MASt3R~\citep{leroy2024mast3r}; \textit{2) End-to-End Feed-Forward} methods, including VGGT~\citep{wang2025vggt}, Fast3R~\citep{yang2025fast3r}, FastVGGT~\citep{shen2025fastvggt}, MUSt3R~\citep{cabon2025must3r}, MapAnything~\citep{keetha2026mapanything}, OmniVGGT~\citep{peng2025omnivggt}, $\pi^3$~\citep{wang2025pi3}, AMB3R~\citep{wang2025amb3r}, DepthAnything3~\citep{lin2025depth}, WorldMirror~\citep{liu2025worldmirror}, and VGGT-Omega~\citep{wang2026vggtomega};  \textit{3) Online/Streaming} methods, including Spann3r~\citep{wang2024spann3r}, CUT3R~\citep{wang2025cut3r}, MonST3R~\citep{zhang2024monst3r}, Point3R~\citep{point3r}, Stream3R~\citep{stream3r2025}, StreamVGGT~\citep{streamVGGT}, PAGE4D~\citep{zhou2025page4d}, InfiniteVGGT~\citep{yuan2026infinitevggt}, WinT3R~\citep{li2025wint3r}, LongStream~\citep{cheng2026longstream} and LingBot-Map~\citep{chen2026lingbotmap}; \textit{4) Chunk-based} methods, including VGGT-Long~\citep{deng2025vggtlong}, Pi-Long~\citep{deng2025vggtlong} and DA3-Streaming~\citep{deng2025vggtlong}; \textit{5) SLAM-based} methods, including MASt3R-SLAM~\citep{murai2024_mast3rslam} and VGGT-SLAM~\citep{maggio2025vggt-slam2}; and \textit{6) Test-Time Training} methods, including TTT3R~\citep{chen2025ttt3r}, Scal3R~\citep{xie2026scal3r} and LoGeR~\citep{zhang2026loger}.
All evaluated models are listed in Table~\ref{tab:main_leaderboard_filtered}. 

\subsection{Task Description and Metrics}
To comprehensively assess the capabilities of each model, we design five general evaluation tasks across different settings as follows. 
Appendix~\ref{appendix:metrics} reports complete metric definitions.

\textbf{Camera Pose Estimation.} Given an input image sequence, we evaluate pairwise camera geometry using Relative Rotation Accuracy (RRA) and Relative Translation Accuracy (RTA), following~\citet{wang2025vggt}.
We use \textit{RAcc}$_{x}$ and \textit{TAcc}$_{x}$ to denote the fraction of pairs with angular errors below threshold $x$, and AUC$_{x}$ as the area under the joint accuracy curve of RRA and RTA up to threshold~$x$.

\textbf{Camera Trajectory Estimation.} 
For continuous image sequences, \emph{i.e.}, our medium and dense input settings, we additionally compute Absolute Trajectory Error~(\textit{ATE}), Relative Translation Error~(\textit{RPE}$_t$), and Relative Rotation Error~(\textit{RPE}$_r$), all evaluated after global Sim(3) alignment with the ground-truth trajectory, following~\citet{teed2021droid-slam}.

\textbf{Depth Estimation.} 
For single/multi-view depth estimation, we compute \textit{AbsRel}, \textit{SqRel}, \textit{RMSE}, \textit{LogRMSE}, and threshold (inlier, $\delta_{\tau}$) metrics over all valid pixels with respect to the ground-truth depth, where the inlier ratio indicates the percentage of pixels with correct predictions.
Predicted depths are aligned to the ground truth via median alignment by default.
For models with metric-scale prediction capability, we additionally report AbsRel 
both before and after alignment.

\textbf{Dense-View Reconstruction.} 
We evaluate scene-level 3D reconstruction on a subset of scenes. 
Given the ground-truth and reconstructed point sets, we compute \textit{Accuracy} and \textit{Completeness}, and report the \textit{F-score} (harmonic mean of Precision and Recall) and the \textit{Overall} score, defined as (Accuracy + Completeness) / 2, following~\citet{lin2025depth}.

\textbf{Prior-Enhanced Prediction.} 
This task targets methods that accept auxiliary prior inputs (e.g., MapAnything, OmniVGGT~\citep{keetha2026mapanything,peng2025omnivggt}).  
We evaluate their performance under two settings: with all ground-truth depth priors provided, and with all ground-truth camera pose priors provided.

\section{Depth-Anything-Next}
\label{dan}
To fill the gap of 3D foundation models in the egocentric and wrist-view domains, we introduce Depth-Anything-Next and our \dataset dataset in this section.

\subsection{\dataset Dataset}

\begin{figure}[t] 
\centering 
\includegraphics[width=\columnwidth]{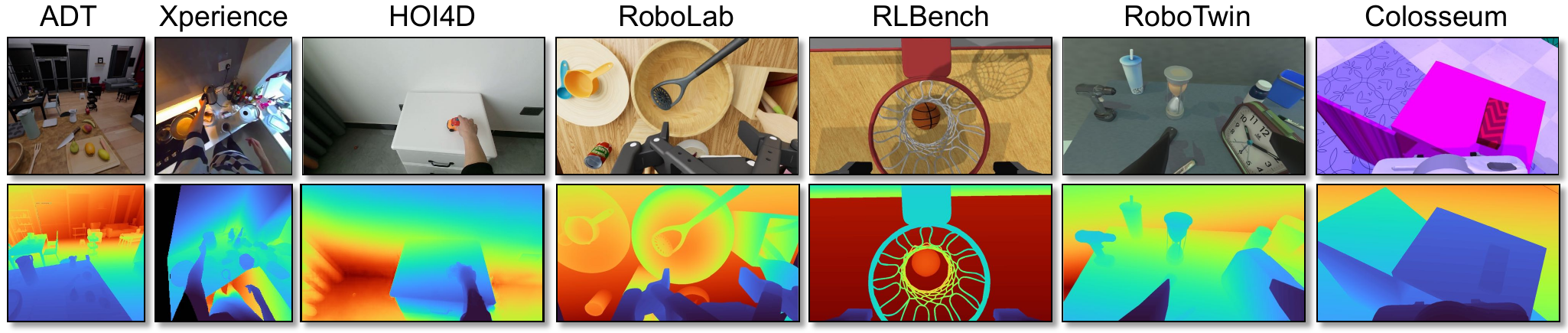}
\caption{\textbf{\dataset Data Samples}. The dataset showcases a diverse array of assets and episodes.} 
\vspace{-10pt}
\label{Fig.egoexpert} %
\end{figure}

We curate \textbf{\dataset}, a large-scale 3D dataset comprising 5.5M high-quality frames across 22K scenes, primarily collected from egocentric and wrist-view perspectives.
This category of views presents unique challenges, including high motion dynamics, frequent occlusions, and ultra-close-range capture, making it a critical data paradigm for embodied intelligence applications.
Tab.~\ref{egoexpert-table} presents the statistics of \dataset, where all datasets provide image sequences, metric depths, camera intrinsics, and extrinsics.
For simulation-based datasets, extensive domain randomization is applied to various factors, including background appearance, object size, color, and wrist camera placement.
Please refer to Appendix~\ref{appendix:datasets} for the data collection pipeline.

\subsection{Model Architecture}

\begin{wrapfigure}{r}{0.7\textwidth}
  \centering
  \vspace{-10pt}  
  \includegraphics[width=\linewidth]{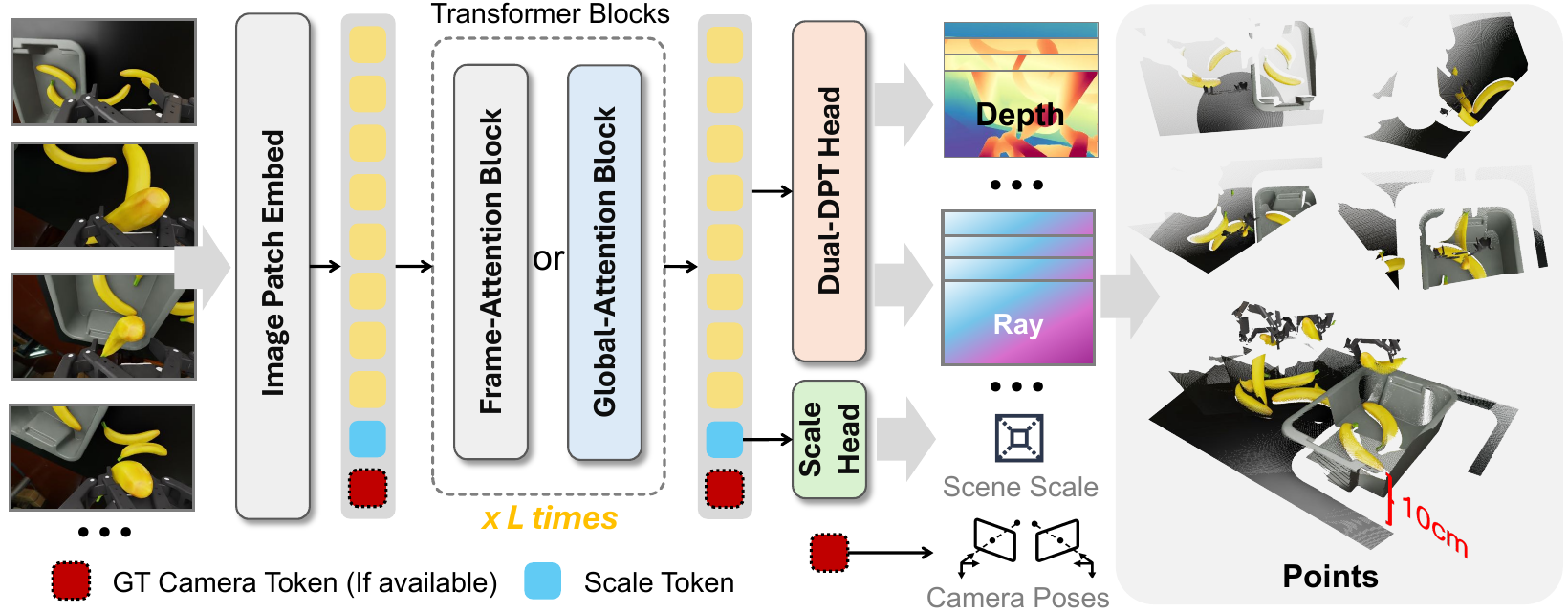}
  \vspace{-15pt}
  \caption{\textbf{Overview of Depth-Anything-Next.} The model incorporates additional scale tokens to learn the scene-level metric scale. Optionally, camera pose information can be embedded as GT camera tokens to serve as auxiliary input to provide geometric guidance.}
  \label{fig:dan}
  \vspace{-10pt}
\end{wrapfigure}

Depth-Anything-Next builds upon the architecture of DA3~\citep{lin2025depth}, while extending it with the capability to predict absolute scale in an end-to-end manner. 
Fig.~\ref{fig:dan} illustrates the overall architecture of \ours.
Specifically, \ours takes a sequence of frames $\mathbf{I} = \{I_i\}^N_{i=1}$ and auxiliary camera information $\mathbf{C} = \{C_i\}^N_{i=1}$ as input, where each $C_i=\{K_i,G_i\}$ comprises the intrinsics $K_i\in \mathbb{R}^{3\times 3}$ and the pose $G_i\in\mathrm{SE}(3)$.
All the frames are first patchified into patch tokens $\mathbf{e}_f$. 
The patch tokens from all frames are then concatenated with camera tokens $\mathbf{e}_c$ and scale tokens $\mathbf{e}_s$, and jointly processed by the transformer encoder 
$\mathcal{E}$:
$
(\hat{\mathbf{e}}_c,\hat{\mathbf{e}}_s,\hat{\mathbf{e}}_f)=\mathcal{E}(\mathbf{e}_c,\mathbf{e}_s,\mathbf{e}_f).
$
Here, $\mathbf{e}_c$ represents GT camera tokens when auxiliary camera information is available, and defaults to learnable camera tokens otherwise.
\ours adopts pure transformer blocks as its encoder, where frame-wise self-attention and global self-attention alternate throughout the layers.
After being processed by the encoder ($L$ layers), the patch tokens $\hat{\mathbf{e}}_f$ and camera tokens $\hat{\mathbf{e}}_c$ are fed into the Dual-DPT heads to produce the final depth map $\hat{\mathbf{D}}$ and ray map $\hat{\mathbf{R}}$ predictions, while the scale tokens $\hat{\mathbf{e}}_s$ are passed through a lightweight MLP to regress a scalar scale factor $\hat{S}$.
Finally, the scene point cloud is reconstructed from the predicted depth map $\hat{\mathbf{D}}$, ray map $\hat{\mathbf{R}}$, and scale factor $\hat{S}$.

\subsection{Implementation Details}

\textbf{Training Objectives.}
Following DA3~\citep{lin2025depth}, our training objective is a multi-task loss comprising five components: depths $\{\hat{\mathbf{D}},\mathbf{D}\}$, depth gradients $\{\nabla\hat{\mathbf{D}},\nabla\mathbf{D}\}$, ray maps $\{\hat{\mathbf{R}},\mathbf{R}\}$, points $\{\hat{\mathbf{P}},\mathbf{P}\}$, and scale $\{\hat{S},S\}$ supervision.
The total loss is defined as $\mathcal{L} = \mathcal{L}_{\text{depth}} + \alpha \mathcal{L}_{\text{grad}}+ \mathcal{L}_{\text{ray}} + \mathcal{L}_{\text{pmap}} + \mathcal{L}_{\text{scale}}$.
Prior to loss computation, all ground-truth signals are canonicalized into the first-camera coordinate frame and then normalized by a per-scene scale factor $S$, defined as the mean $\ell_2$ norm of the valid ground-truth world points $\mathbf{P}$. We divide $\mathbf{P}$, $\mathbf{D}$, and the camera translations by $S$, and use $S$ itself as the regression target of the scale head, so that $\hat{S}$ recovers the absolute metric scale to which the other predictions are invariant.
All loss terms are based on the $\ell_1$ norm with $\alpha=1$, where the scale loss is defined as: $\mathcal{L}_{\text{scale}} = \| f_{\log}(\hat{S}) - f_{\log}(S) \|_1$ and $f_{\log }: \mathbf{x} \rightarrow(\mathbf{x} /\|\mathbf{x}\|) \cdot \log (1+\|\mathbf{x}\|)$.

\textbf{Datasets.} 
We train our model mainly on \dataset.
To mitigate potential generalization degradation, we retain data from 11 general 3D datasets for joint training, including Mapfree~\citep{arnold2022mapfree}, Hypersim~\citep{roberts2021hypersim}, Infinigen~\citep{raistrick2024infinigen}, etc.
The full datasheet and dataset mixture can be found in the Appendix~\ref{appendix:dan}.

\textbf{Training Details.}
Our \ours architecture follows DA3-Giant~\citep{lin2025depth} with $L = 41$ Transformer blocks and is initialized by pre-trained weights.
During training, each batch incorporates ground-truth camera information as auxiliary input with probability $p=20\%$.
The training runs end-to-end on 4 NVIDIA H200 GPUs over seven days.
We provide complete implementation details in Appendix~\ref{appendix:dan}.

\section{Findings: How to Train Your Best Spatial Foundation Models?}
\begin{table*}[!t]
\centering
\setlength{\fboxsep}{1pt}
\caption{\textbf{Main Results on \benchmark.}
We report performance across four input settings:
\textit{Single Frame}, \textit{Sparse}, \textit{Medium}, \textit{Dense}, and their \textit{Average} across all settings.
\textit{Time} is the per-sequence inference time in seconds reported in the sparse regime, averaged over all scenes.
The best, second-best, and third-best results in each column are highlighted in
\colorbox{bestred}{deep blue}, \colorbox{secondorange}{medium blue}, and \colorbox{thirdyellow}{light blue}, respectively.
Out-of-memory (OOM, $>$ 140G) and Timeout (T.O, $>$ 4h per scene) cells are shaded \colorbox{oomred}{light red}.
Within each sub-category, the \textbf{bold} value marks the in-group best.
}
\label{tab:main_leaderboard_filtered}
\resizebox{\textwidth}{!}{%
\renewcommand{\arraystretch}{1.15}
\setlength{\tabcolsep}{3pt}
\begin{tabular}{l cc >{\columncolor{subcol}}c >{\columncolor{subcol}}c >{\columncolor{subcol}}c >{\columncolor{subcol}}c >{\columncolor{subcol}}c >{\columncolor{subcol}}c >{\columncolor{subcol}}c >{\columncolor{subcol}}c >{\columncolor{subcol}}c >{\columncolor{subcol}}c >{\columncolor{subcol}}c >{\columncolor{subcol}}c >{\columncolor{subcol}}c >{\columncolor{subcol}}c >{\columncolor{subcol}}c}
\toprule
\multirow{2.5}{*}{\textbf{Method}}
& \multirow{2.5}{*}{\makecell{\textbf{\#Params}\\\textbf{(M)}}}
& \multirow{2.5}{*}{\makecell{\textbf{Time}\\\textbf{(s)}}}
& \multicolumn{1}{c}{\textbf{Single Frame}}
& \multicolumn{2}{c}{\textbf{Sparse}}
& \multicolumn{4}{c}{\textbf{Medium}}
& \multicolumn{4}{c}{\textbf{Dense}}
& \multicolumn{4}{c}{\textbf{Average}} \\
\cmidrule(lr){4-4} \cmidrule(lr){5-6} \cmidrule(lr){7-10} \cmidrule(lr){11-14} \cmidrule(lr){15-18}
 & & &
 AbsRel$\downarrow$ &
 AbsRel$\downarrow$ & AUC@30$\uparrow$ &
 AbsRel$\downarrow$ & AUC@30$\uparrow$ & ATE$\downarrow$ & F-Score$\uparrow$ &
 AbsRel$\downarrow$ & AUC@30$\uparrow$ & ATE$\downarrow$ & F-Score$\uparrow$ &
 AbsRel$\downarrow$ & AUC@30$\uparrow$ & ATE$\downarrow$ & F-Score$\uparrow$ \\
\midrule
\multicolumn{18}{c}{\cellcolor{catgray}\textbf{Optimization-based}} \\
\midrule
 DUSt3R & 571.17 & \textbf{7.59} & \textbf{0.385} & 0.257 & 0.498 & 0.276 & 0.448 & \textbf{1.691} & 0.343 & \cellcolor{oomred}OOM & \cellcolor{oomred}OOM & \cellcolor{oomred}OOM & \cellcolor{oomred}OOM & (0.306) & (0.473) & (1.691) & (0.343) \\
 MASt3R & 688.64 & 8.17 & 0.456 & \textbf{0.209} & \textbf{0.568} & \textbf{0.259} & \textbf{0.522} & 1.911 & \textbf{0.370} & \cellcolor{oomred}OOM & \cellcolor{oomred}OOM & \cellcolor{oomred}OOM & \cellcolor{oomred}OOM & (0.308) & (0.545) & (1.911) & (0.370) \\
\midrule
\multicolumn{18}{c}{\cellcolor{catgray}\textbf{End-to-End Feed-Forward}} \\
\midrule
 VGGT & 1256.54 & 0.40 & \cellcolor{secondorange}0.184 & 0.105 & 0.700 & 0.125 & 0.687 & 0.727 & 0.661 & \cellcolor{oomred}OOM & \cellcolor{oomred}OOM & \cellcolor{oomred}OOM & \cellcolor{oomred}OOM & (0.138) & (0.694) & (0.727) & (0.661) \\
 Fast3R & 647.55 & 0.90 & 0.350 & 0.260 & 0.392 & 0.255 & 0.386 & 6.582 & 0.300 & 0.331 & 0.232 & \textbf{13.68} & 0.224 & 0.299 & 0.337 & 10.13 & 0.262 \\
 FastVGGT & 1157.94 & 0.24 & \cellcolor{bestred}\textbf{0.183} & 0.113 & 0.631 & 0.105 & 0.662 & 0.738 & 0.576 & \cellcolor{secondorange}0.120 & \textbf{0.588} & 19.23 & \cellcolor{thirdyellow}\textbf{0.479} & \cellcolor{secondorange}\textbf{0.130} & 0.627 & 9.984 & \textbf{0.527} \\
 MUSt3R & 423.43 & 0.96 & 0.429 & 0.165 & 0.614 & 0.162 & 0.643 & 3.097 & 0.507 & \cellcolor{oomred}T.O & \cellcolor{oomred}T.O & \cellcolor{oomred}T.O & \cellcolor{oomred}T.O & (0.252) & (0.629) & (3.097) & (0.507) \\
 MapAnything & 1228.49 & 0.22 & 0.451 & 0.153 & 0.579 & 0.146 & 0.579 & 1.737 & 0.420 & \cellcolor{oomred}OOM & \cellcolor{oomred}OOM & \cellcolor{oomred}OOM & \cellcolor{oomred}OOM & (0.250) & (0.579) & (1.737) & (0.420) \\
 OmniVGGT & 1217.49 & \cellcolor{thirdyellow}0.22 & \cellcolor{thirdyellow}0.188 & 0.117 & 0.665 & 0.111 & 0.665 & 1.491 & 0.595 & \cellcolor{oomred}OOM & \cellcolor{oomred}OOM & \cellcolor{oomred}OOM & \cellcolor{oomred}OOM & (0.139) & (0.665) & (1.491) & (0.595) \\
 $\pi^{3}$ & 958.70 & \cellcolor{secondorange}\textbf{0.20} & 0.478 & 0.092 & 0.742 & \cellcolor{thirdyellow}0.082 & 0.749 & 0.565 & 0.649 & \cellcolor{bestred}\textbf{0.109} & 0.524 & 16.39 & 0.332 & 0.190 & \textbf{0.672} & \textbf{8.478} & 0.491 \\
 $\pi^{3}$-X & 1360.03 & 0.24 & 0.371 & \cellcolor{thirdyellow}0.084 & 0.741 & \cellcolor{secondorange}0.078 & 0.744 & \cellcolor{bestred}\textbf{0.369} & 0.658 & \cellcolor{oomred}OOM & \cellcolor{oomred}OOM & \cellcolor{oomred}OOM & \cellcolor{oomred}OOM & (0.178) & (0.742) & (0.369) & (0.658) \\
 AMB3R & 1563.12 & 0.53 & 0.466 & 0.088 & 0.739 & 0.085 & 0.727 & 0.645 & 0.554 & \cellcolor{oomred}OOM & \cellcolor{oomred}OOM & \cellcolor{oomred}OOM & \cellcolor{oomred}OOM & (0.213) & (0.733) & (0.645) & (0.554) \\
 DA3-Small & 34.30 & 0.39 & 0.385 & 0.191 & 0.476 & 0.176 & 0.479 & 4.850 & 0.432 & 0.208 & 0.368 & 28.12 & 0.325 & 0.240 & 0.441 & 16.48 & 0.379 \\
 DA3-Base & 135.37 & 0.40 & 0.349 & 0.159 & 0.566 & 0.142 & 0.562 & 3.865 & 0.515 & 0.166 & 0.436 & 26.35 & 0.399 & 0.204 & 0.521 & 15.11 & 0.457 \\
 DA3-Large & 410.94 & 0.41 & 0.333 & 0.128 & 0.688 & 0.105 & 0.701 & 2.722 & 0.626 & \cellcolor{oomred}OOM & \cellcolor{oomred}OOM & \cellcolor{oomred}OOM & \cellcolor{oomred}OOM & (0.189) & (0.694) & (2.722) & (0.626) \\
 DA3-Giant & 1355.67 & 0.47 & 0.368 & 0.095 & \cellcolor{thirdyellow}0.785 & 0.086 & \cellcolor{secondorange}0.776 & 1.161 & \cellcolor{bestred}\textbf{0.742} & \cellcolor{oomred}OOM & \cellcolor{oomred}OOM & \cellcolor{oomred}OOM & \cellcolor{oomred}OOM & (0.183) & (0.780) & (1.161) & (0.742) \\
 DA3-Nested & 1689.85 & 0.52 & 0.364 & 0.106 & 0.779 & 0.086 & \cellcolor{thirdyellow}0.770 & 1.980 & \cellcolor{secondorange}0.737 & \cellcolor{oomred}OOM & \cellcolor{oomred}OOM & \cellcolor{oomred}OOM & \cellcolor{oomred}OOM & (0.185) & (0.774) & (1.980) & (0.737) \\
 WorldMirror & 1263.34 & 0.22 & 0.349 & 0.139 & 0.660 & 0.118 & 0.674 & 1.357 & 0.575 & \cellcolor{oomred}OOM & \cellcolor{oomred}OOM & \cellcolor{oomred}OOM & \cellcolor{oomred}OOM & (0.202) & (0.667) & (1.357) & (0.575) \\
 VGGT-Omega & 1143.81 & 0.48 & 0.516 & \cellcolor{bestred}\textbf{0.077} & \cellcolor{bestred}\textbf{0.803} & \cellcolor{bestred}\textbf{0.067} & \cellcolor{bestred}\textbf{0.795} & 0.659 & 0.706 & -- & -- & -- & -- & (0.220) & (0.799) & (0.659) & (0.706) \\ \hdashline
\makecell[l]{\ours$^{\dagger}$ (Ours)} & 1303.76 & 0.50
 & \makecell{0.166\\{\scriptsize\textcolor{gaingreen}{($-$54.9\%)}}}
 & \makecell{0.050\\{\scriptsize\textcolor{gaingreen}{($-$47.4\%)}}}
 & \makecell{0.809\\{\scriptsize\textcolor{gaingreen}{($+$3.1\%)}}}
 & \makecell{0.035\\{\scriptsize\textcolor{gaingreen}{($-$59.3\%)}}}
 & \makecell{0.819\\{\scriptsize\textcolor{gaingreen}{($+$5.5\%)}}}
 & \makecell{1.442\\{\scriptsize\textcolor{lossred}{($+$24.2\%)}}}
 & \makecell{0.727\\{\scriptsize\textcolor{lossred}{($-$2.0\%)}}}
 & \cellcolor{oomred}OOM & \cellcolor{oomred}OOM & \cellcolor{oomred}OOM & \cellcolor{oomred}OOM
 & (0.084) & (0.814) & (1.442) & (0.727) \\
\midrule
\multicolumn{18}{c}{\cellcolor{catgray}\textbf{Online}} \\
\midrule
 Spann3r$^{224}$ & 658.69 & 0.55 & 0.370 & 0.274 & 0.329 & 0.252 & 0.361 & 4.312 & 0.254 & 0.315 & 0.246 & 26.48 & 0.159 & 0.303 & 0.312 & 15.4 & 0.207 \\
 CUT3R & 793.31 & 0.41 & 0.247 & 0.196 & 0.519 & 0.189 & 0.469 & 2.676 & 0.286 & 0.260 & 0.165 & 25.54 & 0.109 & 0.223 & 0.384 & 14.11 & 0.197 \\
 MonST3R & 571.17 & 20.81 & 0.309 & 0.227 & 0.269 & 0.241 & 0.195 & 2.234 & 0.081 & \cellcolor{oomred}OOM & \cellcolor{oomred}OOM & \cellcolor{oomred}OOM & \cellcolor{oomred}OOM & (0.259) & (0.232) & (2.234) & (0.081) \\
 Point3R & 828.01 & 1.05 & 0.379 & 0.221 & 0.339 & 0.228 & 0.303 & 6.512 & 0.211 & 0.285 & 0.212 & 28.09 & 0.139 & 0.278 & 0.285 & 17.3 & 0.175 \\
 Stream3R-S & 1190.60 & 0.62 & 0.409 & 0.114 & 0.603 & 0.204 & 0.427 & 5.717 & 0.348 & \cellcolor{oomred}OOM & \cellcolor{oomred}OOM & \cellcolor{oomred}OOM & \cellcolor{oomred}OOM & (0.242) & (0.515) & (5.717) & (0.348) \\
 Stream3R-W & 1190.60 & 0.62 & 0.409 & 0.117 & 0.597 & 0.240 & 0.364 & 6.756 & 0.323 & \cellcolor{oomred}OOM & \cellcolor{oomred}OOM & \cellcolor{oomred}OOM & \cellcolor{oomred}OOM & (0.255) & (0.480) & (6.756) & (0.323) \\
 StreamVGGT & 1256.54 & 0.85 & 0.219 & 0.154 & 0.598 & 0.171 & 0.562 & 4.940 & 0.397 & 0.198 & 0.413 & 26.9 & 0.251 & 0.185 & 0.524 & 15.92 & 0.324 \\
 Page4D & 1256.81 & 0.56 & 0.228 & \textbf{0.112} & 0.608 & \textbf{0.107} & 0.618 & 0.855 & \textbf{0.423} & \cellcolor{oomred}OOM & \cellcolor{oomred}OOM & \cellcolor{oomred}OOM & \cellcolor{oomred}OOM & (0.149) & (0.613) & (0.855) & (0.423) \\
 InfiniteVGGT & 1256.54 & 0.46 & \textbf{0.217} & 0.154 & 0.596 & 0.170 & 0.563 & 4.964 & 0.402 & 0.197 & 0.416 & 27.01 & 0.254 & 0.184 & 0.525 & 15.99 & 0.328 \\
 Wint3R & 749.46 & 0.41 & 0.619 & 0.157 & 0.499 & 0.144 & 0.444 & 3.944 & 0.401 & 0.234 & 0.202 & 27.8 & 0.114 & 0.288 & 0.382 & 15.87 & 0.258 \\
 LongStream-B & 1190.60 & 0.59 & 0.523 & 0.153 & 0.549 & 0.224 & 0.455 & 0.925 & 0.135 & 0.269 & 0.294 & 5.766 & 0.083 & 0.292 & 0.433 & 3.345 & 0.109 \\
 LongStream-S & 1190.60 & 0.83 & 0.523 & 0.151 & 0.543 & 0.166 & 0.385 & 1.188 & 0.126 & 0.279 & 0.218 & 10.08 & 0.083 & 0.280 & 0.382 & 5.634 & 0.105 \\
 LingbotMap$^{*}$-W & 1157.94 & \textbf{0.30} & 0.333 & 0.138 & \textbf{0.650} & 0.114 & 0.641 & 0.509 & 0.362 & 0.167 & 0.553 & 4.694 & 0.352 & 0.188 & 0.615 & 2.602 & 0.357 \\
 LingbotMap$^{*}$-S & 1157.94 & 0.33 & 0.333 & 0.138 & \textbf{0.650} & 0.114 & \textbf{0.647} & \textbf{0.508} & 0.411 & \cellcolor{thirdyellow}\textbf{0.139} & \cellcolor{bestred}\textbf{0.627} & \cellcolor{secondorange}\textbf{3.470} & \textbf{0.472} & \textbf{0.181} & \textbf{0.641} & \cellcolor{secondorange}\textbf{1.989} & \textbf{0.442} \\
\midrule
\multicolumn{18}{c}{\cellcolor{catgray}\textbf{Chunk-wise}} \\
\midrule
 VGGT-Long & 1256.54 & \cellcolor{bestred}\textbf{0.20} & \cellcolor{secondorange}\textbf{0.184} & 0.105 & 0.700 & 0.131 & 0.679 & 0.512 & 0.633 & 0.222 & 0.507 & 8.467 & 0.467 & \textbf{0.161} & 0.629 & 4.490 & \cellcolor{thirdyellow}0.550 \\
 $\pi^{3}$-Long & 958.70 & 0.23 & 0.478 & \textbf{0.092} & 0.742 & 0.097 & 0.740 & \cellcolor{thirdyellow}\textbf{0.465} & 0.590 & \textbf{0.216} & \cellcolor{secondorange}\textbf{0.614} & \cellcolor{thirdyellow}\textbf{4.021} & 0.251 & 0.221 & \cellcolor{secondorange}0.699 & \cellcolor{thirdyellow}\textbf{2.243} & 0.420 \\
 DA3-Streaming & 1355.67 & 0.51 & 0.368 & 0.095 & \cellcolor{secondorange}\textbf{0.785} & \textbf{0.091} & \textbf{0.767} & 0.563 & \cellcolor{thirdyellow}\textbf{0.725} & 0.245 & 0.546 & 8.575 & \cellcolor{bestred}\textbf{0.516} & 0.200 & \cellcolor{bestred}\textbf{0.699} & 4.569 & \cellcolor{bestred}\textbf{0.621} \\
\midrule
\multicolumn{18}{c}{\cellcolor{catgray}\textbf{SLAM-based}} \\
\midrule
 MASt3R-SLAM & 688.64 & 3.04 & 0.348 & 0.336 & 0.190 & 0.348 & 0.262 & 6.075 & 0.130 & 0.404 & 0.311 & 25.7 & 0.121 & 0.359 & 0.254 & 15.89 & 0.126 \\
 VGGT-SLAM & 1256.54 & \textbf{0.57} & \cellcolor{secondorange}\textbf{0.184} & \textbf{0.105} & \textbf{0.700} & \textbf{0.129} & \textbf{0.645} & \textbf{0.686} & \textbf{0.610} & \textbf{0.211} & \textbf{0.441} & \textbf{9.069} & \textbf{0.384} & \cellcolor{thirdyellow}\textbf{0.157} & \textbf{0.595} & \textbf{4.878} & \textbf{0.497} \\
\midrule
\multicolumn{18}{c}{\cellcolor{catgray}\textbf{Test-Time Training}} \\
\midrule
 TTT3R & 793.31 & 0.61 & 0.247 & 0.202 & 0.469 & 0.179 & 0.493 & 2.343 & 0.294 & 0.222 & 0.321 & 21.07 & 0.173 & 0.212 & 0.428 & 11.71 & 0.233 \\
 Scal3R & 1266.14 & 2.32 & 0.227 & 0.114 & \textbf{0.732} & 0.147 & 0.670 & \cellcolor{secondorange}\textbf{0.400} & \textbf{0.671} & 0.244 & 0.480 & \cellcolor{bestred}\textbf{2.396} & \cellcolor{secondorange}\textbf{0.498} & 0.183 & 0.627 & \cellcolor{bestred}\textbf{1.398} & \cellcolor{secondorange}\textbf{0.585} \\
 LoGeR & 1254.62 & \textbf{0.26} & 0.251 & 0.095 & 0.687 & 0.113 & 0.693 & 0.591 & 0.504 & 0.197 & 0.552 & 5.217 & 0.335 & 0.164 & 0.644 & 2.904 & 0.419 \\
 LoGeR$^*$ & 1254.60 & 0.30 & \textbf{0.200} & \cellcolor{secondorange}\textbf{0.077} & 0.708 & \textbf{0.083} & \textbf{0.714} & 0.566 & 0.574 & \textbf{0.156} & \cellcolor{thirdyellow}\textbf{0.598} & 4.598 & 0.421 & \cellcolor{bestred}\textbf{0.129} & \cellcolor{thirdyellow}\textbf{0.673} & 2.582 & 0.497 \\
\bottomrule
\end{tabular}%
}
{\scriptsize
\setlength{\baselineskip}{8.5pt}%
``S'' denotes \textbf{stream},
``B'' denotes \textbf{batch}, and
``W'' denotes \textbf{window}.
LingbotMap$^{*}$ indicates the best checkpoint is selected in each regime.
\ours$^{\dagger}$: Numbers in (parentheses) below each \ours entry give the relative gap to DA3-Giant.
 Values in parentheses in the \textit{Average} column indicate that the method runs OOM on the dense regime, and the average is therefore computed over fewer settings. Such methods are excluded from per-column rankings in the \textit{Average} column. Note that \ours is excluded from the per-column rankings.
\par
}
\vspace{-10pt}
\end{table*}

Tab.~\ref{tab:main_leaderboard_filtered} presents the main results on \benchmark, reporting depth, camera pose, trajectory, and point cloud metrics across single-frame, sparse, medium, dense, and average settings for 41 models spanning six reconstruction paradigms.
We refer the reader to Appendix~\ref{appendix:completeresults} for per-regime sub-tables and per-dataset metric breakdowns.
\begin{figure}[t]
  \centering
  \begin{minipage}[t]{0.49\textwidth}
    \centering
    \includegraphics[width=\linewidth]{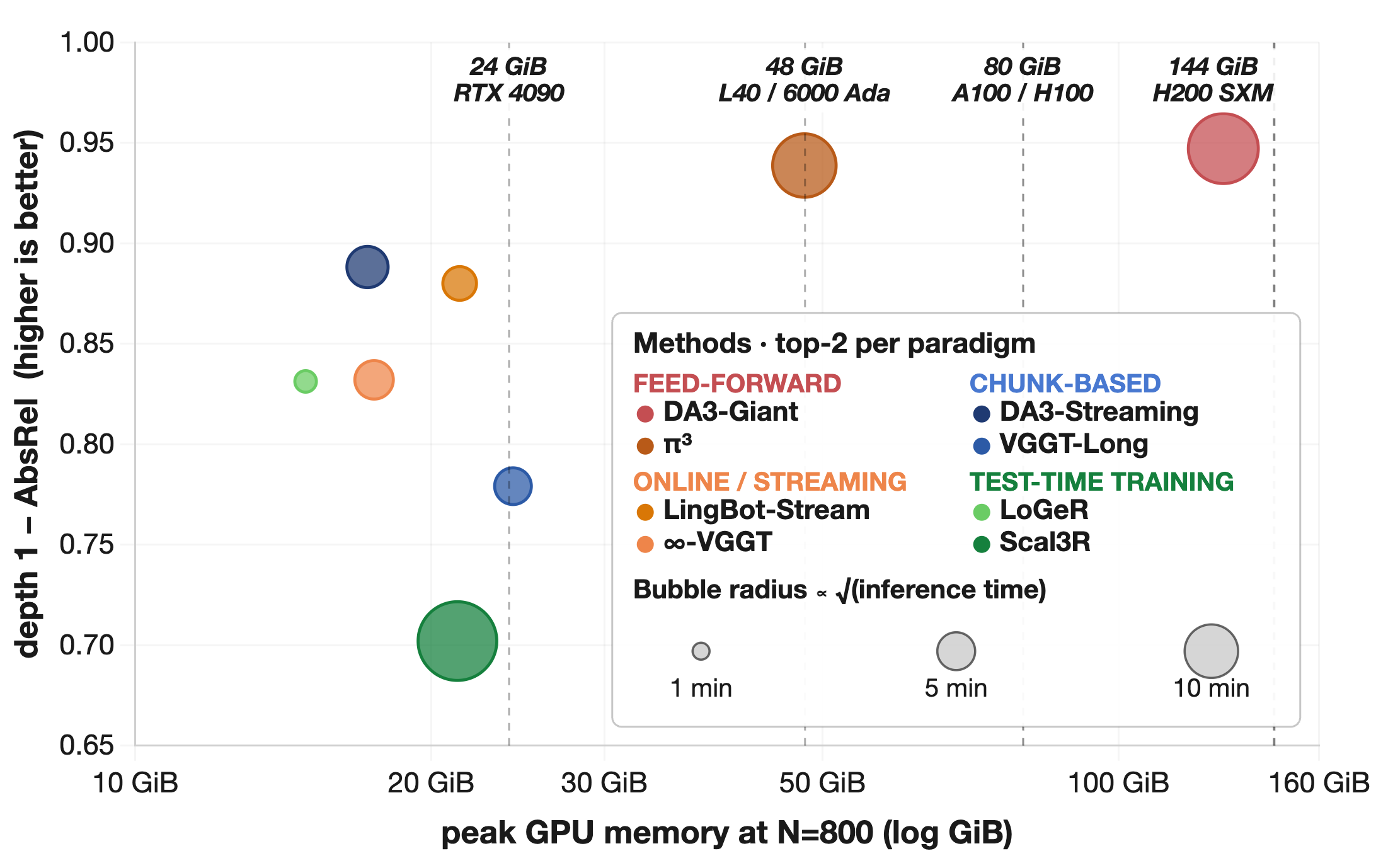}
    \captionof{figure}{\textbf{Operating snapshot.} Memory, depth error, and inference time at $N=800$.}
    \label{fig:geobench_operating_snapshot}
  \end{minipage}
  \hfill
  \begin{minipage}[t]{0.49\textwidth}
    \centering
    \includegraphics[width=\linewidth]{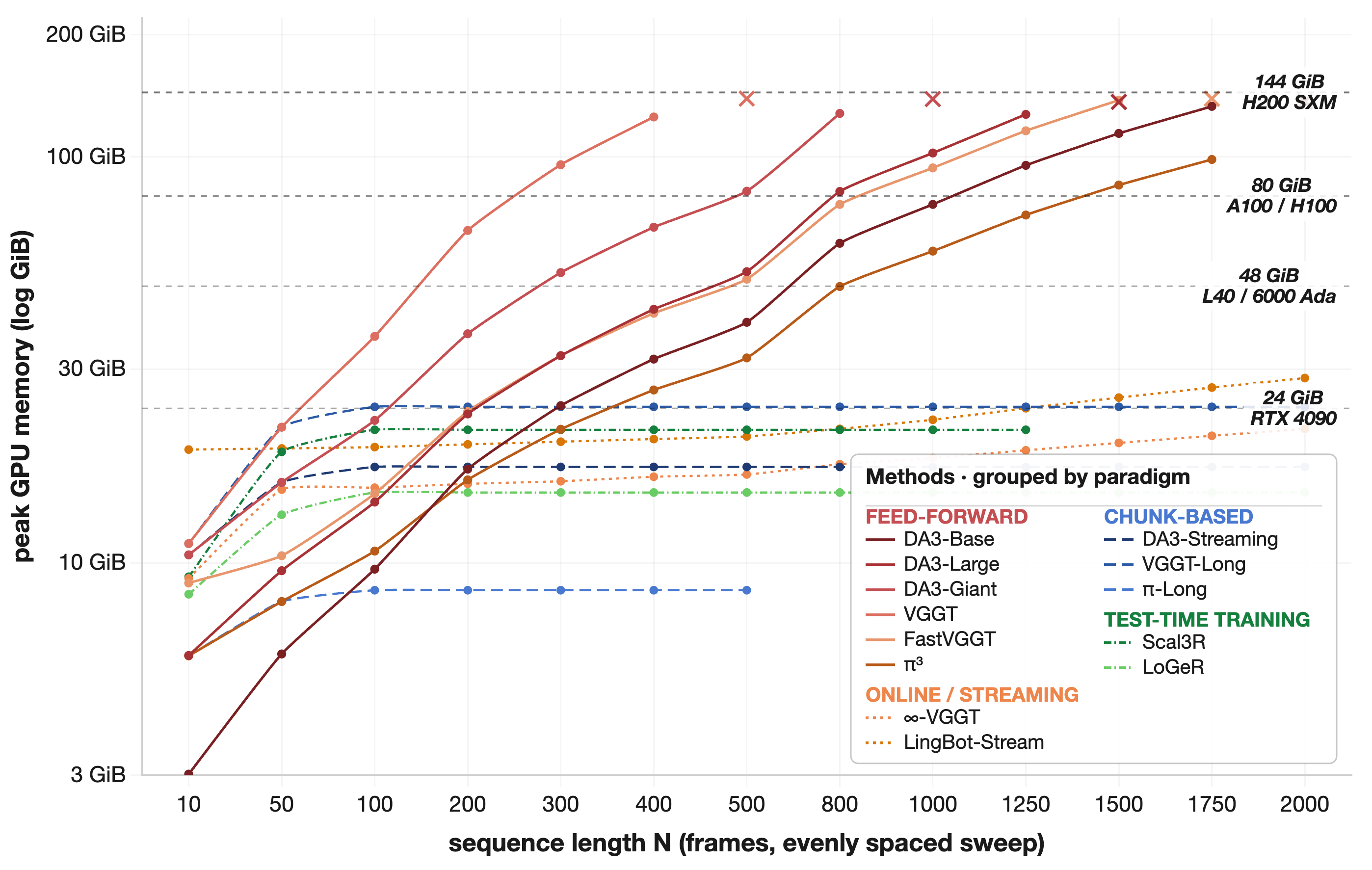}
    \captionof{figure}{\textbf{Memory scaling.} Peak GPU memory versus input sequence length.}
    \label{fig:geobench_memory_scaling}
  \end{minipage}
\end{figure}

\noindent\textbf{Full-Context Attention Sets the Accuracy Upper Bound on High-Memory GPUs.}
Fig.~\ref{fig:geobench_operating_snapshot} compares representative models at a fixed sequence length of $N=800$, where all compared methods successfully complete the input.
Under this shared input budget, full-context feed-forward models occupy the strongest accuracy region of the memory--accuracy Pareto plot.
In particular, DA3-Giant and $\pi^{3}$ achieve the lowest depth errors among the compared paradigms, outperforming streaming and online-map variants in reconstruction accuracy.
This indicates that globally coupled attention over the input sequence remains a highly effective representation mechanism for geometric reasoning.
By jointly resolving cross-view correspondences and enforcing scene-level consistency, full-context models provide stronger reconstruction accuracy and generalization under the same budget.
\observationbox{
\textbf{Takeaway}: \textit{Under the same input budget, the full-context attention models still define the accuracy upper bound, suggesting that globally coupled representations remain crucial for high-fidelity 3D geometry estimation and reconstruction.}
}

\noindent\textbf{Bounded-Memory Modeling Enables Long-Sequence Reconstruction on Limited GPUs.}
The accuracy advantage of full-context models comes with a clear physical constraint.
As shown in Fig.~\ref{fig:geobench_memory_scaling}, their GPU memory consumption grows rapidly with sequence length and eventually reaches the out-of-memory boundary on dense long-horizon inputs.
Streaming, online, chunk-wise, and TTT variants exhibit a different scaling behavior: by restricting the active context through causal updates, sliding windows, or chunked inference, they maintain substantially flatter memory curves and can continue processing sequences that full-context models cannot complete under the same hardware budget.
As sequence length grows, these methods show stronger trajectory-level consistency (lower ATE), reflecting their ability to maintain long-range geometric alignment.
However, their depth estimation accuracy remains below that of full-context models across all input regimes, indicating that bounded-memory inference trades pairwise geometric precision for scalability.
Thus, these two model families occupy complementary operating regimes: full-context models are preferable when accuracy on bounded inputs is the priority, while bounded-memory methods are better suited for long-horizon or resource-constrained deployment.
\observationbox{
\textbf{Takeaway}: \textit{Streaming, chunk-wise, and TTT models trade part 
of the full-context accuracy advantage for bounded-memory inference, enabling 
continuous long-horizon reconstruction under realistic hardware constraints.}
}

\begin{figure}[t]
  \centering
  \begin{minipage}[t]{0.49\textwidth}
    \centering
    \includegraphics[width=\linewidth]{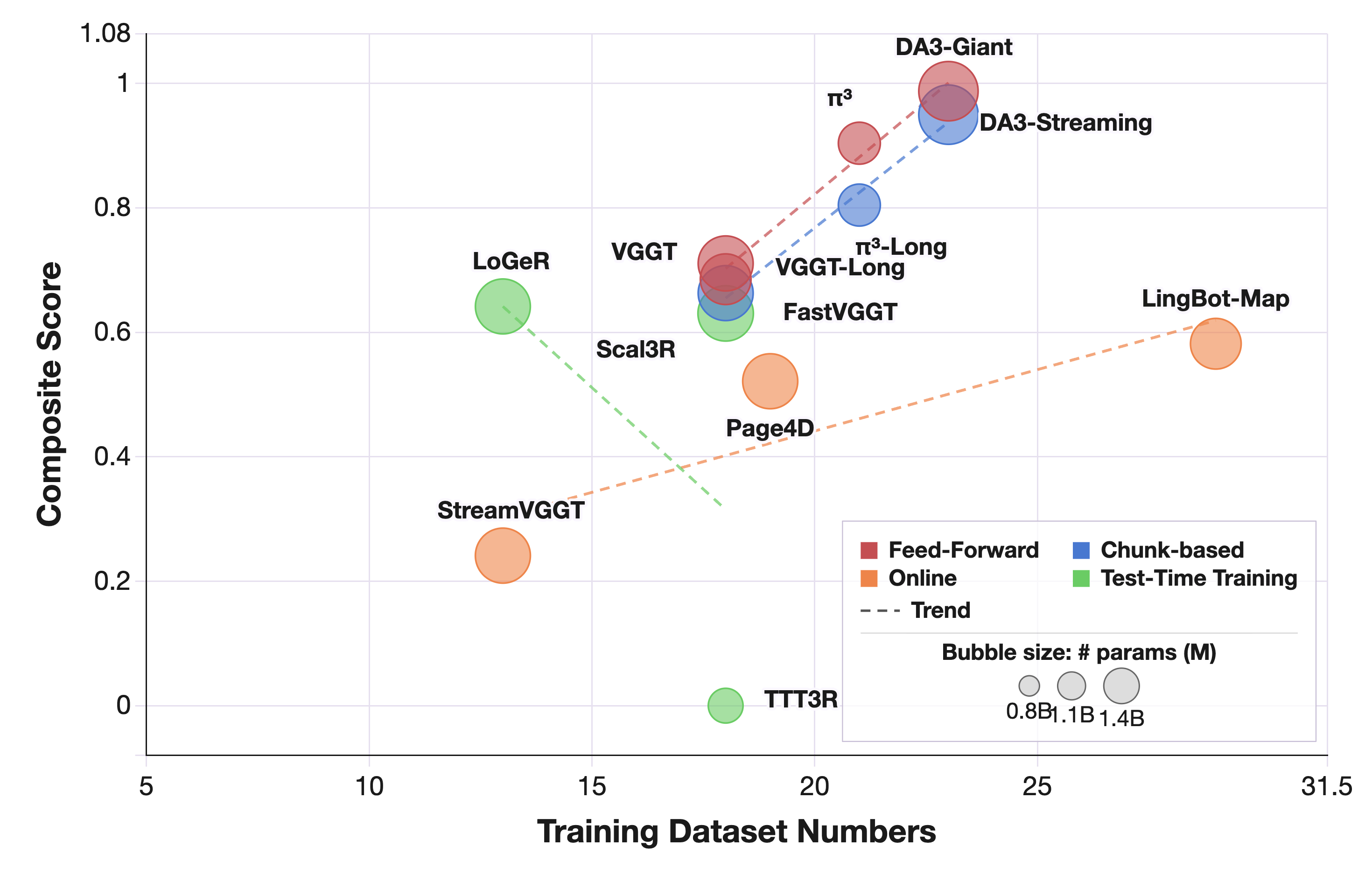}
    \captionof{figure}{\textbf{Training coverage.} Dataset count, parameter scale, and \benchmark accuracy.}
    \label{fig:geobench_training_coverage}
  \end{minipage}
  \hfill
  \begin{minipage}[t]{0.49\textwidth}
    \centering
    \includegraphics[width=\linewidth]{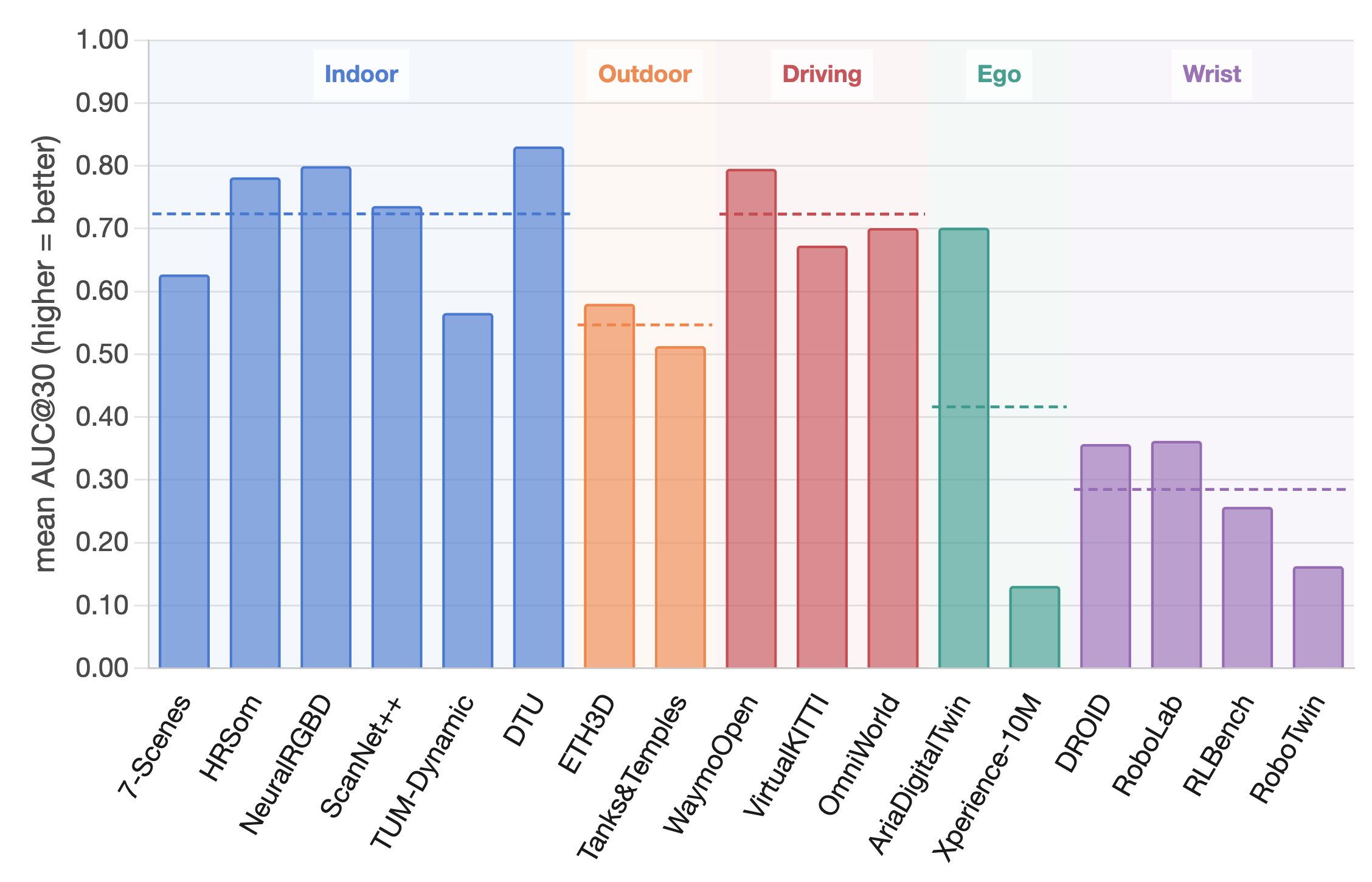}
    \captionof{figure}{\textbf{Domain-level OOD severity.} Mean AUC@30 grouped by evaluation domain.}
    \label{fig:geobench_domain_ood}
  \end{minipage}
\end{figure}

\noindent\textbf{Training Data Volume Correlates with Performance, but Data Quality is the a Critical Factor.}
Tab.~\ref{tab:merged_dataset_usage_training_only} summarizes the training datasets used across all evaluated methods, showing considerable variation in both dataset count and domain coverage.
As shown in Fig.~\ref{fig:geobench_training_coverage}, there is a clear positive correlation between the number of training datasets and composite benchmark score within the end-to-end and online paradigms: methods trained on more diverse data sources consistently achieve higher performance.
However, data volume alone does not tell the full story.
Within comparable dataset counts, data quality plays a more decisive role than sheer dataset quantity.
A representative example is DA3, which leverages synthetic datasets to train a teacher model and subsequently refines noisy depth estimates to generate high-quality pseudo-GT supervision. 
This careful curation strategy enables DA3 to achieve the highest composite scores among feed-forward models, despite not relying on the largest training corpus.
\observationbox{
\textbf{Takeaway}: \textit{Training data volume and performance are positively correlated, but data quality is the more decisive factor: carefully curated, high-quality pseudo-GT supervision consistently outperforms larger but noisier training mixtures under comparable dataset scales.}
}

\noindent\textbf{Egocentric-View and Wrist-View Remain the Dominant OOD Failure Modes.}
The most severe generalization gap exposed by \benchmark is not on standard indoor reconstruction datasets, but on embodied-view domains.
As shown in Fig.~\ref{fig:geobench_domain_ood}, the cross-method average remains relatively strong on indoor datasets, while performance drops sharply on ego-view and especially wrist-view sequences.
This failure is not caused by a single weak model: the OOD curve averages over the full evaluated method pool, indicating a field-level limitation.
The training-data analysis explains this behavior.
Current pre-training mixtures heavily cover standard indoor and outdoor reconstruction data, but real robot wrist-view data is systematically absent, and real egocentric coverage is sparse.

To address this gap, \ours is trained on \dataset, which explicitly incorporates egocentric and wrist-view data into the training mixture.
As shown in Tab.~\ref{tab:main_leaderboard_filtered}, \ours achieves consistent and substantial improvements over DA3-Giant across depth and camera pose metrics: depth AbsRel improves by 47\% from $0.095$ to $0.050$ on sparse inputs and 59\% from $0.086$ to $0.035$ on medium inputs, while AUC@30 improves by $+3.1\%$ and $+5.5\%$ respectively.
These gains confirm that targeted in-domain data curation is an effective and practical strategy for closing the OOD gap.

\observationbox{
\textbf{Takeaway}: \textit{The largest remaining data gap is domain diversity, not only data volume: ego-view and wrist-view data are underrepresented in current training mixtures and produce the strongest OOD failures on \benchmark. %
\ours, trained on \dataset, directly addresses this gap.}
}

\noindent\textbf{Is Test-Time Training a Free Lunch?}

\begin{wraptable}{r}{0.58\textwidth}
\vspace{-20pt}
\centering
\caption{\textbf{TTT vs.\ Base model across input length regimes.}
For each (Base model, TTT) pair, we report aggregate camera-pose \textbf{AUC@30} ($\uparrow$) and global trajectory \textbf{ATE} ($\downarrow$), formatted as \emph{base} $\rightarrow$ \emph{TTT}. {\color{ForestGreen}Green} marks improvement, {\color{red}red} marks regression. $^{\dagger}$ Vanilla VGGT OOMs under \emph{Dense}.
}
\label{tab:ttt_length_ablation}
\footnotesize
\setlength{\tabcolsep}{4pt}
\renewcommand{\arraystretch}{1.12}
\begin{tabular}{l l c c}
\toprule
\textbf{Regime} & \textbf{Pair} & \textbf{AUC@30}~$\uparrow$ & \textbf{ATE}~$\downarrow$ \\
\midrule
\multirow{3}{*}{Sparse}
 & CUT3R / TTT3R & 0.519 $\rightarrow$ {\color{red}0.470}        & --- \\
 & VGGT / Scal3R & 0.700 $\rightarrow$ {\color{ForestGreen}0.732} & --- \\
 & Pi3 / LoGeR   & 0.742 $\rightarrow$ {\color{red}0.708}        & --- \\
\midrule
\multirow{3}{*}{Medium}
 & CUT3R / TTT3R & 0.469 $\rightarrow$ {\color{ForestGreen}0.493} & 2.68 $\rightarrow$ {\color{ForestGreen}2.34} \\
 & VGGT / Scal3R & 0.687 $\rightarrow$ {\color{red}0.670}         & 0.73 $\rightarrow$ {\color{ForestGreen}0.40} \\
 & Pi3 / LoGeR   & 0.749 $\rightarrow$ {\color{red}0.714}         & 0.57 $\rightarrow$ {\color{red}0.57} \\
\midrule
\multirow{3}{*}{Dense}
 & CUT3R / TTT3R & 0.165 $\rightarrow$ {\color{ForestGreen}0.321} & 25.5 $\rightarrow$ {\color{ForestGreen}21.1} \\
 & VGGT$^{\dagger}$ / Scal3R & N/A $\rightarrow$ {\color{ForestGreen}0.480}           & N/A $\rightarrow$ \color{ForestGreen}{2.40} \\
 & Pi3 / LoGeR   & 0.524 $\rightarrow$ {\color{ForestGreen}0.598} & 16.4 $\rightarrow$ {\color{ForestGreen}4.60} \\
\bottomrule
\end{tabular}
\vspace{-12pt}
\end{wraptable}

Tab.~\ref{tab:ttt_length_ablation} contrasts three pairs of feedforward 3D foundation models with their test-time training (TTT) descendants: CUT3R~\citep{wang2025cut3r} / TTT3R~\citep{chen2025ttt3r}, VGGT~\citep{wang2025vggt} / Scal3R~\citep{xie2026scal3r}, and Pi3~\citep{wang2025pi3} / LoGeR~\citep{zhang2026loger} across \emph{Sparse}, \emph{Medium}, and \emph{Dense}, summarized by aggregate camera-pose \textbf{AUC@30} ($\uparrow$) and global trajectory \textbf{ATE} ($\downarrow$).
Specifically, the TTT methods are designed to digest sequences longer than the context their base models were trained on.
For example, Scal3R inserts chunk-wise Global Context Memory into VGGT, and LoGeR augments Pi3 with a parametric TTT memory plus sliding-window attention.
Therefore, the natural prior is that TTT should pay off most when sequences are long, and least when they are short and discontinuous.
The \emph{Dense} block confirms this expectation. 
TTT3R nearly doubles AUC@30 ($0.165\!\to\!0.321$, $+95\%$) and cuts ATE by $17.5\%$; LoGeR lifts AUC@30 by $14.1\%$ and reduces ATE by $72\%$ over Pi3; Scal3R is the \textit{only} VGGT-family model that runs at all, since vanilla VGGT OOMs under thousand-frame inputs.
In the \emph{Medium} setting, inputs contain a mix of short and moderately long segments, which limits the benefit of TTT. 
Gains become inconsistent across metrics: AUC@30 improves only for TTT3R ($+5.1\%$) while regressing for Scal3R ($-2.5\%$) and LoGeR ($-4.7\%$); ATE narrows for TTT3R ($-12.7\%$) and Scal3R ($-45.2\%$) but stays flat for LoGeR ($0.57\!\to\!0.57$).
In the \emph{Sparse} setting, inputs consist of only $4$--$15$ frames with large baselines and limited temporal continuity, making it difficult for TTT methods to perform effective test-time updates. 
As a result, the end-to-end base models outperform their TTT counterparts on AUC@30 in two of the three pairs, with only Scal3R holding even ($+4.7\%$).
\observationbox{
\textbf{Takeaway}: \textit{TTT's gains are concentrated on \emph{dense} long sequences, where it consistently improves both pairwise camera pose accuracy and global trajectory consistency over the base models, which confirms that TTT is engineered as a length-generalization mechanism rather than as a universal free lunch.}
}

\noindent\textbf{Can Injecting GT Priors Drive Performance to a Perfect Level?}

\begin{table*}[!tbp]
\centering
\setlength{\fboxsep}{0pt}
\caption{\textbf{Effect of GT Priors across Sparse / Medium.}
We inject ground-truth depth and/or camera (pose + intrinsic) priors for each prior-aware model.
Trajectory and point-cloud metrics are not reported for the sparse regime.
}
\label{tab:prior_ablation_combined}
\resizebox{\textwidth}{!}{%
\renewcommand{\arraystretch}{1.15}
\setlength{\tabcolsep}{3pt}
\begin{tabular}{l cc >{\columncolor{subcol}}c >{\columncolor{subcol}}c >{\columncolor{subcol}}c >{\columncolor{subcol}}c >{\columncolor{subcol}}c >{\columncolor{subcol}}c >{\columncolor{subcol}}c >{\columncolor{subcol}}c >{\columncolor{subcol}}c >{\columncolor{subcol}}c >{\columncolor{subcol}}c >{\columncolor{subcol}}c >{\columncolor{subcol}}c >{\columncolor{subcol}}c >{\columncolor{subcol}}c >{\columncolor{subcol}}c >{\columncolor{subcol}}c >{\columncolor{subcol}}c}
\toprule
\multirow{2.5}{*}{\textbf{Method}}
& \multicolumn{2}{c}{\textbf{Aux. Prior}}
& \multicolumn{6}{c}{\textbf{Depth}}
& \multicolumn{7}{c}{\textbf{Camera}}
& \multicolumn{3}{c}{\textbf{Trajectory}}
& \multicolumn{2}{c}{\textbf{PointCloud}} \\
\cmidrule(lr){2-3} \cmidrule(lr){4-9} \cmidrule(lr){10-16} \cmidrule(lr){17-19} \cmidrule(lr){20-21}
 & Depth & Camera &
 AbsRel$\downarrow$ & SqRel$\downarrow$ & RMSE$\downarrow$ & $\delta_{1.03}\uparrow$ & $\delta_{1.05}\uparrow$ & $\delta_{1.10}\uparrow$ &
 RAcc$_{3}\uparrow$ & RAcc$_{5}\uparrow$ & TAcc$_{3}\uparrow$ & TAcc$_{5}\uparrow$ & AUC@5$\uparrow$ & AUC@15$\uparrow$ & AUC@30$\uparrow$ &
 ATE$\downarrow$ & RPE$_t\downarrow$ & RPE$_r\downarrow$ &
 F-Score$\uparrow$ & Overall$\downarrow$ \\
\midrule
\multicolumn{21}{c}{\cellcolor{catgray}\textbf{Sparse}} \\
\midrule
 \multirow{2}{*}{DA3-Giant} & \xmark & \xmark & 0.095 & 0.107 & 0.608 & 0.563 & 0.689 & 0.821 & 0.791 & 0.870 & 0.587 & 0.682 & 0.525 & 0.699 & 0.785 & -- & -- & -- & -- & -- \\
  & \xmark & \cmark & \textbf{0.078} & \textbf{0.100} & \textbf{0.586} & \textbf{0.635} & \textbf{0.749} & \textbf{0.859} & \cellcolor{secondorange}\textbf{0.968} & \cellcolor{thirdyellow}\textbf{0.981} & \cellcolor{bestred}\textbf{0.965} & \cellcolor{bestred}\textbf{0.989} & \cellcolor{bestred}\textbf{0.918} & \cellcolor{bestred}\textbf{0.969} & \cellcolor{bestred}\textbf{0.984} & -- & -- & -- & -- & -- \\
\midrule
 \multirow{4}{*}{MapAnything} & \xmark & \xmark & 0.153 & 1.079 & 1.337 & 0.361 & 0.512 & 0.701 & 0.608 & 0.762 & 0.300 & 0.423 & 0.244 & 0.446 & 0.579 & -- & -- & -- & -- & -- \\
  & \cmark & \xmark & 0.029 & 0.048 & 0.415 & 0.726 & 0.845 & 0.940 & 0.651 & 0.783 & 0.404 & 0.507 & 0.317 & 0.547 & 0.674 & -- & -- & -- & -- & -- \\
  & \xmark & \cmark & 0.143 & 1.172 & 1.192 & 0.427 & 0.580 & 0.755 & \cellcolor{bestred}\textbf{1.000} & \cellcolor{bestred}\textbf{1.000} & 0.779 & 0.894 & \cellcolor{thirdyellow}0.741 & \cellcolor{thirdyellow}0.883 & \cellcolor{thirdyellow}0.934 & -- & -- & -- & -- & -- \\
  & \cmark & \cmark & \cellcolor{thirdyellow}\textbf{0.020} & \cellcolor{thirdyellow}\textbf{0.040} & \cellcolor{thirdyellow}\textbf{0.368} & \cellcolor{thirdyellow}\textbf{0.861} & \textbf{0.906} & \textbf{0.949} & \cellcolor{bestred}\textbf{1.000} & \cellcolor{bestred}\textbf{1.000} & \cellcolor{secondorange}\textbf{0.956} & \cellcolor{secondorange}\textbf{0.981} & \cellcolor{secondorange}\textbf{0.913} & \cellcolor{secondorange}\textbf{0.966} & \cellcolor{secondorange}\textbf{0.982} & -- & -- & -- & -- & -- \\
\midrule
 \multirow{4}{*}{OmniVGGT} & \xmark & \xmark & 0.117 & 0.119 & 0.669 & 0.534 & 0.671 & 0.810 & 0.620 & 0.746 & 0.416 & 0.507 & 0.332 & 0.537 & 0.665 & -- & -- & -- & -- & -- \\
  & \cmark & \xmark & 0.023 & 0.061 & 0.479 & 0.840 & 0.913 & 0.963 & 0.613 & 0.723 & 0.385 & 0.479 & 0.313 & 0.505 & 0.631 & -- & -- & -- & -- & -- \\
  & \xmark & \cmark & 0.115 & 0.117 & 0.665 & 0.545 & 0.680 & 0.814 & 0.896 & 0.953 & 0.819 & 0.896 & 0.697 & 0.858 & 0.921 & -- & -- & -- & -- & -- \\
  & \cmark & \cmark & \textbf{0.021} & \textbf{0.044} & \textbf{0.427} & \textbf{0.850} & \cellcolor{thirdyellow}\textbf{0.919} & \cellcolor{thirdyellow}\textbf{0.969} & \textbf{0.907} & \textbf{0.959} & \cellcolor{thirdyellow}\textbf{0.833} & \cellcolor{thirdyellow}\textbf{0.917} & \textbf{0.715} & \textbf{0.876} & \textbf{0.934} & -- & -- & -- & -- & -- \\
\midrule
 \multirow{4}{*}{$\pi^{3}$-X} & \xmark & \xmark & 0.084 & 0.084 & 0.599 & 0.576 & 0.710 & 0.833 & 0.756 & 0.837 & 0.491 & 0.595 & 0.427 & 0.627 & 0.741 & -- & -- & -- & -- & -- \\
  & \cmark & \xmark & \cellcolor{secondorange}0.009 & \cellcolor{bestred}\textbf{0.017} & \cellcolor{secondorange}0.255 & \cellcolor{secondorange}0.959 & \cellcolor{secondorange}0.979 & \cellcolor{secondorange}0.992 & 0.787 & 0.860 & 0.540 & 0.645 & 0.476 & 0.667 & 0.769 & -- & -- & -- & -- & -- \\
  & \xmark & \cmark & 0.080 & 0.082 & 0.589 & 0.640 & 0.757 & 0.858 & 0.936 & 0.979 & 0.715 & 0.834 & 0.636 & 0.827 & 0.901 & -- & -- & -- & -- & -- \\
  & \cmark & \cmark & \cellcolor{bestred}\textbf{0.009} & \cellcolor{secondorange}0.017 & \cellcolor{bestred}\textbf{0.255} & \cellcolor{bestred}\textbf{0.960} & \cellcolor{bestred}\textbf{0.980} & \cellcolor{bestred}\textbf{0.992} & \cellcolor{thirdyellow}\textbf{0.940} & \cellcolor{secondorange}\textbf{0.981} & \textbf{0.758} & \textbf{0.848} & \textbf{0.679} & \textbf{0.843} & \textbf{0.908} & -- & -- & -- & -- & -- \\
\midrule
 \multirow{4}{*}{WorldMirror} & \xmark & \xmark & 0.139 & 0.165 & 0.803 & 0.443 & 0.585 & 0.747 & 0.701 & 0.796 & 0.400 & 0.507 & 0.328 & 0.537 & 0.666 & -- & -- & -- & -- & -- \\
  & \cmark & \xmark & 0.081 & 0.243 & 0.933 & 0.510 & 0.643 & 0.789 & 0.695 & 0.808 & 0.377 & 0.507 & 0.329 & 0.544 & 0.672 & -- & -- & -- & -- & -- \\
  & \xmark & \cmark & 0.127 & \textbf{0.158} & \textbf{0.759} & 0.559 & 0.688 & 0.811 & 0.812 & 0.899 & \textbf{0.629} & \textbf{0.762} & \textbf{0.534} & 0.762 & 0.863 & -- & -- & -- & -- & -- \\
  & \cmark & \cmark & \textbf{0.058} & 0.282 & 0.945 & \textbf{0.674} & \textbf{0.772} & \textbf{0.868} & \textbf{0.826} & \textbf{0.911} & 0.620 & 0.753 & 0.533 & \textbf{0.764} & \textbf{0.864} & -- & -- & -- & -- & -- \\
\midrule
\multicolumn{21}{c}{\cellcolor{catgray}\textbf{Medium}} \\
\midrule
 \multirow{2}{*}{DA3-Giant} & \xmark & \xmark & 0.086 & \textbf{0.088} & \textbf{0.578} & 0.572 & 0.686 & 0.812 & 0.750 & 0.834 & 0.587 & 0.663 & 0.532 & 0.684 & 0.776 & 1.161 & 0.284 & 2.275 & 0.742 & 0.073 \\
  & \xmark & \cmark & \textbf{0.078} & 0.111 & 0.579 & \textbf{0.644} & \textbf{0.756} & \textbf{0.862} & \cellcolor{thirdyellow}\textbf{0.984} & \cellcolor{secondorange}\textbf{0.992} & \cellcolor{bestred}\textbf{0.970} & \cellcolor{bestred}\textbf{0.987} & \cellcolor{bestred}\textbf{0.951} & \cellcolor{bestred}\textbf{0.978} & \cellcolor{bestred}\textbf{0.987} & \cellcolor{bestred}\textbf{0.000} & \cellcolor{bestred}\textbf{0.000} & \cellcolor{bestred}\textbf{0.002} & \cellcolor{secondorange}\textbf{0.772} & \textbf{0.062} \\
\midrule
 \multirow{4}{*}{MapAnything} & \xmark & \xmark & 0.146 & 0.490 & 1.052 & 0.347 & 0.491 & 0.681 & 0.563 & 0.702 & 0.312 & 0.419 & 0.254 & 0.451 & 0.579 & 1.737 & 0.533 & 2.852 & 0.420 & 0.114 \\
  & \cmark & \xmark & 0.032 & 0.037 & 0.380 & 0.683 & 0.811 & 0.933 & 0.597 & 0.733 & 0.393 & 0.499 & 0.316 & 0.529 & 0.664 & 2.053 & 0.512 & 2.521 & 0.539 & 0.086 \\
  & \xmark & \cmark & 0.126 & 0.225 & 0.893 & 0.429 & 0.567 & 0.737 & \cellcolor{secondorange}0.998 & \cellcolor{bestred}\textbf{1.000} & 0.786 & 0.881 & \cellcolor{thirdyellow}0.740 & \cellcolor{thirdyellow}0.879 & \cellcolor{thirdyellow}0.930 & \cellcolor{thirdyellow}0.051 & \cellcolor{thirdyellow}0.029 & \cellcolor{thirdyellow}0.270 & 0.521 & 0.108 \\
  & \cmark & \cmark & \cellcolor{thirdyellow}\textbf{0.019} & \cellcolor{thirdyellow}\textbf{0.034} & \cellcolor{thirdyellow}\textbf{0.356} & \cellcolor{thirdyellow}\textbf{0.857} & \textbf{0.910} & \textbf{0.955} & \cellcolor{bestred}\textbf{1.000} & \cellcolor{bestred}\textbf{1.000} & \cellcolor{secondorange}\textbf{0.940} & \cellcolor{secondorange}\textbf{0.969} & \cellcolor{secondorange}\textbf{0.896} & \cellcolor{secondorange}\textbf{0.953} & \cellcolor{secondorange}\textbf{0.972} & \cellcolor{secondorange}\textbf{0.042} & \cellcolor{secondorange}\textbf{0.023} & \cellcolor{secondorange}\textbf{0.216} & \cellcolor{bestred}\textbf{0.773} & \cellcolor{bestred}\textbf{0.056} \\
\midrule
 \multirow{4}{*}{OmniVGGT} & \xmark & \xmark & 0.111 & 0.096 & 0.649 & 0.518 & 0.645 & 0.780 & 0.609 & 0.726 & 0.426 & 0.527 & 0.340 & 0.542 & 0.665 & 1.491 & 0.355 & 2.768 & 0.595 & 0.104 \\
  & \cmark & \xmark & \textbf{0.023} & \textbf{0.056} & \textbf{0.454} & 0.838 & \cellcolor{thirdyellow}\textbf{0.914} & \cellcolor{thirdyellow}\textbf{0.966} & 0.602 & 0.727 & 0.418 & 0.520 & 0.325 & 0.531 & 0.663 & 1.904 & 0.404 & 2.949 & 0.582 & 0.094 \\
  & \xmark & \cmark & 0.108 & 0.106 & 0.668 & 0.521 & 0.650 & 0.786 & 0.870 & \textbf{0.936} & 0.792 & 0.878 & 0.664 & 0.834 & 0.905 & \textbf{0.513} & \textbf{0.111} & 0.589 & 0.673 & 0.073 \\
  & \cmark & \cmark & 0.023 & 0.064 & 0.479 & \textbf{0.839} & 0.910 & 0.961 & \textbf{0.870} & 0.930 & \cellcolor{thirdyellow}\textbf{0.804} & \cellcolor{thirdyellow}\textbf{0.895} & \textbf{0.671} & \textbf{0.843} & \textbf{0.910} & 0.639 & 0.133 & \textbf{0.576} & \textbf{0.693} & \textbf{0.068} \\
\midrule
 \multirow{4}{*}{$\pi^{3}$-X} & \xmark & \xmark & 0.078 & 0.061 & 0.538 & 0.582 & 0.712 & 0.831 & 0.741 & 0.827 & 0.536 & 0.628 & 0.463 & 0.644 & 0.744 & 0.369 & 0.108 & 1.459 & 0.658 & 0.074 \\
  & \cmark & \xmark & \cellcolor{secondorange}0.008 & \cellcolor{bestred}\textbf{0.013} & \cellcolor{secondorange}0.239 & \cellcolor{secondorange}0.969 & \cellcolor{bestred}\textbf{0.984} & \cellcolor{bestred}\textbf{0.993} & 0.761 & 0.844 & 0.547 & 0.638 & 0.478 & 0.659 & 0.763 & 0.478 & 0.129 & 1.277 & 0.667 & 0.074 \\
  & \xmark & \cmark & 0.070 & 0.060 & 0.536 & 0.652 & 0.759 & 0.856 & 0.912 & \cellcolor{thirdyellow}\textbf{0.960} & 0.721 & 0.827 & 0.641 & 0.818 & 0.889 & 0.373 & 0.075 & \textbf{0.453} & 0.703 & 0.067 \\
  & \cmark & \cmark & \cellcolor{bestred}\textbf{0.008} & \cellcolor{secondorange}0.013 & \cellcolor{bestred}\textbf{0.238} & \cellcolor{bestred}\textbf{0.969} & \cellcolor{secondorange}0.984 & \cellcolor{secondorange}0.993 & \textbf{0.914} & 0.958 & \textbf{0.747} & \textbf{0.840} & \textbf{0.677} & \textbf{0.829} & \textbf{0.890} & \textbf{0.112} & \textbf{0.040} & 0.461 & \textbf{0.748} & \cellcolor{secondorange}\textbf{0.057} \\
\midrule
 \multirow{4}{*}{WorldMirror} & \xmark & \xmark & 0.118 & 0.129 & 0.745 & 0.446 & 0.574 & 0.721 & 0.676 & 0.777 & 0.419 & 0.535 & 0.354 & 0.557 & 0.674 & 1.356 & 0.342 & 2.046 & 0.576 & 0.090 \\
  & \cmark & \xmark & 0.082 & 0.132 & 0.726 & 0.512 & 0.638 & 0.782 & 0.694 & 0.797 & 0.420 & 0.540 & 0.354 & 0.564 & 0.685 & 1.797 & 0.479 & 1.974 & 0.643 & 0.073 \\
  & \xmark & \cmark & 0.101 & \textbf{0.117} & 0.700 & 0.572 & 0.682 & 0.790 & 0.766 & 0.863 & 0.621 & 0.744 & 0.528 & 0.740 & 0.838 & \textbf{0.276} & \textbf{0.093} & 0.937 & 0.664 & 0.077 \\
  & \cmark & \cmark & \textbf{0.057} & 0.126 & \textbf{0.688} & \textbf{0.668} & \textbf{0.755} & \textbf{0.851} & \textbf{0.793} & \textbf{0.872} & \textbf{0.633} & \textbf{0.754} & \textbf{0.542} & \textbf{0.752} & \textbf{0.845} & 0.313 & 0.107 & \textbf{0.894} & \cellcolor{thirdyellow}\textbf{0.755} & \cellcolor{thirdyellow}\textbf{0.058} \\
\bottomrule
\end{tabular}%
}
\end{table*}

Tab.~\ref{tab:prior_ablation_combined} presents an ablation study on the effect of injecting ground-truth depth and camera pose priors for five prior-aware models (DA3-Giant, MapAnything, OmniVGGT, $\pi^3$-X, and WorldMirror) across sparse and medium input settings, evaluated on depth, camera, trajectory, and point cloud metrics.
All prior-aware models benefit from depth prior injection to varying degrees, with depth metrics approaching near-GT-level accuracy across the board.
However, the effect of camera pose priors is more nuanced.
DA3-Giant and MapAnything exhibit strong prior adherence: injecting GT camera poses leads to highly consistent pose predictions, with AUC@15 maintained above 90\% under all settings.
In contrast, OmniVGGT, $\pi^{3}$-X, and WorldMirror show moderate improvements but fail to fully conform to the injected camera poses under challenging conditions (see Fig.~\ref{Fig.bad_case} in Appendix.~\ref{prior_vis}), partially overriding them with their own predictions.
\observationbox{
\textbf{Takeaway}: \textit{Injecting GT depth priors consistently drives depth estimation to near-perfect accuracy, yet camera pose priors yield inconsistent gains. Some models partially override the injected poses with their own predictions and fail under challenging conditions.}
}

\begin{figure*}[!tbp]
\centering
\includegraphics[width=\textwidth]{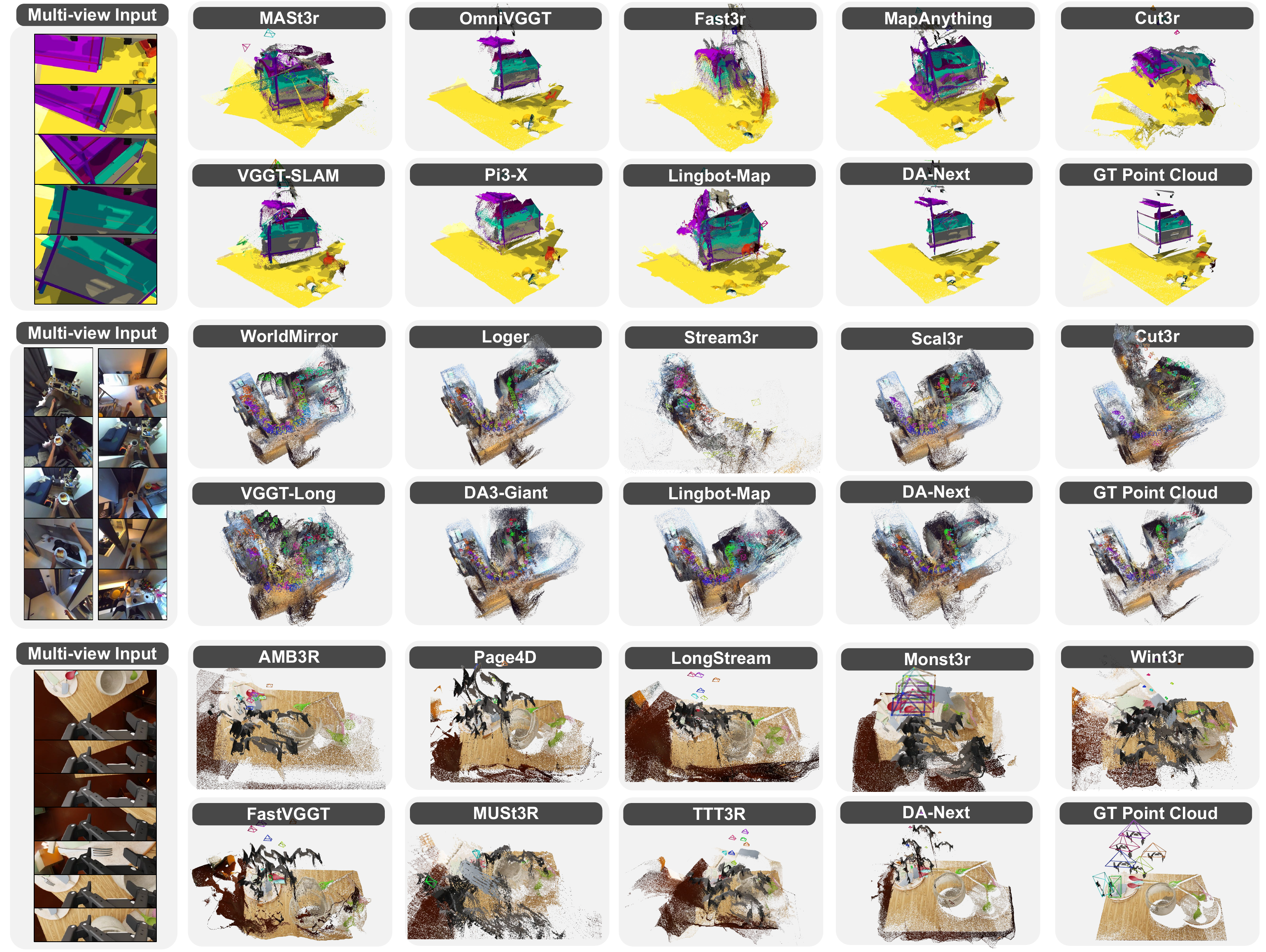}
\caption{\textbf{Qualitative comparison of multi-view 3D reconstruction on \benchmark.}}
\label{Fig.main_vis}
\end{figure*}

Fig.~\ref{Fig.main_vis} presents a qualitative comparison against representative baselines, where \ours yields a more accurate camera trajectory and sharper geometry under challenging viewpoints.
Due to space limitations, we provide additional visualizations in Appendix~\ref{prior_vis} and findings in Appendix~\ref{appendix:other_findings}.

\section{Conclusion}
We presented \benchmark, a comprehensive, reproducible, and cross-paradigm benchmark for evaluating spatial foundation models across diverse domains, input densities, and reconstruction task suites.
Through extensive experiments on 41 models across 6 paradigms, \benchmark reveals that current spatial foundation models are not yet all-round players, exposing critical gaps in domain generalization, input-density robustness.
To address the most significant data gap identified, we introduced \dataset, a large-scale egocentric and wrist-view dataset, and trained \ours as a strong baseline.
We hope \benchmark serves as a rigorous foundation for future research toward more generalizable and robust 3D foundation models.

\newpage
{
\small
\bibliography{custom}
}

\newpage
\appendix
  \startcontents[appendix]
  \begingroup
  \small
  \linespread{0.94}\selectfont
  \printcontents[appendix]{l}{1}{%
    \section*{Appendix Contents}%
    \vspace{-0.75em}%
  }
  \endgroup
  \newpage

\section{\benchmark Data Curation Pipeline}
\label{appendix:pipeline}
\begin{figure}[t] 
\centering 
\includegraphics[width=0.9\columnwidth]{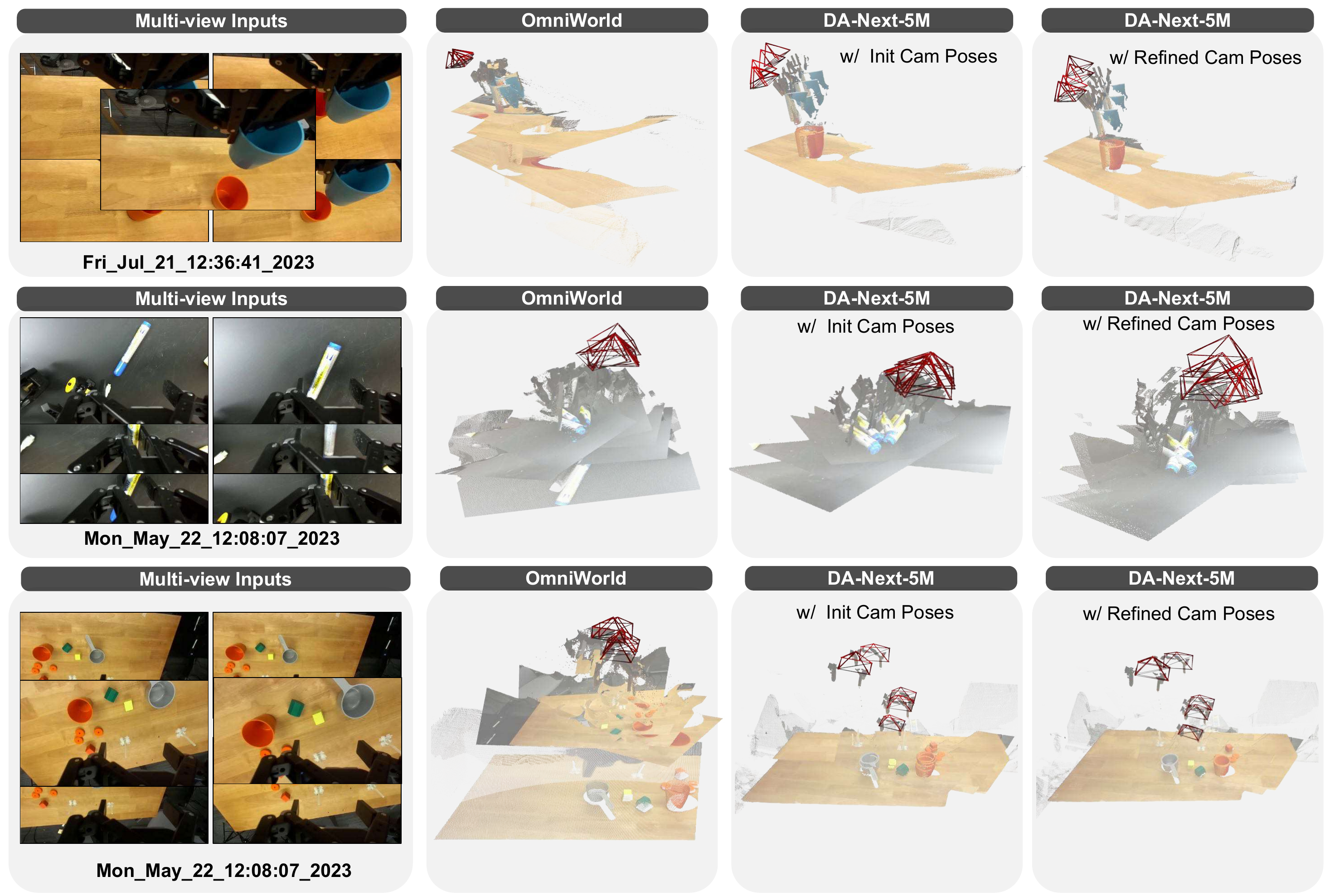}
\caption{\textbf{Visual comparison on the DROID dataset.} We compare the annotations from OmniWorld against our reconstructed point clouds using initial poses and refined poses, respectively.} 
\vspace{-4pt}
\label{Fig.droid_vis} %
\end{figure}

In this section, we detail the data processing pipeline in \benchmark, covering the post-processing pipelines for Xperience (Sec.~\ref{appendix:ropedia}), DROID (Sec.~\ref{appendix:droid}), and our collected simulation datasets (Sec.~\ref{appendix:other_collection}), as well as a unified depth map post-processing pipeline (Sec.~\ref{appendix:DepthPost-Processing}) applied to selected datasets to ensure annotation quality.

\subsection{DROID Curation Pipeline}
\label{appendix:droid}
Accurate geometric annotations are critical for reliable benchmark evaluation.
However, the raw DROID~\citep{khazatsky2024droid} data is highly noisy with 
unreliable camera calibration, precluding direct use of the original annotations.
We first investigate learning-based annotation solutions, including Mega-SaM~\citep{li2025megasam} and VIPE~\citep{huang2025vipe} for visual SLAM, and Camera Depth Model~\citep{liu2025cdm}, PriorDA~\citep{wang2025priorda}, Lingbot-Depth~\citep{lingbot-depth2026}, and PromptDA~\citep{lin2024promptda} for depth completion. However, none of these methods are trained on wrist-view domains, and we observe significant failure cases under the close-range, high-motion conditions characteristic of this setting.

Since DROID provides synchronized stereo pairs, we turn to stereo-based depth estimation as a more reliable annotation source. 
We test S$^2$M$^2$~\citep{min2025s2m2}, FoundationStereo~\citep{wen2025foundationstereo}, IGEV~\citep{xu2025igev++}, DynamicStereo~\citep{Karaev_2023_dynamic_stereo}, and WAFT-Stereo~\citep{wang2026waftstereo}, ultimately selecting S$^2$M$^2$ for its superior performance on wrist-view sequences.
Specifically, we randomly select a subset of DROID sequences and feed the left-right stereo image pairs from each wrist-view sequence into S$^2$M$^2$ to obtain disparity maps. 
Metric depth maps for the left eye are then computed using the ZED camera intrinsics and baseline, with a confidence threshold of 0.999 applied to mask out unreliable depth predictions.
The masked depth maps and camera intrinsics are subsequently injected as priors into MapAnything~\citep{keetha2026mapanything} to obtain initial camera poses.
We then employ SAM3~\citep{carion2025sam3} to annotate masks for moving objects on selected keyframes, which are propagated to the full sequence.
Finally, leveraging the initial camera poses, RGB images, and propagated masks, we perform Bundle Adjustment with both depth and photometric losses to refine the camera parameters and obtain globally aligned point clouds.
Sequences with poor temporal depth consistency are filtered out by computing the mean per-pixel reprojection depth error between adjacent frames.
Fig.~\ref{Fig.droid_vis} presents a comparison between our DROID data curation pipeline 
and the OmniWorld data processing approach.
OmniWorld annotates wrist-view depth using FoundationStereo and relies on the dataset's original inaccurate camera poses, resulting in unreliable depth estimates and misaligned point clouds across frames (\emph{e.g.}, incorrect depth on the marker pen on the table in \texttt{Mon\_May\_22\_12:08:07\_2023}).
In contrast, our annotation pipeline produces temporally consistent depth estimates with well-preserved cross-frame continuity.

\begin{figure}[t] 
\centering 
\includegraphics[width=0.9\columnwidth]{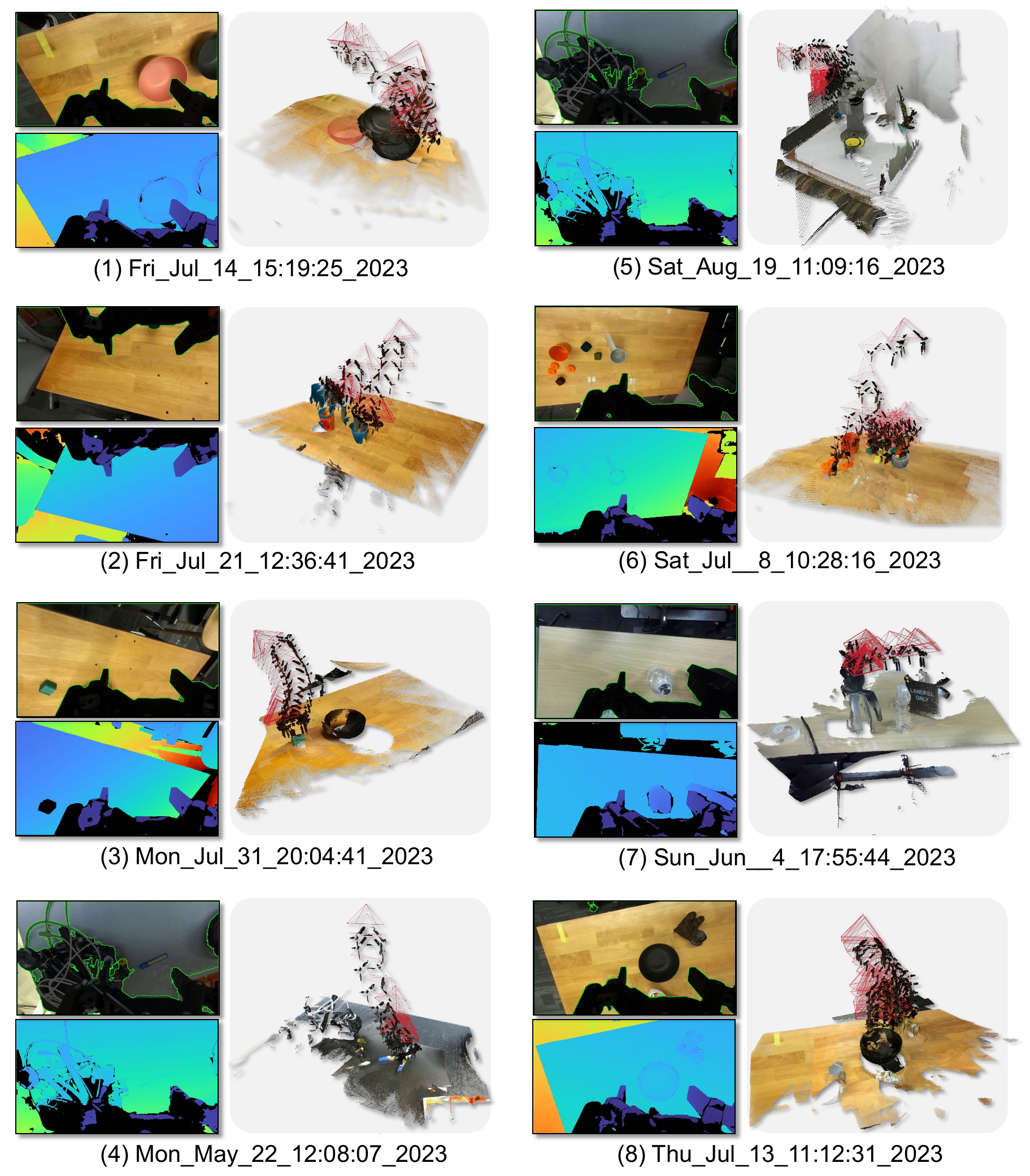}
\caption{\textbf{DROID Gallery 1.}} 
\vspace{-4pt}
\label{Fig.droid_gallery1} %
\end{figure}

\begin{figure}[t] 
\centering 
\includegraphics[width=0.9\columnwidth]{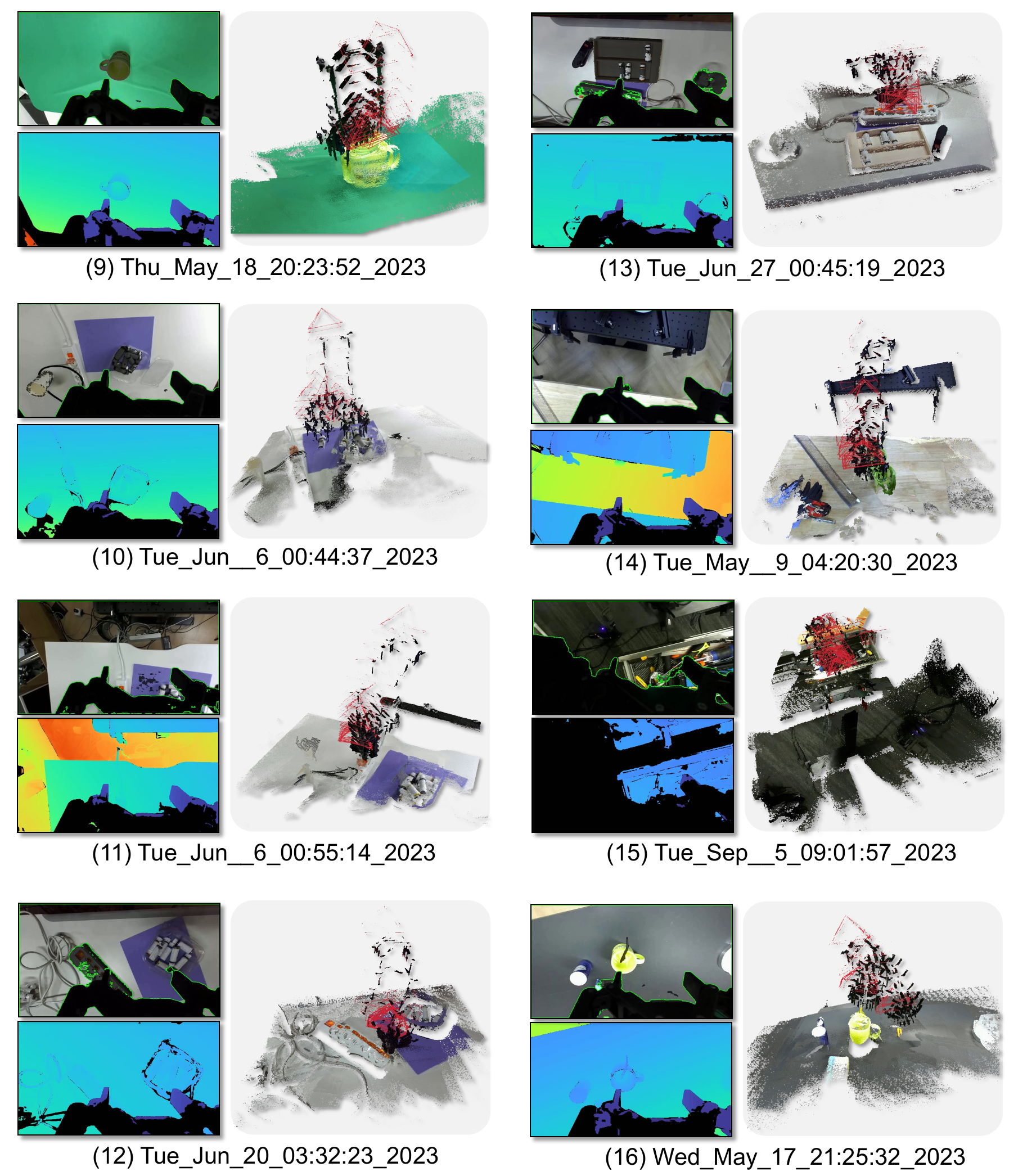}
\caption{\textbf{DROID Gallery 2.} } 
\vspace{-4pt}
\label{Fig.droid_gallery2} %
\end{figure}

We also provide a gallery of the DROID~\citep{khazatsky2024droid} dataset we processed in Fig.~\ref{Fig.droid_gallery1} and \ref{Fig.droid_gallery2}.
For each scene, we visualize the first frame of the RGB sequence overlaid with the SAM3 segmentation mask (top-left), the corresponding depth map (bottom-left), and the reconstructed point cloud (right).

\subsection{Xperience Curation Pipeline}
\label{appendix:ropedia}
\begin{figure*}
    \centering
    \includegraphics[width=1\linewidth]{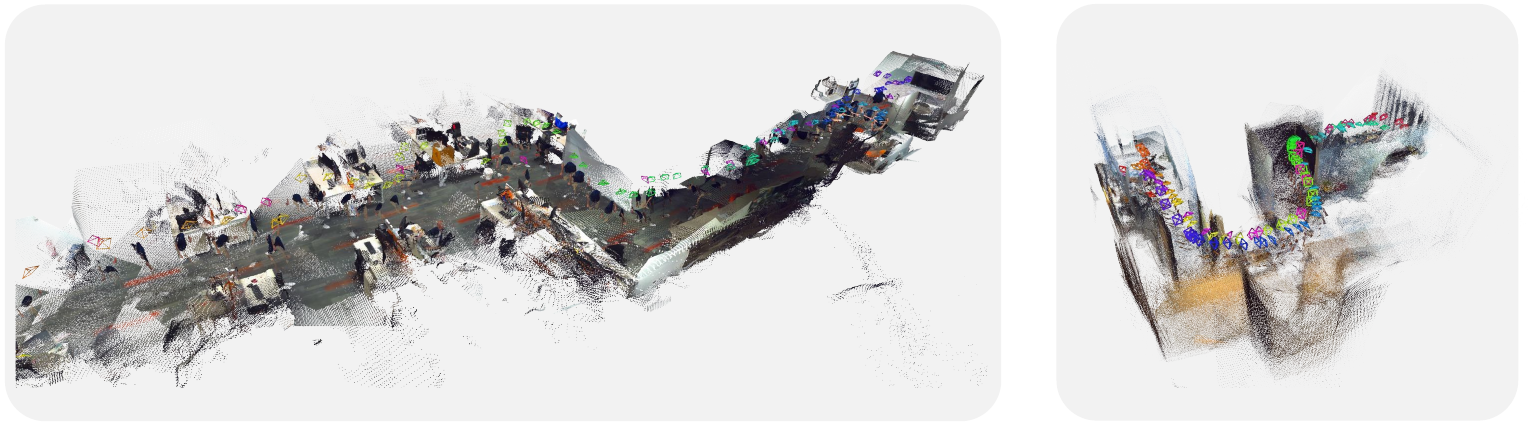}
    \caption{\textbf{Visualization of Xperience Samples.} 
}
    \label{fig:ropedia_vis}
    \vspace{-4pt}
\end{figure*}

Xperience~\citep{xperience_10m} consists of egocentric sequences captured via a head-mounted device during human activity, recorded by four fisheye cameras.
We randomly sample a subset of the original data for processing.
We obtain camera poses using a SLAM system built upon VIPE~\citep{huang2025vipe}, and estimate metric depth maps and their confidence maps from rectified stereo image pairs using FoundationStereo~\citep{wen2025foundationstereo}.
Fig.~\ref{fig:ropedia_vis} presents sample visualizations of Xperience sequences included 
in \benchmark.

\subsection{Other Data Collection}
\label{appendix:other_collection}
We collect robot manipulation sequences from four simulators: RLBench~\citep{james2020rlbench}, Robo Colosseum~\citep{pumacay2024colosseum}, RoboTwin~\citep{chen2025robotwin}, and RoboLab~\citep{yang2026robolab}.
We refer the reader to Sec.~\ref{appendix:datasets} for details on the data collection procedure.

\subsection{Depth Map Post-Processing Pipeline}
\label{appendix:DepthPost-Processing}
Raw depth maps collected from sensors or simulation engines often contain systematic artifacts that degrade geometric evaluation quality. 
We apply a unified five-stage cleaning pipeline to all real-world datasets in \benchmark, producing a per-frame binary validity mask that downstream evaluation uses to ignore unreliable pixels.

\textbf{Stage 1: Depth Range Clipping.}
Depth values are first converted to meters and then clipped to a scene-appropriate valid range $[d_{\min}, d_{\max}]$. 
Pixels falling outside this range, including near-field sensor noise and far-field sensor dropout are marked invalid.
The range bounds are set per dataset according to the typical operating distance of the capture platform (\emph{e.g.}, $[0.05\text{m},\,5\text{m}]$ for egocentric indoor scenes and $[0\text{m},\,3\text{m}]$ for close-range wrist-view sequences).

\textbf{Stage 2: Flying Point Removal.}
Flying points are erroneous depth values that appear at depth discontinuity boundaries, caused by stereo mismatches or sensor mixed-pixel effects.
We detect them by computing the spatial gradient of the depth map and identifying pixels where the gradient magnitude is disproportionately large relative to the local depth value, which indicates a sharp, physically implausible depth jump.
Detected discontinuity pixels and a small surrounding neighborhood (controlled by an erosion radius) are masked out, preventing boundary artifacts from contaminating geometric metrics.

\textbf{Stage 3: Edge-Aware Bilateral Filtering.}
Surviving valid pixels are smoothed using a joint bilateral filter guided by the corresponding RGB image.
The filter suppresses high-frequency depth noise while preserving genuine depth edges aligned with visible object boundaries.
Only already-valid pixels are updated; no hole-filling is performed, ensuring that masked regions remain invalid after filtering.

\textbf{Stage 4: Small Isolated Region Removal.}
After filtering, the valid-pixel mask may contain small, disconnected clusters of surviving pixels that are physically implausible as independent surface patches.
We apply connected-component analysis on the binary valid mask and remove any component whose area falls below a minimum threshold, eliminating residual noise fragments that escaped the earlier stages.

\textbf{Stage 5: Sky Mask.}
For outdoor datasets (\emph{e.g.}, OmniWorld~\citep{zhou2025omniworld}, Lingbot-Depth~\citep{lingbot-depth2026}), depth sensors and stereo algorithms systematically fail on sky regions, producing either missing values or extreme outliers at near-infinite range. 
We generate a per-frame sky mask using a pretrained semantic segmentation model (SegFormer~\citep{xie2021segformer} fine-tuned on ADE20K~\citep{zhou2019ade20k}) and set all sky-classified pixels to invalid regardless of their depth values.
This prevents sky regions from inflating depth error metrics or introducing spurious points at the horizon in point cloud evaluation.

\section{Benchmark Details}
\label{appendix:benchmarkdetails}

In this section, we provide a comprehensive description of \benchmark.
We first detail the composition of our dataset collection (Sec.~\ref{app:regime-details}), covering the 19 source datasets and the three multi-density evaluation regimes (\textit{Sparse}, \textit{Medium}, and \textit{Dense}).
We then describe the hardware configuration and general experimental settings (Sec.~\ref{appendix:metrics}), followed by the complete mathematical definitions of all evaluation metrics used to assess camera pose estimation and depth prediction quality.
Next, we report the implementation details and hyperparameters for each evaluated method, organized by category: optimization-based, feed-forward, streaming, chunk-based, SLAM-based, and test-time training approaches.
Finally, we present qualitative visualizations of prior-enhanced models and summarize the training datasets used by all compared methods.

\subsection{Benchmark Composition}
\label{appendix:full_benchmark}

Our benchmark aggregates \textbf{19} publicly available source datasets, spanning indoor and outdoor scenes, static reconstructions and dynamic sequences, and three viewpoint regimes (third-person, egocentric, and wrist-mounted), captured from both real sensors and simulation.
A complete summary of these datasets, together with their per-scene attribute tags, the supported input regimes (Single Frame, Sparse, Medium, Dense), and the number of evaluated scenes and frames, is provided in Table~\ref{tab:datasets_summary}.
In total, the benchmark contains \textbf{546} scenes and
\textbf{72{,}540} evaluation frames.

%
%
%
\begin{table*}[t]
\centering
\caption{\textbf{Source Datasets and Scene Attributes Used in Our Benchmark.}
We summarize the datasets together with their per-dataset attribute tags.
The four \textit{\#Scenes} sub-columns report the number of scenes evaluated
under each input regime (\textit{Single Frame}, \textit{Sparse}, \textit{Medium},
\textit{Dense}); ``--'' marks regimes in which a dataset does not participate.
\textit{\#Frames} is the total frame count actually consumed across all regimes.
}
\label{tab:datasets_summary}
\resizebox{\textwidth}{!}{%
\renewcommand{\arraystretch}{1}
\setlength{\tabcolsep}{6pt}
\begin{tabular}{c l >{\columncolor{subcol}}c c >{\columncolor{subcol}}c c >{\columncolor{subcol}}c c >{\columncolor{subcol}}c c >{\columncolor{subcol}}r}
\toprule
\multirow{2.5}{*}{\textbf{\#}}
& \multirow{2.5}{*}{\textbf{Dataset}}
& \multirow{2.5}{*}{\textbf{Environment}}
& \multirow{2.5}{*}{\textbf{Dynamics}}
& \multirow{2.5}{*}{\textbf{View Type}}
& \multirow{2.5}{*}{\textbf{Source}}
& \multicolumn{4}{c}{\textbf{\#Scenes}}
& \multirow{2.5}{*}{\textbf{\#Frames}} \\
\cmidrule(lr){7-10}
 & & & & & & \textbf{Single} & \textbf{Sparse} & \textbf{Medium} & \textbf{Dense} & \\
\midrule
1  & 7-Scenes~\citep{shotton20137scenes}          & Indoor  & Static  & Normal & Real       & 7  & 7  & 7  & 7  & 7{,}242  \\
2  & ADT~\citep{pan2023aria}                      & Indoor  & Dynamic & Egocentric    & Simulation & 4  & 4  & 4  & 4  & 2{,}112  \\
3  & DROID~\citep{khazatsky2024droid}             & Indoor  & Dynamic & Wrist  & Real       & 16 & 16 & 16 & 16 & 3{,}440  \\
4  & DTU~\citep{jensen2014dtu}                     & Indoor  & Static  & Normal & Real       & 13 & 13 & 13 & -- & 431      \\
5  & ETH3D~\citep{schops2017eth3d}                 & Both    & Static  & Normal & Real       & 8  & 8  & 8  & -- & 150      \\
6  & Hiroom~\citep{lin2025depth}                  & Indoor  & Static  & Normal & Simulation & 6  & 6  & 6  & -- & 90       \\
7  & KITTI-Odometry~\citep{geiger2012kittiod}     & Outdoor & Dynamic & Normal & Real       & -- & -- & -- & 11 & 12{,}973 \\
8  & LingBot-Depth~\citep{lingbot-depth2026}      & Both    & Static  & Mixed  & Mixed      & 50 & -- & -- & -- & 50       \\
9  & NRGBD~\citep{azinovic2022Neuralrgbd}                 & Indoor  & Static  & Normal & Real       & 8  & 8  & 8  & 8  & 10{,}945 \\
10 & OmniWorld~\citep{zhou2025omniworld}          & Outdoor & Dynamic & Normal & Simulation & 5  & 5  & 5  & 4  & 2{,}033  \\
11 & RLBench~\citep{james2020rlbench}             & Indoor  & Dynamic & Wrist  & Simulation & 10 & 10 & 10 & 8  & 1{,}887  \\
12 & Robolab~\citep{yang2026robolab}              & Indoor  & Dynamic & Wrist  & Simulation & 8  & 8  & 8  & 8  & 9{,}360  \\
13 & RoboTwin~\citep{chen2025robotwin}            & Indoor  & Dynamic & Wrist  & Simulation & 8  & 8  & 8  & 7  & 1{,}695  \\
14 & ScanNet++~\citep{yeshwanth2023scannet++}     & Indoor  & Static  & Normal & Real       & 11 & 11 & 11 & 11 & 4{,}137  \\
15 & Tanks and Temples~\citep{Knapitsch2017tanks} & Both    & Static  & Normal & Real       & 4  & 4  & 4  & 4  & 934      \\
16 & TUM~\citep{sturm12tumdynamic}                & Indoor  & Dynamic & Normal & Real       & 6  & 6  & 6  & 6  & 6{,}657  \\
17 & V-KITTI~\citep{cabon2020vkitti}              & Outdoor & Dynamic & Normal & Simulation & 5  & 5  & 5  & 5  & 2{,}851  \\
18 & Waymo~\citep{sun2020waymo}                   & Outdoor & Dynamic & Normal & Real       & 8  & 8  & 8  & 8  & 2{,}464  \\
19 & Xperience~\citep{xperience_10m}              & Indoor  & Dynamic & Egocentric    & Real       & 2  & 2  & 2  & 2  & 3{,}089  \\
\midrule
\multicolumn{6}{r}{\textbf{Total}} & \textbf{179} & \textbf{129} & \textbf{129} & \textbf{109} & \textbf{72{,}540} \\
\bottomrule
\end{tabular}%
}
\end{table*}

\subsection{Multi-density Evaluation Regime Details}
\label{app:regime-details}

\textbf{\textsc{Sparse regime.}}
Sparse-view selection is implemented as a deterministic set-cover procedure.
Let $\mathcal{V}$ denote the voxel support of a scene and $\mathcal{F}$ the set of candidate frames.
Each frame $f \in \mathcal{F}$ covers a subset of voxels $V_f \subseteq \mathcal{V}$.
Given a frame budget $K$, the selected frame set is defined as
\[
\mathcal{S}
=
\arg\max_{S \subseteq \mathcal{F},\, |S| \le K}
\left|
\bigcup_{f \in S} V_f
\right|.
\]
In practice, we optimize this objective greedily by repeatedly selecting the frame that provides the largest marginal increase in voxel coverage.

\textbf{\textsc{Medium regime.}}
The medium regime uses the same coverage objective but applies a length-adaptive frame budget.
For a scene with $N$ frames, the selected set satisfies
\[
\mathcal{S}
=
\arg\max_{S \subseteq \mathcal{F},\, F_{\min}(N) \le |S| \le F_{\max}(N)}
\left|
\bigcup_{f \in S} V_f
\right|,
\]
where $F_{\min}(N)$ and $F_{\max}(N)$ define the admissible range of selected frames.
This prevents over-pruning short sequences and over-sampling long ones while preserving moderate view overlap.

\textbf{\textsc{Dense regime.}}
For dense evaluation, the goal is to preserve temporal continuity while bounding evaluation cost.
Given a scene with $N$ frames and a target maximum budget of $T=500$ frames, we retain all frames when $N \le T$.
Otherwise, we subsample the trajectory with a stride
\[
s = \left\lceil \frac{N}{T} \right\rceil,
\]
which yields approximately $T$ evenly spaced frames per scene.
Datasets with too few valid frames or too few eligible trajectories after filtering are not included in the \textsc{Dense} regime.
Accordingly, missing dense entries in the per-dataset results tables indicate unavailable evaluation indices rather than failed method runs.

\subsection{General Setting of the Benchmark}
All experiments are run on a single workstation with the following configuration: 2$\times$ Intel Xeon Platinum 8580 processors (60 cores / 120 threads per socket, 240 logical cores in total), 2 TB of system memory, and 8$\times$ NVIDIA H200 GPUs, each with 141 GB of memory and pairwise connected through 18-link NVLink (NV18). 
The system runs Ubuntu 22.04 with CUDA 12.8 and NVIDIA driver 570.148.08.
For all methods, we prioritize using the resolution recommended by each method for inference.
Specifically, for methods that only support a fixed resolution (\emph{e.g.}, Spann3R at $224 \times 224$), we resize the image such that the width matches the target resolution while preserving the original aspect ratio, followed by a center crop. For all other methods, inference is performed at a resolution of $512$ or $518$ pixels, depending on each model's patch stride ($16$ or $14$, respectively).

\subsection{Evaluation Metrics}
\label{appendix:metrics}

We provide the complete mathematical definitions of all evaluation metrics used in \benchmark.
Throughout, $N$ denotes the number of input frames in a scene, indexed by $i\in\{1,\ldots,N\}$.
Each camera pose $G_i\in SE(3)$ is a \emph{world-to-camera} transformation, decomposed as $G_i=[R_i\mid t_i]$ with rotation $R_i\in SO(3)$ and translation $t_i\in\mathbb{R}^3$, so that the camera centre in world coordinates is $\mathbf{c}_i=-R_i^{\top}t_i$.
Ground-truth quantities carry a superscript star (e.g.\ $R_i^*$, $t_i^*$, $\mathbf{c}_i^*=-{R_i^*}^{\!\top}t_i^*$) and predicted quantities carry a hat (e.g.\ $\hat{R}_i$, $\hat{t}_i$, $\hat{\mathbf{c}}_i=-\hat{R}_i^{\top}\hat{t}_i$).
$D_p$ and $\hat{D}_p$ denote the ground-truth and predicted depth at pixel $p$, and $\mathbf{P}_p$, $\hat{\mathbf{P}}_p$ the corresponding 3D world points.

\subsubsection{Camera Pose Estimation}
\label{appendix:metrics:pose}

Camera pose estimation is evaluated via pairwise relative geometry across all $|\mathcal{P}|$ ordered pairs $\mathcal{P}=\{(i,j)\mid 1\!\le\!i\!<\!j\!\le\!N\}$.

\textbf{Relative rotation error.} The relative rotation from camera $i$ to camera $j$ is
\begin{equation}
  R_{ij}^{*} = R_j^{*}\,{R_i^{*}}^{\!\top},
  \qquad
  \hat{R}_{ij} = \hat{R}_j\,\hat{R}_i^{\top}.
\end{equation}
The pairwise rotation error is the geodesic distance on $SO(3)$:
\begin{equation}
  e^{R}_{ij}
  = \arccos\!\left(\frac{\operatorname{tr}\!\bigl({R_{ij}^{*}}^{\!\top}\hat{R}_{ij}\bigr)-1}{2}\right)
  \;\in\; [0^{\circ},\,180^{\circ}].
  \label{eq:rre}
\end{equation}
The \emph{Relative Rotation Accuracy} (\textit{RAcc}) at threshold $x$ is the fraction of pairs whose rotation error falls below $x$:
\begin{equation}
  \mathrm{RAcc}_{x}
  = \frac{1}{|\mathcal{P}|}\sum_{(i,j)\in\mathcal{P}}
    \mathbf{1}\!\left[e^{R}_{ij} < x\right].
  \label{eq:racc}
\end{equation}

\textbf{Relative translation error.}
Since the predicted trajectory may carry an arbitrary global scale, translation quality is measured by the \emph{angular deviation} between the translation components of the predicted and ground-truth pairwise relative poses, with the $180^{\circ}$ direction ambiguity folded out.
Let
\begin{equation}
  t_{ij}^{*}\;=\;\bigl[(G_i^{*})^{-1}G_j^{*}\bigr]_{\mathrm{trans}}\;=\;R_i^{*\top}\bigl(t_j^{*}-t_i^{*}\bigr),
  \qquad
  \hat{t}_{ij}\;=\;\bigl[\hat{G}_i^{-1}\hat{G}_j\bigr]_{\mathrm{trans}}\;=\;\hat{R}_i^{\top}\bigl(\hat{t}_j-\hat{t}_i\bigr)
\end{equation}
denote the translation components of the relative SE(3) poses, expressed in camera $i$'s local frame, with corresponding unit directions
\begin{equation}
  \tau_{ij}^{*} = \frac{t_{ij}^{*}}{\|t_{ij}^{*}\|_2},
  \qquad
  \hat{\tau}_{ij} = \frac{\hat{t}_{ij}}{\|\hat{t}_{ij}\|_2}.
\end{equation}
The pairwise translation error is
\begin{equation}
  e^{t}_{ij}
  = \arccos\!\left(\bigl|\tau_{ij}^{*}\cdot\hat{\tau}_{ij}\bigr|\right)
  \;\in\; [0^{\circ},\,90^{\circ}],
  \label{eq:rte}
\end{equation}
where the absolute value absorbs the inherent $180^{\circ}$ direction ambiguity under unknown global scale.
The \emph{Relative Translation Accuracy} (\textit{TAcc}) at threshold $x$ is
\begin{equation}
  \mathrm{TAcc}_{x}
  = \frac{1}{|\mathcal{P}|}\sum_{(i,j)\in\mathcal{P}}
    \mathbf{1}\!\left[e^{t}_{ij} < x\right].
  \label{eq:tacc}
\end{equation}

\textbf{AUC.}
The joint accuracy curve measures the fraction of pairs for which \emph{both} rotation and translation errors are below threshold $x$:
\begin{equation}
  \mathrm{Acc}_{x}
  = \frac{1}{|\mathcal{P}|}\sum_{(i,j)\in\mathcal{P}}
    \mathbf{1}\!\left[\max\!\left(e^{R}_{ij},\,e^{t}_{ij}\right) < x\right].
  \label{eq:acc_joint}
\end{equation}
The \emph{Area Under the Curve} up to a maximum threshold $x_{\max}$ is
\begin{equation}
  \mathrm{AUC}_{x_{\max}}
  = \frac{1}{x_{\max}}\int_{0}^{x_{\max}}\mathrm{Acc}_{x}\,\mathrm{d}x,
  \label{eq:auc}
\end{equation}
approximated in practice by linear interpolation over a uniform grid of thresholds $x\in[0^{\circ},x_{\max}]$.

\subsubsection{Camera Trajectory Estimation}
\label{appendix:metrics:traj}

For continuous image sequences (medium and dense regimes), the predicted trajectory $\{\hat{R}_i,\hat{\mathbf{c}}_i\}_{i=1}^{N}$ is first aligned to the ground truth $\{R_i^{*},\mathbf{c}_i^{*}\}_{i=1}^{N}$ via a global $\mathrm{Sim}(3)$ transformation.
Specifically, we solve for scale $s^{*}\!>0$, rotation $\bar{R}^{*}\!\in SO(3)$, and translation $\bar{\mathbf{t}}^{*}\!\in\mathbb{R}^{3}$ by minimising
\begin{equation}
  \min_{s,\,\bar{R},\,\bar{\mathbf{t}}}
  \sum_{i=1}^{N}
  \bigl\|s\,\bar{R}\,\hat{\mathbf{c}}_i + \bar{\mathbf{t}} - \mathbf{c}_i^{*}\bigr\|_2^2,
  \label{eq:sim3_align}
\end{equation}
yielding scale-aligned camera centres
$\tilde{\mathbf{c}}_i = s^{*}\bar{R}^{*}\hat{\mathbf{c}}_i + \bar{\mathbf{t}}^{*}$ and correspondingly aligned rotations $\tilde{R}_i = \bar{R}^{*}\hat{R}_i$. Let $\tilde{T}_i\in SE(3)$ denote the resulting aligned camera pose.

\textbf{Absolute Trajectory Error (ATE).}
\begin{equation}
  \mathrm{ATE}
  = \sqrt{\frac{1}{N}\sum_{i=1}^{N}
    \bigl\|\tilde{\mathbf{c}}_i - \mathbf{c}_i^{*}\bigr\|_2^2}\,.
  \label{eq:ate}
\end{equation}

\textbf{Relative Pose Error.}
For consecutive frame pairs with temporal displacement $\Delta=1$, the ground-truth and aligned-predicted relative SE(3) motions are
\begin{align}
  \delta T_i^{*}    &= (G_i^{*})^{-1}\,G_{i+1}^{*},
  &\delta\tilde{T}_i &= \tilde{T}_i^{-1}\,\tilde{T}_{i+1},\\
  \delta R_i^{*}    &= R_{i+1}^{*}\,{R_i^{*}}^{\!\top},
  &\delta\tilde{R}_i &= \tilde{R}_{i+1}\,\tilde{R}_i^{\top},
\end{align}
and the per-step pose-error matrix is $E_i = (\delta T_i^{*})^{-1}\,\delta\tilde{T}_i \in SE(3)$.
The \emph{Relative Translation Error} and \emph{Relative Rotation Error}, each averaged over all $N\!-\!1$ consecutive windows, are
\begin{align}
  \mathrm{RPE}_{t}
  &= \frac{1}{N-1}\sum_{i=1}^{N-1}
    \bigl\|\,[E_i]_{\mathrm{trans}}\,\bigr\|_2,
  \label{eq:rpet}\\
  \mathrm{RPE}_{r}
  &= \frac{1}{N-1}\sum_{i=1}^{N-1}
    \arccos\!\left(\frac{\operatorname{tr}\!\bigl({\delta R_i^{*}}^{\!\top}\delta\tilde{R}_i\bigr)-1}{2}\right),
  \label{eq:rper}
\end{align}
where $[E_i]_{\mathrm{trans}}$ denotes the translation component of $E_i$. This follows the standard \texttt{evo}-library RPE definition adopted by the DROID-SLAM evaluation protocol.
ATE and $\mathrm{RPE}_{t}$ are reported in metres; $\mathrm{RPE}_{r}$ is in degrees.

\subsubsection{Depth Estimation}
\label{appendix:metrics:depth}
Let $\Omega_D = \{p \mid D_p > 0\}$ denote the set of pixels with valid ground-truth depth in a given frame.
For models that do not produce metric-scale output, the predicted depth $\hat{D}_p$ is first rescaled by the per-frame \emph{median alignment}:
\begin{equation}
  \hat{D}_p \;\leftarrow\; s\cdot\hat{D}_p,
  \qquad
  s = \operatorname*{median}_{p\in\Omega_D}\!\left(\frac{D_p}{\hat{D}_p}\right).
  \label{eq:median_align}
\end{equation}
All depth metrics reported in this paper are computed after median-scale alignment, unless otherwise specified.

\textbf{Absolute Relative Error (AbsRel).}
\begin{equation}
  \mathrm{AbsRel}
  = \frac{1}{|\Omega_D|}\sum_{p\in\Omega_D}
    \frac{\left|D_p - \hat{D}_p\right|}{D_p}.
  \label{eq:absrel}
\end{equation}

\textbf{Squared Relative Error (SqRel).}
\begin{equation}
  \mathrm{SqRel}
  = \frac{1}{|\Omega_D|}\sum_{p\in\Omega_D}
    \frac{\left(D_p - \hat{D}_p\right)^{2}}{D_p}.
  \label{eq:sqrel}
\end{equation}

\textbf{Root Mean Squared Error (RMSE).}
\begin{equation}
  \mathrm{RMSE}
  = \sqrt{\frac{1}{|\Omega_D|}\sum_{p\in\Omega_D}\!\left(D_p - \hat{D}_p\right)^{2}}\,.
  \label{eq:rmse}
\end{equation}

\textbf{Log-scale RMSE (LogRMSE).}
\begin{equation}
  \mathrm{LogRMSE}
  = \sqrt{\frac{1}{|\Omega_D|}\sum_{p\in\Omega_D}\!\left(\log D_p - \log\hat{D}_p\right)^{2}}\,.
  \label{eq:logrmse}
\end{equation}

\textbf{Threshold Inlier Ratio ($\delta_{\tau}$).}
\begin{equation}
  \delta_{\tau}
  = \frac{1}{|\Omega_D|}
    \left|\left\{p\in\Omega_D\;\Bigg|\;
      \max\!\left(\frac{D_p}{\hat{D}_p},\;\frac{\hat{D}_p}{D_p}\right) < \tau
    \right\}\right|.
  \label{eq:delta}
\end{equation}
We report $\delta_{1.03}$, $\delta_{1.05}$, and $\delta_{1.10}$, following the RMVD evaluation protocol of~\citet{schroeppel2022rmvd}.
For AbsRel, SqRel, RMSE, and LogRMSE, lower is better; for $\delta_{\tau}$, higher is better.

\subsubsection{Dense-View Reconstruction}
\label{appendix:metrics:recon}

Scene-level reconstruction quality is evaluated by comparing the predicted and ground-truth point clouds on ScanNet++~\citep{yeshwanth2023scannet++}, NRGBD~\citep{azinovic2022Neuralrgbd}, DTU~\citep{jensen2014dtu}, Hiroom~\citep{lin2025depth}, and 7-Scenes~\citep{shotton20137scenes}, following the protocol of DA3-Bench~\citep{lin2025depth}.

The ground-truth point cloud $\mathcal{P}^{*}$ is obtained by uniformly sampling 
$10^{6}$ points from the GT mesh, which is reconstructed via offline TSDF fusion 
of all available RGB-D frames.
The predicted point cloud $\hat{\mathcal{P}}$ is obtained via online TSDF fusion of the model's per-frame predicted depth maps and camera poses, from which a triangle mesh is extracted and uniformly resampled to $10^{6}$ points.

Before computing metrics, both point clouds undergo dataset-specific preprocessing.
For ScanNet++, NRGBD, and DTU, $\hat{\mathcal{P}}$ is cropped to the axis-aligned bounding box of $\mathcal{P}^{*}$ inflated by $0.1$\,m to exclude out-of-range predictions. 
No cropping is applied for 7-Scenes and Hiroom.
Both point clouds are then voxel-downsampled with dataset-specific voxel sizes: $0.02$\,m for ScanNet++ and NRGBD, $0.01$\,m for DTU, and $4/512$\,m (${\approx}7.8$\,mm) for 7-Scenes and Hiroom.

We report two complementary families of Chamfer-based metrics.

\noindent\textbf{Threshold-based Precision and Recall.}
\begin{align}
  \mathrm{Precision}
  &= \frac{1}{|\hat{\mathcal{P}}|}\sum_{\hat{\mathbf{P}}_m\in\hat{\mathcal{P}}}
    \mathbf{1}\!\left[\min_{\mathbf{P}_n^{*}\in\mathcal{P}^{*}}
    \bigl\|\hat{\mathbf{P}}_m - \mathbf{P}_n^{*}\bigr\|_2 < d_{\tau}\right], \\
  \mathrm{Recall}
  &= \frac{1}{|\mathcal{P}^{*}|}\sum_{\mathbf{P}_n^{*}\in\mathcal{P}^{*}}
    \mathbf{1}\!\left[\min_{\hat{\mathbf{P}}_m\in\hat{\mathcal{P}}}
    \bigl\|\mathbf{P}_n^{*} - \hat{\mathbf{P}}_m\bigr\|_2 < d_{\tau}\right].
\end{align}
Precision is the fraction of predicted points within $d_{\tau}$ of any GT point. 
Recall is the fraction of GT points covered by the prediction within $d_{\tau}$. 
Both are reported as percentages.

\noindent\textbf{Mean Chamfer Accuracy and Completeness (metres).}
\begin{align}
  \overline{\mathrm{Acc}}
  &= \frac{1}{|\hat{\mathcal{P}}|}\sum_{\hat{\mathbf{P}}_m\in\hat{\mathcal{P}}}
    \min_{\mathbf{P}_n^{*}\in\mathcal{P}^{*}}\bigl\|\hat{\mathbf{P}}_m - 
    \mathbf{P}_n^{*}\bigr\|_2, \\
  \overline{\mathrm{Comp}}
  &= \frac{1}{|\mathcal{P}^{*}|}\sum_{\mathbf{P}_n^{*}\in\mathcal{P}^{*}}
    \min_{\hat{\mathbf{P}}_m\in\hat{\mathcal{P}}}\bigl\|\mathbf{P}_n^{*} - 
    \hat{\mathbf{P}}_m\bigr\|_2,
\end{align}
which measures the mean nearest-neighbor distances in metres.

\noindent\textbf{F-score and Overall.}
\begin{align}
  \text{F-score} &= \frac{2\cdot\mathrm{Precision}\cdot\mathrm{Recall}}
    {\mathrm{Precision}+\mathrm{Recall}}\quad(\%), \\
  \text{Overall}  &= \frac{\overline{\mathrm{Acc}}+\overline{\mathrm{Comp}}}{2}
    \quad(\mathrm{m}).
\end{align}
F-score is the harmonic mean of threshold-based precision and recall. 
Overall is the mean of the two Chamfer distances, reported in metres. 
The distance threshold is set to $d_{\tau}=0.05$\,m across all datasets.

\subsection{Evaluated Models Details}
\label{sec:model_overview}

We group the 32 methods (41 variants) into six paradigms.
For each model, we briefly summarize its core idea, and then list the key inference-time configuration adopted in our benchmark.
Unless otherwise noted, every model is driven by a unified \texttt{ModelAdapter} interface that feeds a complete scene (RGB tensors, GT intrinsics used only for evaluation) to the model and collects \texttt{pred\_depth}, \texttt{pred\_pose} (cam-to-world), and optionally \texttt{pred\_pointcloud}/\texttt{pred\_confidence}; depth is aligned to GT by median scaling before computing metrics unless the model produces metric depth.

\subsubsection{Optimization-based Methods}

\textbf{DUSt3R}~\citep{wang2024dust3r} regresses per-pair dense point maps in a canonical coordinate frame, and recovers a globally consistent reconstruction by a gradient-based global alignment over all image pairs.
In our benchmark, the ViT-Large 512 checkpoint (\texttt{naver/DUSt3R\_ViTLarge\_BaseDecoder\_512\_dpt}) is used, with \texttt{niter}=300, \texttt{schedule}=\texttt{cosine}, \texttt{lr}=0.01 for the global aligner, and inputs resized to width~512 aligned to a stride of 16.

\textbf{MASt3R}~\citep{leroy2024mast3r} extends DUSt3R with a dense matching head that produces robust 2D correspondences, enabling metric-scale recovery through a sparse global optimizer. We use the metric checkpoint \texttt{naver/MASt3R\_ViTLarge\_BaseDecoder\_512\_catmlpdpt\_metric} with \texttt{niter}=300, \texttt{schedule}=\texttt{linear}, \texttt{lr}=0.01, input width 512 (align 16), and median depth scaling for final evaluation.

\subsubsection{End-to-End Feed-Forward Methods}

\textbf{VGGT}~\citep{wang2025vggt} is a single forward model using a unified transformer that jointly predicts camera poses, depth maps, and point clouds for an arbitrary set of views.
We run \texttt{facebook/VGGT-1B} at the native 518-pixel resolution; no extra iterative refinement is applied.

\textbf{VGGT-Omega}~\citep{wang2026vggtomega} We use official implementation code and \texttt{VGGT-Omega-1B-512} checkpoint and run VGGT-Omega-1B at the native 512-pixel resolution. 

\textbf{Fast3R}~\citep{yang2025fast3r} performs one-shot multi-view prediction for up to a thousand views, using factorized global attention to scale VGGT-style reconstruction to long sequences.
We use \texttt{jedyang97/Fast3R\_ViT\_Large\_512} with input width~512 (align~16) and \texttt{niter\_PnP}=100 for the auxiliary PnP-RANSAC pose estimation.

\textbf{FastVGGT}~\citep{shen2025fastvggt} accelerates VGGT by a training-free token-merging scheme that drops redundant cross-view tokens inside the transformer.
In our runs (checkpoint \texttt{checkpoints/fastvggt/model\_tracker\_fixed\_e20.pt}) we set \texttt{merging}=0 (baseline layer) with \texttt{merge\_ratio}=0.9 and \texttt{enable\_point}=\texttt{false} to save memory.

\textbf{MUSt3R}~\citep{cabon2025must3r} couples DUSt3R-style pair prediction with a multi-view registrar that produces a single globally aligned reconstruction without per-scene optimization.
We use \texttt{MUSt3R\_512.pth} at width~512 (align~16), with \texttt{preserve\_gpu\_mem}=\texttt{true} and \texttt{num\_refinements\_iterations}=1.

\textbf{MapAnything}~\citep{keetha2026mapanything} is a universal feed-forward reconstructor that can ingest any combination of images, poses, intrinsics and depth priors and output metric-scale dense geometry.
We evaluate MapAnything on the \texttt{facebook/map-anything} checkpoint with \texttt{memory\_efficient\_inference}=\texttt{true} and \texttt{use\_amp}=\texttt{true}.

\textbf{OmniVGGT}~\citep{peng2025omnivggt} augments VGGT with a Geo-Adapter so that any subset of camera poses with intrinsics and depths can be injected as priors.
We run \texttt{Livioni/OmniVGGT} in its prior-free configuration for the main comparison, and additionally use its prior-injection branch in Table~\ref{tab:prior_ablation_combined}.

\textbf{Pi3} (\textbf{$\pi^{3}$})~\citep{wang2025pi3} removes the reference-frame bias of VGGT-style models with a permutation-equivariant design, predicting depth and cameras in a symmetric manner across views.
We evaluate \texttt{yyfz233/Pi3} and its metric variant \texttt{yyfz233/Pi3X}.
Pi3X additionally predicts metric-scale depth and is evaluated without scale alignment.
Pi3X also supports causal prior information injection, which is also used in Table~\ref{tab:prior_ablation_combined}.

\textbf{AMB3R}~\citep{wang2025amb3r} introduces an attention-based multi-branch backbone that outputs metric-scale depth and poses in one forward pass.
We use \texttt{checkpoints/amb3r/amb3r.pt} with \texttt{depth\_alignment}=\texttt{none} and \texttt{data\_type}=\texttt{bf16}.

\textbf{Depth Anything 3 (DA3)}~\citep{lin2025depth} is a scalable 3D foundation model based on transformers trained on large-scale data.
We benchmark four sizes: \textbf{DA3-Small} (\texttt{depth-anything/DA3-SMALL}), \textbf{DA3-Base} (\texttt{depth-anything/DA3-BASE}), \textbf{DA3-Large 1.1} (\texttt{depth-anything/DA3-LARGE-1.1}) and \textbf{DA3-Giant 1.1} (\texttt{depth-anything/DA3-GIANT-1.1}), all with reference-view strategy \texttt{ref\_view\_strategy}=\texttt{first} for fair comparison.
The two-branch variant \textbf{DA3Nested} (\texttt{depth-anything/DA3NESTED-GIANT-LARGE-1.1}) outputs metric-scale depth by combining an anyview Giant branch with a metric Large branch.

\textbf{WorldMirror}~\citep{liu2025worldmirror} is a unified feed-forward reconstruction model that can flexibly condition on (pose, depth, intrinsic) priors.
We use \texttt{tencent/HunyuanWorld-Mirror} with \texttt{cond\_flags}=[0,0,0] (no prior) for the main benchmark.

\subsubsection{Online / Streaming Methods}

\textbf{Spann3R}~\citep{wang2024spann3r} builds an external spatial memory on top of DUSt3R so that each incoming frame can be incrementally fused into a shared canonical frame.
We use \texttt{spann3r\_101.pth} at a resolution of 224 (align~16) with \texttt{inference\_mode}=\texttt{online}, \texttt{use\_feat}=\texttt{false} and
\texttt{focal\_mode}=\texttt{weiszfeld}.

\textbf{CUT3R}~\citep{wang2025cut3r} performs continuous updating of a persistent internal scene state, regressing dense geometry frame-by-frame. We use the default CUT3R checkpoint at width~512 (align~16) with \texttt{focal\_mode}=\texttt{weiszfeld} for focal estimation.

\textbf{MonST3R}~\citep{zhang2024monst3r} specializes DUSt3R for dynamic scenes by injecting temporal smoothness, optical-flow, and translation losses into the global optimizer.
We use \texttt{Junyi42/MonST3R\_PO-TA-S-W\_ViTLarge\_BaseDecoder\_512\_dpt} at width~512 (align~16) with \texttt{niter}=300, \texttt{schedule}=\texttt{linear}, \texttt{lr}=0.01, \texttt{batch\_size}=16, \texttt{winsize}=5, \texttt{scenegraph\_type}=\texttt{swinstride}, \texttt{flow\_loss\_weight}=0.01, \texttt{flow\_loss\_threshold}=25, \texttt{flow\_loss\_start\_iter}=0.1, \texttt{sam2\_mask\_refine}=\texttt{true}, \texttt{batchify}=\texttt{true} (\texttt{temporal\_smoothing\_weight}=0.01, \texttt{translation\_weight}=1.0, \texttt{shared\_focal}=\texttt{true}).

\textbf{Point3R}~\citep{point3r} maintains an explicit online 3D point memory that grows as new frames arrive, enabling streaming reconstruction with constant-time per-frame cost. We use \texttt{point3r\_512.pth} at width~512 (align~16) with \texttt{focal\_mode}=\texttt{weiszfeld}.

\textbf{Stream3R}~\citep{stream3r2025} formulates multi-view reconstruction as causal next-token prediction over pointmaps, enabling an efficient StreamVGGT-style decoder. 
We use \texttt{yslan/STream3R} checkpoint and implement \texttt{stream} and \texttt{window} variants using default settings.

\textbf{StreamVGGT}~\citep{streamVGGT} is a streaming counterpart of VGGT with temporally causal cross-view attention and KV-caching.
We use \texttt{lch01/StreamVGGT} in the default online mode.

\textbf{PAGE4D}~\citep{zhou2025page4d} extends VGGT to 4D reconstruction by modeling both static scene geometry and dynamic motion components.
We use \texttt{checkpoints/page4d/checkpoint\_nomask.pt} in its default configuration.

\textbf{InfiniteVGGT}~\citep{yuan2026infinitevggt} augments StreamVGGT with a rolling KV-cache, so that arbitrarily long sequences can be processed with bounded memory.
We use \texttt{lch01/StreamVGGT} as the backbone, image width 518 (align~14) and \texttt{total\_budget}=1{,}200{,}000 KV-tokens.

\textbf{WinT3R}~\citep{li2025wint3r} employs a sliding-window transformer with a camera-token pool to capture long-range geometry cheaply.
We use \texttt{lizizun/WinT3R} at width~512 (align~16) with \texttt{inference\_mode}=\texttt{online}, \texttt{window\_size}=4, \texttt{state\_size}=1024 and \texttt{ret\_first\_pred}=\texttt{false}.

\textbf{LongStream}~\citep{cheng2026longstream} builds on a VGGT backbone with a keyframe-driven long-context memory module to target long-horizon streaming reconstruction.
We use \texttt{NicolasCC/LongStream} at width~518 (align~14) under \texttt{streaming\_mode}=\texttt{causal} with \texttt{keyframe\_stride}=8, \texttt{keyframe\_mode}=\texttt{fixed} and \texttt{rel\_pose\_num\_iterations}=4, and evaluate two inference variants: \emph{batch-refresh} and \emph{streaming-refresh} with a sliding window of size~48.
The \texttt{refresh} parameter is set to 3 for \emph{batch-refresh} and 7 for \emph{streaming-refresh}.
In addition, we disable the scale token prediction of LongStream, as we observe that enabling metric scale prediction introduces erroneous outliers in depth estimation metrics on our benchmark, despite the model natively supporting metric-scale output.

\textbf{LingBot-Map}~\citep{chen2026lingbotmap} is an online mapping model designed for streaming data, supporting both per-frame streaming and window-based inference with KV-cache sliding windows.
We evaluate two variants: \textbf{LingBot-Map (windowed)} with \texttt{mode}=\texttt{windowed}, \texttt{window\_size}=64, \texttt{overlap\_size}=16, \texttt{num\_scale\_frames}=8, \texttt{kv\_cache\_sliding\_window}=64 and \texttt{camera\_num\_iterations}=4; and \textbf{LingBot-Map (streaming)} with adaptive \texttt{keyframe\_interval}, \texttt{kv\_cache\_sliding\_window}=64, \texttt{kv\_cache\_scale\_frames}=8 and SDPA attention (\texttt{use\_sdpa}=\texttt{false}, \texttt{enable\_3d\_rope}=\texttt{true}). Both variants use \texttt{image\_size}=518, \texttt{patch\_size}=14 and \texttt{enable\_point}=\texttt{false}. The medium regime runs use the long-context checkpoint \texttt{lingbot-map-long.pt}, while the sparse and single-frame settings use the short-context \texttt{lingbot-map.pt} checkpoint.

\subsubsection{Chunk-based Methods}

\textbf{VGGT-Long}~\citep{deng2025vggtlong} scales VGGT to long sequences by overlapping-chunk inference and cross-chunk Sim(3) alignment.
We use \texttt{facebook/VGGT-1B} as the backbone with \texttt{chunk\_size}=60 and \texttt{overlap}=30.
The chunked streaming variant \textbf{DA3-Streaming} uses backbone \texttt{depth-anything/DA3-GIANT-1.1} with \texttt{chunk\_size}=60, \texttt{overlap}=30 and \texttt{ref\_view\_strategy}=\texttt{first} for long sequences.
The long-sequence variant \textbf{Pi-Long} (built on the Pi3 backbone) also uses \texttt{chunk\_size}=60, \texttt{overlap}=30.

\subsubsection{SLAM-based Methods}

\textbf{MASt3R-SLAM}~\citep{murai2024_mast3rslam} integrates MASt3R's metric dense
matcher into a SLAM pipeline that jointly estimates camera trajectory and
metric geometry.
We use \texttt{MASt3R\_ViTLarge\_BaseDecoder\_512\_catmlpdpt\_metric.pth} in uncalibrated mode (\texttt{use\_calib}=\texttt{false}), with explicit overrides \texttt{img\_size}=512 and \texttt{c\_conf\_threshold}=1.5 (point-cloud filtering); the remaining tracking hyper-parameters use the official \texttt{config/base.yaml} defaults (\texttt{C\_conf}=0.0, \texttt{Q\_conf}=1.5, \texttt{min\_match\_frac}=0.05, \texttt{max\_iters}=50).

\textbf{VGGT-SLAM}~\citep{maggio2025vggt-slam2} turns VGGT into a SLAM system
by constructing overlapping submaps and performing Sim(3) cross-submap
alignment with loop closures.
We use \texttt{facebook/VGGT-1B} with \texttt{submap\_size}=32, \texttt{overlapping\_window\_size}=3, \texttt{max\_loops}=0 (no loop closure), \texttt{conf\_threshold}=25.0, \texttt{lc\_thres}=0.95 and \texttt{use\_optical\_flow}=\texttt{true}.

\subsubsection{Test-Time Training Methods}

\textbf{TTT3R}~\citep{chen2025ttt3r} performs per-sequence test-time training on top of a CUT3R-style backbone, fine-tuning a small set of parameters to adapt to each scene.
We use \texttt{cut3r\_512\_dpt\_4\_64.pth} at width~512 (align~16) with \texttt{focal\_mode}=\texttt{weiszfeld}.

\textbf{Scal3R}~\citep{xie2026scal3r} introduces a scalable pipeline that processes long videos in blocks with loop-level test-time refinement.
We use \texttt{xbillowy/Scal3R} at width~518 (align~14) with \texttt{block\_size}=60, \texttt{overlap\_size}=30 (20 for dense), \texttt{loop\_size}=20, \texttt{use\_xyz\_align}=1 and \texttt{test\_use\_amp}=\texttt{false}.

\textbf{LoGeR}~\citep{zhang2026loger} is a long-context hybrid-memory geometry model that combines test-time training with sliding-window attention.
We evaluate the \texttt{LoGeR} and \texttt{LoGeR}$^*$ variants from \texttt{Junyi42/LoGeR} at width~518 (align~14) with \texttt{window\_size}=32, \texttt{overlap\_size}=3 and \texttt{reset\_every}=1 (TTT-state reset across windows). For \texttt{LoGeR} we set \texttt{se3}=\texttt{false}, \texttt{sim3}=\texttt{false}; for \texttt{LoGeR}$^*$ we set \texttt{se3}=\texttt{true}, \texttt{sim3}=\texttt{false}.

\subsection{\benchmark Visualizations}
\label{prior_vis}

\begin{figure*}[p]
  \centering
  \includegraphics[width=0.9\textwidth]{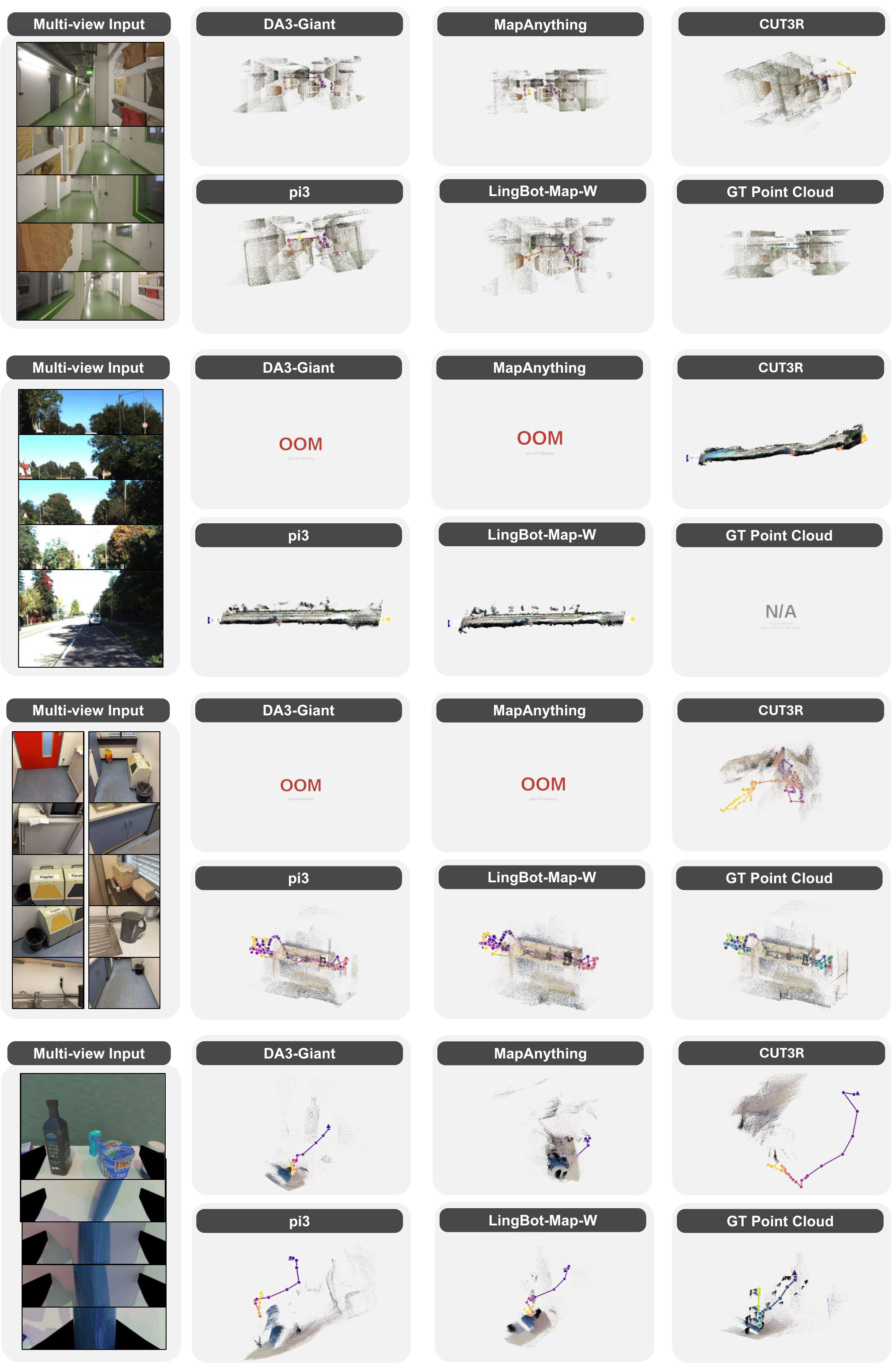}
  \caption{\textbf{Representative benchmark cases.}
  We show input views, per-method point-cloud reconstructions, and depth/camera
  metrics for four representative scenes spanning short indoor reconstruction,
  dense outdoor driving, dense indoor long-horizon reconstruction, and wrist-view
  OOD input.}
  \label{fig:representative_cases}
\end{figure*}

\textbf{Qualitative Comparison on Representative \benchmark Cases.}
Fig.~\ref{fig:representative_cases} visualizes four representative cases from
\benchmark, covering short indoor reconstruction, dense outdoor driving, dense
indoor long-horizon reconstruction, and wrist-view out-of-distribution input.
Each case shows the sampled input views, point-cloud reconstructions from
representative methods, and the corresponding depth and camera metrics.

\begin{figure}[t] 
\centering 
\includegraphics[width=\columnwidth]{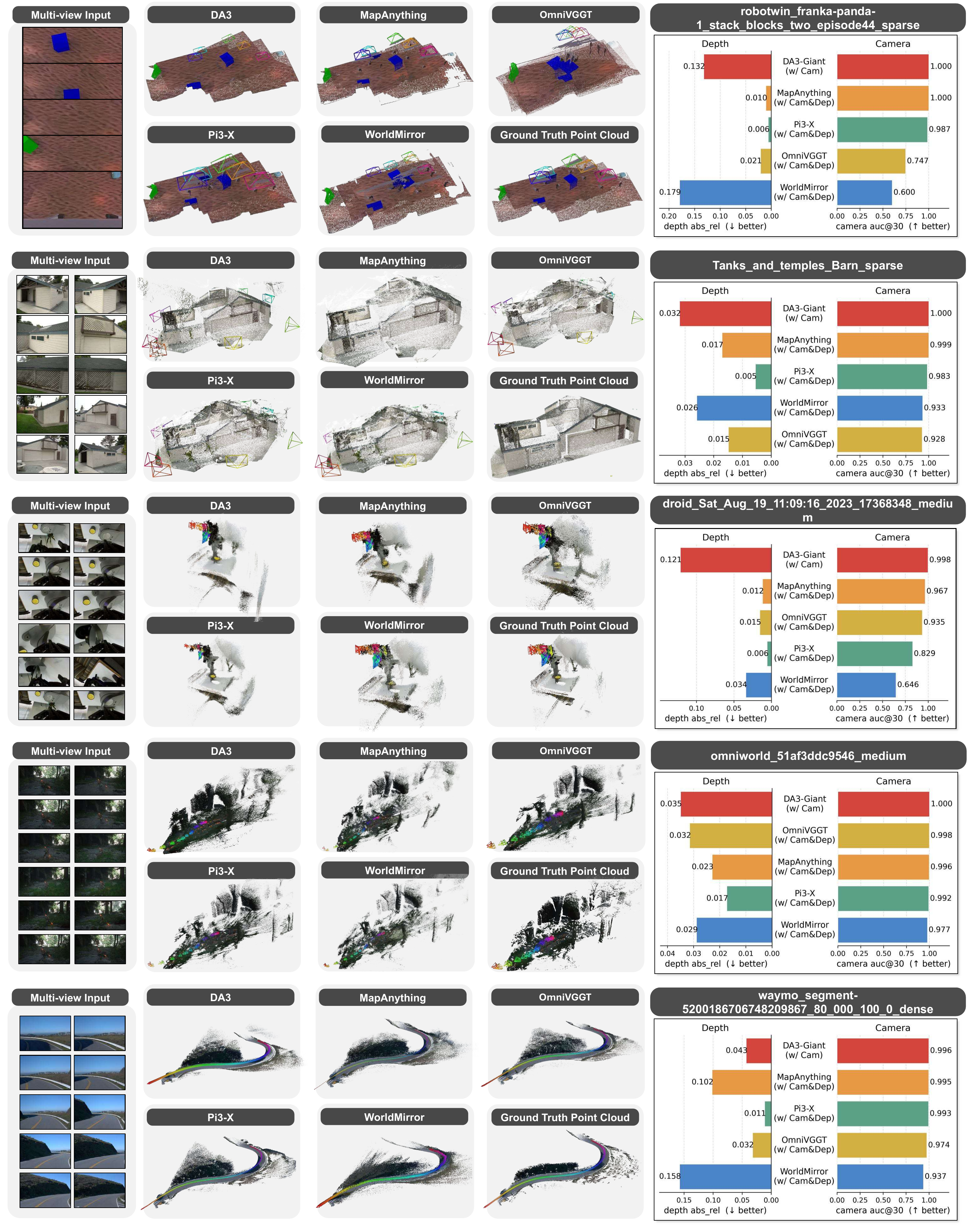}
\caption{\textbf{Qualitative Comparison on Prior-Enhanced Models with Auxiliary Camera and Depth Information}. We visualize the reconstruction quality of DA3-Giant, MapAnything, OmniVGGT, Pi3, and WorldMirror under the setting where both camera and depth prior information are provided as auxiliary inputs.
For each scene, the right panel reports the depth AbsRel and camera AUC@30 metrics 
for each method.} 
\label{Fig.prior_both} %
\end{figure}

\begin{figure}[t] 
\centering 
\includegraphics[width=\columnwidth]{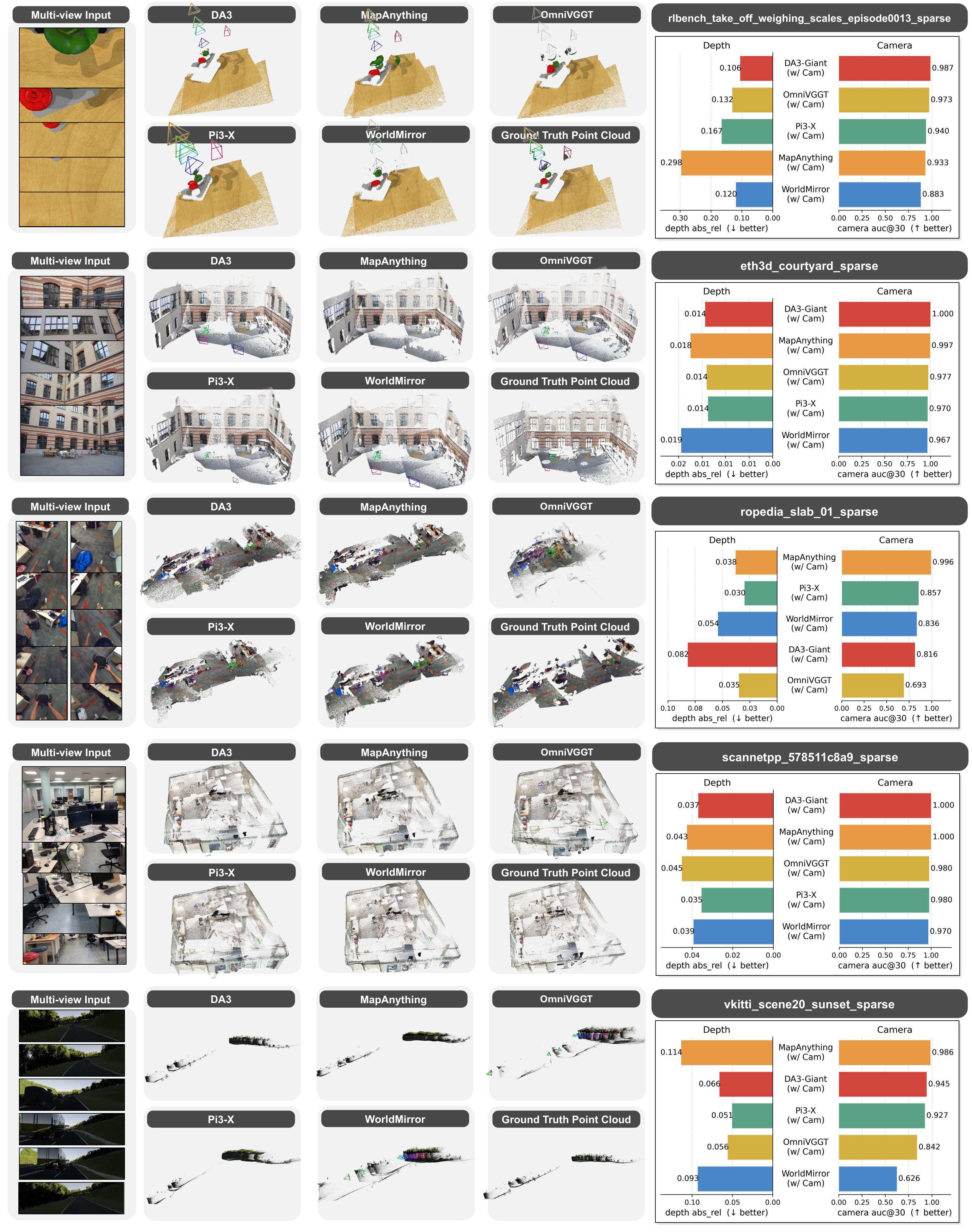}
\caption{\textbf{Qualitative Comparison on Prior-Enhanced Models with Auxiliary Camera Information}. We visualize the reconstruction quality of DA3-Giant, MapAnything, OmniVGGT, Pi3, and WorldMirror under the setting where camera prior information is provided as auxiliary inputs.
For each scene, the right panel reports the depth AbsRel and camera AUC@30 metrics 
for each method.} 
\label{Fig.prior_cam} %
\end{figure}

\begin{figure}[t] 
\centering 
\includegraphics[width=\columnwidth]{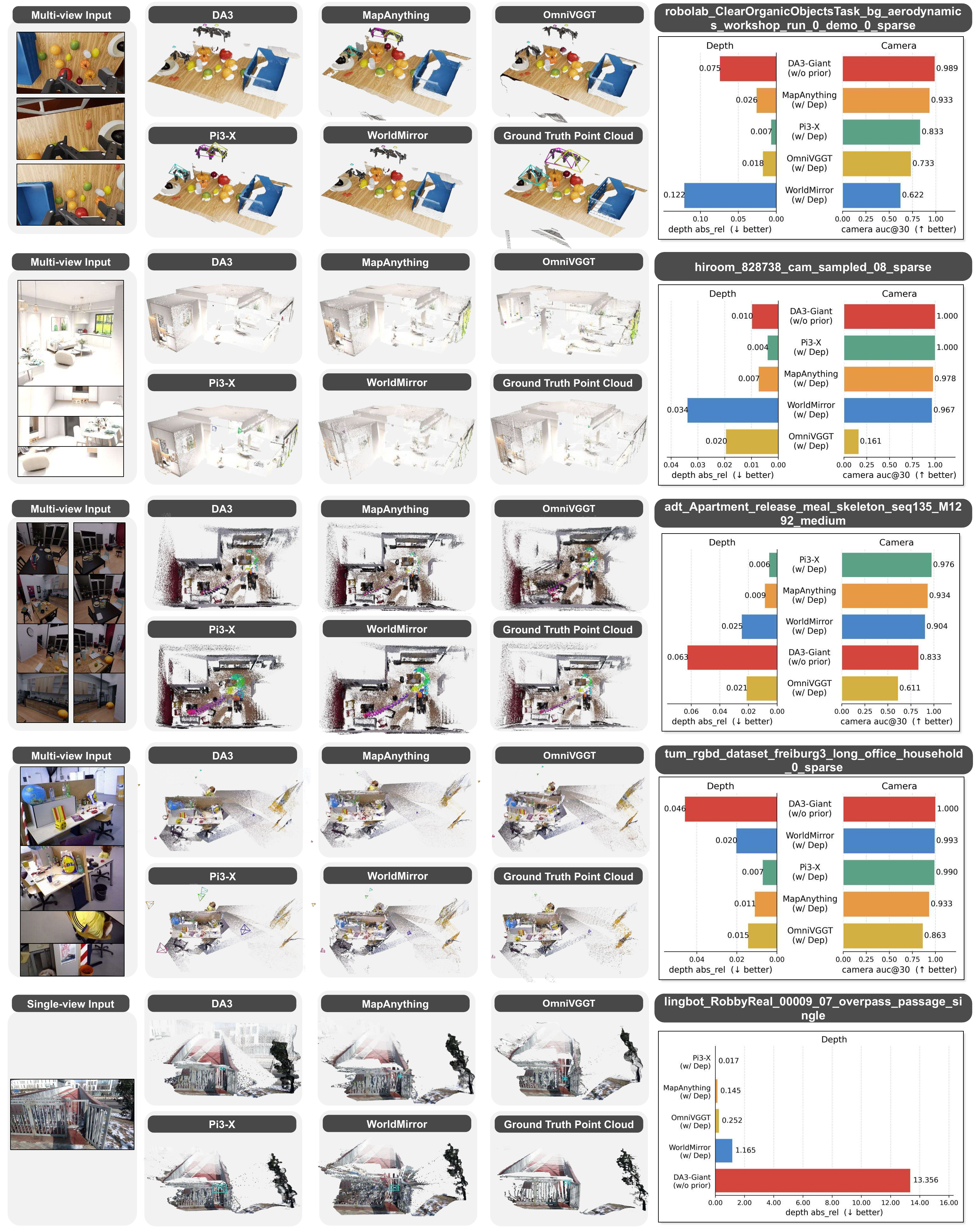}
\caption{\textbf{Qualitative Comparison on Prior-Enhanced Models with Auxiliary Depth Information}. We visualize the reconstruction quality of DA3-Giant, MapAnything, OmniVGGT, Pi3, and WorldMirror under the setting where depth prior information is provided as auxiliary inputs.
For each scene, the right panel reports the depth AbsRel and camera AUC@30 metrics 
for each method.} 
\label{Fig.prior_depth} %
\end{figure}

\textbf{Qualitative Comparison on Prior-Enhanced Models.}
Fig.~\ref{Fig.prior_both}, Fig.~\ref{Fig.prior_cam}, and Fig.~\ref{Fig.prior_depth} present qualitative comparisons of prior-aware models under joint camera and depth prior injection, camera-only prior injection, and depth-only prior injection settings, respectively.

\begin{figure}[t] 
\centering 
\includegraphics[width=\columnwidth]{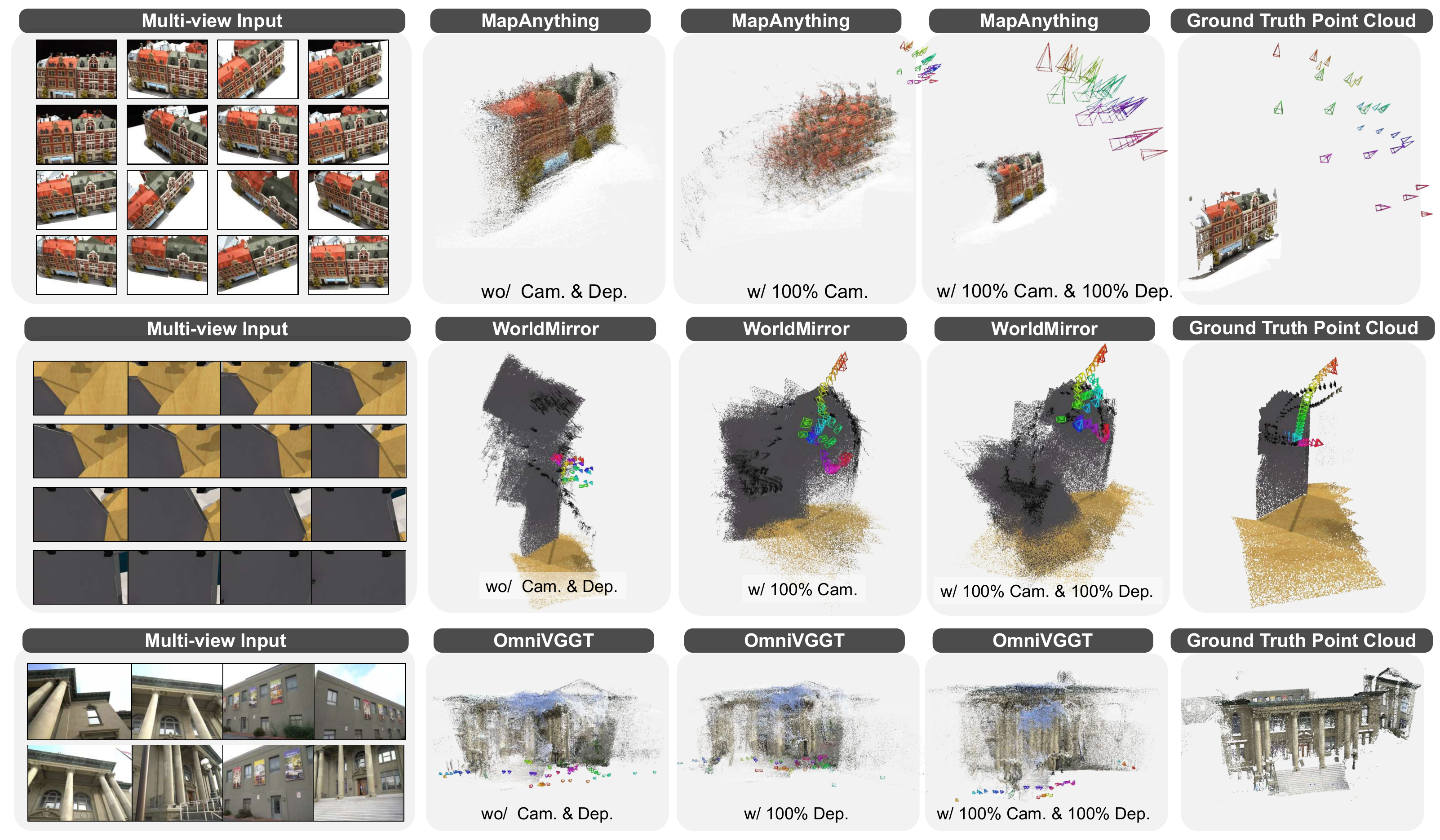}
\caption{\textbf{Failure Cases on Challenging Scenes.} We visualize failure cases 
of MapAnything, WorldMirror, and OmniVGGT across three challenging scenes. } 
\label{Fig.bad_case} 
\end{figure}

\textbf{Bad Cases For Prior-Enhanced Models.}
Fig.~\ref{Fig.bad_case} illustrates representative failure cases of prior-aware 
models under challenging conditions, including object-centric scenes, extreme no-overlap
view, and out-of-distribution wrist-view sequences.
MapAnything fails on object-centric scenes when GT camera priors are injected.  
WorldMirror fails on out-of-distribution scenes such as wrist-view sequences, even with both camera and depth priors provided.
OmniVGGT breaks down under extreme no-overlap conditions.

\subsection{Evaluated Models and Training Datasets in \benchmark}
In this subsection, we summarize the training data and annotations used by all evaluated methods. 
Tab.~\ref{tab:merged_dataset_usage_training_only} presents the training datasets of all compared methods, excluding training-free and chunk-wise methods, as they do not involve explicit training.
Evaluation-only, qualitative-only, and runtime benchmark datasets are not marked.
\textcolor{trainmark}{\checkmark} indicates datasets directly used for model training. 
\textcolor{pretrainmark}{\checkmark} indicates that the method is initialized from a pretrained checkpoint whose backbone has been trained on the corresponding dataset.
Tab.~\ref{tab:dataset_profile_summary_cited} presents the properties of each dataset, including scene type, real-world vs.\ synthetic domain, the presence of dynamic content, and public license information.

\begin{table*}[h]
\definecolor{altcolblue}{RGB}{225,237,250}
\definecolor{pretrainmark}{RGB}{197,90,17}
\providecommand{\cmark}{\checkmark}
\providecommand{\pmark}{\textcolor{pretrainmark}{\checkmark}}
\providecommand{\datasetcount}[2]{\textcolor{pretrainmark}{#1}~(#2)}
\providecommand{\rot}[1]{\rotatebox{60}{#1}}
\caption{\textbf{Dataset usage across all methods}. 
\textcolor{green!55!black}{\checkmark} indicates datasets directly used for model training, fine-tuning, or teacher/student training; \textcolor{pretrainmark}{\checkmark} indicates datasets inherited from a pretrained checkpoint whose backbone has been trained on the corresponding dataset.
The last row reports total used (trained): the first number counts all datasets either directly used by the method or inherited through its pretrained checkpoint, while the number in parentheses counts only datasets used to train or fine-tune the method itself. Identical values, e.g., 42~(42), indicate that no additional datasets are inherited.
Evaluation-only, qualitative-only, and runtime benchmark datasets are not marked.
}
\centering
\tiny
\setlength{\tabcolsep}{1.7pt}
\renewcommand{\arraystretch}{1.4}
\resizebox{\textwidth}{!}{
\begin{tabular}{l>{\columncolor{altcolblue}}cc>{\columncolor{altcolblue}}cc>{\columncolor{altcolblue}}cc>{\columncolor{altcolblue}}cc>{\columncolor{altcolblue}}cc>{\columncolor{altcolblue}}cc>{\columncolor{altcolblue}}cc>{\columncolor{altcolblue}}cc>{\columncolor{altcolblue}}cc>{\columncolor{altcolblue}}cc>{\columncolor{altcolblue}}cc>{\columncolor{altcolblue}}cc}
\toprule
Dataset
& \multicolumn{11}{c}{End-to-end}
& \multicolumn{10}{c}{Online / streaming}
& \multicolumn{3}{c}{Test-Time Training} \\
\cmidrule(lr){2-12}\cmidrule(lr){13-22}\cmidrule(lr){23-25}
& \rot{AMB3R}
& \rot{DA3}
& \rot{DUSt3R}
& \rot{Fast3R}
& \rot{MapAnything}
& \rot{MASt3R}
& \rot{MUSt3R}
& \rot{OmniVGGT}
& \rot{$\pi^3$}
& \rot{VGGT}
& \rot{WorldMirror}
& \rot{CUT3R}
& \rot{LingBot-map}
& \rot{LongStream}
& \rot{MonST3R}
& \rot{PAGE-4D}
& \rot{Point3R}
& \rot{Spann3R}
& \rot{STream3R}
& \rot{StreamVGGT}
& \rot{WinT3R}
& \rot{TTT3R}
& \rot{Scal3R}
& \rot{LoGeR} \\
\midrule
CO3D~\citep{reizenstein2021co3d} & \pmark & \cmark & \cmark & \cmark &  & \cmark & \cmark & \pmark & \cmark & \cmark & \cmark & \cmark & \cmark & \cmark & \pmark & \cmark & \cmark & \cmark & \cmark & \cmark & \cmark & \pmark & \cmark & \pmark \\
RealEstate10K~\citep{zhou2018stereomagnification} &  &  &  &  &  &  &  &  &  &  &  & \cmark &  &  &  &  &  &  &  &  &  & \pmark &  &  \\
ScanNet~\citep{dai2017scannet} & \cmark &  &  &  &  &  &  & \cmark & \cmark & \cmark & \pmark & \cmark & \cmark & \cmark &  & \pmark & \cmark & \cmark & \cmark & \cmark & \cmark & \pmark &  & \cmark \\
ScanNet++~\citep{yeshwanth2023scannet++} & \cmark & \cmark & \cmark & \cmark & \cmark & \cmark & \cmark & \cmark & \cmark &  & \cmark & \cmark & \cmark &  & \pmark &  & \cmark & \cmark & \cmark &  & \cmark & \pmark & \cmark & \cmark \\
ARKitScenes~\citep{dehghan2021arkitscenes} &  & \cmark & \cmark & \cmark &  & \cmark & \cmark & \cmark & \cmark &  & \cmark & \cmark &  &  & \pmark &  & \cmark & \cmark & \cmark & \cmark & \cmark & \pmark &  & \cmark \\
Habitat~\citep{savva2019habitat} & \pmark &  & \cmark & \cmark &  & \cmark & \cmark & \pmark & \cmark & \cmark & \cmark &  &  & \pmark & \pmark & \pmark & \pmark & \cmark & \pmark & \pmark & \pmark &  &  & \pmark \\
BlendedMVS~\citep{yao2020blendedmvs} & \pmark & \cmark & \cmark & \cmark & \cmark & \cmark & \cmark & \cmark & \cmark & \cmark & \cmark & \cmark & \cmark & \cmark & \pmark & \pmark & \cmark & \cmark & \cmark & \cmark & \cmark & \pmark & \cmark & \pmark \\
MegaDepth~\citep{li2018megadepth} & \pmark & \cmark & \cmark & \cmark & \cmark & \cmark & \cmark & \cmark & \cmark & \cmark & \cmark & \cmark & \cmark & \cmark & \pmark & \pmark & \cmark & \pmark & \cmark & \cmark & \cmark & \pmark & \cmark & \cmark \\
DL3DV~\citep{ling2024dl3dv} & \pmark & \cmark &  &  & \cmark & \cmark & \cmark & \cmark & \pmark & \cmark & \cmark & \cmark & \cmark & \cmark &  & \pmark &  &  & \cmark & \pmark &  & \pmark & \cmark & \cmark \\
MapFree~\citep{arnold2022mapfree} & \cmark & \cmark &  &  &  & \cmark & \cmark & \cmark &  &  & \cmark & \cmark & \cmark &  &  &  &  &  & \cmark &  &  & \pmark & \cmark &  \\
WildRGBD~\citep{xia2024wildrgbd} & \cmark & \cmark &  &  &  & \cmark & \cmark & \cmark & \cmark & \cmark & \cmark & \cmark & \cmark & \cmark &  & \pmark & \cmark &  & \cmark & \cmark & \cmark & \pmark & \cmark & \pmark \\
Waymo~\citep{sun2020waymo} & \cmark & \cmark & \cmark & \pmark &  & \cmark &  & \cmark &  &  &  & \cmark & \cmark & \cmark & \cmark & \cmark & \cmark & \pmark & \cmark & \cmark & \pmark & \pmark &  & \cmark \\
VirtualKITTI~\citep{cabon2020vkitti} & \cmark & \cmark &  &  &  & \cmark & \cmark & \cmark & \pmark & \cmark & \pmark & \cmark & \cmark & \cmark &  & \pmark & \cmark &  & \cmark & \cmark &  & \pmark & \cmark & \cmark \\
TartanAir~\citep{wang2020tartanair} &  & \cmark &  & \cmark & \cmark & \cmark & \cmark & \cmark & \cmark &  & \cmark & \cmark & \cmark &  & \cmark &  &  &  & \cmark &  & \cmark & \pmark & \cmark & \cmark \\
Unreal4K~\citep{tosi2021unreal} &  & \cmark &  &  & \cmark & \cmark & \cmark & \cmark &  &  & \cmark & \cmark & \cmark &  &  &  &  &  & \cmark &  &  & \pmark &  & \cmark \\
MVS-Synth~\citep{huang2018mvs} & \cmark & \cmark &  &  & \cmark &  &  & \cmark & \pmark & \cmark & \cmark & \cmark & \cmark & \cmark &  & \pmark & \cmark &  & \cmark & \cmark &  & \pmark & \cmark & \pmark \\
HyperSim~\citep{roberts2021hypersim} & \cmark & \cmark &  &  &  &  &  & \cmark & \cmark & \cmark & \cmark & \cmark & \cmark & \cmark &  & \pmark & \cmark &  & \cmark & \cmark & \cmark & \pmark & \cmark & \cmark \\
Mapillary~\citep{antequera2020mapillary} & \pmark &  &  &  & \cmark &  &  & \pmark & \pmark & \cmark & \pmark &  &  & \cmark &  & \pmark &  &  & \pmark & \pmark &  &  & \cmark & \pmark \\
ASE~\citep{avetisyan2024scenescript} & \pmark & \cmark &  &  & \cmark &  &  & \pmark & \pmark & \cmark & \cmark &  & \cmark & \cmark &  & \pmark &  &  & \cmark & \pmark &  &  & \cmark & \pmark \\
ADT~\citep{pan2023aria} & \cmark & \cmark &  &  &  &  &  & \pmark & \pmark & \cmark & \pmark &  & \cmark & \cmark &  & \pmark &  &  & \pmark & \pmark &  &  & \cmark & \pmark \\
PointOdyssey~\citep{zheng2023pointodyssey} & \pmark & \cmark &  & \cmark &  &  &  & \pmark & \pmark & \cmark & \pmark & \cmark & \cmark & \cmark & \cmark & \cmark & \cmark &  & \cmark & \cmark &  & \pmark &  & \pmark \\
Spring~\citep{mehl2023spring} &  & \cmark &  &  & \cmark &  &  & \cmark &  &  &  & \cmark &  & \cmark & \cmark & \cmark & \cmark &  & \cmark & \cmark &  & \pmark &  & \cmark \\
Dynamic Replica~\citep{Karaev_2023_dynamic_stereo} &  &  &  &  & \cmark &  &  & \cmark &  &  &  & \cmark &  &  &  & \cmark &  &  & \cmark &  &  & \pmark &  &  \\
ParallelDomain-4D~\citep{vanhoorick2024gcd} &  &  &  &  & \cmark &  &  &  &  &  &  &  &  &  &  &  &  &  &  &  &  &  &  &  \\
SAIL-VOS 3D~\citep{hu2021sailvos3d} &  &  &  &  & \cmark &  &  &  &  &  &  &  &  &  &  &  &  &  &  &  &  &  &  &  \\
OmniObject3D~\citep{wu2023omniobject3d} & \cmark & \cmark &  &  &  &  &  &  &  &  &  & \cmark &  &  &  &  & \cmark &  & \cmark & \cmark &  & \pmark &  &  \\
StaticThings3D~\citep{schroeppel2022rmvd} &  &  & \cmark & \pmark &  & \cmark & \cmark &  &  &  &  &  &  &  & \pmark &  & \pmark & \pmark & \pmark &  & \pmark &  &  &  \\
TartanGround~\citep{patel2025tartanground} &  &  &  &  &  &  &  &  &  &  &  &  & \cmark &  &  &  &  &  &  &  &  &  &  & \cmark \\
Matterport3D~\citep{chang2017matterport3d} &  &  &  &  &  &  &  & \cmark &  &  & \cmark & \cmark & \cmark &  &  &  &  &  & \cmark &  &  & \pmark &  &  \\
BEDLAM~\citep{black2023bedlam} &  &  &  &  &  &  &  &  &  &  &  & \cmark &  &  &  &  &  &  & \cmark &  &  & \pmark &  &  \\
UASOL~\citep{bauer2019uasol} &  &  &  &  &  &  &  & \cmark &  &  &  & \cmark &  &  &  &  &  &  & \cmark &  &  & \pmark &  &  \\
MVImgNet~\citep{yu2023mvimgnet} &  &  &  &  &  &  &  &  &  &  &  & \cmark &  &  &  &  &  &  & \cmark &  &  & \pmark &  &  \\
CoP3D~\citep{sinha2023cop3d} &  &  &  &  &  &  &  &  &  &  &  & \cmark &  &  &  &  &  &  & \cmark &  &  & \pmark &  &  \\
EDEN~\citep{le2021eden} &  & \cmark &  &  &  &  &  &  &  &  &  & \cmark &  &  &  &  &  &  & \cmark &  &  & \pmark &  &  \\
IRS~\citep{wang2021irs} &  & \cmark &  &  &  &  &  &  &  &  &  & \cmark &  &  &  &  &  &  &  &  &  & \pmark &  &  \\
Synscapes~\citep{wrenninge2018synscapes} &  &  &  &  &  &  &  &  &  &  &  & \cmark &  &  &  &  &  &  & \cmark &  &  & \pmark &  &  \\
3D Ken Burns~\citep{niklaus2019kenburns} &  & \cmark &  &  &  &  &  &  &  &  &  & \cmark &  &  &  &  &  &  &  &  &  & \pmark &  &  \\
SmartPortraits~\citep{kornilova2022smartportraits} &  &  &  &  &  &  &  &  &  &  &  & \cmark &  &  &  &  &  &  &  &  &  & \pmark &  &  \\
UrbanSyn~\citep{gomez2025urbansyn} &  & \cmark &  &  &  &  &  &  &  &  &  & \cmark &  &  &  &  &  &  & \cmark &  &  & \pmark &  &  \\
HOI4D~\citep{liu2022hoi4d} &  &  &  &  &  &  &  &  &  &  &  & \cmark &  &  &  &  &  &  & \cmark &  &  & \pmark &  &  \\
Kubric~\citep{greff2022kubric} & \pmark &  &  &  &  &  &  & \cmark & \pmark & \cmark & \pmark &  & \cmark & \cmark &  & \cmark &  &  & \pmark & \pmark &  &  &  & \pmark \\
Replica~\citep{straub2019replica} & \pmark & \cmark &  &  &  &  &  & \pmark & \pmark & \cmark & \pmark &  & \cmark & \cmark &  & \pmark &  &  & \pmark & \pmark &  &  & \cmark & \pmark \\
Objaverse~\citep{deitke2023objaverse} & \pmark & \cmark &  &  &  &  &  & \pmark & \pmark & \cmark & \pmark &  & \cmark & \cmark &  & \pmark &  &  & \pmark & \pmark &  &  &  & \pmark \\
Texverse~\citep{zhang2025texverse} &  &  &  &  &  &  &  &  &  &  &  &  & \cmark &  &  &  &  &  &  &  &  &  &  &  \\
GTA-SfM~\citep{wang2020flow} & \cmark & \cmark &  &  &  &  &  &  & \cmark &  &  &  & \cmark &  &  &  &  &  &  &  & \cmark &  &  & \pmark \\
SceneNet RGB-D~\citep{mccormac2016scenenet} &  & \cmark &  &  &  &  &  &  &  &  &  &  & \cmark &  &  &  &  &  &  &  &  &  & \cmark &  \\
MatrixCity~\citep{li2023matrixcity} &  & \cmark &  &  &  &  &  &  & \cmark &  &  &  & \cmark &  &  &  &  &  &  &  & \cmark &  & \cmark & \pmark \\
Mid-Air~\citep{fonder2019midair} &  &  &  &  &  &  &  &  & \cmark &  &  &  & \cmark &  &  &  &  &  &  &  &  &  &  & \pmark \\
KITTI-360~\citep{liao2021kitti360} &  &  &  &  &  &  &  &  &  &  &  &  & \cmark &  &  &  &  &  &  &  &  &  &  &  \\
Gibson~\citep{xia2018gibson} &  &  &  &  &  &  &  &  &  &  &  &  & \cmark &  &  &  &  &  &  &  &  &  &  &  \\
HM3D~\citep{ramakrishnan2021hm3d} &  &  &  &  &  &  &  &  &  &  &  &  & \cmark &  &  &  &  &  &  &  &  &  &  &  \\
Taskonomy~\citep{zamir2018taskonomy} &  & \cmark &  &  &  &  &  &  & \cmark &  &  &  &  &  &  &  &  &  &  &  & \cmark &  & \cmark & \pmark \\
MegaSynth~\citep{jiang2025megasynth} &  & \cmark &  &  &  &  &  &  &  &  &  &  &  &  &  &  &  &  &  &  &  &  &  &  \\
OmniWorld~\citep{zhou2025omniworld} &  & \cmark &  &  &  &  &  &  &  &  &  &  &  &  &  &  &  &  &  &  &  &  &  & \cmark \\
Trellis~\citep{xiang2025structured} &  & \cmark &  &  &  &  &  &  &  &  &  &  &  &  &  &  &  &  &  &  &  &  &  &  \\
TauAgent~\citep{gil2021online} &  & \cmark &  &  &  &  &  &  &  &  &  &  &  &  &  &  &  &  &  &  &  &  &  &  \\
Structured3D~\citep{zheng2020structured3d} &  & \cmark &  &  &  &  &  &  &  &  &  &  &  &  &  &  &  &  &  &  &  &  &  &  \\
DIML Outdoor~\citep{cho2021diml} &  & \cmark &  &  &  &  &  &  &  &  &  &  &  &  &  &  &  &  &  &  &  &  &  &  \\
DDAD~\citep{guizilini2020packnet} &  & \cmark &  &  &  &  &  &  &  &  &  &  &  &  &  &  &  &  &  &  &  &  &  &  \\
Argoverse~\citep{chang2019argoverse} &  & \cmark &  &  &  &  &  &  &  &  &  &  &  &  &  &  &  &  &  &  &  &  &  &  \\
Lyft~\citep{houston2021one} &  & \cmark &  &  &  &  &  &  &  &  &  &  &  &  &  &  &  &  &  &  &  &  &  &  \\
PandaSet~\citep{xiao2021pandaset} &  & \cmark &  &  &  &  &  &  &  &  &  &  &  &  &  &  &  &  &  &  &  &  &  &  \\
DSEC~\citep{gehrig2021dsec} &  & \cmark &  &  &  &  &  &  &  &  &  &  &  &  &  &  &  &  &  &  &  &  &  &  \\
Driving Stereo~\citep{yang2019drivingstereo} &  & \cmark &  &  &  &  &  &  &  &  &  &  &  &  &  &  &  &  &  &  &  &  &  &  \\
Cityscapes~\citep{cordts2016cityscapes} &  & \cmark &  &  &  &  &  &  &  &  &  &  &  &  &  &  &  &  &  &  &  &  &  &  \\
\midrule
\textbf{\#Datasets: total used (trained)} & \datasetcount{22}{11} & \datasetcount{42}{42} & \datasetcount{8}{8} & \datasetcount{10}{8} & \datasetcount{13}{13} & \datasetcount{14}{14} & \datasetcount{13}{13} & \datasetcount{27}{19} & \datasetcount{24}{14} & \datasetcount{17}{17} & \datasetcount{23}{15} & \datasetcount{32}{32} & \datasetcount{30}{30} & \datasetcount{19}{18} & \datasetcount{11}{4} & \datasetcount{20}{6} & \datasetcount{16}{14} & \datasetcount{9}{6} & \datasetcount{36}{29} & \datasetcount{21}{13} & \datasetcount{15}{12} & \datasetcount{32}{0} & \datasetcount{18}{18} & \datasetcount{29}{13} \\
\bottomrule
\end{tabular}
}
\label{tab:merged_dataset_usage_training_only}
\end{table*}


\begin{table*}[h]
\caption{\textbf{Dataset licenses and basic properties.}
We summarize the scene type, real/synthetic domain, dynamic content, and public license or access terms for each dataset.
``Non-Commercial'' denotes non-commercial use, ``Terms'' denotes dataset-specific access terms, and ``Not specified'' indicates that no clear standalone public dataset license was identified.}
\label{tab:dataset_profile_summary_cited}
\centering
\tiny
\setlength{\tabcolsep}{3.2pt}
\renewcommand{\arraystretch}{1.1}
\begin{tabularx}{\textwidth}{@{}c >{\centering\arraybackslash}X c c c >{\centering\arraybackslash}X@{}}
\toprule
\textbf{Idx.} & \textbf{Dataset} & \textbf{Scene Type} & \textbf{Real/Synth.} & \textbf{Dynamic} & \textbf{License / Access Terms} \\
\midrule
1  & CO3D~\cite{reizenstein2021co3d} & Object & Real & Static & CC BY-NC 4.0 \\
2  & RealEstate10K~\cite{zhou2018stereomagnification} & Indoor+Outdoor & Real & Static & CC BY 4.0 \\
3  & ScanNet~\cite{dai2017scannet} & Indoor & Real & Static & Non-Commercial \\
4  & ScanNet++~\cite{yeshwanth2023scannet++} & Indoor & Real & Static & Non-Commercial \\
5  & ARKitScenes~\cite{dehghan2021arkitscenes} & Indoor & Real & Static & Non-Commercial \\
6  & Habitat~\cite{savva2019habitat} & Indoor & Real & Static & Habitat dataset-specific terms \\
7  & BlendedMVS~\cite{yao2020blendedmvs} & Indoor+Outdoor & Synth. & Static & CC BY 4.0 \\
8  & MegaDepth~\cite{li2018megadepth} & Outdoor & Real & Static & CC BY 4.0 + source image licenses \\
9  & DL3DV~\cite{ling2024dl3dv} & Indoor+Outdoor & Real & Mostly Static & DL3DV Terms \\
10 & MapFree~\cite{arnold2022mapfree} & Outdoor & Real & Static & Non-Commercial \\
11 & WildRGBD~\cite{xia2024wildrgbd} & Object & Real & Static & MIT \\
12 & Waymo~\cite{sun2020waymo} & Outdoor & Real & Dynamic & Non-Commercial \\
13 & VirtualKITTI~\cite{cabon2020vkitti} & Outdoor & Synth. & Dynamic & CC BY-NC-SA 3.0 \\
14 & TartanAir~\cite{wang2020tartanair} & Indoor+Outdoor & Synth. & Dynamic & CC BY 4.0 \\
15 & Unreal4K~\cite{tosi2021unreal} & Indoor+Outdoor & Synth. & Static & MIT \\
16 & MVS-Synth~\cite{huang2018mvs} & Outdoor & Synth. & Mostly Static & Non-Commercial \\
17 & HyperSim~\cite{roberts2021hypersim} & Indoor & Synth. & Static & CC BY-SA 3.0 \\
18 & Mapillary~\cite{antequera2020mapillary} & Outdoor & Real & Static & CC BY-NC-SA \\
19 & ASE~\cite{avetisyan2024scenescript} & Indoor & Synth. & Static & Non-Commercial \\
20 & ADT~\cite{pan2023aria} & Indoor & Real & Dynamic & Non-Commercial \\
21 & PointOdyssey~\cite{zheng2023pointodyssey} & Indoor+Outdoor & Synth. & Dynamic & CC BY-NC-SA 4.0 \\
22 & Spring~\cite{mehl2023spring} & Outdoor & Synth. & Dynamic & CC BY 4.0 \\
23 & Dynamic Replica~\cite{Karaev_2023_dynamic_stereo} & Indoor & Synth. & Dynamic & Non-Commercial \\
24 & ParallelDomain-4D~\cite{vanhoorick2024gcd} & Outdoor & Synth. & Dynamic & CC BY-NC 4.0 \\
25 & SAIL-VOS 3D~\cite{hu2021sailvos3d} & Indoor+Outdoor & Synth. & Dynamic & Non-Commercial \\
26 & OmniObject3D~\cite{wu2023omniobject3d} & Object & Real & Static & CC BY 4.0 \\
27 & StaticThings3D~\cite{schroeppel2022rmvd} & Object & Synth. & Static & Non-Commercial \\
28 & TartanGround~\cite{patel2025tartanground} & Outdoor & Synth. & Static & CC BY 4.0 \\
29 & Matterport3D~\cite{chang2017matterport3d} & Indoor & Real & Static & Non-Commercial \\
30 & BEDLAM~\cite{black2023bedlam} & Human-centric & Synth. & Dynamic & Non-Commercial \\
31 & UASOL~\cite{bauer2019uasol} & Outdoor & Real & Static & CC BY-NC-SA 3.0 \\
32 & MVImgNet~\cite{yu2023mvimgnet} & Object & Real & Static & MVImgNet Terms \\
33 & CoP3D~\cite{sinha2023cop3d} & Object & Real & Dynamic & CC BY-NC 4.0 \\
34 & EDEN~\cite{le2021eden} & Outdoor & Synth. & Static & Research Use \\
35 & IRS~\cite{wang2021irs} & Indoor & Synth. & Static & Apache 2.0 \\
36 & Synscapes~\cite{wrenninge2018synscapes} & Outdoor & Synth. & Dynamic & Non-Commercial \\
37 & 3D Ken Burns~\cite{niklaus2019kenburns} & Indoor+Outdoor & Synth. & Static & CC BY-NC-SA 4.0 \\
38 & SmartPortraits~\cite{kornilova2022smartportraits} & Human-centric & Real & Dynamic & Academic Use \\
39 & UrbanSyn~\cite{gomez2025urbansyn} & Outdoor & Synth. & Dynamic & CC BY-SA 4.0 \\
40 & HOI4D~\cite{liu2022hoi4d} & Indoor & Real & Dynamic & CC BY-NC 4.0 \\
41 & Kubric~\cite{greff2022kubric} & Object & Synth. & Dynamic & Apache 2.0 \\
42 & Replica~\cite{straub2019replica} & Indoor & Real & Static & Research Use \\
43 & Objaverse~\cite{deitke2023objaverse} & Object & Synth. & Static & Per-object CC licenses \\
44 & TexVerse~\cite{zhang2025texverse} & Object & Synth. & Static & Per-object CC licenses \\
45 & GTA-SfM~\cite{wang2020flow} & Outdoor & Synth. & Static & Not specified \\
46 & SceneNet RGB-D~\cite{mccormac2016scenenet} & Indoor & Synth. & Static & Non-Commercial \\
47 & MatrixCity~\cite{li2023matrixcity} & Outdoor & Synth. & Static & CC BY-NC 4.0 \\
48 & Mid-Air~\cite{fonder2019midair} & Outdoor & Synth. & Static & CC BY-NC-SA 4.0 \\
49 & KITTI-360~\cite{liao2021kitti360} & Outdoor & Real & Dynamic & CC BY-NC-SA 3.0 \\
50 & Gibson~\cite{xia2018gibson} & Indoor & Real & Static & Non-Commercial \\
51 & HM3D~\cite{ramakrishnan2021hm3d} & Indoor & Real & Static & Non-Commercial \\
52 & Taskonomy~\cite{zamir2018taskonomy} & Indoor & Real & Static & Non-Commercial \\
53 & MegaSynth~\cite{jiang2025megasynth} & Indoor & Synth. & Static & CC BY-NC-SA 4.0 \\
54 & OmniWorld~\cite{zhou2025omniworld} & Outdoor & Synth. & Dynamic & CC BY-NC-SA 4.0 \\
55 & Trellis~\cite{xiang2025structured} & Object & Synth. & Static & MIT \\
56 & TauAgent~\cite{gil2021online} & Indoor+Outdoor & Synth. & Dynamic & Not specified \\
57 & Structured3D~\cite{zheng2020structured3d} & Indoor & Synth. & Static & Structured3D Terms \\
58 & DIML Outdoor~\cite{cho2021diml} & Outdoor & Real & Mostly Static & Non-Commercial \\
59 & DDAD~\cite{guizilini2020packnet} & Outdoor & Real & Dynamic & CC BY-NC-SA 4.0 \\
60 & Argoverse~\cite{chang2019argoverse} & Outdoor & Real & Dynamic & CC BY-NC-SA 4.0 \\
61 & Lyft~\cite{houston2021one} & Outdoor & Real & Dynamic & CC BY-NC-SA 4.0 \\
62 & PandaSet~\cite{xiao2021pandaset} & Outdoor & Real & Dynamic & CC BY 4.0 \\
63 & DSEC~\cite{gehrig2021dsec} & Outdoor & Real & Dynamic & CC BY-SA 4.0 \\
64 & Driving Stereo~\cite{yang2019drivingstereo} & Outdoor & Real & Dynamic & MIT \\
65 & Cityscapes~\cite{cordts2016cityscapes} & Outdoor & Real & Dynamic & Non-Commercial \\
\bottomrule
\end{tabularx}
\end{table*}

\clearpage
\section{The Collection of \dataset}
\label{appendix:datasets}

\begin{wraptable}{r}{0.60\textwidth}
  \vspace{-4pt}
  \footnotesize
  \renewcommand{\arraystretch}{1.03}
  \setlength{\tabcolsep}{2pt}
  \centering
  \caption{\textbf{\dataset dataset statistics.}}
  \label{egoexpert-table}
  \begin{tabularx}{\linewidth}{@{}
    >{\centering\arraybackslash}p{0.42cm}
    >{\raggedright\arraybackslash}X
    >{\centering\arraybackslash}p{0.78cm}
    >{\centering\arraybackslash}p{0.72cm}
    >{\raggedleft\arraybackslash}p{0.76cm}
    >{\raggedleft\arraybackslash}p{0.92cm}
  @{}}
  \toprule
  \textbf{Id} & \textbf{Dataset} & \textbf{View} & \textbf{R/S} & \textbf{Fr.} & \textbf{Sc.} \\ \midrule
  1 & \textbf{Xperience~\citep{xperience_10m}}          & Ego   & Real & 400K & 100    \\
  2 & \textbf{Aria Digital Twin}~\citep{pan2023aria}    & Ego   & Syn. & 86K  & 232    \\
  3 & \textbf{Colosseum}~\citep{pumacay2024colosseum}   & Wrist & Syn. & 334K & 1,837  \\
  4 & \textbf{HOI4D}~\citep{liu2022hoi4d}               & Ego   & Real & 739K & 2,466  \\
  5 & \textbf{RLBench}~\citep{james2020rlbench}         & Wrist & Syn. & 1.2M & 5,120  \\
  6 & \textbf{Robolab}~\citep{yang2026robolab}          & Wrist & Syn. & 158K & 132    \\
  7 & \textbf{RoboTwin}~\citep{chen2025robotwin}        & Wrist & Syn. & 2.6M & 11,923 \\
  \bottomrule
  \end{tabularx}
  \vspace{-6pt}
\end{wraptable}

This section details the data collection pipeline of \dataset.
\dataset consists of 22K sequences with 5.5M frames in total.
Each scene in the dataset contains an image sequence $\mathbf{I}\in\mathbb{R}^{ N\times H\times W\times 3}$, depth maps $\mathbf{D}\in\mathbb{R}^{H\times W\times N}$, camera intrinsics $\mathbf{K}\in\mathbb{R}^{3\times 3}$, and camera extrinsics $\mathbf{G}_{\text{c2w}}\in\mathrm{SE}(3)$ for each frame.
Fig.~\ref{Fig.egoexpert} showcases data samples from \dataset.
Note that all evaluation scenes in \benchmark are held out from the training splits of their respective datasets.

\textbf{Aria Digital Twin (ADT).}
ADT~\citep{pan2023aria} is an egocentric RGB-D dataset captured with Meta Project Aria smart glasses in two instrumented indoor environments.
It comprises 200 sequences of natural daily activities performed by 9 participants interacting with 398 object instances (344 static, 74 dynamic). 
We process the raw data into 232 sequences with 86K frames in total at a resolution of $512 \times 512$ using official preprocessing scripts. 
We observe scattered outlier points along object boundaries in scene visualizations and apply a unified depth post-processing step to refine the depth maps accordingly.

\textbf{HOI4D.}
HOI4D~\citep{liu2022hoi4d} is a large-scale 4D egocentric dataset for category-level human-object interaction, containing 2.4M RGB-D frames across 4,000 sequences collected by 9 participants interacting with 800 object instances from 16 categories in 610 distinct indoor rooms.
Each frame provides RGB and depth from a head-mounted RGBD camera with a resolution of $1920 \times 1080$, together with camera 6-DoF poses, 3D hand poses, category-level object poses, panoptic segmentation, and reconstructed scene point clouds.
We directly use a subset of the raw HOI4D data (2.5K scenes and 739K frames) with only depth post-processing filtering (Appendix~\ref{appendix:DepthPost-Processing}) applied.

\textbf{RLBench.}
RLBench~\citep{james2020rlbench} is a large-scale robot learning benchmark and simulation environment built on CoppeliaSim~\citep{rohmer2013CoppeliaSim}, featuring more than 100 distinct hand-designed manipulation tasks ranging from simple reaching to long-horizon multi-stage operations.
Each task provides an unlimited supply of expert demonstrations via built-in motion planners. 
Visual observations include RGB, depth, and segmentation from both an over-the-shoulder stereo camera and an eye-in-hand monocular camera, with pixel-accurate synthetic depth and ground-truth 6-DoF camera poses available at every frame.
Since the original simulation's wrist-view camera does not capture the gripper, we adjust the wrist camera pose relative to the end-effector and increase the resolution to $1280 \times 720$ to broaden the field of view.
We collect around 50 episodes per task using motion planners with random seeds across 103 tasks to form our dataset, with 1.2M frames in total.

\textbf{Robo Colosseum.}
The Robo Colosseum~\citep{pumacay2024colosseum} is a robotic manipulation benchmark built on top of RLBench that introduces systematic environmental perturbations for evaluating generalization.
It provides 20 diverse manipulation tasks evaluated across 14 perturbation axes, including object color, texture, and size, background and tabletop appearance, lighting conditions, number of distractors, physical properties, and camera pose, with each perturbation applied independently and in combination.
RGB, depth, and camera pose annotations are identical in format to RLBench.
We similarly adjust the initial wrist camera placement and set the rendering resolution 
to $1280 \times 720$. 
We collect 100 episodes per task with random seeds across 19 tasks, which is 1.8K episodes with 334K frames.
Domain randomization is configured as follows: the number of distractor objects is set to 5--8; camera pose perturbations follow  \texttt{euler\_range}~$= [[-0.05,-0.05,-0.05],\,[0.05,0.05,0.05]]$ and \texttt{position\_range}~$= [[0,0,0],\,[0,0,0.3]]$; surface colors (\emph{e.g.}, table and objects) are randomized within \texttt{color\_range}~$= [[0.25,0.25,0.25],\,[1.0,1.0,1.0]]$; lighting color is randomized within $[[0,0,0],\,[0.5,0.5,0.5]]$; object scale is perturbed in the range $[0.8,\,1.0]$; and texture randomization is enabled for objects, the table surface, and the background.

\textbf{RoboTwin.}
RoboTwin~\citep{chen2025robotwin} is a large-scale simulation-based dataset and benchmark for bimanual robotic manipulation, built on top of the RoboTwin Object Dataset (RoboTwin-OD) comprising 731 object instances across 147 categories with rich manipulation annotations (grasp points, functional points, object axes) and diverse language descriptions.
It spans 50 dual-arm collaborative manipulation tasks executed across 5 heterogeneous robot embodiments.
For \dataset, we collect data using five robot embodiments: Aloha-AgileX, ARX-X5, Franka, Piper, and UR5 across 50 bimanual tasks, which is 11K scenes with 2.6M frames in total.
The wrist camera resolution is set to $1280 \times 720$ with a field of view of $60^\circ$, and approximately 25 episodes are collected per task. 
Domain randomization is enabled with random backgrounds, cluttered table setups, a clean background rate of 0.2, random table height variation of 0.1, and random lighting. 
The wrist camera position is randomly switched between two fixed placements across episodes. For each episode, the image sequence from either the left or right arm is randomly selected for collection.
The RoboTwin dataset has 12K episodes, comprising 2.6M frames in total.

\textbf{Robolab.}
Robolab~\citep{yang2026robolab} is a high-fidelity simulation benchmarking framework for task-generalist robotic policies, developed by NVIDIA and built on Isaac Sim.
It introduces the Robolab-120 benchmark comprising 120 tasks organized along three competency axes: \textit{visual}, \textit{procedural}, and \textit{relational}, each at three difficulty levels, with LLM-enabled generation of novel scenes and tasks.
Scenes feature photorealistic assets and physically accurate simulation, with ground-truth RGB, depth, and 6-DoF camera poses from wrist-mounted and external cameras.
For \dataset, we use $\pi_{0.5}$~\citep{intelligence2025pi05} as the trajectory generator to collect data across 108 tasks under 2 background settings, yielding a total of 158K frames at a resolution of $1280 \times 720$.

\section{Detail of \ours}
\label{appendix:dan}
\subsection{Model Architecture}
\ours is built upon the Giant variant of Depth-Anything-3~\citep{lin2025depth}.
The backbone adopts the DINOv2~\citep{oquab2023dinov2} ViT-Giant (\texttt{vitg}) architecture, in which Alternating Attention, QK-Norm, and Rotary Position Embedding (RoPE) are enabled starting from the 13th layer, while the \texttt{cat\_token} and \texttt{scale\_token} are retained. 
Multi-scale features are extracted from the 19th, 27th, 33rd, and 39th layers. 
The depth and ray prediction heads are instantiated as a DualDPT module with an input dimension of $3072$, a feature dimension of $256$, output channels of $[256, 512, 1024, 1024]$, and an output dimension of $2$.
The Scale Head is implemented as a 3-layer MLP (input $1536$, hidden $1024$, ReLU activation, Softplus output), and the Camera Encoder has an output dimension of $1536$. 
All weights are initialized from the officially released \texttt{da3-giant-1.1} checkpoint.

\subsection{Training Objective}
In \ours, all input images $\mathbf{I} = \{I_i\}^N_{i=1}$, together with the available camera parameters $\mathbf{C} = \{C_i\}^N_{i=1}=\{K_i,G_i\}^N_{i=1}$ (if provided), are fed into the network $\mathcal{G}$, which predicts the complete ray maps $\hat{\mathbf{R}}$, depth maps $\hat{\mathbf{D}}$, scale $\hat{S}$, and their corresponding confidence maps $\hat{\mathbf{Y}}_d, \hat{\mathbf{Y}}_r$ in an end-to-end manner:
\begin{equation}
\mathcal{G}\left(\mathbf{I},\mathbf{C}\right)=(\hat{\mathbf{R}},\hat{\mathbf{D}}, \hat{\mathbf{Y}}_d,\hat{\mathbf{Y}}_r,\hat{S}).
\end{equation}
The total training objective is a weighted sum of five task-specific $\ell_1$ terms,
\begin{equation}
\mathcal{L} = \mathcal{L}_{\text{depth}} + \alpha\,\mathcal{L}_{\text{grad}} + \mathcal{L}_{\text{ray}} + \mathcal{L}_{\text{pmap}} + \mathcal{L}_{\text{scale}},
\label{eq:total_loss}
\end{equation}
with $\alpha=1$.
Let $m_p\!\in\!\{0,1\}$ denote the per-pixel validity mask of the ground truth and $\Omega=\{p\mid m_p=1\}$ the set of valid pixels across all $N$ views, with $|\Omega|$ its cardinality. We detail each term below.

\noindent\textbf{Confidence-weighted regression.}
Following~\citet{wang2024dust3r,leroy2024mast3r}, each dense prediction is supervised with a confidence-weighted $\ell_1$ regression term. 
Given a prediction $\hat{x}$, its confidence $\hat{y}$, and the ground truth $x$, the per-pixel regression and confidence terms are
\begin{equation}
\ell_{\text{reg}}(\hat{x},x\,;p) = \| \hat{x}_p-x_p \|_1,\qquad
\ell_{\text{conf}}(\hat{x},x,\hat{y}\,;p) = \gamma\,\hat{y}_p\,\ell_{\text{reg}}(\hat{x},x\,;p) - \beta\log \hat{y}_p,
\label{eq:conf_reg}
\end{equation}
where $\gamma$ and $\beta$ balance the data term and the confidence regularizer that prevents $\hat{y}\!\to\!0$.

\noindent\textbf{Depth loss.}
For the predicted depth map $\hat{\mathbf{D}}$ with confidence $\hat{\mathbf{Y}}_d$, the depth term aggregates the confidence-weighted and plain $\ell_1$ terms over valid pixels:
\begin{equation}
\mathcal{L}_{\text{depth}}(\hat{\mathbf{D}},\mathbf{D}\,;\hat{\mathbf{Y}}_d)
= \tfrac{1}{|\Omega|}\sum_{p\in\Omega}\!\Big(
\gamma\,\hat{Y}_{d,p}\,\big|\hat{D}_p - D_p\big| - \beta\log \hat{Y}_{d,p} + \big|\hat{D}_p - D_p\big|
\Big).
\end{equation}

\noindent\textbf{Depth-gradient loss.}
To preserve sharp edges while enforcing smoothness on planar regions, we penalize the discrepancy between the depth gradients of the prediction and the ground truth:
\begin{equation}
\mathcal{L}_{\text{grad}}(\hat{\mathbf{D}},\mathbf{D}) = \|\nabla_x \hat{\mathbf{D}} - \nabla_x \mathbf{D}\|_1 + \|\nabla_y \hat{\mathbf{D}} - \nabla_y \mathbf{D}\|_1,
\end{equation}
where $\nabla_x$ and $\nabla_y$ are the horizontal and vertical finite difference operators. In practice, this term is evaluated in a multi-scale manner over $J$ dyadic scales (stride $2^{j}$, $j=0,\dots,J\!-\!1$) and averaged.

\noindent\textbf{Ray loss.}
The ray map $\hat{\mathbf{R}}\in\mathbb{R}^{N\times H\times W\times 6}$ encodes per-pixel camera origins $\hat{\mathbf{o}}$ and viewing directions $\hat{\mathbf{d}}$. With its confidence $\hat{\mathbf{Y}}_r$, the ray loss mirrors the depth term:
\begin{equation}
\mathcal{L}_{\text{ray}}(\hat{\mathbf{R}},\mathbf{R}\,;\hat{\mathbf{Y}}_r)
= \tfrac{1}{|\Omega|}\sum_{p\in\Omega}\!\Big(
\gamma\,\hat{Y}_{r,p}\,\|\hat{\mathbf{R}}_p - \mathbf{R}_p\|_1 - \beta\log \hat{Y}_{r,p} + \|\hat{\mathbf{R}}_p - \mathbf{R}_p\|_1
\Big).
\end{equation}

\noindent\textbf{Point-map loss.}
We recover the 3D world points by back-projecting depth along the predicted rays, $\hat{\mathbf{P}} = \hat{\mathbf{D}}\odot\hat{\mathbf{d}} + \hat{\mathbf{o}}$, and supervise them with a masked $\ell_1$ against the ground-truth world points $\mathbf{P}$:
\begin{equation}
\mathcal{L}_{\text{pmap}}(\hat{\mathbf{D}}\odot\hat{\mathbf{d}}+\hat{\mathbf{o}},\,\mathbf{P}) = \frac{1}{|\Omega|}\sum_{p\in\Omega}\big\|\hat{\mathbf{P}}_p - \mathbf{P}_p \big\|_1.
\end{equation}

\noindent\textbf{Scale loss.}
The predicted global scale $\hat{S}$ is supervised against the ground-truth scale factor $S$ via a log-space $\ell_1$ term,
\begin{equation}
\mathcal{L}_{\text{scale}}(\hat{S},S) = \big\| f_{\log}(\hat{S}) - f_{\log}(S) \big\|_1,\qquad
f_{\log}:\mathbf{x}\mapsto \tfrac{\mathbf{x}}{\|\mathbf{x}\|}\,\log(1+\|\mathbf{x}\|),
\end{equation}
which compresses the dynamic range of absolute scale and yields stable gradients across scenes of drastically different sizes.

\noindent\textbf{Ground-truth preprocessing and scale target.}
To remove the inherent scene-scale ambiguity across heterogeneous training data and keep the magnitudes of different modalities consistent, all ground-truth signals are canonicalized in a two-step procedure before loss computation.
\emph{(i) Coordinate canonicalization.} The first frame is taken as the reference: we apply the \mbox{cam-to-world} transform of the first camera to all extrinsics, and accordingly transform the ground-truth world points $\mathbf{P}$ into the first-camera coordinate frame.
\emph{(ii) Scale normalization.} We then compute the per-scene scale factor as the mean $\ell_2$ norm of the valid reprojected world points,
\begin{equation}
S \;=\; \frac{1}{|\Omega|}\sum_{p\in\Omega} \|\mathbf{P}_p\|_2,
\label{eq:gt_scale}
\end{equation}
and divide the ground-truth world points, depths and camera translations by $S$:
\begin{equation}
\mathbf{P}\!\leftarrow\!\mathbf{P}/S,\qquad
\mathbf{D}\!\leftarrow\!\mathbf{D}/S,\qquad
\mathbf{t}\!\leftarrow\!\mathbf{t}/S.
\end{equation}
The resulting scalar $S$ is also kept as the regression target of the scale head, so that $\hat{S}$ in $\mathcal{L}_{\text{scale}}$ learns to recover the absolute metric scale that the other dense predictions are invariant to. During inference, the predicted $\hat{S}$ is multiplied back to lift the normalized geometry into its metric space.
We set $\gamma=1.0$ and $\beta=0.2$ for all confidence-weighted terms, use $J=3$ scales for the multi-scale gradient loss, and apply quantile filtering with $\tau=0.98$ to suppress outliers.

\subsection{Camera Conditioning Training}
\label{sec:pose_cond}
We adopt a \emph{stochastic pose-conditioning} scheme at training time: for each mini-batch we flip a Bernoulli coin with probability $p\in[0,1]$, and inject the ground-truth camera as an auxiliary input only when the coin turns up, i.e.\
\begin{equation}
u \sim \mathrm{Bernoulli}(p),\qquad
(\hat{\mathbf{R}},\hat{\mathbf{D}},\hat{\mathbf{Y}}_d,\hat{\mathbf{Y}}_r,\hat{S}) =
\begin{cases}
\mathcal{G}(\mathbf{I},\tilde{\mathbf{G}},\mathbf{K}), & u=1,\\[2pt]
\mathcal{G}(\mathbf{I}), & u=0,
\end{cases}
\end{equation}
where $\tilde{\mathbf{G}}$ denotes the canonicalized ground-truth extrinsics described below, and $\mathbf{K}=\{K_i\}_{i=1}^{N}$ the intrinsics. When $u=0$, the encoder instead receives a learnable placeholder camera token $\mathbf{e}_c$; when $u=1$, $(\tilde{\mathbf{G}},\mathbf{K})$ is tokenized and replaces $\mathbf{e}_c$, so that the encoder attends to geometrically grounded pose cues.

\noindent\textbf{Extrinsic Normalization for Conditioning.}
The extrinsics fed into the network are preprocessed independently from the supervision pipeline of Eq.~\eqref{eq:gt_scale}, so as to remain agnostic to the absolute metric scale. Specifically, given the world-to-camera matrices $\{G_i\}_{i=1}^{N}$, we (i) right-multiply by $G_1^{-1}$ so that the first view becomes the canonical frame, and (ii) divide the translation components by the median camera-centre distance $\bar{d}$:
\begin{equation}
\tilde{G}_i \;=\; G_i\,G_1^{-1},\qquad
\tilde{G}_i[{:}3,3] \;\leftarrow\; \tilde{G}_i[{:}3,3]\,/\,\max(\bar{d},\,\epsilon),
\end{equation}
with $\bar{d}=\mathrm{median}_i\,\|\mathbf{c}_i\|_2$ and $\mathbf{c}_i$ the camera centre extracted from $\tilde{G}_i^{-1}$. This keeps the conditioning signal in a scale-invariant canonical frame and prevents trivial leakage of the ground-truth scale through the pose input (which is instead supervised via $\mathcal{L}_{\text{scale}}$).

\subsection{Training Datasets}

\begin{table*}[t]
\caption{\textbf{Training Datasets Statistics.} Each training epoch is composed of 18 datasets, and their dataset mixture ratios are reported as the training prob.}
\label{dataset-table}
\small
\renewcommand{\arraystretch}{0.98}
\centering
\setlength{\tabcolsep}{2pt}
\begin{tabularx}{0.96\textwidth}{@{}
  >{\centering\arraybackslash}p{0.78cm}
  >{\centering\arraybackslash}p{0.52cm}
  >{\raggedright\arraybackslash}X
  >{\centering\arraybackslash}p{1.18cm}
  >{\centering\arraybackslash}p{1.28cm}
  >{\centering\arraybackslash}p{1.08cm}
  >{\raggedleft\arraybackslash}p{0.86cm}
  >{\centering\arraybackslash}p{0.78cm}
  >{\centering\arraybackslash}p{0.98cm}
@{}}
\toprule
\textbf{Cat.} & \textbf{Id.} & \textbf{Dataset} & \textbf{Scene} & \textbf{R/S} & \textbf{Dyn.} & \textbf{\# Fr.} & \textbf{Met.} & \textbf{Prob.} \\ \midrule
\multirow{11}{*}{\rotatebox{90}{\textbf{General 3D Datasets}}}
& 1  & \textbf{HyperSim}~\citep{roberts2021hypersim}         & Indoor  & Syn. & Stat. & 70K  & \cmark & 2.8 \\
& 2  & \textbf{Infinigen}~\citep{raistrick2024infinigen}    & Indoor  & Syn. & Stat. & 142K & \cmark & 2.5 \\
& 3  & \textbf{MapFree}~\citep{arnold2022mapfree}           & Outdoor & Real & Stat. & 2.6M & \cmark & 6.5 \\
& 4  & \textbf{Matterport 3D}~\citep{ramakrishnan2021hm3d}  & Indoor  & Real & Stat. & 1.9M & \cmark & 7.3 \\
& 5  & \textbf{MVS-Synth}~\citep{huang2018mvs}              & Outdoor & Syn. & Stat. & 12K  & \cmark & 0.4 \\
& 6  & \textbf{ScanNet++}~\citep{yeshwanth2023scannet++}    & Indoor  & Real & Stat. & 7.8M & \cmark & 4.1 \\
& 7  & \textbf{Spring}~\citep{mehl2023spring}               & Outdoor & Syn. & Dyn.  & 5K   & \cmark & 2.3 \\
& 8  & \textbf{Tartanair}~\citep{wang2020tartanair}         & Mixed   & Syn. & Stat. & 3M   & \cmark & 6.8 \\
& 9  & \textbf{Unreal 4K}~\citep{tosi2021unreal}           & Outdoor & Syn. & Stat. & 16K  & \cmark & 0.4 \\
& 10 & \textbf{Vkitti}~\citep{cabon2020vkitti}             & Outdoor & Syn. & Dyn.  & 42K  & \cmark & 1.5 \\
& 11 & \textbf{Waymo}~\citep{sun2020waymo}                 & Outdoor & Real & Dyn.  & 7.9M & \cmark & 5.1 \\ \midrule
\multirow{6}{*}{\rotatebox{90}{\textbf{\dataset}}}
& 12 & \textbf{Aria Digital Twin}~\citep{pan2023aria}      & Indoor  & Syn. & Dyn. & 87K  & \cmark & 8.0 \\
& 13 & \textbf{Colosseum}~\citep{pumacay2024colosseum}     & Indoor  & Syn. & Dyn. & 334K & \cmark & 5.5  \\
& 14 & \textbf{HOI4D}~\citep{liu2022hoi4d}                 & Indoor  & Real & Dyn. & 739K & \cmark & 11.1 \\
& 15 & \textbf{RLBench}~\citep{james2020rlbench}           & Indoor  & Syn. & Dyn. & 1.2M & \cmark & 5.5  \\
& 16 & \textbf{Robolab}~\citep{yang2026robolab}            & Indoor  & Syn. & Dyn. & 158K & \cmark & 11.1 \\
& 17 & \textbf{RoboTwin}~\citep{chen2025robotwin}          & Indoor  & Syn. & Dyn. & 1.8M & \cmark & 11.1 \\
& 18 & \textbf{Xperience}~\citep{xperience_10m}          & Mixed  & Real & Dyn. & 400K & \cmark & 8.0 \\
\bottomrule
\end{tabularx}
\end{table*}

We train \ours model using images from 18 datasets, including: \dataset (7 datasets in total), HyperSim~\citep{roberts2021hypersim}, Infinigen~\citep{raistrick2024infinigen}, Spring~\citep{mehl2023spring}, MapFree~\citep{arnold2022mapfree}, Matterport 3D~\citep{ramakrishnan2021hm3d}, MVS-Synth~\citep{huang2018mvs},  ScanNet++~\citep{yeshwanth2023scannet++}, TartanAir~\citep{wang2020tartanair}, Unreal 4K~\citep{tosi2021unreal}, Virtual KITTI~\citep{cabon2020vkitti}, Waymo~\citep{sun2020waymo}.
These datasets cover normal, egocentric, and wrist view perspectives, both synthetic and real-world content, indoor and outdoor environments, as well as static and dynamic scenes. 
Such a diverse composition preserves a strong generalization capability for \ours.
Table~\ref{dataset-table} summarizes the statistics of the datasets we used. 
In each epoch, we sample a fixed total number of samples from the training datasets, with their mixture ratio indicated by the ``Training Prob'' column in the table.

\subsection{Frame Sampling Strategy} For every batch, we select between 2 and 18 frames from multiple random training scenes while maintaining a constant total of 18 frames within each batch.
We sample each batch of images based on the Euclidean distance to the camera pose. 
For each frame, all other frames are ranked according to their pose distance, and the top $N$ ($N$ depends on the frame density in each dataset) closest frames are selected as its valid range. 
Then, for each sequence, we randomly choose one frame as the anchor frame and sample the remaining frames from its valid range.

\subsection{Training Configuration}
Our \ours architecture follows DA3-Giant~\citep{lin2025depth} with $L = 40$ Transformer blocks and is initialized by pre-trained weights.
During training, each batch incorporates ground-truth camera information as auxiliary input with probability $p=20\%$.
The training runs end-to-end on 4 NVIDIA H200 GPUs over seven days.
Training is conducted in BF16 mixed precision with a fixed random seed of $42$ for $10$ epochs in total. 
The per-device batch size is $18$ frames, and gradient accumulation is performed every $2$ steps.
During training, the backbone, depth and ray prediction head, camera encoder, and scale head are jointly optimized, whereas the camera decoder is kept frozen (we use the ray map branch).
Since the 3D Gaussian Splatting branch is not involved in this work, its corresponding GS Head is also frozen.
Gradient checkpointing is further enabled to reduce GPU memory consumption.
We employ the AdamW optimizer with $\beta_1 = 0.9$, $\beta_2 = 0.95$, $\epsilon = 1 \times 10^{-8}$, and a weight decay of $0.01$. A layer-wise learning rate scheme is adopted: $1 \times 10^{-5}$ for the backbone, $2 \times 10^{-5}$ for the prediction head, camera encoder. 
The learning rate follows a cosine decay schedule with linear warm-up, where the warm-up stage lasts $5{,}000$ steps and the minimum learning rate factor is $0.1$. Gradient clipping is applied with a maximum norm of $1.0$.
To enhance robustness across varying aspect ratios, multi-scale training is performed
over $17$ resolutions, where the width is fixed at $504$ and the height ranges
from $280$ to $504$ with a step of $14$.

\section{Additional Findings}
\label{appendix:other_findings}

\subsection{Do More Input Frames Always Lead to Better Results?}
A natural assumption in 3D reconstruction is that more input frames yield better geometric estimates. 
However, Tab.~\ref{tab:main_leaderboard_filtered} reveals a more nuanced picture.
Under the sparse regime, where inputs consist of only a few frames with large inter-frame baselines and limited overlap, most models struggle with geometric estimation due to insufficient visual correspondence.
Increasing the number of input frames from sparse to medium consistently improves both depth and camera pose metrics across paradigms, confirming that a moderate level of multi-view overlap is beneficial.
However, further extending to the dense regime does not uniformly bring additional gains. 
For many feed-forward models, dense inputs introduce redundant or near-duplicate frames that add memory pressure without providing useful new geometric constraints, leading to stagnating or even degraded performance on reconstruction.
This non-monotonic behavior suggests that the relationship between input density and reconstruction quality is task- and model-dependent, rather than strictly increasing.
\observationbox{
\textbf{Takeaway}: \textit{For bounded 3D reconstruction tasks, there exists an optimal input density range that maximizes reconstruction quality. Too few frames leave geometric constraints underspecified, while excessive inputs can introduce redundancy and hurt performance, making careful input density selection as important as model choice.}
}

\begin{figure}[t] 
\centering 
\includegraphics[width=0.9\columnwidth]{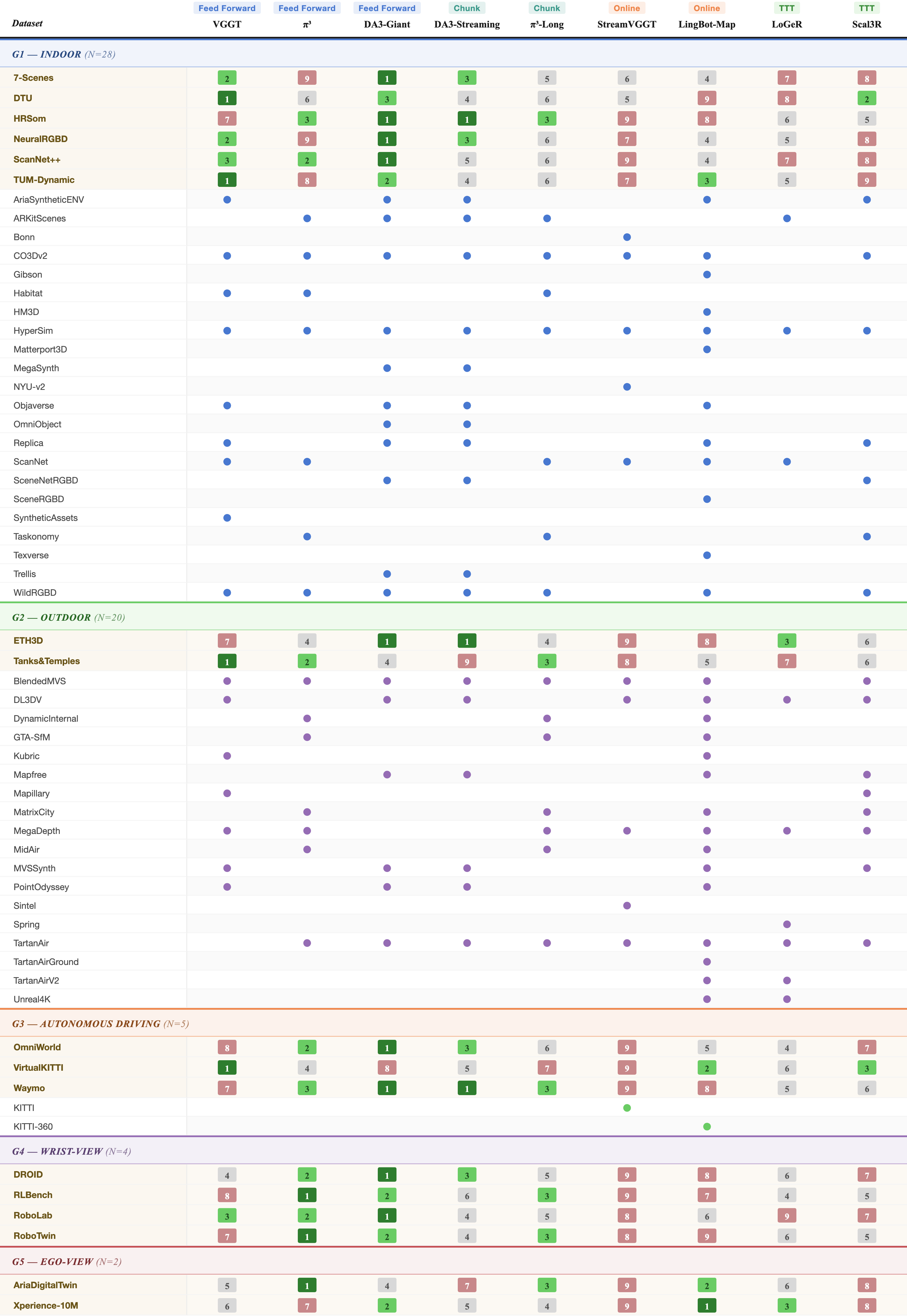}
\caption{\textbf{Training Data Domain Coverage vs.\ Overall Ranking Across Domain Groups.}
We compare the best-performing methods from each paradigm on five domain groups in \benchmark: \textit{Indoor}, \textit{Outdoor}, \textit{Autonomous Driving}, \textit{Wrist-view}, and \textit{Ego-view}, where $N$ denotes the number of datasets in each group.
Colored cells report the per-domain ranking of each method, while scatter points indicate whether the corresponding training set includes in-domain data from that group.
}
\vspace{-4pt}
\label{Fig.corelation_average} %
\end{figure}

\begin{figure}[t] 
\centering 
\includegraphics[width=0.9\columnwidth]{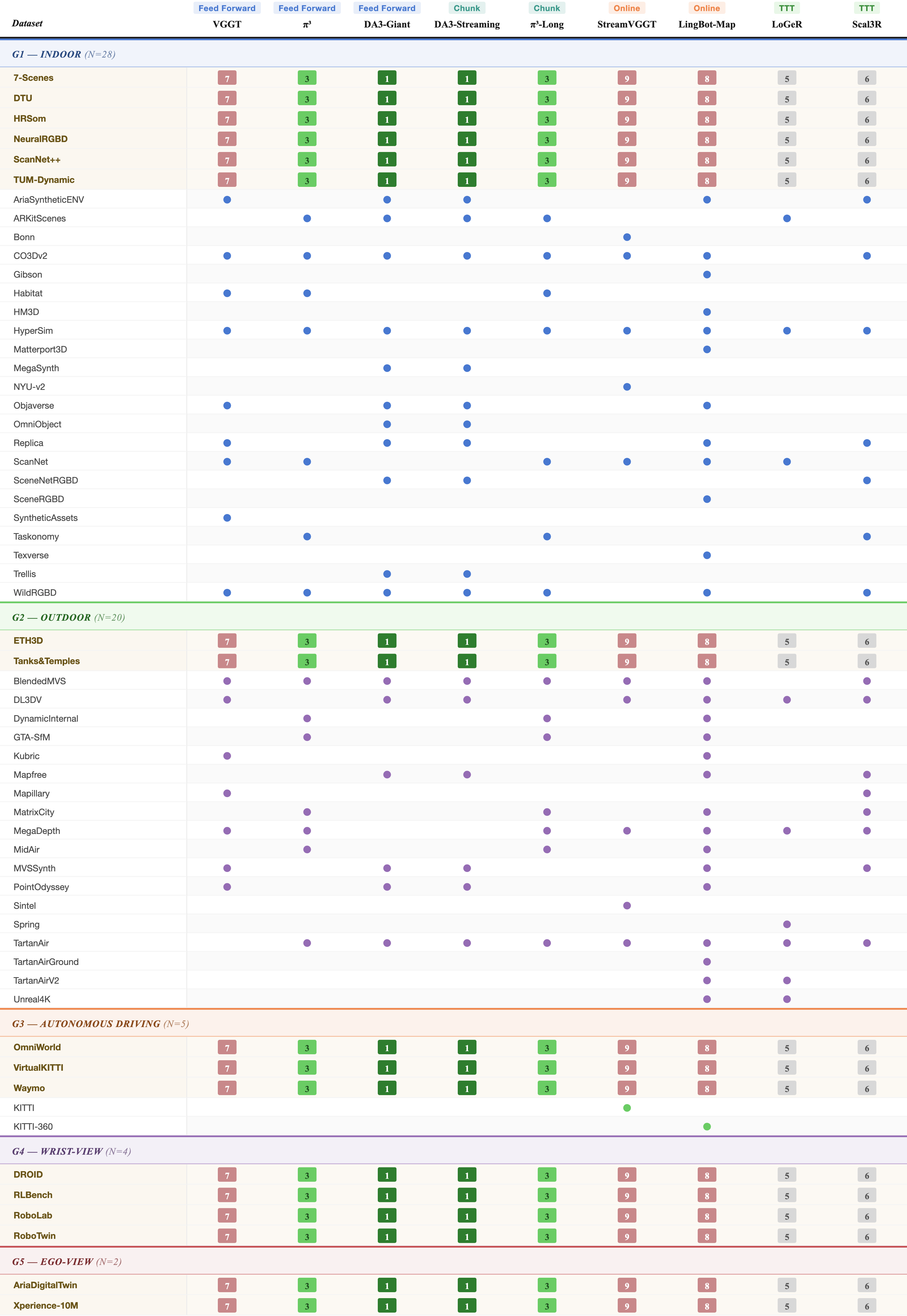}
\caption{\textbf{Training Data Domain Coverage vs.\ Overall Ranking Across Domain Groups (Sparse).}
We compare the best-performing methods from each paradigm on five domain groups in \benchmark under the sparse regime, where $N$ denotes the number of datasets in each group.
Colored cells report the per-domain ranking of each method, while scatter points indicate whether the corresponding training set includes in-domain data from that group.
}
\vspace{-4pt}
\label{Fig.corelation_sparse} %
\end{figure}

\begin{figure}[t] 
\centering 
\includegraphics[width=0.9\columnwidth]{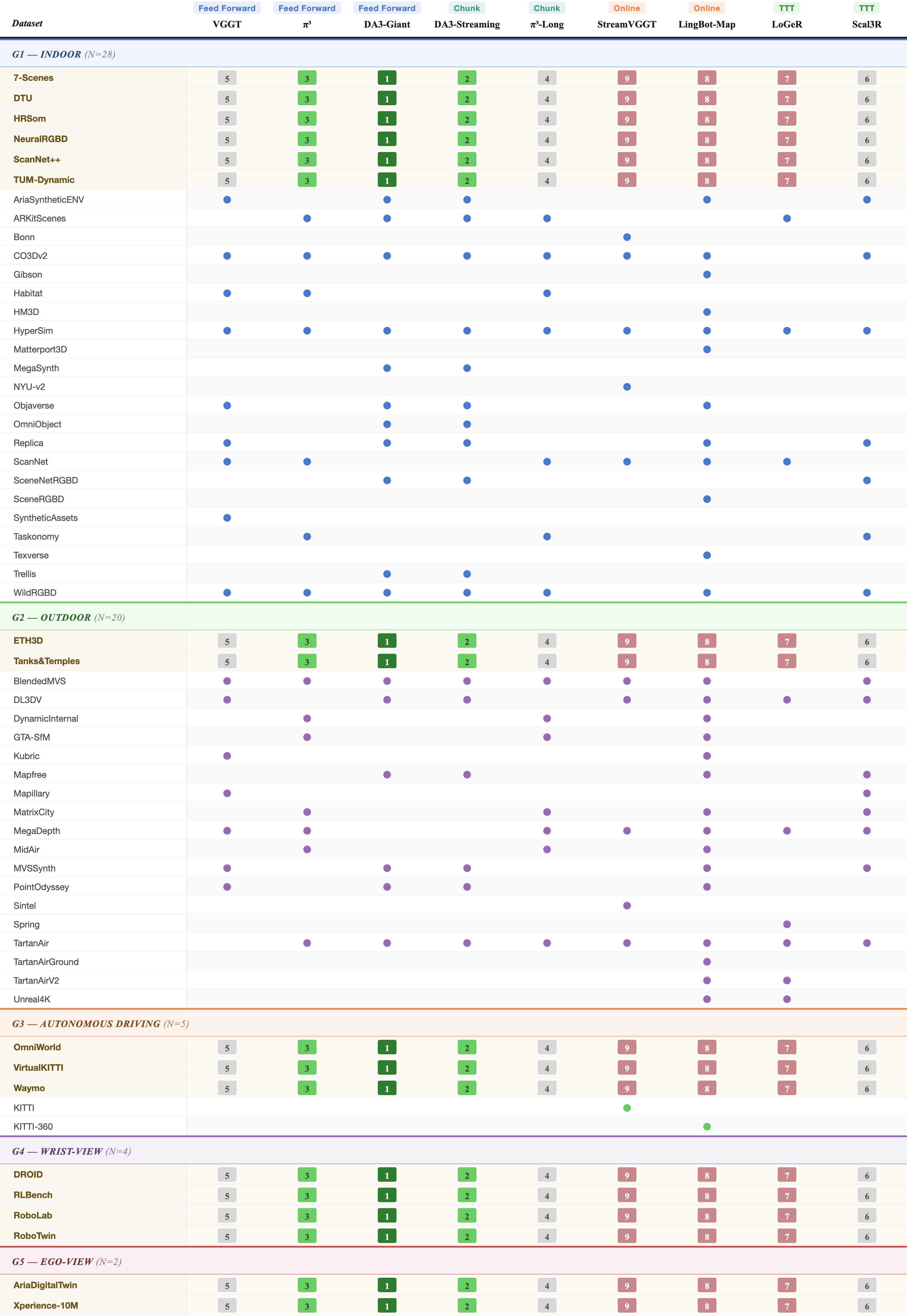}
\caption{\textbf{Training Data Domain Coverage vs.\ Overall Ranking Across Domain Groups (Medium).}
We compare the best-performing methods from each paradigm on five domain groups in \benchmark under the medium regime, where $N$ denotes the number of datasets in each group.
Colored cells report the per-domain ranking of each method, while scatter points indicate whether the corresponding training set includes in-domain data from that group.
}
\vspace{-4pt}
\label{Fig.corelation_medium} %
\end{figure}
\begin{figure}[t] 
\centering 
\includegraphics[width=0.9\columnwidth]{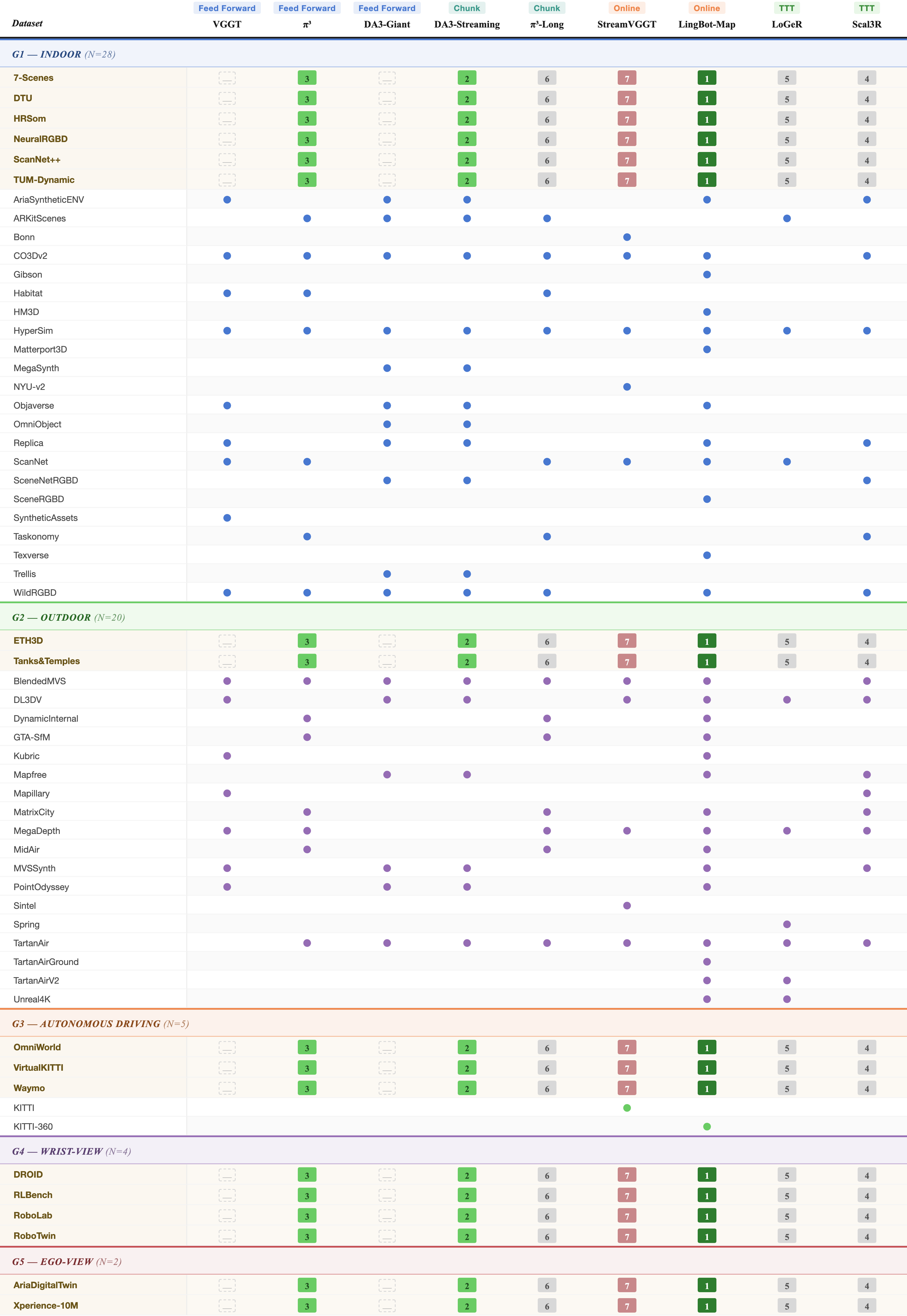}
\caption{\textbf{Training Data Domain Coverage vs.\ Overall Ranking Across Domain Groups (Dense).}
We compare the best-performing methods from each paradigm on five domain groups in \benchmark under the dense regime, where 
$N$ denotes the number of datasets in each group.
Colored cells report the per-domain ranking of each method, while scatter points indicate whether the corresponding training set includes in-domain data from that group.
Empty cells indicate that the method runs out of memory under the corresponding input regime.
}
\vspace{-4pt}
\label{Fig.corelation_dense} %
\end{figure}

\subsection{How to Select the Right Spatial Foundation Model for Your Task?}
Fig.~\ref{Fig.corelation_average} presents the overall ranking of the best-performing methods from each paradigm (Feed-Forward, Chunk-wise, Online, and TTT) across different domain subsets of \benchmark.
The overall ranking is computed from the weighted Average column of each per-domain sub-table, aggregating four complementary metrics: AbsRel, AUC@30, ATE, and F-Score.
We partition \benchmark into five domain groups: G1: Indoor, G2: Outdoor, G3: Driving, G4: Wrist-view, and G5: Ego-view.
The colored cells report the ranking of each method on each domain, while the scatter points indicate whether the method's training data includes in-domain samples from the corresponding group.
Fig.~\ref{Fig.corelation_sparse}, Fig.~\ref{Fig.corelation_medium}, and Fig.~\ref{Fig.corelation_dense} present the per-domain ranking breakdowns under the sparse, medium, and dense input settings, respectively.

Fig.~\ref{Fig.corelation_average} reveals that model rankings are highly 
inconsistent across domain groups and input density settings.
This inconsistency stems from two factors: the training data coverage of each method, and their architectural design choices that favor different operating regimes (\emph{e.g.}, long-sequence streaming methods naturally excel on dense, 
extended trajectories but may underperform on sparse multi-view inputs).
This finding gives a guideline for downstream deployment: selecting a spatial foundation model for a target application requires verifying not only its general benchmark performance, but also whether its training mixture covers the relevant domain.
Furthermore, for practitioners preparing fine-tuning datasets, these results suggest that domain match matters more than dataset volume: prioritizing data from the target domain, even in modest quantities, is likely to yield greater gains than simply increasing the total number of training scenes from unrelated distributions.
\observationbox{
\textbf{Takeaway}: \textit{No single model dominates across all domains and input regimes. Domain coverage in training data is the primary predictor of per-domain ranking, and should be the first consideration when selecting or fine-tuning a spatial foundation model for a specific downstream application.}
}
\clearpage

\section{The Complete \benchmark Results}
\label{appendix:completeresults}

\textbf{Per-regime Breakdowns of the Main Table.} We present detailed per-regime results for the single-frame, sparse, medium, and dense input settings in Tab.~\ref{tab:sub_single}, Tab.~\ref{tab:sub_sparse}, Tab.~\ref{tab:sub_medium}, and Tab.~\ref{tab:sub_dense}, respectively, complementing the aggregated results in Tab.~\ref{tab:main_leaderboard_filtered}.

\noindent\textbf{Performance Comparison on Each Dataset.}
We present per-dataset metric breakdowns for all evaluated methods in this section.

\begin{itemize}
  \item \textbf{7-Scenes.} Per-dataset results on 7-Scenes are reported in Table~\ref{tab:dataset_7scenes}.
  \item \textbf{ADT.} Per-dataset results on Aria Digital Twin (ADT) are reported in Table~\ref{tab:dataset_adt}.
  \item \textbf{DROID.} Per-dataset results on DROID are reported in Table~\ref{tab:dataset_droid}.
  \item \textbf{DTU.} Per-dataset results on DTU are reported in Table~\ref{tab:dataset_dtu}.
  \item \textbf{ETH3D.} Per-dataset results on ETH3D are reported in Table~\ref{tab:dataset_eth3d}.
  \item \textbf{HiRoom.} Per-dataset results on HiRoom are reported in Table~\ref{tab:dataset_hiroom}.
  \item \textbf{KITTI Odometry.} Per-dataset results on KITTI Odometry are reported in Table~\ref{tab:dataset_kittiodometry}.
  \item \textbf{Lingbot-Depth.} Per-dataset results on Lingbot are reported in Table~\ref{tab:dataset_lingbot}.
  \item \textbf{NRGBD.} Per-dataset results on NRGBD are reported in Table~\ref{tab:dataset_nrgbd}.
  \item \textbf{OmniWorld.} Per-dataset results on OmniWorld are reported in Table~\ref{tab:dataset_omniworld}.
  \item \textbf{RLBench.} Per-dataset results on RLBench are reported in Table~\ref{tab:dataset_rlbench}.
  \item \textbf{Robolab.} Per-dataset results on Robolab are reported in Table~\ref{tab:dataset_robolab}.
  \item \textbf{RoboTwin.} Per-dataset results on RoboTwin are reported in Table~\ref{tab:dataset_robotwin}.
  \item \textbf{Xperience.} Per-dataset results on Xperience are reported in Table~\ref{tab:dataset_ropedia}.
  \item \textbf{ScanNet++.} Per-dataset results on ScanNet++ are reported in Table~\ref{tab:dataset_scannetpp}.
  \item \textbf{Tanks and Temples.} Per-dataset results on Tanks and Temples are reported in Table~\ref{tab:dataset_tanksandtemples}.
  \item \textbf{TUM RGB-D.} Per-dataset results on TUM RGB-D are reported in Table~\ref{tab:dataset_tum}.
  \item \textbf{Virtual KITTI.} Per-dataset results on Virtual KITTI are reported in Table~\ref{tab:dataset_vkitti}.
  \item \textbf{Waymo.} Per-dataset results on Waymo are reported in Table~\ref{tab:dataset_waymo}.
\end{itemize}

\noindent\textbf{Metric Depth Evaluation.}
Tab.~\ref{tab:metric_depth} compares the native metric-depth prediction quality of all six metric-scale-capable methods on \benchmark, evaluated without median/scale alignment across single-frame, sparse, and medium input settings.


\begin{table}[t]
\centering
\caption{\textbf{Detailed Results on the Single-Frame Setting.}
The best, second-best, and third-best results in each column are highlighted in
\colorbox{bestred}{deep blue}, \colorbox{secondorange}{medium blue}, and \colorbox{thirdyellow}{light blue}, respectively.
Out-of-memory cells are shaded \colorbox{oomred}{light red}.
Within each sub-category, the \textbf{bold} value marks the in-group best.
}
\label{tab:sub_single}
\resizebox{0.55\columnwidth}{!}{%
\renewcommand{\arraystretch}{1.15}
\setlength{\tabcolsep}{3pt}
\begin{tabular}{l c >{\columncolor{subcol}}c >{\columncolor{subcol}}c >{\columncolor{subcol}}c >{\columncolor{subcol}}c >{\columncolor{subcol}}c >{\columncolor{subcol}}c}
\toprule
\multirow{2.5}{*}{\textbf{Method}}
& \multirow{2.5}{*}{\makecell{\textbf{\#Params}\\\textbf{(M)}}}
& \multicolumn{6}{c}{\textbf{Depth}} \\
\cmidrule(lr){3-8}
 & &
 AbsRel$\downarrow$ & SqRel$\downarrow$ & RMSE$\downarrow$ & $\delta_{1.03}\uparrow$ & $\delta_{1.05}\uparrow$ & $\delta_{1.10}\uparrow$ \\
\midrule
\multicolumn{8}{c}{\cellcolor{catgray}\textbf{Optimization-based}} \\
\midrule
 DUSt3R & 571.17 & \textbf{0.385} & \textbf{5.337} & 0.992 & \textbf{0.391} & \textbf{0.524} & \textbf{0.685} \\
 MASt3R & 688.64 & 0.456 & 10.14 & \textbf{0.973} & 0.373 & 0.508 & 0.684 \\
\midrule
\multicolumn{8}{c}{\cellcolor{catgray}\textbf{End-to-End Feed-Forward}} \\
\midrule
 VGGT & 1256.54 & \cellcolor{secondorange}0.184 & \cellcolor{secondorange}0.779 & \cellcolor{thirdyellow}\textbf{0.608} & 0.517 & 0.639 & 0.773 \\
 Fast3R & 647.55 & 0.350 & 3.454 & 1.139 & 0.339 & 0.463 & 0.631 \\
 FastVGGT & 1157.94 & \cellcolor{bestred}\textbf{0.183} & \cellcolor{bestred}\textbf{0.761} & 0.608 & 0.517 & 0.640 & 0.773 \\
 MUSt3R & 423.43 & 0.429 & 8.637 & 0.934 & 0.393 & 0.524 & 0.699 \\
 MapAnything & 1228.49 & 0.451 & 11.99 & 0.854 & 0.441 & 0.583 & 0.747 \\
 OmniVGGT & 1217.49 & \cellcolor{thirdyellow}0.188 & \cellcolor{thirdyellow}0.795 & 0.620 & 0.525 & 0.649 & 0.791 \\
 $\pi^{3}$ & 958.70 & 0.478 & 16.85 & 0.983 & 0.542 & 0.665 & 0.794 \\
 $\pi^{3}$-X & 1360.03 & 0.371 & 8.834 & 0.631 & 0.522 & 0.651 & 0.802 \\
 AMB3R & 1563.12 & 0.466 & 15.32 & 0.658 & 0.544 & \cellcolor{secondorange}\textbf{0.671} & 0.802 \\
 DA3-Small & 34.30 & 0.385 & 6.358 & 0.948 & 0.300 & 0.431 & 0.625 \\
 DA3-Base & 135.37 & 0.349 & 5.281 & 0.842 & 0.359 & 0.501 & 0.691 \\
 DA3-Large & 410.94 & 0.333 & 5.649 & 0.763 & 0.449 & 0.579 & 0.742 \\
 DA3-Giant & 1355.67 & 0.368 & 7.494 & 0.724 & 0.504 & 0.628 & 0.771 \\
 DA3-Nested & 1689.85 & 0.364 & 7.358 & 0.800 & 0.500 & 0.633 & 0.778 \\
 WorldMirror & 1263.34 & 0.349 & 7.695 & 0.707 & 0.485 & 0.619 & 0.776 \\
 VGGT-Omega & 1143.81 & 0.516 & 20.72 & 1.086 & \cellcolor{thirdyellow}\textbf{0.552} & \cellcolor{thirdyellow}0.671 & \cellcolor{bestred}\textbf{0.812} \\ \hdashline
 DANext$^{\dagger}$ (Ours) & 1303.76 & 0.166 & 0.985 & 0.646 & 0.511 & 0.636 & 0.802 \\
\midrule
\multicolumn{8}{c}{\cellcolor{catgray}\textbf{Online}} \\
\midrule
 Spann3r$^{224}$ & 658.69 & 0.370 & 12.39 & 2.345 & 0.288 & 0.417 & 0.596 \\
 CUT3R & 793.31 & 0.247 & \textbf{1.356} & 0.712 & 0.409 & 0.555 & 0.720 \\
 MonST3R & 571.17 & 0.309 & 2.500 & 0.999 & 0.348 & 0.470 & 0.633 \\
 Point3R & 828.01 & 0.379 & 6.806 & 0.783 & 0.400 & 0.534 & 0.703 \\
 Stream3R-S & 1190.60 & 0.409 & 11.3 & 0.672 & \cellcolor{bestred}\textbf{0.564} & \cellcolor{bestred}\textbf{0.691} & \cellcolor{secondorange}\textbf{0.811} \\
 Stream3R-W & 1190.60 & 0.409 & 11.3 & 0.672 & \cellcolor{secondorange}0.564 & \cellcolor{bestred}\textbf{0.691} & \cellcolor{secondorange}\textbf{0.811} \\
 StreamVGGT & 1256.54 & 0.219 & 1.696 & \cellcolor{secondorange}0.583 & 0.523 & 0.646 & 0.784 \\
 Page4D & 1256.81 & 0.228 & 1.977 & 0.636 & 0.467 & 0.606 & 0.762 \\
 InfiniteVGGT & 1256.54 & \textbf{0.217} & 1.643 & \cellcolor{bestred}\textbf{0.583} & 0.524 & 0.645 & 0.783 \\
 Wint3R & 749.46 & 0.619 & 25.47 & 1.053 & 0.434 & 0.555 & 0.719 \\
 LongStream-B & 1190.60 & 0.523 & 17.1 & 0.682 & 0.458 & 0.597 & 0.770 \\
 LongStream-S & 1190.60 & 0.523 & 17.1 & 0.682 & 0.458 & 0.597 & 0.770 \\
 LingbotMap$^{*}$-W & 1157.94 & 0.333 & 4.470 & 0.755 & 0.437 & 0.572 & 0.741 \\
 LingbotMap$^{*}$-S & 1157.94 & 0.333 & 4.470 & 0.755 & 0.437 & 0.572 & 0.741 \\
\midrule
\multicolumn{8}{c}{\cellcolor{catgray}\textbf{Chunk-wise}} \\
\midrule
 VGGT-Long & 1256.54 & \cellcolor{secondorange}\textbf{0.184} & \cellcolor{secondorange}\textbf{0.779} & \cellcolor{thirdyellow}\textbf{0.608} & 0.517 & 0.639 & 0.773 \\
 $\pi^{3}$-Long & 958.70 & 0.478 & 16.85 & 0.983 & \textbf{0.542} & \textbf{0.665} & \textbf{0.794} \\
 DA3-Streaming & 1355.67 & 0.368 & 7.494 & 0.724 & 0.504 & 0.628 & 0.771 \\
\midrule
\multicolumn{8}{c}{\cellcolor{catgray}\textbf{SLAM-based}} \\
\midrule
 MASt3R-SLAM & 688.64 & 0.348 & 1.553 & 1.237 & 0.165 & 0.268 & 0.471 \\
 VGGT-SLAM & 1256.54 & \cellcolor{secondorange}\textbf{0.184} & \cellcolor{secondorange}\textbf{0.779} & \cellcolor{thirdyellow}\textbf{0.608} & \textbf{0.517} & \textbf{0.639} & \textbf{0.773} \\
\midrule
\multicolumn{8}{c}{\cellcolor{catgray}\textbf{Test-Time Training}} \\
\midrule
 TTT3R & 793.31 & 0.247 & 1.357 & 0.712 & 0.409 & 0.555 & 0.720 \\
 Scal3R & 1266.14 & 0.227 & 1.583 & \textbf{0.613} & 0.519 & 0.644 & 0.775 \\
 LoGeR & 1254.62 & 0.251 & 2.718 & 0.635 & 0.527 & 0.666 & 0.800 \\
 LoGeR$^*$ & 1254.60 & \textbf{0.200} & \textbf{1.320} & 0.637 & \textbf{0.540} & \textbf{0.668} & \cellcolor{thirdyellow}\textbf{0.803} \\
\bottomrule
\end{tabular}%
}
\end{table}

\begin{table*}[t]
\centering
\caption{\textbf{Detailed Results on the Sparse Setting.}
The best, second-best, and third-best results in each column are highlighted in
\colorbox{bestred}{deep blue}, \colorbox{secondorange}{medium blue}, and \colorbox{thirdyellow}{light blue}, respectively.
Out-of-memory cells are shaded \colorbox{oomred}{light red}.
Within each sub-category, the \textbf{bold} value marks the in-group best.
}
\label{tab:sub_sparse}
\resizebox{\textwidth}{!}{%
\renewcommand{\arraystretch}{1.15}
\setlength{\tabcolsep}{3pt}
\begin{tabular}{l c >{\columncolor{subcol}}c >{\columncolor{subcol}}c >{\columncolor{subcol}}c >{\columncolor{subcol}}c >{\columncolor{subcol}}c >{\columncolor{subcol}}c >{\columncolor{subcol}}c >{\columncolor{subcol}}c >{\columncolor{subcol}}c >{\columncolor{subcol}}c >{\columncolor{subcol}}c >{\columncolor{subcol}}c >{\columncolor{subcol}}c}
\toprule
\multirow{2.5}{*}{\textbf{Method}}
& \multirow{2.5}{*}{\makecell{\textbf{\#Params}\\\textbf{(M)}}}
& \multicolumn{6}{c}{\textbf{Depth}}
& \multicolumn{7}{c}{\textbf{Camera}} \\
\cmidrule(lr){3-8} \cmidrule(lr){9-15}
 & &
 AbsRel$\downarrow$ & SqRel$\downarrow$ & RMSE$\downarrow$ & $\delta_{1.03}\uparrow$ & $\delta_{1.05}\uparrow$ & $\delta_{1.10}\uparrow$ &
 RAcc$_{3}\uparrow$ & RAcc$_{5}\uparrow$ & TAcc$_{3}\uparrow$ & TAcc$_{5}\uparrow$ & AUC@5$\uparrow$ & AUC@15$\uparrow$ & AUC@30$\uparrow$ \\
\midrule
\multicolumn{15}{c}{\cellcolor{catgray}\textbf{Optimization-based}} \\
\midrule
 DUSt3R & 571.17 & 0.257 & 2.693 & 1.567 & 0.295 & 0.420 & 0.600 & 0.527 & 0.654 & 0.242 & 0.350 & 0.184 & 0.374 & 0.498 \\
 MASt3R & 688.64 & \textbf{0.209} & \textbf{0.606} & \textbf{1.246} & \textbf{0.309} & \textbf{0.456} & \textbf{0.639} & \textbf{0.587} & \textbf{0.701} & \textbf{0.301} & \textbf{0.411} & \textbf{0.244} & \textbf{0.438} & \textbf{0.568} \\
\midrule
\multicolumn{15}{c}{\cellcolor{catgray}\textbf{End-to-End Feed-Forward}} \\
\midrule
 VGGT & 1256.54 & 0.105 & 0.105 & 0.673 & 0.498 & 0.627 & 0.778 & 0.734 & 0.833 & 0.454 & 0.550 & 0.405 & 0.585 & 0.700 \\
 Fast3R & 647.55 & 0.260 & 0.760 & 1.576 & 0.228 & 0.340 & 0.508 & 0.347 & 0.486 & 0.156 & 0.235 & 0.109 & 0.264 & 0.392 \\
 FastVGGT & 1157.94 & 0.113 & 0.098 & 0.666 & 0.457 & 0.590 & 0.752 & 0.642 & 0.773 & 0.384 & 0.472 & 0.314 & 0.503 & 0.631 \\
 MUSt3R & 423.43 & 0.165 & 0.276 & 0.999 & 0.366 & 0.502 & 0.686 & 0.622 & 0.726 & 0.343 & 0.435 & 0.273 & 0.482 & 0.614 \\
 MapAnything & 1228.49 & 0.153 & 1.079 & 1.337 & 0.361 & 0.512 & 0.701 & 0.608 & 0.762 & 0.300 & 0.423 & 0.244 & 0.446 & 0.579 \\
 OmniVGGT & 1217.49 & 0.117 & 0.119 & 0.669 & 0.534 & 0.671 & 0.810 & 0.620 & 0.746 & 0.416 & 0.507 & 0.332 & 0.537 & 0.665 \\
 $\pi^{3}$ & 958.70 & 0.092 & 0.258 & 0.855 & 0.547 & 0.686 & 0.826 & 0.769 & 0.854 & 0.474 & 0.586 & 0.412 & 0.619 & 0.742 \\
 $\pi^{3}$-X & 1360.03 & \cellcolor{thirdyellow}0.084 & 0.084 & \textbf{0.599} & \cellcolor{secondorange}\textbf{0.576} & \cellcolor{secondorange}\textbf{0.710} & \cellcolor{thirdyellow}0.833 & 0.756 & 0.837 & 0.491 & 0.595 & 0.427 & 0.627 & 0.741 \\
 AMB3R & 1563.12 & 0.088 & \cellcolor{thirdyellow}\textbf{0.083} & 0.617 & 0.534 & 0.668 & 0.812 & 0.743 & 0.844 & 0.475 & 0.591 & 0.412 & 0.619 & 0.739 \\
 DA3-Small & 34.30 & 0.191 & 0.270 & 1.057 & 0.308 & 0.440 & 0.633 & 0.392 & 0.568 & 0.172 & 0.300 & 0.127 & 0.336 & 0.476 \\
 DA3-Base & 135.37 & 0.159 & 0.210 & 0.937 & 0.386 & 0.520 & 0.690 & 0.528 & 0.678 & 0.292 & 0.399 & 0.222 & 0.427 & 0.566 \\
 DA3-Large & 410.94 & 0.128 & 0.198 & 0.839 & 0.470 & 0.601 & 0.753 & 0.662 & 0.776 & 0.443 & 0.528 & 0.361 & 0.559 & 0.688 \\
 DA3-Giant & 1355.67 & 0.095 & 0.107 & 0.608 & 0.563 & 0.689 & 0.821 & \cellcolor{secondorange}0.792 & \cellcolor{thirdyellow}0.871 & \cellcolor{secondorange}\textbf{0.586} & \cellcolor{thirdyellow}0.682 & \cellcolor{secondorange}\textbf{0.525} & \cellcolor{secondorange}\textbf{0.699} & \cellcolor{thirdyellow}0.785 \\
 DA3-Nested & 1689.85 & 0.106 & 0.199 & 0.805 & \cellcolor{thirdyellow}0.564 & 0.692 & 0.814 & 0.784 & 0.855 & 0.577 & \cellcolor{secondorange}\textbf{0.682} & \cellcolor{thirdyellow}0.519 & 0.691 & 0.779 \\
 WorldMirror & 1263.34 & 0.139 & 0.162 & 0.798 & 0.440 & 0.583 & 0.748 & 0.697 & 0.793 & 0.394 & 0.506 & 0.320 & 0.532 & 0.660 \\
 VGGT-Omega & 1143.81 & \cellcolor{bestred}\textbf{0.077} & 0.362 & 1.089 & 0.556 & \cellcolor{thirdyellow}0.702 & \cellcolor{secondorange}\textbf{0.843} & \cellcolor{bestred}\textbf{0.808} & \cellcolor{bestred}\textbf{0.892} & \cellcolor{thirdyellow}0.585 & 0.678 & 0.504 & \cellcolor{thirdyellow}0.695 & \cellcolor{bestred}\textbf{0.803} \\ \hdashline
 DANext$^{\dagger}$ (Ours) & 1303.76 & 0.050 & 0.073 & 0.554 & 0.647 & 0.772 & 0.893 & 0.815 & 0.900 & 0.587 & 0.718 & 0.518 & 0.718 & 0.809 \\
\midrule
\multicolumn{15}{c}{\cellcolor{catgray}\textbf{Online}} \\
\midrule
 Spann3r$^{224}$ & 658.69 & 0.274 & 212.3 & 10.47 & 0.209 & 0.317 & 0.496 & 0.265 & 0.423 & 0.080 & 0.163 & 0.052 & 0.198 & 0.329 \\
 CUT3R & 793.31 & 0.196 & 0.218 & 0.917 & 0.300 & 0.445 & 0.638 & 0.478 & 0.637 & 0.223 & 0.356 & 0.177 & 0.388 & 0.519 \\
 MonST3R & 571.17 & 0.227 & 0.399 & 1.338 & 0.220 & 0.332 & 0.521 & 0.243 & 0.311 & 0.107 & 0.177 & 0.069 & 0.172 & 0.269 \\
 Point3R & 828.01 & 0.221 & 0.280 & 1.044 & 0.251 & 0.379 & 0.559 & 0.194 & 0.372 & 0.058 & 0.123 & 0.028 & 0.177 & 0.339 \\
 Stream3R-S & 1190.60 & 0.114 & 0.120 & 0.705 & \textbf{0.450} & \textbf{0.597} & \textbf{0.765} & 0.618 & 0.774 & 0.351 & 0.437 & 0.278 & 0.471 & 0.603 \\
 Stream3R-W & 1190.60 & 0.117 & 0.131 & 0.739 & 0.442 & 0.584 & 0.753 & 0.610 & 0.762 & 0.344 & 0.430 & 0.272 & 0.464 & 0.597 \\
 StreamVGGT & 1256.54 & 0.154 & 0.362 & 1.243 & 0.314 & 0.424 & 0.609 & 0.638 & 0.771 & 0.339 & 0.435 & 0.283 & 0.472 & 0.598 \\
 Page4D & 1256.81 & \textbf{0.112} & \textbf{0.089} & \textbf{0.646} & 0.438 & 0.587 & 0.758 & 0.551 & 0.730 & 0.295 & 0.428 & 0.224 & 0.456 & 0.608 \\
 InfiniteVGGT & 1256.54 & 0.154 & 0.363 & 1.245 & 0.314 & 0.423 & 0.610 & 0.639 & 0.767 & 0.346 & 0.435 & 0.286 & 0.471 & 0.596 \\
 Wint3R & 749.46 & 0.157 & 0.232 & 0.909 & 0.399 & 0.532 & 0.702 & 0.438 & 0.600 & 0.214 & 0.334 & 0.157 & 0.366 & 0.499 \\
 LongStream-B & 1190.60 & 0.153 & 0.152 & 0.771 & 0.380 & 0.526 & 0.712 & 0.502 & 0.678 & 0.263 & 0.383 & 0.197 & 0.417 & 0.549 \\
 LongStream-S & 1190.60 & 0.151 & 0.144 & 0.754 & 0.383 & 0.531 & 0.716 & 0.501 & 0.677 & 0.258 & 0.377 & 0.196 & 0.413 & 0.543 \\
 LingbotMap$^{*}$-W & 1157.94 & 0.138 & 0.138 & 0.759 & 0.376 & 0.536 & 0.730 & \textbf{0.696} & \textbf{0.811} & \textbf{0.359} & \textbf{0.469} & \textbf{0.303} & \textbf{0.520} & \textbf{0.650} \\
 LingbotMap$^{*}$-S & 1157.94 & 0.138 & 0.138 & 0.759 & 0.376 & 0.536 & 0.730 & \textbf{0.696} & \textbf{0.811} & \textbf{0.359} & \textbf{0.469} & \textbf{0.303} & \textbf{0.520} & \textbf{0.650} \\
\midrule
\multicolumn{15}{c}{\cellcolor{catgray}\textbf{Chunk-wise}} \\
\midrule
 VGGT-Long & 1256.54 & 0.105 & \textbf{0.105} & 0.673 & 0.498 & 0.627 & 0.778 & 0.734 & 0.833 & 0.454 & 0.550 & 0.405 & 0.585 & 0.700 \\
 $\pi^{3}$-Long & 958.70 & \textbf{0.092} & 0.258 & 0.855 & 0.547 & 0.686 & \textbf{0.826} & 0.769 & 0.854 & 0.474 & 0.586 & 0.412 & 0.619 & 0.742 \\
 DA3-Streaming & 1355.67 & 0.095 & 0.107 & \textbf{0.608} & \textbf{0.563} & \textbf{0.689} & 0.821 & \cellcolor{thirdyellow}\textbf{0.790} & \cellcolor{secondorange}\textbf{0.871} & \cellcolor{bestred}\textbf{0.588} & \cellcolor{bestred}\textbf{0.682} & \cellcolor{bestred}\textbf{0.525} & \cellcolor{bestred}\textbf{0.699} & \cellcolor{secondorange}\textbf{0.785} \\
\midrule
\multicolumn{15}{c}{\cellcolor{catgray}\textbf{SLAM-based}} \\
\midrule
 MASt3R-SLAM & 688.64 & 0.336 & 1.715 & 2.282 & 0.120 & 0.194 & 0.347 & 0.317 & 0.401 & 0.066 & 0.095 & 0.047 & 0.118 & 0.190 \\
 VGGT-SLAM & 1256.54 & \textbf{0.105} & \textbf{0.105} & \textbf{0.673} & \textbf{0.498} & \textbf{0.627} & \textbf{0.778} & \textbf{0.734} & \textbf{0.833} & \textbf{0.454} & \textbf{0.550} & \textbf{0.405} & \textbf{0.585} & \textbf{0.700} \\
\midrule
\multicolumn{15}{c}{\cellcolor{catgray}\textbf{Test-Time Training}} \\
\midrule
 TTT3R & 793.31 & 0.202 & 0.257 & 0.966 & 0.282 & 0.428 & 0.628 & 0.452 & 0.591 & 0.197 & 0.302 & 0.153 & 0.342 & 0.469 \\
 Scal3R & 1266.14 & 0.114 & 0.091 & \cellcolor{thirdyellow}0.593 & 0.535 & 0.669 & 0.810 & \textbf{0.734} & \textbf{0.833} & \textbf{0.484} & \textbf{0.581} & \textbf{0.418} & \textbf{0.610} & \textbf{0.732} \\
 LoGeR & 1254.62 & 0.095 & \cellcolor{secondorange}0.077 & \cellcolor{secondorange}0.569 & 0.546 & 0.700 & 0.833 & 0.695 & 0.821 & 0.411 & 0.538 & 0.348 & 0.564 & 0.687 \\
 LoGeR$^*$ & 1254.60 & \cellcolor{secondorange}\textbf{0.077} & \cellcolor{bestred}\textbf{0.071} & \cellcolor{bestred}\textbf{0.558} & \cellcolor{bestred}\textbf{0.581} & \cellcolor{bestred}\textbf{0.725} & \cellcolor{bestred}\textbf{0.854} & 0.715 & 0.824 & 0.424 & 0.547 & 0.353 & 0.576 & 0.708 \\
\bottomrule
\end{tabular}%
}
\end{table*}

\begin{table*}[t]
\centering
\caption{\textbf{Detailed Results on the Medium Setting.}
The best, second-best, and third-best results in each column are highlighted in
\colorbox{bestred}{deep blue}, \colorbox{secondorange}{medium blue}, and \colorbox{thirdyellow}{light blue}, respectively.
Out-of-memory cells are shaded \colorbox{oomred}{light red}.
Within each sub-category, the \textbf{bold} value marks the in-group best.
}
\label{tab:sub_medium}
\resizebox{\textwidth}{!}{%
\renewcommand{\arraystretch}{1.15}
\setlength{\tabcolsep}{3pt}
\begin{tabular}{l c >{\columncolor{subcol}}c >{\columncolor{subcol}}c >{\columncolor{subcol}}c >{\columncolor{subcol}}c >{\columncolor{subcol}}c >{\columncolor{subcol}}c >{\columncolor{subcol}}c >{\columncolor{subcol}}c >{\columncolor{subcol}}c >{\columncolor{subcol}}c >{\columncolor{subcol}}c >{\columncolor{subcol}}c >{\columncolor{subcol}}c >{\columncolor{subcol}}c >{\columncolor{subcol}}c >{\columncolor{subcol}}c >{\columncolor{subcol}}c >{\columncolor{subcol}}c}
\toprule
\multirow{2.5}{*}{\textbf{Method}}
& \multirow{2.5}{*}{\makecell{\textbf{\#Params}\\\textbf{(M)}}}
& \multicolumn{6}{c}{\textbf{Depth}}
& \multicolumn{7}{c}{\textbf{Camera}}
& \multicolumn{3}{c}{\textbf{Trajectory}}
& \multicolumn{2}{c}{\textbf{PointCloud}} \\
\cmidrule(lr){3-8} \cmidrule(lr){9-15} \cmidrule(lr){16-18} \cmidrule(lr){19-20}
 & &
 AbsRel$\downarrow$ & SqRel$\downarrow$ & RMSE$\downarrow$ & $\delta_{1.03}\uparrow$ & $\delta_{1.05}\uparrow$ & $\delta_{1.10}\uparrow$ &
 RAcc$_{3}\uparrow$ & RAcc$_{5}\uparrow$ & TAcc$_{3}\uparrow$ & TAcc$_{5}\uparrow$ & AUC@5$\uparrow$ & AUC@15$\uparrow$ & AUC@30$\uparrow$ &
 ATE$\downarrow$ & RPE$_t\downarrow$ & RPE$_r\downarrow$ &
 F-Score$\uparrow$ & Overall$\downarrow$ \\
\midrule
\multicolumn{20}{c}{\cellcolor{catgray}\textbf{Optimization-based}} \\
\midrule
 DUSt3R & 571.17 & 0.276 & \textbf{0.860} & \textbf{1.447} & 0.247 & 0.361 & 0.529 & 0.402 & 0.543 & 0.172 & 0.273 & 0.130 & 0.315 & 0.448 & \textbf{1.691} & 0.386 & 4.096 & 0.343 & \textbf{0.142} \\
 MASt3R & 688.64 & \textbf{0.259} & 1.871 & 1.898 & \textbf{0.259} & \textbf{0.379} & \textbf{0.554} & \textbf{0.477} & \textbf{0.615} & \textbf{0.230} & \textbf{0.344} & \textbf{0.176} & \textbf{0.381} & \textbf{0.522} & 1.911 & \textbf{0.236} & \textbf{2.997} & \textbf{0.370} & 0.382 \\
\midrule
\multicolumn{20}{c}{\cellcolor{catgray}\textbf{End-to-End Feed-Forward}} \\
\midrule
 VGGT & 1256.54 & 0.125 & 0.602 & 0.688 & 0.499 & 0.613 & 0.747 & 0.695 & 0.782 & 0.484 & 0.562 & 0.432 & 0.586 & 0.687 & 0.727 & 0.216 & 2.202 & 0.661 & 0.087 \\
 Fast3R & 647.55 & 0.255 & 0.567 & 1.539 & 0.241 & 0.344 & 0.496 & 0.296 & 0.422 & 0.164 & 0.248 & 0.117 & 0.272 & 0.386 & 6.582 & 2.055 & 13.65 & 0.300 & 0.210 \\
 FastVGGT & 1157.94 & 0.105 & 0.086 & 0.611 & 0.479 & 0.597 & 0.737 & 0.678 & 0.771 & 0.448 & 0.542 & 0.379 & 0.554 & 0.662 & 0.738 & 0.259 & 3.260 & 0.576 & 0.121 \\
 MUSt3R & 423.43 & 0.162 & 0.327 & 0.966 & 0.388 & 0.522 & 0.696 & 0.618 & 0.734 & 0.361 & 0.475 & 0.296 & 0.507 & 0.643 & 3.097 & 0.613 & 2.686 & 0.507 & 0.230 \\
 MapAnything & 1228.49 & 0.146 & 0.490 & 1.052 & 0.347 & 0.491 & 0.681 & 0.563 & 0.702 & 0.312 & 0.419 & 0.254 & 0.451 & 0.579 & 1.737 & 0.533 & 2.852 & 0.420 & 0.114 \\
 OmniVGGT & 1217.49 & 0.111 & 0.096 & 0.649 & 0.518 & 0.645 & 0.780 & 0.609 & 0.726 & 0.426 & 0.527 & 0.340 & 0.542 & 0.665 & 1.491 & 0.355 & 2.768 & 0.595 & 0.104 \\
 $\pi^{3}$ & 958.70 & \cellcolor{thirdyellow}0.082 & 0.229 & 0.822 & 0.563 & 0.689 & \cellcolor{thirdyellow}0.814 & 0.742 & 0.830 & 0.501 & 0.613 & 0.433 & 0.636 & 0.749 & 0.565 & 0.128 & \cellcolor{thirdyellow}\textbf{1.240} & 0.649 & 0.080 \\
 $\pi^{3}$-X & 1360.03 & \cellcolor{secondorange}0.078 & \cellcolor{secondorange}\textbf{0.061} & \cellcolor{secondorange}\textbf{0.538} & \cellcolor{secondorange}0.582 & \cellcolor{secondorange}0.712 & \cellcolor{secondorange}0.831 & 0.741 & 0.827 & 0.536 & 0.628 & 0.463 & 0.644 & 0.744 & \cellcolor{bestred}\textbf{0.369} & \cellcolor{thirdyellow}\textbf{0.108} & 1.459 & 0.658 & 0.074 \\
 AMB3R & 1563.12 & 0.085 & 0.074 & 0.580 & 0.539 & 0.661 & 0.795 & 0.704 & 0.799 & 0.496 & 0.588 & 0.429 & 0.613 & 0.727 & 0.645 & 0.223 & 1.881 & 0.554 & 0.123 \\
 DA3-Small & 34.30 & 0.176 & 0.226 & 1.008 & 0.312 & 0.435 & 0.617 & 0.370 & 0.535 & 0.192 & 0.301 & 0.136 & 0.338 & 0.479 & 4.850 & 1.356 & 5.534 & 0.432 & 0.123 \\
 DA3-Base & 135.37 & 0.142 & 0.171 & 0.884 & 0.395 & 0.521 & 0.683 & 0.494 & 0.626 & 0.289 & 0.407 & 0.227 & 0.430 & 0.562 & 3.865 & 0.976 & 3.928 & 0.515 & 0.100 \\
 DA3-Large & 410.94 & 0.105 & 0.155 & 0.779 & 0.498 & 0.618 & 0.767 & 0.655 & 0.758 & 0.487 & 0.575 & 0.411 & 0.588 & 0.701 & 2.722 & 0.460 & 2.800 & 0.626 & 0.130 \\
 DA3-Giant & 1355.67 & 0.086 & 0.088 & 0.578 & 0.572 & 0.686 & 0.812 & \cellcolor{secondorange}0.749 & \cellcolor{secondorange}0.834 & \cellcolor{secondorange}0.587 & \cellcolor{secondorange}0.663 & \cellcolor{bestred}\textbf{0.532} & \cellcolor{secondorange}0.684 & \cellcolor{secondorange}0.776 & 1.161 & 0.285 & 2.253 & \cellcolor{bestred}\textbf{0.742} & \cellcolor{bestred}\textbf{0.073} \\
 DA3-Nested & 1689.85 & 0.086 & 0.182 & 0.801 & \cellcolor{thirdyellow}0.577 & \cellcolor{thirdyellow}0.690 & 0.810 & 0.742 & \cellcolor{thirdyellow}0.833 & 0.569 & \cellcolor{thirdyellow}0.658 & 0.510 & \cellcolor{thirdyellow}0.676 & \cellcolor{thirdyellow}0.770 & 1.980 & 0.394 & 2.708 & \cellcolor{secondorange}0.737 & \cellcolor{secondorange}0.073 \\
 WorldMirror & 1263.34 & 0.118 & 0.126 & 0.741 & 0.445 & 0.573 & 0.721 & 0.674 & 0.775 & 0.417 & 0.533 & 0.352 & 0.556 & 0.674 & 1.357 & 0.347 & 2.056 & 0.575 & 0.090 \\
 VGGT-Omega & 1143.81 & \cellcolor{bestred}\textbf{0.067} & 0.304 & 1.049 & \cellcolor{bestred}\textbf{0.589} & \cellcolor{bestred}\textbf{0.721} & \cellcolor{bestred}\textbf{0.845} & \cellcolor{bestred}\textbf{0.787} & \cellcolor{bestred}\textbf{0.870} & \cellcolor{bestred}\textbf{0.589} & \cellcolor{bestred}\textbf{0.692} & \cellcolor{thirdyellow}0.512 & \cellcolor{bestred}\textbf{0.700} & \cellcolor{bestred}\textbf{0.795} & 0.659 & 0.160 & 1.369 & 0.706 & 0.078 \\ \hdashline
 DANext$^{\dagger}$ (Ours) & 1303.76 & 0.035 & 0.062 & 0.520 & 0.715 & 0.830 & 0.928 & 0.806 & 0.880 & 0.630 & 0.733 & 0.553 & 0.733 & 0.819 & 1.442 & 0.251 & 1.602 & 0.727 & 0.072 \\
\midrule
\multicolumn{20}{c}{\cellcolor{catgray}\textbf{Online}} \\
\midrule
 Spann3r$^{224}$ & 658.69 & 0.252 & 145 & 9.194 & 0.208 & 0.315 & 0.486 & 0.198 & 0.358 & 0.085 & 0.160 & 0.049 & 0.205 & 0.361 & 4.312 & 1.118 & 6.784 & 0.254 & 0.249 \\
 CUT3R & 793.31 & 0.189 & 0.423 & 1.095 & 0.271 & 0.396 & 0.576 & 0.335 & 0.506 & 0.160 & 0.273 & 0.110 & 0.316 & 0.469 & 2.676 & 0.373 & 4.271 & 0.286 & 0.385 \\
 MonST3R & 571.17 & 0.241 & 0.595 & 1.551 & 0.162 & 0.252 & 0.413 & 0.161 & 0.221 & 0.037 & 0.078 & 0.026 & 0.100 & 0.195 & 2.234 & 0.526 & 12.99 & 0.081 & 0.502 \\
 Point3R & 828.01 & 0.228 & 0.261 & 1.044 & 0.247 & 0.368 & 0.551 & 0.178 & 0.340 & 0.055 & 0.116 & 0.027 & 0.161 & 0.303 & 6.512 & 0.847 & 8.545 & 0.211 & 0.213 \\
 Stream3R-S & 1190.60 & 0.204 & 0.725 & 1.414 & 0.295 & 0.395 & 0.544 & 0.392 & 0.514 & 0.223 & 0.299 & 0.178 & 0.317 & 0.427 & 5.717 & 0.621 & 5.650 & 0.348 & 0.594 \\
 Stream3R-W & 1190.60 & 0.240 & 1.099 & 1.717 & 0.260 & 0.350 & 0.500 & 0.330 & 0.441 & 0.173 & 0.251 & 0.136 & 0.265 & 0.364 & 6.756 & 0.896 & 7.827 & 0.323 & 0.565 \\
 StreamVGGT & 1256.54 & 0.171 & 0.509 & 1.458 & 0.271 & 0.372 & 0.539 & 0.566 & 0.679 & 0.305 & 0.405 & 0.251 & 0.437 & 0.562 & 4.940 & 0.526 & 3.820 & 0.397 & 0.154 \\
 Page4D & 1256.81 & \textbf{0.107} & \textbf{0.079} & \textbf{0.613} & \textbf{0.454} & \textbf{0.589} & \textbf{0.748} & 0.531 & 0.712 & 0.334 & 0.464 & 0.246 & 0.479 & 0.618 & 0.855 & 0.255 & 2.574 & \textbf{0.423} & \textbf{0.118} \\
 InfiniteVGGT & 1256.54 & 0.170 & 0.504 & 1.453 & 0.271 & 0.372 & 0.539 & 0.566 & 0.679 & 0.304 & 0.407 & 0.251 & 0.439 & 0.563 & 4.964 & 0.536 & 3.908 & 0.402 & 0.151 \\
 Wint3R & 749.46 & 0.144 & 0.174 & 0.856 & 0.356 & 0.494 & 0.677 & 0.336 & 0.508 & 0.157 & 0.267 & 0.108 & 0.303 & 0.444 & 3.944 & 0.625 & 3.351 & 0.401 & 0.201 \\
 LongStream-B & 1190.60 & 0.224 & 0.182 & 0.854 & 0.213 & 0.324 & 0.504 & 0.376 & 0.550 & 0.165 & 0.270 & 0.118 & 0.315 & 0.455 & 0.925 & 0.282 & 4.074 & 0.135 & 0.303 \\
 LongStream-S & 1190.60 & 0.166 & 0.147 & 0.796 & 0.288 & 0.418 & 0.609 & 0.266 & 0.412 & 0.128 & 0.217 & 0.085 & 0.249 & 0.385 & 1.188 & 0.345 & 4.945 & 0.126 & 0.345 \\
 LingbotMap$^{*}$-W & 1157.94 & 0.114 & 0.177 & 0.799 & 0.385 & 0.535 & 0.725 & 0.610 & 0.745 & 0.349 & 0.471 & 0.290 & 0.502 & 0.641 & 0.509 & \textbf{0.176} & \textbf{1.896} & 0.362 & 0.196 \\
 LingbotMap$^{*}$-S & 1157.94 & 0.114 & 0.179 & 0.807 & 0.405 & 0.549 & 0.730 & \textbf{0.621} & \textbf{0.753} & \textbf{0.369} & \textbf{0.479} & \textbf{0.304} & \textbf{0.511} & \textbf{0.647} & \textbf{0.508} & 0.193 & 2.025 & 0.411 & 0.175 \\
\midrule
\multicolumn{20}{c}{\cellcolor{catgray}\textbf{Chunk-wise}} \\
\midrule
 VGGT-Long & 1256.54 & 0.131 & 0.601 & 0.679 & 0.479 & 0.593 & 0.735 & 0.673 & 0.773 & 0.465 & 0.548 & 0.410 & 0.573 & 0.679 & 0.512 & 0.164 & 2.199 & 0.633 & 0.090 \\
 $\pi^{3}$-Long & 958.70 & 0.097 & 0.248 & 0.897 & 0.446 & 0.583 & 0.742 & 0.734 & 0.825 & 0.501 & 0.609 & 0.426 & 0.626 & 0.740 & \cellcolor{thirdyellow}\textbf{0.465} & \cellcolor{secondorange}\textbf{0.106} & \cellcolor{secondorange}\textbf{1.149} & 0.590 & 0.087 \\
 DA3-Streaming & 1355.67 & \textbf{0.091} & \textbf{0.110} & \textbf{0.618} & \textbf{0.560} & \textbf{0.674} & \textbf{0.794} & \cellcolor{thirdyellow}\textbf{0.743} & \textbf{0.828} & \cellcolor{thirdyellow}\textbf{0.578} & \textbf{0.656} & \cellcolor{secondorange}\textbf{0.520} & \textbf{0.675} & \textbf{0.767} & 0.563 & 0.135 & 1.922 & \cellcolor{thirdyellow}\textbf{0.725} & \cellcolor{thirdyellow}\textbf{0.074} \\
\midrule
\multicolumn{20}{c}{\cellcolor{catgray}\textbf{SLAM-based}} \\
\midrule
 MASt3R-SLAM & 688.64 & 0.348 & 1.504 & 2.278 & 0.102 & 0.169 & 0.312 & 0.321 & 0.438 & 0.084 & 0.135 & 0.064 & 0.168 & 0.262 & 6.075 & 0.682 & 10.98 & 0.130 & 0.870 \\
 VGGT-SLAM & 1256.54 & \textbf{0.129} & \textbf{0.212} & \textbf{0.691} & \textbf{0.448} & \textbf{0.562} & \textbf{0.703} & \textbf{0.617} & \textbf{0.723} & \textbf{0.416} & \textbf{0.502} & \textbf{0.363} & \textbf{0.531} & \textbf{0.645} & \textbf{0.686} & \textbf{0.145} & \textbf{2.186} & \textbf{0.610} & \textbf{0.091} \\
\midrule
\multicolumn{20}{c}{\cellcolor{catgray}\textbf{Test-Time Training}} \\
\midrule
 TTT3R & 793.31 & 0.179 & 0.242 & 0.970 & 0.295 & 0.424 & 0.604 & 0.412 & 0.573 & 0.188 & 0.307 & 0.142 & 0.351 & 0.493 & 2.343 & 0.462 & 5.740 & 0.294 & 0.373 \\
 Scal3R & 1266.14 & 0.147 & 0.170 & 0.818 & 0.327 & 0.423 & 0.602 & 0.629 & 0.753 & 0.435 & 0.535 & 0.349 & 0.546 & 0.670 & \cellcolor{secondorange}\textbf{0.400} & 0.154 & 1.713 & \textbf{0.671} & 0.201 \\
 LoGeR & 1254.62 & 0.113 & \cellcolor{thirdyellow}0.065 & \cellcolor{thirdyellow}0.542 & 0.435 & 0.576 & 0.741 & 0.690 & 0.793 & 0.405 & 0.540 & 0.340 & 0.564 & 0.693 & 0.591 & 0.123 & 1.254 & 0.504 & 0.096 \\
 LoGeR$^*$ & 1254.60 & \textbf{0.083} & \cellcolor{bestred}\textbf{0.057} & \cellcolor{bestred}\textbf{0.515} & \textbf{0.523} & \textbf{0.667} & \textbf{0.799} & \textbf{0.704} & \textbf{0.801} & \textbf{0.452} & \textbf{0.569} & \textbf{0.373} & \textbf{0.589} & \textbf{0.714} & 0.566 & \cellcolor{bestred}\textbf{0.097} & \cellcolor{bestred}\textbf{1.080} & 0.574 & \textbf{0.086} \\
\bottomrule
\end{tabular}%
}
\end{table*}

\begin{table*}[t]
\setlength{\fboxsep}{0pt}
\centering
\caption{\textbf{Detailed Results on the Dense Regime.}
The best, second-best, and third-best results in each column are highlighted in
\colorbox{bestred}{deep blue}, \colorbox{secondorange}{medium blue}, and \colorbox{thirdyellow}{light blue}, respectively.
Out-of-memory (OOM) and Timeout (T.O) cells are shaded \colorbox{oomred}{light red}.
Within each sub-category, the \textbf{bold} value marks the in-group best.
}
\label{tab:sub_dense}
\resizebox{\textwidth}{!}{%
\renewcommand{\arraystretch}{1.15}
\setlength{\tabcolsep}{3pt}
\begin{tabular}{l c >{\columncolor{subcol}}c c >{\columncolor{subcol}}c c >{\columncolor{subcol}}c c >{\columncolor{subcol}}c c >{\columncolor{subcol}}c c >{\columncolor{subcol}}c c >{\columncolor{subcol}}c c >{\columncolor{subcol}}c c >{\columncolor{subcol}}c c}
\toprule
\multirow{2.5}{*}{\textbf{Method}}
& \multirow{2.5}{*}{\makecell{\textbf{\#Params}\\\textbf{(M)}}}
& \multicolumn{6}{c}{\textbf{Depth}}
& \multicolumn{7}{c}{\textbf{Camera}}
& \multicolumn{3}{c}{\textbf{Trajectory}}
& \multicolumn{2}{c}{\textbf{PointCloud}} \\
\cmidrule(lr){3-8} \cmidrule(lr){9-15} \cmidrule(lr){16-18} \cmidrule(lr){19-20}
 & &
 AbsRel$\downarrow$ & SqRel$\downarrow$ & RMSE$\downarrow$ & $\delta_{1.03}\uparrow$ & $\delta_{1.05}\uparrow$ & $\delta_{1.10}\uparrow$ &
 RAcc$_{3}\uparrow$ & RAcc$_{5}\uparrow$ & TAcc$_{3}\uparrow$ & TAcc$_{5}\uparrow$ & AUC@5$\uparrow$ & AUC@15$\uparrow$ & AUC@30$\uparrow$ &
 ATE$\downarrow$ & RPE$_t\downarrow$ & RPE$_r\downarrow$ &
 F-Score$\uparrow$ & Overall$\downarrow$ \\
\midrule

\multicolumn{20}{c}{\cellcolor{catgray}\textbf{Optimization-based}} \\
\midrule
DUSt3R   & 571.17 & \cellcolor{oomred}OOM & \cellcolor{oomred}OOM & \cellcolor{oomred}OOM & \cellcolor{oomred}OOM & \cellcolor{oomred}OOM & \cellcolor{oomred}OOM & \cellcolor{oomred}OOM & \cellcolor{oomred}OOM & \cellcolor{oomred}OOM & \cellcolor{oomred}OOM & \cellcolor{oomred}OOM & \cellcolor{oomred}OOM & \cellcolor{oomred}OOM & \cellcolor{oomred}OOM & \cellcolor{oomred}OOM & \cellcolor{oomred}OOM & \cellcolor{oomred}OOM & \cellcolor{oomred}OOM \\
MASt3R   & 688.64 & \cellcolor{oomred}OOM & \cellcolor{oomred}OOM & \cellcolor{oomred}OOM & \cellcolor{oomred}OOM & \cellcolor{oomred}OOM & \cellcolor{oomred}OOM & \cellcolor{oomred}OOM & \cellcolor{oomred}OOM & \cellcolor{oomred}OOM & \cellcolor{oomred}OOM & \cellcolor{oomred}OOM & \cellcolor{oomred}OOM & \cellcolor{oomred}OOM & \cellcolor{oomred}OOM & \cellcolor{oomred}OOM & \cellcolor{oomred}OOM & \cellcolor{oomred}OOM & \cellcolor{oomred}OOM \\
\midrule

\multicolumn{20}{c}{\cellcolor{catgray}\textbf{End-to-End Feed-Forward}} \\
\midrule
VGGT          & 1256.54 & \cellcolor{oomred}OOM & \cellcolor{oomred}OOM & \cellcolor{oomred}OOM & \cellcolor{oomred}OOM & \cellcolor{oomred}OOM & \cellcolor{oomred}OOM & \cellcolor{oomred}OOM & \cellcolor{oomred}OOM & \cellcolor{oomred}OOM & \cellcolor{oomred}OOM & \cellcolor{oomred}OOM & \cellcolor{oomred}OOM & \cellcolor{oomred}OOM & \cellcolor{oomred}OOM & \cellcolor{oomred}OOM & \cellcolor{oomred}OOM & \cellcolor{oomred}OOM & \cellcolor{oomred}OOM \\
Fast3R        & 647.55 & 0.331 & 0.814 & 2.162 & 0.152 & 0.232 & 0.376 & 0.191 & 0.304 & 0.071 & 0.116 & 0.047 & 0.135 & 0.232 & \textbf{13.68} & 3.064 & 12.04 & 0.224 & 0.289 \\
FastVGGT      & 1157.94 & \cellcolor{secondorange}0.120 & \cellcolor{secondorange}\textbf{0.102} & \cellcolor{secondorange}\textbf{0.685} & \cellcolor{bestred}\textbf{0.421} & \cellcolor{secondorange}0.552 & \cellcolor{secondorange}0.704 & \cellcolor{bestred}\textbf{0.609} & \textbf{0.716} & \cellcolor{secondorange}\textbf{0.368} & \cellcolor{secondorange}\textbf{0.456} & \cellcolor{secondorange}\textbf{0.306} & \cellcolor{secondorange}\textbf{0.473} & \textbf{0.588} & 19.23 & 1.145 & \textbf{1.899} & \cellcolor{thirdyellow}\textbf{0.479} & 0.185 \\
MUSt3R        & 423.43 & \cellcolor{oomred}T.O & \cellcolor{oomred}T.O & \cellcolor{oomred}T.O & \cellcolor{oomred}T.O & \cellcolor{oomred}T.O & \cellcolor{oomred}T.O & \cellcolor{oomred}T.O & \cellcolor{oomred}T.O & \cellcolor{oomred}T.O & \cellcolor{oomred}T.O & \cellcolor{oomred}T.O & \cellcolor{oomred}T.O & \cellcolor{oomred}T.O & \cellcolor{oomred}T.O & \cellcolor{oomred}T.O & \cellcolor{oomred}T.O & \cellcolor{oomred}T.O & \cellcolor{oomred}T.O \\
MapAnything   & 1228.49 & \cellcolor{oomred}OOM & \cellcolor{oomred}OOM & \cellcolor{oomred}OOM & \cellcolor{oomred}OOM & \cellcolor{oomred}OOM & \cellcolor{oomred}OOM & \cellcolor{oomred}OOM & \cellcolor{oomred}OOM & \cellcolor{oomred}OOM & \cellcolor{oomred}OOM & \cellcolor{oomred}OOM & \cellcolor{oomred}OOM & \cellcolor{oomred}OOM & \cellcolor{oomred}OOM & \cellcolor{oomred}OOM & \cellcolor{oomred}OOM & \cellcolor{oomred}OOM & \cellcolor{oomred}OOM \\
OmniVGGT      & 1217.49 & \cellcolor{oomred}OOM & \cellcolor{oomred}OOM & \cellcolor{oomred}OOM & \cellcolor{oomred}OOM & \cellcolor{oomred}OOM & \cellcolor{oomred}OOM & \cellcolor{oomred}OOM & \cellcolor{oomred}OOM & \cellcolor{oomred}OOM & \cellcolor{oomred}OOM & \cellcolor{oomred}OOM & \cellcolor{oomred}OOM & \cellcolor{oomred}OOM & \cellcolor{oomred}OOM & \cellcolor{oomred}OOM & \cellcolor{oomred}OOM & \cellcolor{oomred}OOM & \cellcolor{oomred}OOM \\
$\pi^{3}$     & 958.70 & \cellcolor{bestred}\textbf{0.109} & 0.268 & 1.008 & \cellcolor{secondorange}0.403 & \cellcolor{bestred}\textbf{0.553} & \cellcolor{bestred}\textbf{0.738} & 0.528 & 0.627 & 0.300 & 0.379 & 0.254 & 0.410 & 0.524 & 16.39 & \textbf{1.132} & 2.237 & 0.332 & 0.757 \\
$\pi^{3}$-X   & 1360.03 & \cellcolor{oomred}OOM & \cellcolor{oomred}OOM & \cellcolor{oomred}OOM & \cellcolor{oomred}OOM & \cellcolor{oomred}OOM & \cellcolor{oomred}OOM & \cellcolor{oomred}OOM & \cellcolor{oomred}OOM & \cellcolor{oomred}OOM & \cellcolor{oomred}OOM & \cellcolor{oomred}OOM & \cellcolor{oomred}OOM & \cellcolor{oomred}OOM & \cellcolor{oomred}OOM & \cellcolor{oomred}OOM & \cellcolor{oomred}OOM & \cellcolor{oomred}OOM & \cellcolor{oomred}OOM \\
AMB3R         & 1563.12 & \cellcolor{oomred}OOM & \cellcolor{oomred}OOM & \cellcolor{oomred}OOM & \cellcolor{oomred}OOM & \cellcolor{oomred}OOM & \cellcolor{oomred}OOM & \cellcolor{oomred}OOM & \cellcolor{oomred}OOM & \cellcolor{oomred}OOM & \cellcolor{oomred}OOM & \cellcolor{oomred}OOM & \cellcolor{oomred}OOM & \cellcolor{oomred}OOM & \cellcolor{oomred}OOM & \cellcolor{oomred}OOM & \cellcolor{oomred}OOM & \cellcolor{oomred}OOM & \cellcolor{oomred}OOM \\
DA3-Small     & 34.30 & 0.208 & 0.280 & 1.226 & 0.249 & 0.368 & 0.556 & 0.297 & 0.435 & 0.130 & 0.215 & 0.088 & 0.243 & 0.368 & 28.12 & 3.996 & 5.154 & 0.325 & 0.187 \\
DA3-Base      & 135.37 & 0.166 & 0.220 & 1.093 & 0.327 & 0.454 & 0.632 & 0.378 & 0.509 & 0.198 & 0.285 & 0.151 & 0.310 & 0.436 & 26.35 & 3.818 & 6.089 & 0.399 & \textbf{0.146} \\
DA3-Large     & 410.94 & \cellcolor{oomred}OOM & \cellcolor{oomred}OOM & \cellcolor{oomred}OOM & \cellcolor{oomred}OOM & \cellcolor{oomred}OOM & \cellcolor{oomred}OOM & \cellcolor{oomred}OOM & \cellcolor{oomred}OOM & \cellcolor{oomred}OOM & \cellcolor{oomred}OOM & \cellcolor{oomred}OOM & \cellcolor{oomred}OOM & \cellcolor{oomred}OOM & \cellcolor{oomred}OOM & \cellcolor{oomred}OOM & \cellcolor{oomred}OOM & \cellcolor{oomred}OOM & \cellcolor{oomred}OOM \\
DA3-Giant     & 1355.67 & \cellcolor{oomred}OOM & \cellcolor{oomred}OOM & \cellcolor{oomred}OOM & \cellcolor{oomred}OOM & \cellcolor{oomred}OOM & \cellcolor{oomred}OOM & \cellcolor{oomred}OOM & \cellcolor{oomred}OOM & \cellcolor{oomred}OOM & \cellcolor{oomred}OOM & \cellcolor{oomred}OOM & \cellcolor{oomred}OOM & \cellcolor{oomred}OOM & \cellcolor{oomred}OOM & \cellcolor{oomred}OOM & \cellcolor{oomred}OOM & \cellcolor{oomred}OOM & \cellcolor{oomred}OOM \\
DA3-Nested    & 1689.85 & \cellcolor{oomred}OOM & \cellcolor{oomred}OOM & \cellcolor{oomred}OOM & \cellcolor{oomred}OOM & \cellcolor{oomred}OOM & \cellcolor{oomred}OOM & \cellcolor{oomred}OOM & \cellcolor{oomred}OOM & \cellcolor{oomred}OOM & \cellcolor{oomred}OOM & \cellcolor{oomred}OOM & \cellcolor{oomred}OOM & \cellcolor{oomred}OOM & \cellcolor{oomred}OOM & \cellcolor{oomred}OOM & \cellcolor{oomred}OOM & \cellcolor{oomred}OOM & \cellcolor{oomred}OOM \\
WorldMirror   & 1263.34 & \cellcolor{oomred}OOM & \cellcolor{oomred}OOM & \cellcolor{oomred}OOM & \cellcolor{oomred}OOM & \cellcolor{oomred}OOM & \cellcolor{oomred}OOM & \cellcolor{oomred}OOM & \cellcolor{oomred}OOM & \cellcolor{oomred}OOM & \cellcolor{oomred}OOM & \cellcolor{oomred}OOM & \cellcolor{oomred}OOM & \cellcolor{oomred}OOM & \cellcolor{oomred}OOM & \cellcolor{oomred}OOM & \cellcolor{oomred}OOM & \cellcolor{oomred}OOM & \cellcolor{oomred}OOM \\
\ours   & 1303.76 & \cellcolor{oomred}OOM & \cellcolor{oomred}OOM & \cellcolor{oomred}OOM & \cellcolor{oomred}OOM & \cellcolor{oomred}OOM & \cellcolor{oomred}OOM & \cellcolor{oomred}OOM & \cellcolor{oomred}OOM & \cellcolor{oomred}OOM & \cellcolor{oomred}OOM & \cellcolor{oomred}OOM & \cellcolor{oomred}OOM & \cellcolor{oomred}OOM & \cellcolor{oomred}OOM & \cellcolor{oomred}OOM & \cellcolor{oomred}OOM & \cellcolor{oomred}OOM & \cellcolor{oomred}OOM \\
\midrule

\multicolumn{20}{c}{\cellcolor{catgray}\textbf{Online}} \\
\midrule
Spann3r       & 658.69 & 0.315 & 97.78 & 5.152 & 0.136 & 0.217 & 0.373 & 0.143 & 0.260 & 0.050 & 0.099 & 0.023 & 0.125 & 0.246 & 26.48 & 3.257 & 5.064 & 0.159 & 0.322 \\
CUT3R         & 793.31 & 0.260 & 0.497 & 1.346 & 0.192 & 0.290 & 0.458 & 0.130 & 0.200 & 0.043 & 0.079 & 0.023 & 0.080 & 0.165 & 25.54 & 0.484 & 1.180 & 0.109 & 0.497 \\
MonST3R       & 571.17 & \cellcolor{oomred}OOM & \cellcolor{oomred}OOM & \cellcolor{oomred}OOM & \cellcolor{oomred}OOM & \cellcolor{oomred}OOM & \cellcolor{oomred}OOM & \cellcolor{oomred}OOM & \cellcolor{oomred}OOM & \cellcolor{oomred}OOM & \cellcolor{oomred}OOM & \cellcolor{oomred}OOM & \cellcolor{oomred}OOM & \cellcolor{oomred}OOM & \cellcolor{oomred}OOM & \cellcolor{oomred}OOM & \cellcolor{oomred}OOM & \cellcolor{oomred}OOM & \cellcolor{oomred}OOM \\
Point3R       & 828.01 & 0.285 & 1.066 & 1.450 & 0.230 & 0.339 & 0.504 & 0.148 & 0.263 & 0.035 & 0.080 & 0.015 & 0.104 & 0.212 & 28.09 & 1.286 & 3.057 & 0.139 & 0.299 \\
Stream3R-S    & 1190.60 & \cellcolor{oomred}OOM & \cellcolor{oomred}OOM & \cellcolor{oomred}OOM & \cellcolor{oomred}OOM & \cellcolor{oomred}OOM & \cellcolor{oomred}OOM & \cellcolor{oomred}OOM & \cellcolor{oomred}OOM & \cellcolor{oomred}OOM & \cellcolor{oomred}OOM & \cellcolor{oomred}OOM & \cellcolor{oomred}OOM & \cellcolor{oomred}OOM & \cellcolor{oomred}OOM & \cellcolor{oomred}OOM & \cellcolor{oomred}OOM & \cellcolor{oomred}OOM & \cellcolor{oomred}OOM \\
Stream3R-W    & 1190.60 & \cellcolor{oomred}OOM & \cellcolor{oomred}OOM & \cellcolor{oomred}OOM & \cellcolor{oomred}OOM & \cellcolor{oomred}OOM & \cellcolor{oomred}OOM & \cellcolor{oomred}OOM & \cellcolor{oomred}OOM & \cellcolor{oomred}OOM & \cellcolor{oomred}OOM & \cellcolor{oomred}OOM & \cellcolor{oomred}OOM & \cellcolor{oomred}OOM & \cellcolor{oomred}OOM & \cellcolor{oomred}OOM & \cellcolor{oomred}OOM & \cellcolor{oomred}OOM & \cellcolor{oomred}OOM \\
StreamVGGT    & 1256.54 & 0.198 & 0.590 & 1.715 & 0.179 & 0.279 & 0.460 & 0.414 & 0.565 & 0.152 & 0.244 & 0.111 & 0.278 & 0.413 & 26.9 & 1.719 & 1.785 & 0.251 & 0.197 \\
Page4D        & 1256.81 & \cellcolor{oomred}OOM & \cellcolor{oomred}OOM & \cellcolor{oomred}OOM & \cellcolor{oomred}OOM & \cellcolor{oomred}OOM & \cellcolor{oomred}OOM & \cellcolor{oomred}OOM & \cellcolor{oomred}OOM & \cellcolor{oomred}OOM & \cellcolor{oomred}OOM & \cellcolor{oomred}OOM & \cellcolor{oomred}OOM & \cellcolor{oomred}OOM & \cellcolor{oomred}OOM & \cellcolor{oomred}OOM & \cellcolor{oomred}OOM & \cellcolor{oomred}OOM & \cellcolor{oomred}OOM \\
InfiniteVGGT  & 1256.54 & 0.197 & 0.585 & 1.703 & 0.179 & 0.280 & 0.461 & 0.416 & 0.566 & 0.154 & 0.246 & 0.112 & 0.280 & 0.416 & 27.01 & 1.717 & 1.750 & 0.254 & \textbf{0.197} \\
Wint3R        & 749.46 & 0.234 & 0.280 & 1.215 & 0.179 & 0.278 & 0.466 & 0.167 & 0.262 & 0.042 & 0.083 & 0.021 & 0.098 & 0.202 & 27.8 & 0.874 & 1.213 & 0.114 & 0.475 \\
LongStream-B  & 1190.60 & 0.269 & 0.225 & 1.013 & 0.158 & 0.245 & 0.411 & 0.165 & 0.273 & 0.097 & 0.158 & 0.039 & 0.149 & 0.294 & 5.766 & 0.105 & 0.702 & 0.083 & 0.410 \\
LongStream-S  & 1190.60 & 0.279 & 0.230 & 0.998 & 0.170 & 0.261 & 0.426 & 0.130 & 0.210 & 0.076 & 0.126 & 0.026 & 0.107 & 0.218 & 10.08 & 0.186 & 1.038 & 0.083 & 0.438 \\
LingbotMap$^{*}$-W    & 1157.94 & 0.167 & 0.211 & 0.976 & 0.265 & 0.392 & 0.572 & 0.524 & 0.678 & 0.294 & 0.393 & 0.229 & 0.414 & 0.553 & 4.694 & \textbf{0.098} & \textbf{0.502} & 0.352 & 0.383 \\
LingbotMap$^{*}$-S    & 1157.94 & \cellcolor{thirdyellow}\textbf{0.139} & \textbf{0.209} & \textbf{0.958} & \cellcolor{thirdyellow}\textbf{0.384} & \cellcolor{thirdyellow}\textbf{0.516} & \cellcolor{thirdyellow}\textbf{0.677} & \cellcolor{secondorange}\textbf{0.602} & \cellcolor{thirdyellow}\textbf{0.724} & \cellcolor{bestred}\textbf{0.382} & \cellcolor{bestred}\textbf{0.477} & \cellcolor{bestred}\textbf{0.308} & \cellcolor{bestred}\textbf{0.496} & \cellcolor{bestred}\textbf{0.627} & \cellcolor{secondorange}\textbf{3.470} & 0.328 & 0.749 & \textbf{0.472} & 0.296 \\
\midrule

\multicolumn{20}{c}{\cellcolor{catgray}\textbf{Chunk-wise}} \\
\midrule
VGGT-Long       & 1256.54 & 0.222 & 1.006 & \textbf{0.986} & 0.310 & 0.428 & 0.582 & 0.464 & 0.593 & 0.257 & 0.347 & 0.211 & 0.379 & 0.507 & 8.467 & 0.149 & 0.693 & 0.467 & \cellcolor{bestred}\textbf{0.142} \\
$\pi^{3}$-Long  & 958.70 & \textbf{0.216} & \textbf{0.375} & 1.464 & 0.121 & 0.203 & 0.363 & \cellcolor{thirdyellow}\textbf{0.595} & \cellcolor{bestred}\textbf{0.741} & 0.321 & \cellcolor{thirdyellow}\textbf{0.437} & 0.253 & \cellcolor{thirdyellow}\textbf{0.469} & \cellcolor{secondorange}\textbf{0.614} & \cellcolor{thirdyellow}\textbf{4.021} & \cellcolor{thirdyellow}\textbf{0.093} & \cellcolor{thirdyellow}\textbf{0.396} & 0.251 & 0.223 \\
DA3-Streaming   & 1355.67 & 0.245 & 22.56 & 2.475 & \textbf{0.379} & \textbf{0.502} & \textbf{0.666} & 0.513 & 0.625 & \cellcolor{thirdyellow}\textbf{0.331} & 0.405 & \cellcolor{thirdyellow}\textbf{0.277} & 0.427 & 0.546 & 8.575 & 0.119 & 0.588 & \cellcolor{bestred}\textbf{0.516} & 0.162 \\
\midrule

\multicolumn{20}{c}{\cellcolor{catgray}\textbf{SLAM-based}} \\
\midrule
MASt3R-SLAM   & 688.64 & 0.404 & 2.171 & 2.920 & 0.086 & 0.140 & 0.265 & 0.357 & 0.475 & 0.117 & 0.180 & 0.088 & 0.207 & 0.311 & 25.7 & 0.413 & 1.983 & 0.121 & 0.493 \\
VGGT-SLAM     & 1256.54 & \textbf{0.211} & \textbf{0.366} & \textbf{1.051} & \textbf{0.266} & \textbf{0.373} & \textbf{0.530} & \textbf{0.381} & \textbf{0.510} & \textbf{0.206} & \textbf{0.284} & \textbf{0.159} & \textbf{0.309} & \textbf{0.441} & \textbf{9.069} & \textbf{0.152} & \textbf{0.626} & \textbf{0.384} & \textbf{0.160} \\
\midrule

\multicolumn{20}{c}{\cellcolor{catgray}\textbf{Test-Time Training}} \\
\midrule
TTT3R    & 793.31 & 0.222 & 0.353 & 1.222 & 0.220 & 0.331 & 0.507 & 0.221 & 0.351 & 0.102 & 0.172 & 0.064 & 0.193 & 0.321 & 21.07 & 0.569 & 1.234 & 0.173 & 0.283 \\
Scal3R   & 1266.14 & 0.244 & 0.280 & 1.127 & 0.124 & 0.197 & 0.353 & 0.407 & 0.552 & 0.247 & 0.335 & 0.161 & 0.328 & 0.480 & \cellcolor{bestred}\textbf{2.396} & 0.111 & 0.864 & \cellcolor{secondorange}\textbf{0.498} & \cellcolor{secondorange}\textbf{0.142} \\
LoGeR    & 1254.62 & 0.197 & \cellcolor{thirdyellow}0.112 & \cellcolor{thirdyellow}0.741 & 0.225 & 0.345 & 0.524 & 0.535 & 0.677 & 0.249 & 0.345 & 0.197 & 0.391 & 0.552 & 5.217 & \cellcolor{secondorange}0.090 & \cellcolor{secondorange}0.385 & 0.335 & 0.165 \\
LoGeR$^*$ & 1254.60 & \textbf{0.156} & \cellcolor{bestred}\textbf{0.086} & \cellcolor{bestred}\textbf{0.684} & \textbf{0.304} & \textbf{0.435} & \textbf{0.611} & \textbf{0.590} & \cellcolor{secondorange}\textbf{0.725} & \textbf{0.317} & \textbf{0.411} & \textbf{0.256} & \textbf{0.446} & \cellcolor{thirdyellow}\textbf{0.598} & 4.598 & \cellcolor{bestred}\textbf{0.077} & \cellcolor{bestred}\textbf{0.347} & 0.421 & \cellcolor{thirdyellow}0.145 \\
\bottomrule
\end{tabular}%
}
\end{table*}

\clearpage

%
%
%
%

\begin{table*}[t]
\centering
\setlength{\fboxsep}{0pt}
\caption{\textbf{Per-Dataset Results on 7Scenes}.
Performance across different input regimes: \textit{Single Frame}, \textit{Sparse}, \textit{Medium}, \textit{Dense}, and the \textit{Average}.
The best, second-best, and third-best results in each column are highlighted in
\colorbox{bestred}{deep blue}, \colorbox{secondorange}{medium blue}, and \colorbox{thirdyellow}{light blue}, respectively.
Out-of-memory (OOM) and Timeout (T.O) cells are shaded \colorbox{oomred}{light red};
\textit{Average} values for those rows are wrapped in parentheses and excluded from per-column ranking.
Within each sub-category, the \textbf{bold} value marks the in-group best.
Note that \ours (Ours) is excluded from the per-column rankings.
}
\label{tab:dataset_7scenes}
\resizebox{\textwidth}{!}{%
\renewcommand{\arraystretch}{1.15}
\setlength{\tabcolsep}{3pt}
\begin{tabular}{l c >{\columncolor{subcol}}c >{\columncolor{subcol}}c >{\columncolor{subcol}}c >{\columncolor{subcol}}c >{\columncolor{subcol}}c >{\columncolor{subcol}}c >{\columncolor{subcol}}c >{\columncolor{subcol}}c >{\columncolor{subcol}}c >{\columncolor{subcol}}c >{\columncolor{subcol}}c >{\columncolor{subcol}}c >{\columncolor{subcol}}c >{\columncolor{subcol}}c >{\columncolor{subcol}}c}
\toprule
\multirow{2.5}{*}{\textbf{Method}}
& \multirow{2.5}{*}{\makecell{\textbf{\#Params}\\\textbf{(M)}}}
& \multicolumn{1}{c}{\textbf{Single Frame}}
& \multicolumn{2}{c}{\textbf{Sparse}}
& \multicolumn{4}{c}{\textbf{Medium}}
& \multicolumn{4}{c}{\textbf{Dense}}
& \multicolumn{4}{c}{\textbf{Average}} \\
\cmidrule(lr){3-3} \cmidrule(lr){4-5} \cmidrule(lr){6-9} \cmidrule(lr){10-13} \cmidrule(lr){14-17}
 & &
 AbsRel$\downarrow$ &
 AbsRel$\downarrow$ & AUC@30$\uparrow$ &
 AbsRel$\downarrow$ & AUC@30$\uparrow$ & ATE$\downarrow$ & F-Score$\uparrow$ &
 AbsRel$\downarrow$ & AUC@30$\uparrow$ & ATE$\downarrow$ & F-Score$\uparrow$ &
 AbsRel$\downarrow$ & AUC@30$\uparrow$ & ATE$\downarrow$ & F-Score$\uparrow$ \\
\midrule
\multicolumn{17}{c}{\cellcolor{catgray}\textbf{Optimization-based}} \\
\midrule
 DUSt3R & 571.2 & \textbf{0.084} & \textbf{0.075} & \textbf{0.641} & \textbf{0.083} & 0.591 & \textbf{0.107} & 0.281 & \cellcolor{oomred}OOM & \cellcolor{oomred}OOM & \cellcolor{oomred}OOM & \cellcolor{oomred}OOM & (0.079) & (0.616) & (0.107) & (0.281) \\
 MASt3R & 688.6 & 0.111 & 0.085 & 0.591 & 0.105 & \textbf{0.614} & 0.121 & \textbf{0.294} & \cellcolor{oomred}OOM & \cellcolor{oomred}OOM & \cellcolor{oomred}OOM & \cellcolor{oomred}OOM & (0.095) & (0.603) & (0.121) & (0.294) \\
\midrule
\multicolumn{17}{c}{\cellcolor{catgray}\textbf{End-to-End Feed-Forward}} \\
\midrule
 VGGT & 1257 & 0.074 & 0.068 & 0.715 & 0.064 & 0.777 & 0.070 & 0.463 & \cellcolor{oomred}OOM & \cellcolor{oomred}OOM & \cellcolor{oomred}OOM & \cellcolor{oomred}OOM & (0.066) & (0.746) & (0.070) & (0.463) \\
 Fast3R & 647.5 & 0.104 & 0.218 & 0.428 & 0.087 & 0.603 & 0.201 & 0.216 & 0.124 & 0.341 & 0.321 & 0.075 & 0.143 & 0.457 & 0.261 & 0.146 \\
 FastVGGT & 1158 & 0.075 & 0.072 & 0.604 & 0.062 & 0.771 & 0.071 & 0.431 & \cellcolor{secondorange}0.066 & \cellcolor{thirdyellow}0.732 & 0.115 & \cellcolor{secondorange}0.401 & \cellcolor{thirdyellow}0.067 & 0.703 & 0.093 & \cellcolor{bestred}\textbf{0.416} \\
 MUSt3R & 423.4 & 0.072 & 0.077 & 0.557 & 0.063 & 0.748 & 0.077 & 0.383 & \cellcolor{oomred}T.O & \cellcolor{oomred}T.O & \cellcolor{oomred}T.O & \cellcolor{oomred}T.O & (0.070) & (0.652) & (0.077) & (0.383) \\
 MapAnything & 1228 & 0.076 & 0.073 & 0.642 & 0.066 & 0.732 & 0.072 & 0.378 & \cellcolor{oomred}OOM & \cellcolor{oomred}OOM & \cellcolor{oomred}OOM & \cellcolor{oomred}OOM & (0.070) & (0.687) & (0.072) & (0.378) \\
 OmniVGGT & 1217 & \cellcolor{thirdyellow}\textbf{0.062} & 0.063 & 0.673 & \cellcolor{secondorange}0.058 & 0.753 & 0.081 & \cellcolor{thirdyellow}\textbf{0.469} & \cellcolor{oomred}OOM & \cellcolor{oomred}OOM & \cellcolor{oomred}OOM & \cellcolor{oomred}OOM & (0.061) & (0.713) & (0.081) & (0.469) \\
 $\pi^{3}$ & 958.7 & 0.065 & 0.063 & 0.663 & \cellcolor{thirdyellow}0.059 & 0.769 & 0.063 & 0.393 & 0.074 & 0.161 & 0.427 & 0.052 & \cellcolor{secondorange}0.065 & 0.531 & 0.245 & 0.222 \\
 $\pi^{3}$-X & 1360 & 0.069 & 0.064 & 0.693 & 0.061 & 0.776 & 0.063 & 0.442 & \cellcolor{oomred}OOM & \cellcolor{oomred}OOM & \cellcolor{oomred}OOM & \cellcolor{oomred}OOM & (0.062) & (0.735) & (0.063) & (0.442) \\
 AMB3R & 1563 & 0.063 & \cellcolor{secondorange}0.060 & 0.680 & \cellcolor{secondorange}0.058 & 0.782 & 0.063 & \cellcolor{secondorange}0.490 & \cellcolor{oomred}OOM & \cellcolor{oomred}OOM & \cellcolor{oomred}OOM & \cellcolor{oomred}OOM & (0.059) & (0.731) & (0.063) & (0.490) \\
 DA3-Small & 34.3 & 0.090 & 0.082 & 0.549 & 0.067 & 0.699 & 0.096 & 0.404 & \cellcolor{thirdyellow}0.070 & 0.671 & 0.103 & \cellcolor{thirdyellow}0.399 & 0.073 & 0.640 & 0.099 & \cellcolor{thirdyellow}0.401 \\
 DA3-Base & 135.4 & 0.088 & 0.075 & 0.675 & 0.061 & 0.743 & 0.079 & 0.395 & \cellcolor{bestred}\textbf{0.060} & \cellcolor{secondorange}\textbf{0.735} & \textbf{0.086} & \cellcolor{bestred}\textbf{0.405} & \cellcolor{secondorange}\textbf{0.065} & \textbf{0.718} & \textbf{0.082} & 0.400 \\
 DA3-Large & 410.9 & 0.077 & 0.074 & 0.760 & \cellcolor{thirdyellow}0.059 & \cellcolor{thirdyellow}0.789 & 0.062 & 0.464 & \cellcolor{oomred}OOM & \cellcolor{oomred}OOM & \cellcolor{oomred}OOM & \cellcolor{oomred}OOM & (0.066) & (0.774) & (0.062) & (0.464) \\
 DA3-Giant & 1356 & 0.066 & 0.066 & \cellcolor{thirdyellow}0.780 & \cellcolor{secondorange}\textbf{0.058} & \cellcolor{secondorange}\textbf{0.790} & \cellcolor{thirdyellow}\textbf{0.059} & 0.439 & \cellcolor{oomred}OOM & \cellcolor{oomred}OOM & \cellcolor{oomred}OOM & \cellcolor{oomred}OOM & (0.062) & (0.785) & (0.059) & (0.439) \\
 DA3-Nested & 1690 & 0.067 & 0.064 & \cellcolor{secondorange}\textbf{0.784} & \cellcolor{secondorange}0.058 & \cellcolor{secondorange}0.790 & 0.060 & 0.435 & \cellcolor{oomred}OOM & \cellcolor{oomred}OOM & \cellcolor{oomred}OOM & \cellcolor{oomred}OOM & (0.061) & (0.787) & (0.060) & (0.435) \\
 WorldMirror & 1263 & 0.068 & 0.066 & 0.695 & \cellcolor{thirdyellow}0.059 & 0.770 & 0.068 & 0.463 & \cellcolor{oomred}OOM & \cellcolor{oomred}OOM & \cellcolor{oomred}OOM & \cellcolor{oomred}OOM & (0.063) & (0.733) & (0.068) & (0.463) \\
 VGGT-Omega & 1144 & \cellcolor{secondorange}\textbf{0.059} & \cellcolor{bestred}\textbf{0.056} & \cellcolor{bestred}\textbf{0.808} & \cellcolor{bestred}\textbf{0.049} & \cellcolor{bestred}\textbf{0.858} & \cellcolor{bestred}\textbf{0.035} & \cellcolor{bestred}\textbf{0.582} & -- & -- & -- & -- & (0.052) & (0.833) & (0.035) & (0.582) \\ \hdashline
 \ours (Ours) & 1304 & 0.066 & 0.063 & 0.744 & 0.059 & 0.790 & 0.063 & 0.446 & \cellcolor{oomred}OOM & \cellcolor{oomred}OOM & \cellcolor{oomred}OOM & \cellcolor{oomred}OOM & (0.063) & (0.767) & (0.063) & (0.446) \\
\midrule
\multicolumn{17}{c}{\cellcolor{catgray}\textbf{Online}} \\
\midrule
 Spann3r$^{224}$ & 658.7 & 0.076 & 0.096 & 0.429 & 0.086 & 0.477 & 0.195 & 0.147 & 0.102 & 0.452 & 0.224 & 0.138 & 0.095 & 0.452 & 0.209 & 0.143 \\
 CUT3R & 793.3 & 0.072 & 0.073 & 0.640 & 0.081 & 0.648 & 0.123 & 0.318 & 0.105 & 0.069 & 0.498 & 0.055 & 0.087 & 0.452 & 0.311 & 0.187 \\
 MonST3R & 571.2 & 0.104 & 0.097 & 0.270 & 0.092 & 0.345 & 0.192 & 0.151 & \cellcolor{oomred}OOM & \cellcolor{oomred}OOM & \cellcolor{oomred}OOM & \cellcolor{oomred}OOM & (0.095) & (0.307) & (0.192) & (0.151) \\
 Point3R & 828 & 0.072 & 0.080 & 0.537 & 0.083 & 0.588 & 0.120 & 0.225 & 0.093 & 0.427 & 0.239 & 0.094 & 0.086 & 0.517 & 0.179 & 0.159 \\
 Stream3R-S & 1191 & 0.079 & 0.078 & 0.614 & 0.437 & 0.236 & 0.499 & 0.040 & \cellcolor{oomred}OOM & \cellcolor{oomred}OOM & \cellcolor{oomred}OOM & \cellcolor{oomred}OOM & (0.257) & (0.425) & (0.499) & (0.040) \\
 Stream3R-W & 1191 & 0.079 & 0.078 & 0.614 & 0.308 & 0.214 & 0.537 & 0.048 & \cellcolor{oomred}OOM & \cellcolor{oomred}OOM & \cellcolor{oomred}OOM & \cellcolor{oomred}OOM & (0.193) & (0.414) & (0.537) & (0.048) \\
 StreamVGGT & 1257 & 0.069 & 0.081 & 0.598 & 0.073 & 0.740 & 0.086 & 0.307 & \textbf{0.085} & 0.659 & 0.130 & 0.278 & 0.080 & 0.666 & 0.108 & 0.292 \\
 Page4D & 1257 & 0.069 & \textbf{0.065} & 0.632 & \textbf{0.062} & 0.743 & 0.079 & 0.373 & \cellcolor{oomred}OOM & \cellcolor{oomred}OOM & \cellcolor{oomred}OOM & \cellcolor{oomred}OOM & (0.064) & (0.687) & (0.079) & (0.373) \\
 InfiniteVGGT & 1257 & 0.069 & 0.081 & 0.600 & 0.073 & 0.741 & 0.086 & 0.310 & 0.085 & 0.658 & 0.131 & 0.283 & 0.080 & 0.666 & 0.109 & 0.297 \\
 Wint3R & 749.5 & 0.070 & 0.075 & 0.598 & 0.065 & 0.676 & 0.097 & 0.358 & 0.106 & 0.241 & 0.318 & 0.054 & 0.082 & 0.505 & 0.207 & 0.206 \\
 LongStream-B & 1191 & \cellcolor{bestred}\textbf{0.057} & 0.067 & 0.605 & 0.087 & 0.683 & 0.094 & 0.100 & 0.087 & 0.362 & 0.151 & 0.060 & 0.081 & 0.550 & 0.123 & 0.080 \\
 LongStream-S & 1191 & \cellcolor{bestred}\textbf{0.057} & 0.067 & 0.605 & 0.071 & 0.442 & 0.150 & 0.040 & 0.085 & 0.293 & 0.180 & 0.042 & \textbf{0.075} & 0.447 & 0.165 & 0.041 \\
 LingbotMap$^{*}$-W & 1158 & 0.076 & 0.069 & \textbf{0.702} & 0.085 & 0.758 & 0.070 & 0.340 & 0.111 & 0.674 & 0.107 & 0.220 & 0.088 & 0.711 & 0.088 & 0.280 \\
 LingbotMap$^{*}$-S & 1158 & 0.076 & 0.069 & \textbf{0.702} & 0.081 & \textbf{0.770} & \textbf{0.068} & \textbf{0.375} & 0.092 & \cellcolor{bestred}\textbf{0.769} & \cellcolor{bestred}\textbf{0.069} & \textbf{0.300} & 0.081 & \cellcolor{secondorange}\textbf{0.747} & \cellcolor{bestred}\textbf{0.069} & \textbf{0.337} \\
\midrule
\multicolumn{17}{c}{\cellcolor{catgray}\textbf{Chunk-wise}} \\
\midrule
 VGGT-Long & 1257 & 0.074 & 0.068 & 0.715 & 0.065 & 0.777 & 0.063 & 0.420 & 0.078 & 0.645 & 0.101 & 0.332 & 0.070 & 0.712 & 0.082 & 0.376 \\
 $\pi^{3}$-Long & 958.7 & \textbf{0.065} & \textbf{0.063} & 0.663 & 0.078 & 0.782 & \cellcolor{secondorange}\textbf{0.058} & 0.310 & 0.118 & \textbf{0.679} & 0.091 & 0.213 & 0.086 & 0.708 & \cellcolor{thirdyellow}0.074 & 0.261 \\
 DA3-Streaming & 1356 & 0.066 & 0.066 & \cellcolor{thirdyellow}\textbf{0.780} & \cellcolor{secondorange}\textbf{0.058} & \textbf{0.788} & \cellcolor{thirdyellow}0.059 & \textbf{0.439} & \cellcolor{secondorange}\textbf{0.066} & 0.675 & \cellcolor{thirdyellow}\textbf{0.084} & \textbf{0.336} & \cellcolor{bestred}\textbf{0.063} & \cellcolor{bestred}\textbf{0.748} & \cellcolor{secondorange}\textbf{0.071} & \textbf{0.388} \\
\midrule
\multicolumn{17}{c}{\cellcolor{catgray}\textbf{SLAM-based}} \\
\midrule
 MASt3R-SLAM & 688.6 & 0.141 & 0.185 & 0.193 & 0.181 & 0.662 & 0.095 & 0.209 & 0.196 & \textbf{0.648} & \textbf{0.099} & 0.178 & 0.187 & 0.501 & \textbf{0.097} & 0.194 \\
 VGGT-SLAM & 1257 & \textbf{0.074} & \textbf{0.068} & \textbf{0.715} & \textbf{0.069} & \textbf{0.733} & \textbf{0.075} & \textbf{0.382} & \textbf{0.084} & 0.576 & 0.125 & \textbf{0.270} & \textbf{0.074} & \textbf{0.675} & 0.100 & \textbf{0.326} \\
\midrule
\multicolumn{17}{c}{\cellcolor{catgray}\textbf{Test-Time Training}} \\
\midrule
 TTT3R & 793.3 & 0.072 & 0.077 & 0.552 & 0.070 & 0.701 & 0.097 & 0.339 & 0.097 & 0.304 & 0.333 & 0.151 & 0.081 & 0.519 & 0.215 & 0.245 \\
 Scal3R & 1266 & 0.073 & 0.069 & 0.668 & 0.133 & 0.688 & 0.068 & 0.402 & 0.140 & 0.531 & 0.097 & 0.288 & 0.114 & 0.629 & 0.083 & 0.345 \\
 LoGeR & 1255 & \textbf{0.066} & \cellcolor{thirdyellow}0.062 & 0.686 & 0.081 & 0.758 & 0.079 & 0.282 & 0.106 & 0.651 & 0.116 & 0.238 & 0.083 & 0.698 & 0.097 & 0.260 \\
 LoGeR$^*$ & 1255 & 0.070 & \cellcolor{secondorange}\textbf{0.060} & \textbf{0.735} & \textbf{0.061} & \textbf{0.765} & \textbf{0.063} & \textbf{0.455} & \textbf{0.079} & \textbf{0.701} & \cellcolor{secondorange}\textbf{0.074} & \textbf{0.370} & \cellcolor{thirdyellow}\textbf{0.067} & \cellcolor{thirdyellow}\textbf{0.734} & \cellcolor{bestred}\textbf{0.069} & \cellcolor{secondorange}\textbf{0.413} \\
\bottomrule
\end{tabular}%
}
\end{table*}

\begin{table*}[t]
\centering
\setlength{\fboxsep}{0pt}
\caption{\textbf{Per-Dataset Results on ADT}.
Performance across different input regimes: \textit{Single Frame}, \textit{Sparse}, \textit{Medium}, \textit{Dense}, and the \textit{Average}.
The best, second-best, and third-best results in each column are highlighted in
\colorbox{bestred}{deep blue}, \colorbox{secondorange}{medium blue}, and \colorbox{thirdyellow}{light blue}, respectively.
Out-of-memory (OOM) and Timeout (T.O) cells are shaded \colorbox{oomred}{light red};
\textit{Average} values for those rows are wrapped in parentheses and excluded from per-column ranking.
Within each sub-category, the \textbf{bold} value marks the in-group best.
Note that \ours (Ours) is excluded from the per-column rankings.
}
\label{tab:dataset_adt}
\resizebox{\textwidth}{!}{%
\renewcommand{\arraystretch}{1.15}
\setlength{\tabcolsep}{3pt}
\begin{tabular}{l c >{\columncolor{subcol}}c >{\columncolor{subcol}}c >{\columncolor{subcol}}c >{\columncolor{subcol}}c >{\columncolor{subcol}}c >{\columncolor{subcol}}c >{\columncolor{subcol}}c >{\columncolor{subcol}}c >{\columncolor{subcol}}c >{\columncolor{subcol}}c >{\columncolor{subcol}}c >{\columncolor{subcol}}c}
\toprule
\multirow{2.5}{*}{\textbf{Method}}
& \multirow{2.5}{*}{\makecell{\textbf{\#Params}\\\textbf{(M)}}}
& \multicolumn{1}{c}{\textbf{Single Frame}}
& \multicolumn{2}{c}{\textbf{Sparse}}
& \multicolumn{3}{c}{\textbf{Medium}}
& \multicolumn{3}{c}{\textbf{Dense}}
& \multicolumn{3}{c}{\textbf{Average}} \\
\cmidrule(lr){3-3} \cmidrule(lr){4-5} \cmidrule(lr){6-8} \cmidrule(lr){9-11} \cmidrule(lr){12-14}
 & &
 AbsRel$\downarrow$ &
 AbsRel$\downarrow$ & AUC@30$\uparrow$ &
 AbsRel$\downarrow$ & AUC@30$\uparrow$ & ATE$\downarrow$ &
 AbsRel$\downarrow$ & AUC@30$\uparrow$ & ATE$\downarrow$ &
 AbsRel$\downarrow$ & AUC@30$\uparrow$ & ATE$\downarrow$ \\
\midrule
\multicolumn{14}{c}{\cellcolor{catgray}\textbf{Optimization-based}} \\
\midrule
 DUSt3R & 571.2 & 0.160 & 0.117 & 0.882 & 0.120 & 0.762 & 0.117 & \cellcolor{oomred}OOM & \cellcolor{oomred}OOM & \cellcolor{oomred}OOM & (0.119) & (0.822) & (0.117) \\
 MASt3R & 688.6 & \textbf{0.155} & \textbf{0.079} & \textbf{0.905} & \textbf{0.103} & \textbf{0.861} & \textbf{0.065} & \cellcolor{oomred}OOM & \cellcolor{oomred}OOM & \cellcolor{oomred}OOM & (0.091) & (0.883) & (0.065) \\
\midrule
\multicolumn{14}{c}{\cellcolor{catgray}\textbf{End-to-End Feed-Forward}} \\
\midrule
 VGGT & 1257 & 0.133 & 0.072 & 0.787 & 0.073 & 0.840 & 0.122 & \cellcolor{oomred}OOM & \cellcolor{oomred}OOM & \cellcolor{oomred}OOM & (0.072) & (0.813) & (0.122) \\
 Fast3R & 647.5 & 0.172 & 0.147 & 0.597 & 0.189 & 0.611 & 0.710 & 0.220 & 0.380 & 1.032 & 0.185 & 0.529 & 0.871 \\
 FastVGGT & 1158 & 0.133 & 0.064 & 0.776 & 0.068 & 0.873 & 0.097 & \cellcolor{thirdyellow}0.065 & \cellcolor{thirdyellow}0.825 & 0.083 & 0.066 & 0.825 & 0.090 \\
 MUSt3R & 423.4 & 0.168 & 0.074 & \cellcolor{thirdyellow}0.915 & 0.126 & 0.941 & 0.029 & \cellcolor{oomred}T.O & \cellcolor{oomred}T.O & \cellcolor{oomred}T.O & (0.100) & (0.928) & (0.029) \\
 MapAnything & 1228 & 0.108 & 0.066 & 0.833 & 0.073 & 0.867 & 0.093 & \cellcolor{oomred}OOM & \cellcolor{oomred}OOM & \cellcolor{oomred}OOM & (0.069) & (0.850) & (0.093) \\
 OmniVGGT & 1217 & 0.114 & 0.078 & 0.741 & 0.081 & 0.742 & 0.254 & \cellcolor{oomred}OOM & \cellcolor{oomred}OOM & \cellcolor{oomred}OOM & (0.080) & (0.741) & (0.254) \\
 $\pi^{3}$ & 958.7 & \cellcolor{bestred}\textbf{0.048} & \cellcolor{bestred}0.045 & \cellcolor{bestred}\textbf{0.954} & \cellcolor{bestred}\textbf{0.041} & \cellcolor{secondorange}0.957 & \cellcolor{thirdyellow}0.025 & \cellcolor{bestred}\textbf{0.042} & \cellcolor{bestred}\textbf{0.877} & \cellcolor{secondorange}\textbf{0.042} & \cellcolor{bestred}\textbf{0.043} & \cellcolor{bestred}\textbf{0.929} & \cellcolor{bestred}\textbf{0.033} \\
 $\pi^{3}$-X & 1360 & 0.080 & \cellcolor{secondorange}0.046 & \cellcolor{bestred}0.954 & \cellcolor{bestred}0.041 & \cellcolor{bestred}\textbf{0.965} & \cellcolor{bestred}\textbf{0.021} & \cellcolor{oomred}OOM & \cellcolor{oomred}OOM & \cellcolor{oomred}OOM & (0.044) & (0.959) & (0.021) \\
 AMB3R & 1563 & 0.081 & \cellcolor{bestred}\textbf{0.045} & 0.905 & \cellcolor{thirdyellow}0.045 & 0.934 & 0.035 & \cellcolor{oomred}OOM & \cellcolor{oomred}OOM & \cellcolor{oomred}OOM & (0.045) & (0.920) & (0.035) \\
 DA3-Small & 34.3 & 0.135 & 0.105 & 0.571 & 0.097 & 0.598 & 0.326 & 0.092 & 0.640 & 0.255 & 0.098 & 0.603 & 0.290 \\
 DA3-Base & 135.4 & 0.159 & 0.090 & 0.583 & 0.092 & 0.632 & 0.275 & 0.091 & 0.661 & 0.242 & 0.091 & 0.625 & 0.259 \\
 DA3-Large & 410.9 & 0.127 & 0.072 & 0.726 & 0.059 & 0.783 & 0.130 & \cellcolor{oomred}OOM & \cellcolor{oomred}OOM & \cellcolor{oomred}OOM & (0.066) & (0.755) & (0.130) \\
 DA3-Giant & 1356 & 0.095 & 0.063 & 0.775 & 0.057 & 0.851 & 0.101 & \cellcolor{oomred}OOM & \cellcolor{oomred}OOM & \cellcolor{oomred}OOM & (0.060) & (0.813) & (0.101) \\
 DA3-Nested & 1690 & 0.089 & 0.064 & 0.754 & 0.059 & 0.850 & 0.098 & \cellcolor{oomred}OOM & \cellcolor{oomred}OOM & \cellcolor{oomred}OOM & (0.061) & (0.802) & (0.098) \\
 WorldMirror & 1263 & 0.105 & 0.078 & 0.807 & 0.089 & 0.871 & 0.078 & \cellcolor{oomred}OOM & \cellcolor{oomred}OOM & \cellcolor{oomred}OOM & (0.084) & (0.839) & (0.078) \\
 VGGT-Omega & 1144 & 0.081 & 0.064 & 0.851 & \cellcolor{secondorange}0.042 & 0.937 & 0.039 & -- & -- & -- & (0.053) & (0.894) & (0.039) \\ \hdashline
 \ours (Ours) & 1304 & 0.010 & 0.017 & 0.963 & 0.013 & 0.980 & 0.013 & \cellcolor{oomred}OOM & \cellcolor{oomred}OOM & \cellcolor{oomred}OOM & (0.013) & (0.972) & (0.013) \\
\midrule
\multicolumn{14}{c}{\cellcolor{catgray}\textbf{Online}} \\
\midrule
 Spann3r$^{224}$ & 658.7 & 0.230 & 0.110 & 0.731 & 0.095 & 0.807 & 0.122 & 0.116 & 0.699 & 0.202 & 0.107 & 0.746 & 0.162 \\
 CUT3R & 793.3 & 0.130 & 0.106 & 0.657 & 0.131 & 0.576 & 0.272 & 0.159 & 0.255 & 0.968 & 0.132 & 0.496 & 0.620 \\
 MonST3R & 571.2 & 0.212 & 0.153 & 0.146 & 0.182 & 0.093 & 0.631 & \cellcolor{oomred}OOM & \cellcolor{oomred}OOM & \cellcolor{oomred}OOM & (0.168) & (0.119) & (0.631) \\
 Point3R & 828 & 0.142 & 0.140 & 0.421 & 0.143 & 0.494 & 0.320 & 0.159 & 0.397 & 0.420 & 0.147 & 0.437 & 0.370 \\
 Stream3R-S & 1191 & 0.112 & 0.075 & 0.732 & 0.449 & 0.060 & 1.494 & \cellcolor{oomred}OOM & \cellcolor{oomred}OOM & \cellcolor{oomred}OOM & (0.262) & (0.396) & (1.494) \\
 Stream3R-W & 1191 & 0.112 & 0.078 & 0.720 & 0.655 & 0.054 & 1.512 & \cellcolor{oomred}OOM & \cellcolor{oomred}OOM & \cellcolor{oomred}OOM & (0.367) & (0.387) & (1.512) \\
 StreamVGGT & 1257 & 0.124 & 0.117 & 0.722 & 0.100 & 0.779 & 0.208 & 0.113 & 0.672 & 0.229 & 0.110 & 0.725 & 0.219 \\
 Page4D & 1257 & 0.163 & 0.091 & 0.657 & 0.079 & 0.677 & 0.199 & \cellcolor{oomred}OOM & \cellcolor{oomred}OOM & \cellcolor{oomred}OOM & (0.085) & (0.667) & (0.199) \\
 InfiniteVGGT & 1257 & 0.124 & 0.117 & 0.719 & 0.100 & 0.779 & 0.209 & 0.114 & 0.673 & 0.229 & 0.111 & 0.724 & 0.219 \\
 Wint3R & 749.5 & 0.136 & 0.089 & 0.529 & 0.079 & 0.731 & 0.137 & 0.094 & 0.496 & 0.375 & 0.087 & 0.585 & 0.256 \\
 LongStream-B & 1191 & \cellcolor{secondorange}\textbf{0.056} & 0.068 & 0.776 & 0.148 & 0.529 & 0.266 & 0.138 & 0.327 & 0.281 & 0.118 & 0.544 & 0.274 \\
 LongStream-S & 1191 & \cellcolor{secondorange}\textbf{0.056} & \textbf{0.068} & 0.776 & 0.141 & 0.238 & 0.512 & 0.152 & 0.145 & 0.565 & 0.120 & 0.386 & 0.539 \\
 LingbotMap$^{*}$-W & 1158 & 0.199 & 0.070 & \cellcolor{thirdyellow}\textbf{0.915} & 0.071 & 0.909 & 0.046 & 0.068 & 0.728 & 0.069 & 0.069 & 0.851 & 0.057 \\
 LingbotMap$^{*}$-S & 1158 & 0.199 & 0.070 & \cellcolor{thirdyellow}\textbf{0.915} & \textbf{0.067} & \textbf{0.919} & \textbf{0.038} & \cellcolor{secondorange}\textbf{0.058} & \cellcolor{secondorange}\textbf{0.835} & \cellcolor{thirdyellow}\textbf{0.044} & \textbf{0.065} & \cellcolor{thirdyellow}\textbf{0.890} & \cellcolor{thirdyellow}\textbf{0.041} \\
\midrule
\multicolumn{14}{c}{\cellcolor{catgray}\textbf{Chunk-wise}} \\
\midrule
 VGGT-Long & 1257 & 0.133 & 0.072 & 0.787 & 0.079 & 0.792 & 0.139 & 0.109 & 0.574 & 0.219 & 0.087 & 0.718 & 0.179 \\
 $\pi^{3}$-Long & 958.7 & \cellcolor{bestred}\textbf{0.048} & \cellcolor{bestred}\textbf{0.045} & \cellcolor{bestred}\textbf{0.954} & 0.115 & \cellcolor{thirdyellow}\textbf{0.953} & \cellcolor{secondorange}\textbf{0.022} & 0.135 & \cellcolor{secondorange}\textbf{0.835} & \textbf{0.045} & 0.098 & \cellcolor{secondorange}\textbf{0.914} & \cellcolor{bestred}\textbf{0.033} \\
 DA3-Streaming & 1356 & 0.095 & 0.063 & 0.775 & \textbf{0.059} & 0.837 & 0.102 & \textbf{0.071} & 0.633 & 0.162 & \cellcolor{thirdyellow}\textbf{0.064} & 0.748 & 0.132 \\
\midrule
\multicolumn{14}{c}{\cellcolor{catgray}\textbf{SLAM-based}} \\
\midrule
 MASt3R-SLAM & 688.6 & 0.425 & 0.319 & 0.118 & 0.354 & 0.481 & 0.544 & 0.371 & \textbf{0.747} & \textbf{0.076} & 0.348 & 0.449 & 0.310 \\
 VGGT-SLAM & 1257 & \textbf{0.133} & \textbf{0.072} & \textbf{0.787} & \textbf{0.103} & \textbf{0.653} & \textbf{0.208} & \textbf{0.111} & 0.529 & 0.233 & \textbf{0.095} & \textbf{0.656} & \textbf{0.221} \\
\midrule
\multicolumn{14}{c}{\cellcolor{catgray}\textbf{Test-Time Training}} \\
\midrule
 TTT3R & 793.3 & 0.130 & 0.130 & 0.531 & 0.121 & 0.589 & 0.445 & 0.136 & 0.481 & 0.383 & 0.129 & 0.534 & 0.414 \\
 Scal3R & 1266 & 0.114 & 0.072 & 0.903 & 0.168 & 0.784 & \textbf{0.027} & 0.178 & 0.589 & \cellcolor{secondorange}0.042 & 0.139 & 0.759 & \cellcolor{secondorange}\textbf{0.035} \\
 LoGeR & 1255 & 0.095 & \cellcolor{thirdyellow}0.053 & 0.904 & 0.101 & 0.831 & 0.095 & 0.106 & 0.682 & 0.111 & 0.087 & 0.806 & 0.103 \\
 LoGeR$^*$ & 1255 & \cellcolor{thirdyellow}\textbf{0.069} & \cellcolor{secondorange}\textbf{0.046} & \cellcolor{secondorange}\textbf{0.933} & \textbf{0.054} & \textbf{0.904} & 0.045 & \cellcolor{thirdyellow}\textbf{0.065} & \textbf{0.816} & \cellcolor{bestred}\textbf{0.039} & \cellcolor{secondorange}\textbf{0.055} & \textbf{0.884} & 0.042 \\
\bottomrule
\end{tabular}%
}
\end{table*}

\begin{table*}[t]
\centering
\setlength{\fboxsep}{0pt}
\caption{\textbf{Per-Dataset Results on DROID}.
Performance across different input regimes: \textit{Single Frame}, \textit{Sparse}, \textit{Medium}, \textit{Dense}, and the \textit{Average}.
The best, second-best, and third-best results in each column are highlighted in
\colorbox{bestred}{deep blue}, \colorbox{secondorange}{medium blue}, and \colorbox{thirdyellow}{light blue}, respectively.
Out-of-memory (OOM) and Timeout (T.O) cells are shaded \colorbox{oomred}{light red};
\textit{Average} values for those rows are wrapped in parentheses and excluded from per-column ranking.
Within each sub-category, the \textbf{bold} value marks the in-group best.
Note that \ours (Ours) is excluded from the per-column rankings.
}
\label{tab:dataset_droid}
\resizebox{\textwidth}{!}{%
\renewcommand{\arraystretch}{1.15}
\setlength{\tabcolsep}{3pt}
\begin{tabular}{l c >{\columncolor{subcol}}c >{\columncolor{subcol}}c >{\columncolor{subcol}}c >{\columncolor{subcol}}c >{\columncolor{subcol}}c >{\columncolor{subcol}}c >{\columncolor{subcol}}c >{\columncolor{subcol}}c >{\columncolor{subcol}}c >{\columncolor{subcol}}c >{\columncolor{subcol}}c >{\columncolor{subcol}}c}
\toprule
\multirow{2.5}{*}{\textbf{Method}}
& \multirow{2.5}{*}{\makecell{\textbf{\#Params}\\\textbf{(M)}}}
& \multicolumn{1}{c}{\textbf{Single Frame}}
& \multicolumn{2}{c}{\textbf{Sparse}}
& \multicolumn{3}{c}{\textbf{Medium}}
& \multicolumn{3}{c}{\textbf{Dense}}
& \multicolumn{3}{c}{\textbf{Average}} \\
\cmidrule(lr){3-3} \cmidrule(lr){4-5} \cmidrule(lr){6-8} \cmidrule(lr){9-11} \cmidrule(lr){12-14}
 & &
 AbsRel$\downarrow$ &
 AbsRel$\downarrow$ & AUC@30$\uparrow$ &
 AbsRel$\downarrow$ & AUC@30$\uparrow$ & ATE$\downarrow$ &
 AbsRel$\downarrow$ & AUC@30$\uparrow$ & ATE$\downarrow$ &
 AbsRel$\downarrow$ & AUC@30$\uparrow$ & ATE$\downarrow$ \\
\midrule
\multicolumn{14}{c}{\cellcolor{catgray}\textbf{Optimization-based}} \\
\midrule
 DUSt3R & 571.2 & 0.218 & 0.301 & 0.177 & 0.305 & 0.124 & 0.080 & \cellcolor{oomred}OOM & \cellcolor{oomred}OOM & \cellcolor{oomred}OOM & (0.303) & (0.151) & (0.080) \\
 MASt3R & 688.6 & \textbf{0.167} & \textbf{0.203} & \textbf{0.269} & \textbf{0.244} & \textbf{0.188} & \textbf{0.072} & \cellcolor{oomred}OOM & \cellcolor{oomred}OOM & \cellcolor{oomred}OOM & (0.224) & (0.229) & (0.072) \\
\midrule
\multicolumn{14}{c}{\cellcolor{catgray}\textbf{End-to-End Feed-Forward}} \\
\midrule
 VGGT & 1257 & 0.121 & \cellcolor{thirdyellow}0.111 & 0.525 & 0.131 & 0.450 & 0.025 & \cellcolor{oomred}OOM & \cellcolor{oomred}OOM & \cellcolor{oomred}OOM & (0.121) & (0.487) & (0.025) \\
 Fast3R & 647.5 & 0.199 & 0.320 & 0.200 & 0.282 & 0.150 & 0.074 & 0.303 & 0.134 & 0.076 & 0.301 & 0.161 & 0.075 \\
 FastVGGT & 1158 & 0.121 & 0.123 & 0.506 & 0.125 & 0.443 & 0.028 & \cellcolor{secondorange}0.129 & \cellcolor{secondorange}0.447 & \cellcolor{secondorange}0.026 & \cellcolor{thirdyellow}0.126 & 0.465 & \cellcolor{thirdyellow}0.027 \\
 MUSt3R & 423.4 & 0.185 & 0.190 & 0.412 & 0.176 & 0.389 & 0.037 & \cellcolor{oomred}T.O & \cellcolor{oomred}T.O & \cellcolor{oomred}T.O & (0.183) & (0.401) & (0.037) \\
 MapAnything & 1228 & 0.100 & 0.167 & 0.266 & 0.144 & 0.272 & 0.044 & \cellcolor{oomred}OOM & \cellcolor{oomred}OOM & \cellcolor{oomred}OOM & (0.155) & (0.269) & (0.044) \\
 OmniVGGT & 1217 & 0.136 & 0.262 & 0.551 & 0.121 & 0.509 & 0.020 & \cellcolor{oomred}OOM & \cellcolor{oomred}OOM & \cellcolor{oomred}OOM & (0.192) & (0.530) & (0.020) \\
 $\pi^{3}$ & 958.7 & 0.106 & 0.151 & 0.534 & 0.105 & 0.522 & 0.022 & \cellcolor{bestred}\textbf{0.105} & \cellcolor{bestred}\textbf{0.508} & \cellcolor{bestred}\textbf{0.022} & \cellcolor{secondorange}\textbf{0.120} & \cellcolor{secondorange}\textbf{0.521} & \cellcolor{bestred}\textbf{0.022} \\
 $\pi^{3}$-X & 1360 & 0.100 & \cellcolor{secondorange}0.103 & 0.607 & 0.094 & 0.563 & \cellcolor{thirdyellow}0.019 & \cellcolor{oomred}OOM & \cellcolor{oomred}OOM & \cellcolor{oomred}OOM & (0.099) & (0.585) & (0.019) \\
 AMB3R & 1563 & \textbf{0.098} & \cellcolor{bestred}\textbf{0.091} & 0.561 & \cellcolor{secondorange}\textbf{0.080} & 0.481 & 0.024 & \cellcolor{oomred}OOM & \cellcolor{oomred}OOM & \cellcolor{oomred}OOM & (0.085) & (0.521) & (0.024) \\
 DA3-Small & 34.3 & 0.198 & 0.238 & 0.168 & 0.219 & 0.185 & 0.052 & 0.224 & 0.162 & 0.058 & 0.227 & 0.172 & 0.055 \\
 DA3-Base & 135.4 & 0.158 & 0.190 & 0.294 & 0.159 & 0.252 & 0.040 & 0.161 & 0.237 & 0.042 & 0.170 & 0.261 & 0.041 \\
 DA3-Large & 410.9 & 0.136 & 0.162 & 0.523 & 0.095 & 0.477 & 0.020 & \cellcolor{oomred}OOM & \cellcolor{oomred}OOM & \cellcolor{oomred}OOM & (0.129) & (0.500) & (0.020) \\
 DA3-Giant & 1356 & 0.114 & 0.125 & \cellcolor{secondorange}0.642 & \cellcolor{thirdyellow}0.092 & \cellcolor{bestred}\textbf{0.593} & \cellcolor{bestred}\textbf{0.016} & \cellcolor{oomred}OOM & \cellcolor{oomred}OOM & \cellcolor{oomred}OOM & (0.108) & (0.618) & (0.016) \\
 DA3-Nested & 1690 & 0.123 & 0.146 & \cellcolor{bestred}\textbf{0.675} & 0.110 & \cellcolor{secondorange}0.578 & \cellcolor{secondorange}0.018 & \cellcolor{oomred}OOM & \cellcolor{oomred}OOM & \cellcolor{oomred}OOM & (0.128) & (0.627) & (0.018) \\
 WorldMirror & 1263 & 0.125 & 0.256 & 0.539 & 0.125 & 0.477 & 0.026 & \cellcolor{oomred}OOM & \cellcolor{oomred}OOM & \cellcolor{oomred}OOM & (0.190) & (0.508) & (0.026) \\
 VGGT-Omega & 1144 & \cellcolor{bestred}\textbf{0.073} & \cellcolor{bestred}0.091 & \cellcolor{thirdyellow}0.621 & \cellcolor{bestred}\textbf{0.059} & 0.533 & 0.021 & -- & -- & -- & (0.075) & (0.577) & (0.021) \\ \hdashline
 \ours (Ours) & 1304 & 0.098 & 0.090 & 0.627 & 0.044 & 0.582 & 0.018 & \cellcolor{oomred}OOM & \cellcolor{oomred}OOM & \cellcolor{oomred}OOM & (0.077) & (0.605) & (0.018) \\
\midrule
\multicolumn{14}{c}{\cellcolor{catgray}\textbf{Online}} \\
\midrule
 Spann3r$^{224}$ & 658.7 & 0.189 & 0.246 & 0.087 & 0.259 & 0.103 & 0.076 & 0.326 & 0.077 & 0.091 & 0.277 & 0.089 & 0.084 \\
 CUT3R & 793.3 & 0.190 & 0.253 & 0.305 & 0.294 & 0.199 & 0.057 & 0.399 & 0.066 & 0.086 & 0.315 & 0.190 & 0.071 \\
 MonST3R & 571.2 & 0.195 & 0.204 & 0.183 & 0.269 & 0.168 & 0.073 & \cellcolor{oomred}OOM & \cellcolor{oomred}OOM & \cellcolor{oomred}OOM & (0.237) & (0.176) & (0.073) \\
 Point3R & 828 & 0.213 & 0.369 & 0.167 & 0.353 & 0.094 & 0.081 & 0.348 & 0.077 & 0.082 & 0.357 & 0.113 & 0.081 \\
 Stream3R-S & 1191 & \cellcolor{secondorange}\textbf{0.093} & \textbf{0.112} & 0.467 & 0.139 & 0.339 & 0.050 & \cellcolor{oomred}OOM & \cellcolor{oomred}OOM & \cellcolor{oomred}OOM & (0.125) & (0.403) & (0.050) \\
 Stream3R-W & 1191 & \cellcolor{secondorange}\textbf{0.093} & \textbf{0.112} & 0.467 & 0.220 & 0.217 & 0.086 & \cellcolor{oomred}OOM & \cellcolor{oomred}OOM & \cellcolor{oomred}OOM & (0.166) & (0.342) & (0.086) \\
 StreamVGGT & 1257 & 0.105 & 0.130 & 0.488 & 0.161 & 0.380 & 0.038 & 0.177 & 0.327 & 0.048 & \textbf{0.156} & 0.398 & 0.043 \\
 Page4D & 1257 & 0.135 & 0.127 & 0.322 & \textbf{0.113} & 0.345 & \textbf{0.028} & \cellcolor{oomred}OOM & \cellcolor{oomred}OOM & \cellcolor{oomred}OOM & (0.120) & (0.334) & (0.028) \\
 InfiniteVGGT & 1257 & 0.105 & 0.131 & \textbf{0.491} & 0.162 & 0.379 & 0.038 & 0.178 & \textbf{0.328} & 0.048 & 0.157 & \textbf{0.400} & 0.043 \\
 Wint3R & 749.5 & 0.189 & 0.187 & 0.262 & 0.179 & 0.159 & 0.059 & 0.321 & 0.058 & 0.084 & 0.229 & 0.160 & 0.072 \\
 LongStream-B & 1191 & 0.176 & 0.306 & 0.367 & 0.559 & 0.174 & 0.070 & 0.481 & 0.204 & 0.067 & 0.449 & 0.249 & 0.069 \\
 LongStream-S & 1191 & 0.176 & 0.306 & 0.368 & 0.302 & 0.217 & 0.061 & 0.512 & 0.167 & 0.062 & 0.373 & 0.251 & 0.062 \\
 LingbotMap$^{*}$-W & 1158 & 0.142 & 0.176 & 0.308 & 0.145 & \textbf{0.391} & 0.033 & 0.209 & 0.269 & 0.048 & 0.177 & 0.322 & 0.041 \\
 LingbotMap$^{*}$-S & 1158 & 0.142 & 0.176 & 0.308 & 0.145 & 0.376 & 0.035 & \textbf{0.163} & 0.323 & \textbf{0.036} & 0.161 & 0.336 & \textbf{0.036} \\
\midrule
\multicolumn{14}{c}{\cellcolor{catgray}\textbf{Chunk-wise}} \\
\midrule
 VGGT-Long & 1257 & 0.121 & \cellcolor{thirdyellow}\textbf{0.111} & 0.525 & 0.135 & 0.435 & 0.028 & 0.200 & 0.284 & 0.045 & 0.149 & 0.415 & 0.036 \\
 $\pi^{3}$-Long & 958.7 & \textbf{0.106} & 0.151 & 0.534 & 0.113 & 0.515 & 0.022 & 0.232 & 0.352 & 0.036 & 0.166 & \cellcolor{thirdyellow}0.467 & 0.029 \\
 DA3-Streaming & 1356 & 0.114 & 0.125 & \cellcolor{secondorange}\textbf{0.642} & \textbf{0.094} & \cellcolor{thirdyellow}\textbf{0.567} & \cellcolor{secondorange}\textbf{0.018} & \cellcolor{thirdyellow}\textbf{0.136} & \cellcolor{thirdyellow}\textbf{0.414} & \cellcolor{thirdyellow}\textbf{0.031} & \cellcolor{bestred}\textbf{0.118} & \cellcolor{bestred}\textbf{0.541} & \cellcolor{secondorange}\textbf{0.025} \\
\midrule
\multicolumn{14}{c}{\cellcolor{catgray}\textbf{SLAM-based}} \\
\midrule
 MASt3R-SLAM & 688.6 & 0.261 & 0.342 & 0.239 & 0.392 & 0.126 & 0.078 & 0.378 & 0.108 & 0.076 & 0.370 & 0.158 & 0.077 \\
 VGGT-SLAM & 1257 & \textbf{0.121} & \cellcolor{thirdyellow}\textbf{0.111} & \textbf{0.525} & \textbf{0.141} & \textbf{0.383} & \textbf{0.034} & \textbf{0.232} & \textbf{0.179} & \textbf{0.059} & \textbf{0.161} & \textbf{0.362} & \textbf{0.047} \\
\midrule
\multicolumn{14}{c}{\cellcolor{catgray}\textbf{Test-Time Training}} \\
\midrule
 TTT3R & 793.3 & 0.190 & 0.263 & 0.268 & 0.253 & 0.244 & 0.043 & 0.304 & 0.194 & 0.062 & 0.274 & 0.235 & 0.053 \\
 Scal3R & 1266 & 0.159 & 0.252 & \textbf{0.571} & 0.186 & \textbf{0.495} & \textbf{0.022} & 0.302 & 0.279 & 0.035 & 0.247 & \textbf{0.448} & \textbf{0.028} \\
 LoGeR & 1255 & 0.102 & 0.140 & 0.490 & 0.142 & 0.445 & 0.032 & 0.269 & \textbf{0.394} & \textbf{0.033} & 0.183 & 0.443 & 0.032 \\
 LoGeR$^*$ & 1255 & \cellcolor{thirdyellow}\textbf{0.095} & \textbf{0.120} & 0.518 & \textbf{0.105} & 0.437 & 0.028 & \textbf{0.214} & 0.387 & 0.036 & \textbf{0.146} & 0.447 & 0.032 \\
\bottomrule
\end{tabular}%
}
\end{table*}

\begin{table*}[t]
\centering
\setlength{\fboxsep}{0pt}
\caption{\textbf{Per-Dataset Results on DTU}.
Performance across different input regimes: \textit{Single Frame}, \textit{Sparse}, \textit{Medium}, \textit{Dense}, and the \textit{Average}.
The best, second-best, and third-best results in each column are highlighted in
\colorbox{bestred}{deep blue}, \colorbox{secondorange}{medium blue}, and \colorbox{thirdyellow}{light blue}, respectively.
This dataset is not evaluated under the \textit{Dense} regime, so the corresponding cells are marked as ``-''.
Within each sub-category, the \textbf{bold} value marks the in-group best.
Note that \ours (Ours) is excluded from the per-column rankings.
}
\label{tab:dataset_dtu}
\resizebox{\textwidth}{!}{%
\renewcommand{\arraystretch}{1.15}
\setlength{\tabcolsep}{3pt}
\begin{tabular}{l c >{\columncolor{subcol}}c >{\columncolor{subcol}}c >{\columncolor{subcol}}c >{\columncolor{subcol}}c >{\columncolor{subcol}}c >{\columncolor{subcol}}c >{\columncolor{subcol}}c >{\columncolor{subcol}}c >{\columncolor{subcol}}c >{\columncolor{subcol}}c >{\columncolor{subcol}}c >{\columncolor{subcol}}c >{\columncolor{subcol}}c >{\columncolor{subcol}}c >{\columncolor{subcol}}c}
\toprule
\multirow{2.5}{*}{\textbf{Method}}
& \multirow{2.5}{*}{\makecell{\textbf{\#Params}\\\textbf{(M)}}}
& \multicolumn{1}{c}{\textbf{Single Frame}}
& \multicolumn{2}{c}{\textbf{Sparse}}
& \multicolumn{4}{c}{\textbf{Medium}}
& \multicolumn{4}{c}{\textbf{Dense}}
& \multicolumn{4}{c}{\textbf{Average}} \\
\cmidrule(lr){3-3} \cmidrule(lr){4-5} \cmidrule(lr){6-9} \cmidrule(lr){10-13} \cmidrule(lr){14-17}
  & &
 AbsRel$\downarrow$ &
 AbsRel$\downarrow$ &AUC@30$\uparrow$ &
 AbsRel$\downarrow$ &AUC@30$\uparrow$ &ATE$\downarrow$ &F-Score$\uparrow$ &
 AbsRel$\downarrow$ &AUC@30$\uparrow$ &ATE$\downarrow$ &F-Score$\uparrow$ &
 AbsRel$\downarrow$ &AUC@30$\uparrow$ &ATE$\downarrow$ &F-Score$\uparrow$ \\
\midrule
\multicolumn{17}{c}{\cellcolor{catgray}\textbf{Optimization-based}} \\
\midrule
 DUSt3R & 571.2 & \textbf{0.049} & \textbf{0.064} & 0.627 & 0.077 & 0.613 & 0.045 & 0.385 & - & - & - & - & 0.070 & 0.620 & 0.045 & 0.385 \\
 MASt3R & 688.6 & 0.078 & 0.069 & \textbf{0.723} & \textbf{0.065} & \textbf{0.686} & \textbf{0.034} & \textbf{0.436} & - & - & - & - & \textbf{0.067} & \textbf{0.704} & \textbf{0.034} & \textbf{0.436} \\
\midrule
\multicolumn{17}{c}{\cellcolor{catgray}\textbf{End-to-End Feed-Forward}} \\
\midrule
 VGGT & 1257 & \cellcolor{bestred}\textbf{0.009} & \cellcolor{bestred}\textbf{0.008} & \cellcolor{bestred}\textbf{0.997} & \cellcolor{bestred}\textbf{0.005} & \cellcolor{bestred}\textbf{0.995} & \cellcolor{bestred}\textbf{0.002} & \cellcolor{thirdyellow}0.797 & - & - & - & - & \cellcolor{bestred}\textbf{0.007} & \cellcolor{bestred}\textbf{0.996} & \cellcolor{bestred}\textbf{0.002} & \cellcolor{thirdyellow}0.797 \\
 Fast3R & 647.5 & 0.059 & 0.074 & 0.764 & 0.064 & 0.716 & 0.038 & 0.318 & - & - & - & - & 0.069 & 0.740 & 0.038 & 0.318 \\
 FastVGGT & 1158 & \cellcolor{bestred}0.009 & \cellcolor{secondorange}0.009 & 0.968 & \cellcolor{secondorange}0.007 & 0.960 & 0.005 & \cellcolor{bestred}\textbf{0.809} & - & - & - & - & \cellcolor{secondorange}0.008 & 0.964 & 0.005 & \cellcolor{bestred}\textbf{0.809} \\
 MUSt3R & 423.4 & 0.059 & 0.058 & 0.797 & 0.042 & 0.823 & 0.023 & 0.468 & - & - & - & - & 0.050 & 0.810 & 0.023 & 0.468 \\
 MapAnything & 1228 & 0.090 & 0.084 & 0.624 & 0.083 & 0.621 & 0.044 & 0.290 & - & - & - & - & 0.084 & 0.622 & 0.044 & 0.290 \\
 OmniVGGT & 1217 & 0.022 & 0.013 & 0.962 & 0.012 & 0.962 & 0.005 & 0.735 & - & - & - & - & 0.012 & 0.962 & 0.005 & 0.735 \\
 $\pi^{3}$ & 958.7 & 0.027 & 0.016 & 0.945 & 0.013 & 0.950 & 0.006 & 0.669 & - & - & - & - & 0.015 & 0.947 & 0.006 & 0.669 \\
 $\pi^{3}$-X & 1360 & 0.027 & 0.018 & 0.943 & 0.014 & 0.958 & 0.005 & 0.649 & - & - & - & - & 0.016 & 0.951 & 0.005 & 0.649 \\
 AMB3R & 1563 & 0.012 & \cellcolor{secondorange}0.009 & \cellcolor{secondorange}0.994 & \cellcolor{secondorange}0.007 & \cellcolor{secondorange}0.990 & \cellcolor{secondorange}0.003 & 0.581 & - & - & - & - & \cellcolor{secondorange}0.008 & \cellcolor{secondorange}0.992 & \cellcolor{secondorange}0.003 & 0.581 \\
 DA3-Small & 34.3 & 0.069 & 0.034 & 0.831 & 0.028 & 0.798 & 0.019 & 0.625 & - & - & - & - & 0.031 & 0.814 & 0.019 & 0.625 \\
 DA3-Base & 135.4 & 0.055 & 0.021 & 0.898 & 0.017 & 0.898 & 0.010 & 0.689 & - & - & - & - & 0.019 & 0.898 & 0.010 & 0.689 \\
 DA3-Large & 410.9 & 0.035 & 0.017 & 0.972 & 0.012 & 0.966 & \cellcolor{thirdyellow}0.004 & 0.727 & - & - & - & - & 0.015 & 0.969 & \cellcolor{thirdyellow}0.004 & 0.727 \\
 DA3-Giant & 1356 & 0.033 & 0.014 & 0.992 & 0.011 & \cellcolor{thirdyellow}0.987 & \cellcolor{bestred}0.002 & 0.747 & - & - & - & - & 0.013 & \cellcolor{thirdyellow}0.990 & \cellcolor{bestred}0.002 & 0.747 \\
 DA3-Nested & 1690 & 0.028 & 0.019 & \cellcolor{thirdyellow}0.993 & 0.013 & 0.986 & \cellcolor{bestred}0.002 & 0.739 & - & - & - & - & 0.016 & 0.989 & \cellcolor{bestred}0.002 & 0.739 \\
 WorldMirror & 1263 & 0.039 & 0.024 & 0.898 & 0.021 & 0.899 & 0.011 & 0.571 & - & - & - & - & 0.022 & 0.898 & 0.011 & 0.571 \\
 VGGT-Omega & 1144 & 0.018 & 0.014 & 0.989 & 0.011 & 0.973 & \cellcolor{secondorange}0.003 & 0.679 & -- & -- & -- & -- & (0.012) & (0.981) & (0.003) & (0.679) \\ \hdashline
 \ours (Ours) & 1304 & 0.121 & 0.054 & 0.898 & 0.020 & 0.901 & 0.011 & 0.643 & - & - & - & - & (0.065) & (0.900) & (0.011) & (0.643) \\
\midrule
\multicolumn{17}{c}{\cellcolor{catgray}\textbf{Online}} \\
\midrule
 Spann3r$^{224}$ & 658.7 & 0.040 & 0.047 & 0.602 & 0.044 & 0.634 & 0.043 & 0.473 & - & - & - & - & 0.045 & 0.618 & 0.043 & 0.473 \\
 CUT3R & 793.3 & 0.050 & 0.056 & 0.763 & 0.053 & 0.724 & 0.028 & 0.375 & - & - & - & - & 0.054 & 0.744 & 0.028 & 0.375 \\
 MonST3R & 571.2 & 0.104 & 0.101 & 0.470 & 0.123 & 0.014 & 0.214 & 0.007 & - & - & - & - & 0.112 & 0.242 & 0.214 & 0.007 \\
 Point3R & 828 & 0.038 & 0.061 & 0.596 & 0.063 & 0.499 & 0.048 & 0.239 & - & - & - & - & 0.062 & 0.547 & 0.048 & 0.239 \\
 Stream3R-S & 1191 & \cellcolor{secondorange}\textbf{0.010} & \cellcolor{thirdyellow}\textbf{0.011} & 0.965 & \cellcolor{thirdyellow}\textbf{0.010} & 0.937 & 0.008 & 0.726 & - & - & - & - & \cellcolor{thirdyellow}\textbf{0.011} & 0.951 & 0.008 & 0.726 \\
 Stream3R-W & 1191 & \cellcolor{secondorange}\textbf{0.010} & \cellcolor{thirdyellow}\textbf{0.011} & 0.965 & 0.011 & 0.896 & 0.013 & 0.720 & - & - & - & - & \cellcolor{thirdyellow}0.011 & 0.931 & 0.013 & 0.720 \\
 StreamVGGT & 1257 & 0.012 & 0.012 & 0.971 & 0.011 & \textbf{0.963} & \textbf{0.005} & 0.795 & - & - & - & - & 0.012 & 0.967 & \textbf{0.005} & 0.795 \\
 Page4D & 1257 & 0.019 & 0.022 & 0.877 & 0.015 & 0.860 & 0.011 & 0.568 & - & - & - & - & 0.018 & 0.869 & 0.011 & 0.568 \\
 InfiniteVGGT & 1257 & \cellcolor{thirdyellow}0.011 & 0.012 & \textbf{0.972} & 0.011 & 0.962 & 0.006 & \cellcolor{secondorange}\textbf{0.804} & - & - & - & - & 0.012 & \textbf{0.967} & 0.006 & \cellcolor{secondorange}\textbf{0.804} \\
 Wint3R & 749.5 & 0.033 & 0.028 & 0.791 & 0.026 & 0.785 & 0.020 & 0.610 & - & - & - & - & 0.027 & 0.788 & 0.020 & 0.610 \\
 LongStream-B & 1191 & 0.037 & 0.038 & 0.851 & 0.062 & 0.715 & 0.031 & 0.247 & - & - & - & - & 0.050 & 0.783 & 0.031 & 0.247 \\
 LongStream-S & 1191 & 0.037 & 0.038 & 0.851 & 0.032 & 0.707 & 0.030 & 0.267 & - & - & - & - & 0.035 & 0.779 & 0.030 & 0.267 \\
 LingbotMap$^{*}$-W & 1158 & 0.058 & 0.048 & 0.776 & 0.048 & 0.719 & 0.032 & 0.300 & - & - & - & - & 0.048 & 0.748 & 0.032 & 0.300 \\
 LingbotMap$^{*}$-S & 1158 & 0.058 & 0.048 & 0.776 & 0.048 & 0.719 & 0.032 & 0.300 & - & - & - & - & 0.048 & 0.748 & 0.032 & 0.300 \\
\midrule
\multicolumn{17}{c}{\cellcolor{catgray}\textbf{Chunk-wise}} \\
\midrule
 VGGT-Long & 1257 & \cellcolor{bestred}\textbf{0.009} & \cellcolor{bestred}\textbf{0.008} & \cellcolor{bestred}\textbf{0.997} & \cellcolor{bestred}\textbf{0.005} & \cellcolor{bestred}\textbf{0.995} & \cellcolor{bestred}\textbf{0.002} & \cellcolor{thirdyellow}\textbf{0.797} & - & - & - & - & \cellcolor{bestred}\textbf{0.007} & \cellcolor{bestred}\textbf{0.996} & \cellcolor{bestred}\textbf{0.002} & \cellcolor{thirdyellow}\textbf{0.797} \\
 $\pi^{3}$-Long & 958.7 & 0.027 & 0.016 & 0.945 & 0.013 & 0.950 & 0.006 & 0.669 & - & - & - & - & 0.015 & 0.947 & 0.006 & 0.669 \\
 DA3-Streaming & 1356 & 0.033 & 0.014 & \cellcolor{thirdyellow}0.993 & 0.011 & \cellcolor{thirdyellow}0.987 & \cellcolor{bestred}0.002 & 0.746 & - & - & - & - & 0.013 & \cellcolor{thirdyellow}0.990 & \cellcolor{bestred}0.002 & 0.746 \\
\midrule
\multicolumn{17}{c}{\cellcolor{catgray}\textbf{SLAM-based}} \\
\midrule
 MASt3R-SLAM & 688.6 & 0.126 & 0.120 & 0.251 & 0.140 & 0.320 & 0.093 & 0.189 & - & - & - & - & 0.130 & 0.285 & 0.093 & 0.189 \\
 VGGT-SLAM & 1257 & \cellcolor{bestred}\textbf{0.009} & \cellcolor{bestred}\textbf{0.008} & \cellcolor{bestred}\textbf{0.997} & \cellcolor{bestred}\textbf{0.005} & \cellcolor{bestred}\textbf{0.995} & \cellcolor{bestred}\textbf{0.002} & \cellcolor{thirdyellow}\textbf{0.797} & - & - & - & - & \cellcolor{bestred}\textbf{0.007} & \cellcolor{bestred}\textbf{0.996} & \cellcolor{bestred}\textbf{0.002} & \cellcolor{thirdyellow}\textbf{0.797} \\
\midrule
\multicolumn{17}{c}{\cellcolor{catgray}\textbf{Test-Time Training}} \\
\midrule
 TTT3R & 793.3 & 0.050 & 0.059 & 0.696 & 0.055 & 0.702 & 0.042 & 0.345 & - & - & - & - & 0.057 & 0.699 & 0.042 & 0.345 \\
 Scal3R & 1266 & \cellcolor{secondorange}\textbf{0.010} & \cellcolor{bestred}\textbf{0.008} & \textbf{0.991} & \cellcolor{secondorange}\textbf{0.007} & \textbf{0.981} & \cellcolor{bestred}\textbf{0.002} & \textbf{0.773} & - & - & - & - & \cellcolor{secondorange}\textbf{0.008} & \textbf{0.986} & \cellcolor{bestred}\textbf{0.002} & \textbf{0.773} \\
 LoGeR & 1255 & 0.031 & 0.031 & 0.890 & 0.020 & 0.897 & 0.011 & 0.639 & - & - & - & - & 0.026 & 0.893 & 0.011 & 0.639 \\
 LoGeR$^*$ & 1255 & 0.038 & 0.021 & 0.888 & 0.018 & 0.889 & 0.011 & 0.658 & - & - & - & - & 0.020 & 0.889 & 0.011 & 0.658 \\
\bottomrule
\end{tabular}%
}
\end{table*}

\begin{table*}[t]
\centering
\setlength{\fboxsep}{0pt}
\caption{\textbf{Per-Dataset Results on ETH3D}.
Performance across different input regimes: \textit{Single Frame}, \textit{Sparse}, \textit{Medium}, \textit{Dense}, and the \textit{Average}.
The best, second-best, and third-best results in each column are highlighted in
\colorbox{bestred}{deep blue}, \colorbox{secondorange}{medium blue}, and \colorbox{thirdyellow}{light blue}, respectively.
This dataset is not evaluated under the \textit{Dense} regime, so the corresponding cells are marked as ``-''.
Within each sub-category, the \textbf{bold} value marks the in-group best.
Note that \ours (Ours) is excluded from the per-column rankings.
}
\label{tab:dataset_eth3d}
\resizebox{\textwidth}{!}{%
\renewcommand{\arraystretch}{1.15}
\setlength{\tabcolsep}{3pt}
\begin{tabular}{l c >{\columncolor{subcol}}c >{\columncolor{subcol}}c >{\columncolor{subcol}}c >{\columncolor{subcol}}c >{\columncolor{subcol}}c >{\columncolor{subcol}}c >{\columncolor{subcol}}c >{\columncolor{subcol}}c >{\columncolor{subcol}}c >{\columncolor{subcol}}c >{\columncolor{subcol}}c >{\columncolor{subcol}}c}
\toprule
\multirow{2.5}{*}{\textbf{Method}}
& \multirow{2.5}{*}{\makecell{\textbf{\#Params}\\\textbf{(M)}}}
& \multicolumn{1}{c}{\textbf{Single Frame}}
& \multicolumn{2}{c}{\textbf{Sparse}}
& \multicolumn{3}{c}{\textbf{Medium}}
& \multicolumn{3}{c}{\textbf{Dense}}
& \multicolumn{3}{c}{\textbf{Average}} \\
\cmidrule(lr){3-3} \cmidrule(lr){4-5} \cmidrule(lr){6-8} \cmidrule(lr){9-11} \cmidrule(lr){12-14}
  & &
 AbsRel$\downarrow$ &
 AbsRel$\downarrow$ &AUC@30$\uparrow$ &
 AbsRel$\downarrow$ &AUC@30$\uparrow$ &ATE$\downarrow$ &
 AbsRel$\downarrow$ &AUC@30$\uparrow$ &ATE$\downarrow$ &
 AbsRel$\downarrow$ &AUC@30$\uparrow$ &ATE$\downarrow$ \\
\midrule
\multicolumn{14}{c}{\cellcolor{catgray}\textbf{Optimization-based}} \\
\midrule
 DUSt3R & 571.2 & \textbf{0.046} & 0.099 & 0.482 & 0.084 & 0.528 & 1.297 & - & - & - & 0.092 & 0.505 & 1.297 \\
 MASt3R & 688.6 & 0.051 & \textbf{0.086} & \textbf{0.690} & \textbf{0.057} & \textbf{0.793} & \textbf{0.597} & - & - & - & \textbf{0.071} & \textbf{0.742} & \textbf{0.597} \\
\midrule
\multicolumn{14}{c}{\cellcolor{catgray}\textbf{End-to-End Feed-Forward}} \\
\midrule
 VGGT & 1257 & 0.033 & 0.050 & 0.686 & 0.039 & 0.715 & 0.460 & - & - & - & 0.044 & 0.701 & 0.460 \\
 Fast3R & 647.5 & 0.083 & 0.163 & 0.267 & 0.208 & 0.285 & 2.747 & - & - & - & 0.186 & 0.276 & 2.747 \\
 FastVGGT & 1158 & 0.034 & 0.055 & 0.568 & 0.053 & 0.601 & 0.791 & - & - & - & 0.054 & 0.584 & 0.791 \\
 MUSt3R & 423.4 & 0.037 & 0.071 & 0.587 & 0.056 & 0.762 & 0.554 & - & - & - & 0.063 & 0.675 & 0.554 \\
 MapAnything & 1228 & 0.038 & 0.047 & 0.670 & 0.048 & 0.736 & 0.314 & - & - & - & 0.048 & 0.703 & 0.314 \\
 OmniVGGT & 1217 & 0.026 & 0.042 & 0.556 & 0.039 & 0.638 & 0.952 & - & - & - & 0.040 & 0.597 & 0.952 \\
 $\pi^{3}$ & 958.7 & 0.027 & 0.034 & 0.706 & 0.030 & 0.784 & 0.199 & - & - & - & 0.032 & 0.745 & 0.199 \\
 $\pi^{3}$-X & 1360 & 0.028 & 0.032 & 0.692 & 0.030 & 0.779 & 0.207 & - & - & - & 0.031 & 0.735 & 0.207 \\
 AMB3R & 1563 & \cellcolor{secondorange}\textbf{0.022} & 0.050 & 0.684 & 0.037 & 0.779 & 0.280 & - & - & - & 0.043 & 0.731 & 0.280 \\
 DA3-Small & 34.3 & 0.063 & 0.108 & 0.484 & 0.102 & 0.511 & 1.120 & - & - & - & 0.105 & 0.498 & 1.120 \\
 DA3-Base & 135.4 & 0.045 & 0.080 & 0.655 & 0.068 & 0.677 & 0.685 & - & - & - & 0.074 & 0.666 & 0.685 \\
 DA3-Large & 410.9 & 0.042 & 0.056 & 0.689 & 0.041 & 0.795 & 0.552 & - & - & - & 0.049 & 0.742 & 0.552 \\
 DA3-Giant & 1356 & 0.031 & 0.035 & 0.802 & 0.028 & \cellcolor{secondorange}\textbf{0.880} & 0.196 & - & - & - & 0.031 & \cellcolor{secondorange}0.841 & \cellcolor{thirdyellow}0.196 \\
 DA3-Nested & 1690 & 0.041 & \cellcolor{secondorange}\textbf{0.027} & \cellcolor{secondorange}\textbf{0.805} & \cellcolor{secondorange}\textbf{0.025} & \cellcolor{thirdyellow}0.878 & \cellcolor{thirdyellow}\textbf{0.192} & - & - & - & \cellcolor{bestred}\textbf{0.026} & \cellcolor{bestred}\textbf{0.842} & \cellcolor{secondorange}\textbf{0.192} \\
 WorldMirror & 1263 & 0.031 & 0.040 & 0.709 & 0.039 & 0.785 & 1.411 & - & - & - & 0.040 & 0.747 & 1.411 \\
 VGGT-Omega & 1144 & \cellcolor{bestred}\textbf{0.019} & \cellcolor{bestred}\textbf{0.023} & \cellcolor{bestred}\textbf{0.863} & \cellcolor{bestred}\textbf{0.021} & \cellcolor{bestred}\textbf{0.899} & \cellcolor{bestred}\textbf{0.126} & -- & -- & -- & (0.022) & (0.881) & (0.126) \\ \hdashline
 \ours (Ours) & 1304 & 0.030 & 0.037 & 0.765 & 0.025 & 0.845 & 0.222 & - & - & - & (0.031) & (0.805) & (0.222) \\
\midrule
\multicolumn{14}{c}{\cellcolor{catgray}\textbf{Online}} \\
\midrule
 Spann3r$^{224}$ & 658.7 & 0.156 & 0.256 & 0.256 & 0.264 & 0.256 & 2.833 & - & - & - & 0.260 & 0.256 & 2.833 \\
 CUT3R & 793.3 & 0.040 & 0.100 & 0.428 & 0.104 & 0.449 & 1.354 & - & - & - & 0.102 & 0.439 & 1.354 \\
 MonST3R & 571.2 & 0.051 & 0.131 & 0.207 & 0.196 & 0.078 & 2.126 & - & - & - & 0.163 & 0.142 & 2.126 \\
 Point3R & 828 & 0.046 & 0.173 & 0.223 & 0.177 & 0.181 & 2.042 & - & - & - & 0.175 & 0.202 & 2.042 \\
 Stream3R-S & 1191 & \cellcolor{thirdyellow}\textbf{0.023} & \textbf{0.043} & 0.455 & 0.064 & 0.518 & 1.241 & - & - & - & 0.053 & 0.486 & 1.241 \\
 Stream3R-W & 1191 & \cellcolor{thirdyellow}\textbf{0.023} & \textbf{0.043} & 0.455 & 0.069 & 0.493 & 1.290 & - & - & - & 0.056 & 0.474 & 1.290 \\
 StreamVGGT & 1257 & 0.031 & 0.092 & 0.473 & 0.138 & 0.443 & 1.126 & - & - & - & 0.115 & 0.458 & 1.126 \\
 Page4D & 1257 & 0.031 & 0.046 & 0.527 & \textbf{0.050} & 0.565 & \textbf{0.696} & - & - & - & \textbf{0.048} & 0.546 & \textbf{0.696} \\
 InfiniteVGGT & 1257 & 0.031 & 0.092 & 0.470 & 0.138 & 0.442 & 1.129 & - & - & - & 0.115 & 0.456 & 1.129 \\
 Wint3R & 749.5 & 0.043 & 0.093 & 0.416 & 0.093 & 0.321 & 1.556 & - & - & - & 0.093 & 0.369 & 1.556 \\
 LongStream-B & 1191 & 0.031 & 0.071 & 0.472 & 0.081 & 0.389 & 1.802 & - & - & - & 0.076 & 0.431 & 1.802 \\
 LongStream-S & 1191 & 0.031 & 0.071 & 0.472 & 0.081 & 0.385 & 1.798 & - & - & - & 0.076 & 0.429 & 1.798 \\
 LingbotMap$^{*}$-W & 1158 & 0.038 & 0.052 & \textbf{0.711} & 0.054 & \textbf{0.673} & 1.293 & - & - & - & 0.053 & \textbf{0.692} & 1.293 \\
 LingbotMap$^{*}$-S & 1158 & 0.038 & 0.052 & \textbf{0.711} & 0.054 & 0.673 & 1.293 & - & - & - & 0.053 & \textbf{0.692} & 1.293 \\
\midrule
\multicolumn{14}{c}{\cellcolor{catgray}\textbf{Chunk-wise}} \\
\midrule
 VGGT-Long & 1257 & 0.033 & 0.050 & 0.686 & 0.039 & 0.715 & 0.460 & - & - & - & 0.044 & 0.701 & 0.460 \\
 $\pi^{3}$-Long & 958.7 & \textbf{0.027} & \textbf{0.034} & 0.706 & 0.030 & 0.784 & 0.199 & - & - & - & 0.032 & 0.745 & 0.199 \\
 DA3-Streaming & 1356 & 0.031 & 0.035 & \cellcolor{thirdyellow}\textbf{0.803} & \textbf{0.028} & \cellcolor{secondorange}\textbf{0.880} & \textbf{0.196} & - & - & - & \textbf{0.031} & \cellcolor{secondorange}\textbf{0.841} & \cellcolor{thirdyellow}\textbf{0.196} \\
\midrule
\multicolumn{14}{c}{\cellcolor{catgray}\textbf{SLAM-based}} \\
\midrule
 MASt3R-SLAM & 688.6 & 0.104 & 0.165 & 0.145 & 0.165 & 0.098 & 2.659 & - & - & - & 0.165 & 0.121 & 2.659 \\
 VGGT-SLAM & 1257 & \textbf{0.033} & \textbf{0.050} & \textbf{0.686} & \textbf{0.039} & \textbf{0.715} & \textbf{0.460} & - & - & - & \textbf{0.044} & \textbf{0.701} & \textbf{0.460} \\
\midrule
\multicolumn{14}{c}{\cellcolor{catgray}\textbf{Test-Time Training}} \\
\midrule
 TTT3R & 793.3 & 0.040 & 0.114 & 0.450 & 0.100 & 0.395 & 2.009 & - & - & - & 0.107 & 0.422 & 2.009 \\
 Scal3R & 1266 & \textbf{0.027} & 0.035 & 0.717 & 0.033 & 0.772 & 0.234 & - & - & - & 0.034 & 0.744 & 0.234 \\
 LoGeR & 1255 & 0.032 & 0.035 & 0.717 & \cellcolor{thirdyellow}0.026 & 0.793 & 0.217 & - & - & - & \cellcolor{thirdyellow}0.030 & 0.755 & 0.217 \\
 LoGeR$^*$ & 1255 & 0.030 & \cellcolor{thirdyellow}\textbf{0.031} & \textbf{0.730} & \cellcolor{secondorange}\textbf{0.025} & \textbf{0.828} & \cellcolor{secondorange}\textbf{0.176} & - & - & - & \cellcolor{secondorange}\textbf{0.028} & \cellcolor{thirdyellow}\textbf{0.779} & \cellcolor{bestred}\textbf{0.176} \\
\bottomrule
\end{tabular}%
}
\end{table*}

\begin{table*}[t]
\centering
\setlength{\fboxsep}{0pt}
\caption{\textbf{Per-Dataset Results on Hiroom}.
Performance across different input regimes: \textit{Single Frame}, \textit{Sparse}, \textit{Medium}, \textit{Dense}, and the \textit{Average}.
The best, second-best, and third-best results in each column are highlighted in
\colorbox{bestred}{deep blue}, \colorbox{secondorange}{medium blue}, and \colorbox{thirdyellow}{light blue}, respectively.
This dataset is not evaluated under the \textit{Dense} regime, so the corresponding cells are marked as ``-''.
Within each sub-category, the \textbf{bold} value marks the in-group best.
Note that \ours (Ours) is excluded from the per-column rankings.
}
\label{tab:dataset_hiroom}
\resizebox{\textwidth}{!}{%
\renewcommand{\arraystretch}{1.15}
\setlength{\tabcolsep}{3pt}
\begin{tabular}{l c >{\columncolor{subcol}}c >{\columncolor{subcol}}c >{\columncolor{subcol}}c >{\columncolor{subcol}}c >{\columncolor{subcol}}c >{\columncolor{subcol}}c >{\columncolor{subcol}}c >{\columncolor{subcol}}c >{\columncolor{subcol}}c >{\columncolor{subcol}}c >{\columncolor{subcol}}c >{\columncolor{subcol}}c >{\columncolor{subcol}}c >{\columncolor{subcol}}c >{\columncolor{subcol}}c}
\toprule
\multirow{2.5}{*}{\textbf{Method}}
& \multirow{2.5}{*}{\makecell{\textbf{\#Params}\\\textbf{(M)}}}
& \multicolumn{1}{c}{\textbf{Single Frame}}
& \multicolumn{2}{c}{\textbf{Sparse}}
& \multicolumn{4}{c}{\textbf{Medium}}
& \multicolumn{4}{c}{\textbf{Dense}}
& \multicolumn{4}{c}{\textbf{Average}} \\
\cmidrule(lr){3-3} \cmidrule(lr){4-5} \cmidrule(lr){6-9} \cmidrule(lr){10-13} \cmidrule(lr){14-17}
  & &
 AbsRel$\downarrow$ &
 AbsRel$\downarrow$ &AUC@30$\uparrow$ &
 AbsRel$\downarrow$ &AUC@30$\uparrow$ &ATE$\downarrow$ &F-Score$\uparrow$ &
 AbsRel$\downarrow$ &AUC@30$\uparrow$ &ATE$\downarrow$ &F-Score$\uparrow$ &
 AbsRel$\downarrow$ &AUC@30$\uparrow$ &ATE$\downarrow$ &F-Score$\uparrow$ \\
\midrule
\multicolumn{17}{c}{\cellcolor{catgray}\textbf{Optimization-based}} \\
\midrule
 DUSt3R & 571.2 & \textbf{0.024} & \textbf{0.028} & 0.786 & \textbf{0.031} & 0.853 & 0.079 & 0.289 & - & - & - & - & \textbf{0.030} & 0.820 & 0.079 & 0.289 \\
 MASt3R & 688.6 & 0.056 & 0.075 & \textbf{0.879} & 0.051 & \textbf{0.904} & \textbf{0.054} & \textbf{0.306} & - & - & - & - & 0.063 & \textbf{0.891} & \textbf{0.054} & \textbf{0.306} \\
\midrule
\multicolumn{17}{c}{\cellcolor{catgray}\textbf{End-to-End Feed-Forward}} \\
\midrule
 VGGT & 1257 & 0.024 & 0.020 & 0.792 & 0.017 & 0.848 & 0.091 & 0.558 & - & - & - & - & 0.018 & 0.820 & 0.091 & 0.558 \\
 Fast3R & 647.5 & 0.038 & 0.079 & 0.646 & 0.050 & 0.729 & 0.222 & 0.256 & - & - & - & - & 0.065 & 0.688 & 0.222 & 0.256 \\
 FastVGGT & 1158 & 0.024 & 0.024 & 0.621 & 0.024 & 0.752 & 0.224 & 0.345 & - & - & - & - & 0.024 & 0.687 & 0.224 & 0.345 \\
 MUSt3R & 423.4 & 0.033 & 0.021 & 0.933 & 0.020 & 0.942 & 0.037 & 0.651 & - & - & - & - & 0.021 & 0.938 & 0.037 & 0.651 \\
 MapAnything & 1228 & 0.021 & 0.028 & 0.890 & 0.027 & 0.898 & 0.063 & 0.584 & - & - & - & - & 0.028 & 0.894 & 0.063 & 0.584 \\
 OmniVGGT & 1217 & 0.023 & 0.025 & 0.811 & 0.021 & 0.778 & 0.137 & 0.387 & - & - & - & - & 0.023 & 0.795 & 0.137 & 0.387 \\
 $\pi^{3}$ & 958.7 & 0.020 & 0.022 & 0.932 & 0.013 & 0.948 & 0.033 & 0.679 & - & - & - & - & 0.017 & 0.940 & 0.033 & \cellcolor{thirdyellow}0.679 \\
 $\pi^{3}$-X & 1360 & 0.021 & 0.018 & 0.919 & 0.013 & 0.955 & 0.028 & 0.651 & - & - & - & - & \cellcolor{thirdyellow}0.015 & 0.937 & 0.028 & 0.651 \\
 AMB3R & 1563 & 0.027 & 0.020 & 0.871 & 0.021 & 0.897 & 0.057 & 0.390 & - & - & - & - & 0.021 & 0.884 & 0.057 & 0.390 \\
 DA3-Small & 34.3 & 0.039 & 0.042 & 0.650 & 0.044 & 0.801 & 0.103 & 0.363 & - & - & - & - & 0.043 & 0.725 & 0.103 & 0.363 \\
 DA3-Base & 135.4 & 0.036 & 0.032 & 0.848 & 0.029 & 0.869 & 0.086 & 0.454 & - & - & - & - & 0.031 & 0.859 & 0.086 & 0.454 \\
 DA3-Large & 410.9 & 0.020 & 0.017 & 0.947 & 0.014 & 0.972 & 0.027 & 0.629 & - & - & - & - & 0.016 & \cellcolor{thirdyellow}0.959 & \cellcolor{thirdyellow}0.027 & 0.629 \\
 DA3-Giant & 1356 & \cellcolor{secondorange}\textbf{0.017} & \cellcolor{bestred}\textbf{0.009} & \cellcolor{thirdyellow}0.966 & \cellcolor{bestred}\textbf{0.005} & \cellcolor{bestred}\textbf{0.996} & \cellcolor{bestred}\textbf{0.008} & \cellcolor{bestred}\textbf{0.960} & - & - & - & - & \cellcolor{bestred}\textbf{0.007} & \cellcolor{secondorange}0.981 & \cellcolor{bestred}\textbf{0.008} & \cellcolor{bestred}\textbf{0.960} \\
 DA3-Nested & 1690 & 0.020 & \cellcolor{secondorange}0.010 & \cellcolor{secondorange}\textbf{0.971} & \cellcolor{secondorange}0.006 & \cellcolor{secondorange}0.994 & \cellcolor{secondorange}0.009 & \cellcolor{secondorange}0.947 & - & - & - & - & \cellcolor{secondorange}0.008 & \cellcolor{bestred}\textbf{0.983} & \cellcolor{secondorange}0.009 & \cellcolor{secondorange}0.947 \\
 WorldMirror & 1263 & 0.026 & 0.030 & 0.912 & 0.025 & 0.923 & 0.047 & 0.474 & - & - & - & - & 0.027 & 0.918 & 0.047 & 0.474 \\
 VGGT-Omega & 1144 & \cellcolor{bestred}\textbf{0.008} & \cellcolor{thirdyellow}0.015 & \cellcolor{bestred}\textbf{0.985} & \cellcolor{thirdyellow}0.009 & \cellcolor{thirdyellow}0.980 & \cellcolor{thirdyellow}0.017 & \cellcolor{thirdyellow}0.868 & -- & -- & -- & -- & (0.012) & (0.983) & (0.017) & (0.868) \\ \hdashline
 \ours (Ours) & 1304 & 0.018 & 0.010 & 0.982 & 0.007 & 0.990 & 0.011 & 0.948 & - & - & - & - & (0.012) & (0.986) & (0.011) & (0.948) \\
\midrule
\multicolumn{17}{c}{\cellcolor{catgray}\textbf{Online}} \\
\midrule
 Spann3r$^{224}$ & 658.7 & 0.041 & 0.079 & 0.594 & 0.071 & 0.659 & 0.328 & 0.214 & - & - & - & - & 0.075 & 0.627 & 0.328 & 0.214 \\
 CUT3R & 793.3 & 0.035 & 0.082 & 0.531 & 0.076 & 0.634 & 0.359 & 0.143 & - & - & - & - & 0.079 & 0.583 & 0.359 & 0.143 \\
 MonST3R & 571.2 & 0.033 & 0.094 & 0.327 & 0.070 & 0.167 & 0.702 & 0.050 & - & - & - & - & 0.082 & 0.247 & 0.702 & 0.050 \\
 Point3R & 828 & 0.038 & 0.088 & 0.417 & 0.074 & 0.422 & 0.456 & 0.138 & - & - & - & - & 0.081 & 0.419 & 0.456 & 0.138 \\
 Stream3R-S & 1191 & 0.025 & 0.030 & 0.781 & 0.030 & 0.777 & 0.158 & 0.269 & - & - & - & - & 0.030 & 0.779 & 0.158 & 0.269 \\
 Stream3R-W & 1191 & 0.025 & 0.030 & 0.781 & 0.030 & 0.771 & 0.159 & 0.256 & - & - & - & - & 0.030 & 0.776 & 0.159 & 0.256 \\
 StreamVGGT & 1257 & 0.024 & 0.076 & 0.736 & 0.057 & 0.688 & 0.274 & 0.079 & - & - & - & - & 0.066 & 0.712 & 0.274 & 0.079 \\
 Page4D & 1257 & 0.028 & \textbf{0.029} & 0.736 & \textbf{0.025} & \textbf{0.858} & \textbf{0.076} & \textbf{0.314} & - & - & - & - & \textbf{0.027} & 0.797 & \textbf{0.076} & \textbf{0.314} \\
 InfiniteVGGT & 1257 & 0.023 & 0.075 & 0.731 & 0.057 & 0.687 & 0.274 & 0.078 & - & - & - & - & 0.066 & 0.709 & 0.274 & 0.078 \\
 Wint3R & 749.5 & 0.028 & 0.035 & 0.728 & 0.034 & 0.648 & 0.268 & 0.252 & - & - & - & - & 0.035 & 0.688 & 0.268 & 0.252 \\
 LongStream-B & 1191 & \cellcolor{thirdyellow}\textbf{0.019} & 0.033 & 0.819 & 0.030 & 0.759 & 0.224 & 0.052 & - & - & - & - & 0.031 & 0.789 & 0.224 & 0.052 \\
 LongStream-S & 1191 & \cellcolor{thirdyellow}\textbf{0.019} & 0.033 & 0.819 & 0.030 & 0.759 & 0.224 & 0.053 & - & - & - & - & 0.031 & 0.789 & 0.224 & 0.053 \\
 LingbotMap$^{*}$-W & 1158 & 0.027 & 0.034 & \textbf{0.870} & 0.036 & 0.769 & 0.127 & 0.278 & - & - & - & - & 0.035 & \textbf{0.820} & 0.127 & 0.278 \\
 LingbotMap$^{*}$-S & 1158 & 0.027 & 0.034 & \textbf{0.870} & 0.036 & 0.769 & 0.127 & 0.278 & - & - & - & - & 0.035 & \textbf{0.820} & 0.127 & 0.278 \\
\midrule
\multicolumn{17}{c}{\cellcolor{catgray}\textbf{Chunk-wise}} \\
\midrule
 VGGT-Long & 1257 & 0.024 & 0.020 & 0.792 & 0.017 & 0.848 & 0.091 & 0.559 & - & - & - & - & 0.018 & 0.820 & 0.091 & 0.559 \\
 $\pi^{3}$-Long & 958.7 & 0.020 & 0.022 & 0.932 & 0.013 & 0.948 & 0.033 & 0.679 & - & - & - & - & 0.017 & 0.940 & 0.033 & \cellcolor{thirdyellow}0.679 \\
 DA3-Streaming & 1356 & \cellcolor{secondorange}\textbf{0.017} & \cellcolor{bestred}\textbf{0.009} & \cellcolor{thirdyellow}\textbf{0.966} & \cellcolor{bestred}\textbf{0.005} & \cellcolor{bestred}\textbf{0.996} & \cellcolor{bestred}\textbf{0.008} & \cellcolor{bestred}\textbf{0.960} & - & - & - & - & \cellcolor{bestred}\textbf{0.007} & \cellcolor{secondorange}\textbf{0.981} & \cellcolor{bestred}\textbf{0.008} & \cellcolor{bestred}\textbf{0.960} \\
\midrule
\multicolumn{17}{c}{\cellcolor{catgray}\textbf{SLAM-based}} \\
\midrule
 MASt3R-SLAM & 688.6 & 0.159 & 0.202 & 0.194 & 0.171 & 0.238 & 0.845 & 0.021 & - & - & - & - & 0.187 & 0.216 & 0.845 & 0.021 \\
 VGGT-SLAM & 1257 & \textbf{0.024} & \textbf{0.020} & \textbf{0.792} & \textbf{0.017} & \textbf{0.848} & \textbf{0.091} & \textbf{0.558} & - & - & - & - & \textbf{0.018} & \textbf{0.820} & \textbf{0.091} & \textbf{0.558} \\
\midrule
\multicolumn{17}{c}{\cellcolor{catgray}\textbf{Test-Time Training}} \\
\midrule
 TTT3R & 793.3 & 0.035 & 0.086 & 0.486 & 0.080 & 0.608 & 0.380 & 0.136 & - & - & - & - & 0.083 & 0.547 & 0.380 & 0.136 \\
 Scal3R & 1266 & \textbf{0.022} & 0.019 & \textbf{0.859} & \textbf{0.014} & \textbf{0.958} & \textbf{0.034} & \textbf{0.651} & - & - & - & - & \textbf{0.016} & \textbf{0.909} & \textbf{0.034} & \textbf{0.651} \\
 LoGeR & 1255 & 0.025 & \textbf{0.018} & 0.836 & 0.017 & 0.915 & 0.046 & 0.342 & - & - & - & - & 0.017 & 0.875 & 0.046 & 0.342 \\
 LoGeR$^*$ & 1255 & 0.023 & 0.020 & 0.856 & 0.019 & 0.908 & 0.049 & 0.402 & - & - & - & - & 0.019 & 0.882 & 0.049 & 0.402 \\
\bottomrule
\end{tabular}%
}
\end{table*}

\begin{table*}[t]
\centering
\caption{\textbf{Per-Dataset Results on KITTI-Odometry}.
Only metrics available for the \textit{Dense} regime are reported.
The best, second-best, and third-best results in each column are highlighted in
\colorbox{bestred}{deep blue}, \colorbox{secondorange}{medium blue}, and \colorbox{thirdyellow}{light blue}, respectively.
Out-of-memory (OOM) and Timeout (T.O) cells are shaded \colorbox{oomred}{light red}; methods without dense-regime results use ``--''.
Within each sub-category, the \textbf{bold} value marks the in-group best.
Note that \ours (Ours) is excluded from the per-column rankings.
}
\label{tab:dataset_kittiodometry}
\resizebox{0.45\textwidth}{!}{%
\renewcommand{\arraystretch}{1.15}
\setlength{\tabcolsep}{3pt}
\begin{tabular}{l c >{\columncolor{subcol}}c >{\columncolor{subcol}}c >{\columncolor{subcol}}c >{\columncolor{subcol}}c}
\toprule
\multirow{2.5}{*}{\textbf{Method}}
& \multirow{2.5}{*}{\makecell{\textbf{\#Params}\\\textbf{(M)}}}
& \multicolumn{1}{c}{\textbf{Camera}}
& \multicolumn{3}{c}{\textbf{Trajectory}} \\
\cmidrule(lr){3-3} \cmidrule(lr){4-6}
 & & AUC@30$\uparrow$ & ATE$\downarrow$ & RPE$_t\downarrow$ & RPE$_r\downarrow$ \\
\midrule
\multicolumn{6}{c}{\cellcolor{catgray}\textbf{Optimization-based}} \\
\midrule
 DUSt3R & 571.2 & \cellcolor{oomred}OOM & \cellcolor{oomred}OOM & \cellcolor{oomred}OOM & \cellcolor{oomred}OOM \\
 MASt3R & 688.6 & \cellcolor{oomred}OOM & \cellcolor{oomred}OOM & \cellcolor{oomred}OOM & \cellcolor{oomred}OOM \\
\midrule
\multicolumn{6}{c}{\cellcolor{catgray}\textbf{End-to-End Feed-Forward}} \\
\midrule
 VGGT & 1257 & \cellcolor{oomred}OOM & \cellcolor{oomred}OOM & \cellcolor{oomred}OOM & \cellcolor{oomred}OOM \\
 Fast3R & 647.5 & 0.000 & \textbf{134.5} & 29.12 & 27.34 \\
 FastVGGT & 1158 & 0.307 & 181.9 & \textbf{9.827} & 3.338 \\
 MUSt3R & 423.4 & \cellcolor{oomred}T.O & \cellcolor{oomred}T.O & \cellcolor{oomred}T.O & \cellcolor{oomred}T.O \\
 MapAnything & 1228 & \cellcolor{oomred}OOM & \cellcolor{oomred}OOM & \cellcolor{oomred}OOM & \cellcolor{oomred}OOM \\
 OmniVGGT & 1217 & \cellcolor{oomred}OOM & \cellcolor{oomred}OOM & \cellcolor{oomred}OOM & \cellcolor{oomred}OOM \\
 $\pi^{3}$ & 958.7 & \textbf{0.412} & 153 & 10.07 & \textbf{1.369} \\
 $\pi^{3}$-X & 1360 & \cellcolor{oomred}OOM & \cellcolor{oomred}OOM & \cellcolor{oomred}OOM & \cellcolor{oomred}OOM \\
 AMB3R & 1563 & \cellcolor{oomred}OOM & \cellcolor{oomred}OOM & \cellcolor{oomred}OOM & \cellcolor{oomred}OOM \\
 DA3-Small & 34.3 & 0.089 & 220.3 & 29 & 17.12 \\
 DA3-Base & 135.4 & 0.067 & 211.7 & 30.08 & 40.69 \\
 DA3-Large & 410.9 & \cellcolor{oomred}OOM & \cellcolor{oomred}OOM & \cellcolor{oomred}OOM & \cellcolor{oomred}OOM \\
 DA3-Giant & 1356 & \cellcolor{oomred}OOM & \cellcolor{oomred}OOM & \cellcolor{oomred}OOM & \cellcolor{oomred}OOM \\
 DA3-Nested & 1690 & \cellcolor{oomred}OOM & \cellcolor{oomred}OOM & \cellcolor{oomred}OOM & \cellcolor{oomred}OOM \\
 WorldMirror & 1263 & \cellcolor{oomred}OOM & \cellcolor{oomred}OOM & \cellcolor{oomred}OOM & \cellcolor{oomred}OOM \\
 VGGT-Omega & 1144 & -- & -- & -- & -- \\
 \ours (Ours) & 1304 & \cellcolor{oomred}OOM & \cellcolor{oomred}OOM & \cellcolor{oomred}OOM & \cellcolor{oomred}OOM \\
\midrule
\multicolumn{6}{c}{\cellcolor{catgray}\textbf{Online}} \\
\midrule
 Spann3r$^{224}$ & 658.7 & 0.036 & 214.9 & 23.33 & 18.99 \\
 CUT3R & 793.3 & 0.065 & 206.6 & 2.763 & 1.830 \\
 MonST3R & 571.2 & \cellcolor{oomred}OOM & \cellcolor{oomred}OOM & \cellcolor{oomred}OOM & \cellcolor{oomred}OOM \\
 Point3R & 828 & 0.051 & 206.5 & 7.671 & 4.421 \\
 Stream3R-S & 1191 & \cellcolor{oomred}OOM & \cellcolor{oomred}OOM & \cellcolor{oomred}OOM & \cellcolor{oomred}OOM \\
 Stream3R-W & 1191 & \cellcolor{oomred}OOM & \cellcolor{oomred}OOM & \cellcolor{oomred}OOM & \cellcolor{oomred}OOM \\
 StreamVGGT & 1257 & 0.168 & 206.2 & 14 & 4.174 \\
 Page4D & 1257 & \cellcolor{oomred}OOM & \cellcolor{oomred}OOM & \cellcolor{oomred}OOM & \cellcolor{oomred}OOM \\
 InfiniteVGGT & 1257 & 0.169 & 206.7 & 13.87 & 4.164 \\
 Wint3R & 749.5 & 0.074 & 211.3 & 5.458 & 1.624 \\
 LongStream-B & 1191 & 0.317 & 47.33 & \cellcolor{secondorange}\textbf{0.348} & 0.331 \\
 LongStream-S & 1191 & 0.211 & 88.6 & 0.885 & 0.620 \\
 LingbotMap$^{*}$-W & 1158 & 0.615 & 41.16 & 0.368 & \textbf{0.286} \\
 LingbotMap$^{*}$-S & 1158 & \cellcolor{bestred}\textbf{0.729} & \cellcolor{secondorange}\textbf{29.31} & 2.440 & 0.663 \\
\midrule
\multicolumn{6}{c}{\cellcolor{catgray}\textbf{Chunk-wise}} \\
\midrule
 VGGT-Long & 1257 & 0.465 & 76.14 & 0.798 & 0.468 \\
 $\pi^{3}$-Long & 958.7 & \cellcolor{secondorange}\textbf{0.702} & \cellcolor{thirdyellow}\textbf{31.91} & \cellcolor{thirdyellow}\textbf{0.362} & \cellcolor{bestred}\textbf{0.181} \\
 DA3-Streaming & 1356 & 0.221 & 79.26 & 0.633 & 0.689 \\
\midrule
\multicolumn{6}{c}{\cellcolor{catgray}\textbf{SLAM-based}} \\
\midrule
 MASt3R-SLAM & 688.6 & 0.158 & 196.5 & 1.737 & 0.561 \\
 VGGT-SLAM & 1257 & \textbf{0.436} & \textbf{80.06} & \textbf{0.607} & \cellcolor{thirdyellow}\textbf{0.226} \\
\midrule
\multicolumn{6}{c}{\cellcolor{catgray}\textbf{Test-Time Training}} \\
\midrule
 TTT3R & 793.3 & 0.086 & 182.1 & 3.824 & 2.146 \\
 Scal3R & 1266 & \cellcolor{thirdyellow}\textbf{0.655} & \cellcolor{bestred}\textbf{18.17} & 0.457 & 0.649 \\
 LoGeR & 1255 & 0.457 & 45.03 & 0.382 & 0.238 \\
 LoGeR$^*$ & 1255 & 0.456 & 39.62 & \cellcolor{bestred}\textbf{0.286} & \cellcolor{secondorange}\textbf{0.220} \\
\bottomrule
\end{tabular}%
}
\end{table*}

\begin{table*}[t]
\centering
\caption{\textbf{Per-Dataset Results on Lingbot-Depth}.
Only metrics available for the \textit{Single Frame} regime are reported.
The best, second-best, and third-best results in each column are highlighted in
\colorbox{bestred}{deep blue}, \colorbox{secondorange}{medium blue}, and \colorbox{thirdyellow}{light blue}, respectively.
Within each sub-category, the \textbf{bold} value marks the in-group best.
Note that \ours (Ours) is excluded from the per-column rankings.
}
\label{tab:dataset_lingbot}
\resizebox{0.5\textwidth}{!}{%
\renewcommand{\arraystretch}{1.15}
\setlength{\tabcolsep}{3pt}
\begin{tabular}{l c >{\columncolor{subcol}}c >{\columncolor{subcol}}c >{\columncolor{subcol}}c >{\columncolor{subcol}}c >{\columncolor{subcol}}c >{\columncolor{subcol}}c}
\toprule
\multirow{2.5}{*}{\textbf{Method}}
& \multirow{2.5}{*}{\makecell{\textbf{\#Params}\\\textbf{(M)}}}
& \multicolumn{6}{c}{\textbf{Depth}} \\
\cmidrule(lr){3-8}
 & & AbsRel$\downarrow$ & SqRel$\downarrow$ & RMSE$\downarrow$ & $\delta_{1.03}\uparrow$ & $\delta_{1.05}\uparrow$ & $\delta_{1.10}\uparrow$ \\
\midrule
\multicolumn{8}{c}{\cellcolor{catgray}\textbf{Optimization-based}} \\
\midrule
 DUSt3R & 571.2 & \textbf{0.949} & \textbf{18.34} & \textbf{0.836} & 0.308 & 0.439 & 0.636 \\
 MASt3R & 688.6 & 1.227 & 35.58 & 0.911 & \textbf{0.349} & \textbf{0.485} & \textbf{0.682} \\
\midrule
\multicolumn{8}{c}{\cellcolor{catgray}\textbf{End-to-End Feed-Forward}} \\
\midrule
 VGGT & 1257 & \cellcolor{thirdyellow}0.428 & \cellcolor{secondorange}2.572 & 0.692 & 0.387 & 0.519 & 0.709 \\
 Fast3R & 647.5 & 0.804 & 11.42 & 0.826 & 0.285 & 0.390 & 0.543 \\
 FastVGGT & 1158 & \cellcolor{secondorange}0.424 & \cellcolor{bestred}\textbf{2.505} & 0.692 & 0.388 & 0.520 & 0.710 \\
 MUSt3R & 423.4 & 1.110 & 30.18 & 0.831 & 0.368 & 0.484 & 0.681 \\
 MapAnything & 1228 & 1.326 & 42.33 & 0.784 & 0.423 & 0.571 & 0.749 \\
 OmniVGGT & 1217 & \cellcolor{bestred}\textbf{0.419} & \cellcolor{thirdyellow}2.593 & \textbf{0.614} & 0.428 & 0.557 & 0.751 \\
 $\pi^{3}$ & 958.7 & 1.485 & 59.37 & 0.793 & \cellcolor{secondorange}\textbf{0.480} & \cellcolor{bestred}\textbf{0.622} & \cellcolor{thirdyellow}0.771 \\
 $\pi^{3}$-X & 1360 & 1.118 & 31.36 & 0.674 & 0.461 & 0.592 & \cellcolor{bestred}\textbf{0.780} \\
 AMB3R & 1563 & 1.453 & 54.6 & 0.762 & 0.439 & 0.572 & 0.750 \\
 DA3-Small & 34.3 & 0.973 & 22.16 & 0.796 & 0.240 & 0.359 & 0.578 \\
 DA3-Base & 135.4 & 0.875 & 18.38 & 0.661 & 0.338 & 0.480 & 0.694 \\
 DA3-Large & 410.9 & 0.881 & 19.73 & 0.628 & 0.382 & 0.534 & 0.726 \\
 DA3-Giant & 1356 & 1.034 & 26.41 & 0.693 & 0.440 & 0.586 & 0.744 \\
 DA3-Nested & 1690 & 1.015 & 25.81 & 0.707 & 0.448 & 0.594 & 0.758 \\
 WorldMirror & 1263 & 0.962 & 27.24 & 0.681 & 0.434 & 0.567 & 0.746 \\
 VGGT-Omega & 1144 & 1.647 & 73.16 & 0.889 & 0.450 & 0.582 & 0.768 \\
 \ours (Ours) & 1304 & 0.405 & 3.183 & 0.611 & 0.399 & 0.529 & 0.715 \\
\midrule
\multicolumn{8}{c}{\cellcolor{catgray}\textbf{Online}} \\
\midrule
 Spann3r$^{224}$ & 658.7 & 0.843 & 13.87 & 1.055 & 0.203 & 0.321 & 0.509 \\
 CUT3R & 793.3 & \textbf{0.543} & \textbf{4.487} & 0.611 & 0.347 & 0.482 & 0.677 \\
 MonST3R & 571.2 & 0.660 & 8.235 & 0.792 & 0.332 & 0.455 & 0.636 \\
 Point3R & 828 & 0.988 & 23.95 & 0.766 & 0.300 & 0.422 & 0.632 \\
 Stream3R-S & 1191 & 1.228 & 40.11 & 0.741 & \cellcolor{thirdyellow}\textbf{0.474} & \cellcolor{secondorange}\textbf{0.615} & \cellcolor{secondorange}\textbf{0.773} \\
 Stream3R-W & 1191 & 1.228 & 40.11 & 0.741 & \cellcolor{thirdyellow}\textbf{0.474} & \cellcolor{secondorange}\textbf{0.615} & \cellcolor{secondorange}\textbf{0.773} \\
 StreamVGGT & 1257 & 0.553 & 5.858 & \cellcolor{thirdyellow}0.611 & 0.399 & 0.528 & 0.727 \\
 Page4D & 1257 & 0.566 & 6.851 & 0.734 & 0.355 & 0.474 & 0.667 \\
 InfiniteVGGT & 1257 & 0.547 & 5.667 & \cellcolor{secondorange}\textbf{0.610} & 0.399 & 0.528 & 0.727 \\
 Wint3R & 749.5 & 1.869 & 90.54 & 1.231 & 0.385 & 0.502 & 0.682 \\
 LongStream-B & 1191 & 1.572 & 60.93 & 0.831 & 0.372 & 0.510 & 0.722 \\
 LongStream-S & 1191 & 1.572 & 60.93 & 0.831 & 0.372 & 0.510 & 0.722 \\
 LingbotMap$^{*}$-W & 1158 & 0.883 & 15.64 & 0.768 & 0.385 & 0.516 & 0.693 \\
 LingbotMap$^{*}$-S & 1158 & 0.883 & 15.64 & 0.768 & 0.385 & 0.516 & 0.693 \\
\midrule
\multicolumn{8}{c}{\cellcolor{catgray}\textbf{Chunk-wise}} \\
\midrule
 VGGT-Long & 1257 & \cellcolor{thirdyellow}\textbf{0.428} & \cellcolor{secondorange}\textbf{2.572} & \textbf{0.692} & 0.387 & 0.519 & 0.709 \\
 $\pi^{3}$-Long & 958.7 & 1.485 & 59.37 & 0.793 & \cellcolor{secondorange}\textbf{0.480} & \cellcolor{bestred}\textbf{0.622} & \cellcolor{thirdyellow}\textbf{0.771} \\
 DA3-Streaming & 1356 & 1.034 & 26.41 & 0.693 & 0.440 & 0.586 & 0.744 \\
\midrule
\multicolumn{8}{c}{\cellcolor{catgray}\textbf{SLAM-based}} \\
\midrule
 MASt3R-SLAM & 688.6 & 0.619 & 4.389 & 0.784 & 0.150 & 0.241 & 0.438 \\
 VGGT-SLAM & 1257 & \cellcolor{thirdyellow}\textbf{0.428} & \cellcolor{secondorange}\textbf{2.572} & \textbf{0.692} & \textbf{0.387} & \textbf{0.519} & \textbf{0.709} \\
\midrule
\multicolumn{8}{c}{\cellcolor{catgray}\textbf{Test-Time Training}} \\
\midrule
 TTT3R & 793.3 & 0.543 & 4.488 & 0.611 & 0.347 & 0.482 & 0.677 \\
 Scal3R & 1266 & 0.546 & 5.438 & 0.682 & 0.424 & 0.555 & 0.733 \\
 LoGeR & 1255 & 0.662 & 9.432 & 0.617 & 0.465 & \cellcolor{thirdyellow}\textbf{0.607} & 0.764 \\
 LoGeR$^*$ & 1255 & \textbf{0.506} & \textbf{4.440} & \cellcolor{bestred}\textbf{0.607} & \cellcolor{bestred}\textbf{0.487} & 0.605 & \textbf{0.769} \\
\bottomrule
\end{tabular}%
}
\end{table*}

\begin{table*}[t]
\centering
\setlength{\fboxsep}{0pt}
\caption{\textbf{Per-Dataset Results on NRGBD}.
Performance across different input regimes: \textit{Single Frame}, \textit{Sparse}, \textit{Medium}, \textit{Dense}, and the \textit{Average}.
The best, second-best, and third-best results in each column are highlighted in
\colorbox{bestred}{deep blue}, \colorbox{secondorange}{medium blue}, and \colorbox{thirdyellow}{light blue}, respectively.
Out-of-memory (OOM) and Timeout (T.O) cells are shaded \colorbox{oomred}{light red};
\textit{Average} values for those rows are wrapped in parentheses and excluded from per-column ranking.
Within each sub-category, the \textbf{bold} value marks the in-group best.
Note that \ours (Ours) is excluded from the per-column rankings.
}
\label{tab:dataset_nrgbd}
\resizebox{\textwidth}{!}{%
\renewcommand{\arraystretch}{1.15}
\setlength{\tabcolsep}{3pt}
\begin{tabular}{l c >{\columncolor{subcol}}c >{\columncolor{subcol}}c >{\columncolor{subcol}}c >{\columncolor{subcol}}c >{\columncolor{subcol}}c >{\columncolor{subcol}}c >{\columncolor{subcol}}c >{\columncolor{subcol}}c >{\columncolor{subcol}}c >{\columncolor{subcol}}c >{\columncolor{subcol}}c >{\columncolor{subcol}}c >{\columncolor{subcol}}c >{\columncolor{subcol}}c >{\columncolor{subcol}}c}
\toprule
\multirow{2.5}{*}{\textbf{Method}}
& \multirow{2.5}{*}{\makecell{\textbf{\#Params}\\\textbf{(M)}}}
& \multicolumn{1}{c}{\textbf{Single Frame}}
& \multicolumn{2}{c}{\textbf{Sparse}}
& \multicolumn{4}{c}{\textbf{Medium}}
& \multicolumn{4}{c}{\textbf{Dense}}
& \multicolumn{4}{c}{\textbf{Average}} \\
\cmidrule(lr){3-3} \cmidrule(lr){4-5} \cmidrule(lr){6-9} \cmidrule(lr){10-13} \cmidrule(lr){14-17}
 & &
 AbsRel$\downarrow$ &
 AbsRel$\downarrow$ & AUC@30$\uparrow$ &
 AbsRel$\downarrow$ & AUC@30$\uparrow$ & ATE$\downarrow$ & F-Score$\uparrow$ &
 AbsRel$\downarrow$ & AUC@30$\uparrow$ & ATE$\downarrow$ & F-Score$\uparrow$ &
 AbsRel$\downarrow$ & AUC@30$\uparrow$ & ATE$\downarrow$ & F-Score$\uparrow$ \\
\midrule
\multicolumn{17}{c}{\cellcolor{catgray}\textbf{Optimization-based}} \\
\midrule
 DUSt3R & 571.2 & \textbf{0.028} & \textbf{0.038} & 0.843 & \textbf{0.036} & \textbf{0.783} & 0.265 & \textbf{0.327} & \cellcolor{oomred}OOM & \cellcolor{oomred}OOM & \cellcolor{oomred}OOM & \cellcolor{oomred}OOM & (0.037) & (0.813) & (0.265) & (0.327) \\
 MASt3R & 688.6 & 0.036 & 0.043 & \textbf{0.891} & 0.048 & 0.778 & \textbf{0.205} & 0.290 & \cellcolor{oomred}OOM & \cellcolor{oomred}OOM & \cellcolor{oomred}OOM & \cellcolor{oomred}OOM & (0.046) & (0.834) & (0.205) & (0.290) \\
\midrule
\multicolumn{17}{c}{\cellcolor{catgray}\textbf{End-to-End Feed-Forward}} \\
\midrule
 VGGT & 1257 & 0.024 & 0.016 & 0.926 & 0.012 & 0.956 & 0.039 & 0.696 & \cellcolor{oomred}OOM & \cellcolor{oomred}OOM & \cellcolor{oomred}OOM & \cellcolor{oomred}OOM & (0.014) & (0.941) & (0.039) & (0.696) \\
 Fast3R & 647.5 & 0.042 & 0.073 & 0.779 & 0.041 & 0.858 & 0.102 & 0.345 & 0.039 & 0.592 & 0.157 & 0.351 & 0.051 & 0.743 & 0.130 & 0.348 \\
 FastVGGT & 1158 & 0.026 & 0.023 & 0.888 & 0.014 & 0.947 & 0.044 & 0.569 & \cellcolor{bestred}\textbf{0.019} & 0.831 & 0.162 & \textbf{0.470} & \cellcolor{thirdyellow}\textbf{0.019} & 0.889 & 0.103 & \textbf{0.519} \\
 MUSt3R & 423.4 & 0.024 & 0.026 & 0.912 & 0.021 & 0.932 & 0.053 & 0.569 & \cellcolor{oomred}T.O & \cellcolor{oomred}T.O & \cellcolor{oomred}T.O & \cellcolor{oomred}T.O & (0.024) & (0.922) & (0.053) & (0.569) \\
 MapAnything & 1228 & 0.026 & 0.035 & 0.872 & 0.031 & 0.886 & 0.093 & 0.437 & \cellcolor{oomred}OOM & \cellcolor{oomred}OOM & \cellcolor{oomred}OOM & \cellcolor{oomred}OOM & (0.033) & (0.879) & (0.093) & (0.437) \\
 OmniVGGT & 1217 & \cellcolor{thirdyellow}\textbf{0.016} & 0.020 & 0.913 & 0.013 & 0.948 & 0.039 & 0.728 & \cellcolor{oomred}OOM & \cellcolor{oomred}OOM & \cellcolor{oomred}OOM & \cellcolor{oomred}OOM & (0.016) & (0.931) & (0.039) & (0.728) \\
 $\pi^{3}$ & 958.7 & 0.021 & 0.017 & 0.930 & 0.012 & 0.958 & 0.030 & 0.735 & 0.074 & 0.158 & 1.158 & 0.039 & 0.034 & 0.682 & 0.594 & 0.387 \\
 $\pi^{3}$-X & 1360 & 0.023 & 0.016 & \cellcolor{thirdyellow}0.940 & 0.012 & \cellcolor{thirdyellow}0.962 & \cellcolor{thirdyellow}0.029 & \cellcolor{thirdyellow}0.770 & \cellcolor{oomred}OOM & \cellcolor{oomred}OOM & \cellcolor{oomred}OOM & \cellcolor{oomred}OOM & (0.014) & (0.951) & (0.029) & (0.770) \\
 AMB3R & 1563 & 0.022 & 0.024 & 0.933 & 0.015 & 0.953 & 0.038 & 0.688 & \cellcolor{oomred}OOM & \cellcolor{oomred}OOM & \cellcolor{oomred}OOM & \cellcolor{oomred}OOM & (0.020) & (0.943) & (0.038) & (0.688) \\
 DA3-Small & 34.3 & 0.043 & 0.034 & 0.822 & 0.031 & 0.888 & 0.080 & 0.428 & 0.032 & 0.861 & \cellcolor{secondorange}0.094 & 0.412 & 0.033 & 0.857 & \cellcolor{thirdyellow}0.087 & 0.420 \\
 DA3-Base & 135.4 & 0.031 & 0.024 & 0.891 & 0.020 & 0.937 & 0.048 & 0.561 & \cellcolor{bestred}0.019 & \cellcolor{secondorange}\textbf{0.929} & \cellcolor{bestred}\textbf{0.055} & 0.467 & 0.021 & \cellcolor{secondorange}\textbf{0.919} & \cellcolor{bestred}\textbf{0.051} & 0.514 \\
 DA3-Large & 410.9 & 0.024 & 0.019 & 0.918 & 0.014 & \cellcolor{thirdyellow}0.962 & 0.030 & 0.624 & \cellcolor{oomred}OOM & \cellcolor{oomred}OOM & \cellcolor{oomred}OOM & \cellcolor{oomred}OOM & (0.017) & (0.940) & (0.030) & (0.624) \\
 DA3-Giant & 1356 & 0.026 & \cellcolor{bestred}\textbf{0.009} & \cellcolor{bestred}\textbf{0.944} & \cellcolor{secondorange}\textbf{0.009} & \cellcolor{bestred}0.971 & \cellcolor{bestred}0.025 & \cellcolor{bestred}\textbf{0.811} & \cellcolor{oomred}OOM & \cellcolor{oomred}OOM & \cellcolor{oomred}OOM & \cellcolor{oomred}OOM & (0.009) & (0.957) & (0.025) & (0.811) \\
 DA3-Nested & 1690 & 0.023 & \cellcolor{thirdyellow}0.013 & \cellcolor{secondorange}0.942 & \cellcolor{thirdyellow}0.010 & \cellcolor{bestred}\textbf{0.971} & \cellcolor{bestred}\textbf{0.025} & \cellcolor{bestred}0.811 & \cellcolor{oomred}OOM & \cellcolor{oomred}OOM & \cellcolor{oomred}OOM & \cellcolor{oomred}OOM & (0.011) & (0.957) & (0.025) & (0.811) \\
 WorldMirror & 1263 & 0.020 & 0.019 & 0.916 & 0.014 & 0.948 & 0.041 & 0.603 & \cellcolor{oomred}OOM & \cellcolor{oomred}OOM & \cellcolor{oomred}OOM & \cellcolor{oomred}OOM & (0.017) & (0.932) & (0.041) & (0.603) \\
 VGGT-Omega & 1144 & \cellcolor{bestred}\textbf{0.008} & \cellcolor{secondorange}0.010 & 0.935 & \cellcolor{bestred}\textbf{0.007} & 0.961 & 0.031 & 0.732 & -- & -- & -- & -- & (0.008) & (0.948) & (0.031) & (0.732) \\ \hdashline
 \ours (Ours) & 1304 & 0.028 & 0.014 & 0.943 & 0.011 & 0.966 & 0.028 & 0.837 & \cellcolor{oomred}OOM & \cellcolor{oomred}OOM & \cellcolor{oomred}OOM & \cellcolor{oomred}OOM & (0.018) & (0.955) & (0.028) & (0.837) \\
\midrule
\multicolumn{17}{c}{\cellcolor{catgray}\textbf{Online}} \\
\midrule
 Spann3r$^{224}$ & 658.7 & 0.040 & 0.083 & 0.600 & 0.053 & 0.711 & 0.241 & 0.149 & 0.067 & 0.687 & 0.258 & 0.185 & 0.068 & 0.666 & 0.250 & 0.167 \\
 CUT3R & 793.3 & 0.044 & 0.044 & 0.847 & 0.056 & 0.738 & 0.208 & 0.250 & 0.089 & 0.085 & 1.308 & 0.074 & 0.063 & 0.557 & 0.758 & 0.162 \\
 MonST3R & 571.2 & 0.026 & 0.064 & 0.260 & 0.050 & 0.377 & 0.574 & 0.151 & \cellcolor{oomred}OOM & \cellcolor{oomred}OOM & \cellcolor{oomred}OOM & \cellcolor{oomred}OOM & (0.057) & (0.318) & (0.574) & (0.151) \\
 Point3R & 828 & 0.024 & 0.055 & 0.582 & 0.048 & 0.724 & 0.191 & 0.238 & 0.061 & 0.618 & 0.430 & 0.149 & 0.054 & 0.641 & 0.311 & 0.193 \\
 Stream3R-S & 1191 & 0.024 & 0.031 & 0.893 & 0.105 & 0.384 & 1.039 & 0.100 & \cellcolor{oomred}OOM & \cellcolor{oomred}OOM & \cellcolor{oomred}OOM & \cellcolor{oomred}OOM & (0.068) & (0.639) & (1.039) & (0.100) \\
 Stream3R-W & 1191 & 0.024 & 0.031 & \textbf{0.895} & 0.251 & 0.354 & 1.170 & 0.088 & \cellcolor{oomred}OOM & \cellcolor{oomred}OOM & \cellcolor{oomred}OOM & \cellcolor{oomred}OOM & (0.141) & (0.625) & (1.170) & (0.088) \\
 StreamVGGT & 1257 & 0.026 & 0.046 & 0.871 & 0.051 & 0.890 & 0.136 & 0.196 & 0.051 & 0.739 & 0.241 & 0.211 & 0.049 & 0.834 & 0.188 & 0.204 \\
 Page4D & 1257 & 0.030 & \textbf{0.029} & 0.877 & 0.024 & 0.898 & 0.063 & 0.364 & \cellcolor{oomred}OOM & \cellcolor{oomred}OOM & \cellcolor{oomred}OOM & \cellcolor{oomred}OOM & (0.026) & (0.887) & (0.063) & (0.364) \\
 InfiniteVGGT & 1257 & 0.026 & 0.046 & 0.871 & 0.051 & 0.889 & 0.138 & 0.194 & 0.051 & 0.740 & 0.239 & 0.211 & 0.049 & 0.833 & 0.189 & 0.202 \\
 Wint3R & 749.5 & \textbf{0.023} & 0.032 & 0.779 & 0.027 & 0.796 & 0.254 & 0.472 & 0.083 & 0.264 & 0.750 & 0.091 & 0.047 & 0.613 & 0.502 & 0.281 \\
 LongStream-B & 1191 & 0.025 & 0.031 & 0.876 & 0.082 & 0.656 & 0.376 & 0.060 & 0.091 & 0.277 & 0.692 & 0.053 & 0.068 & 0.603 & 0.534 & 0.057 \\
 LongStream-S & 1191 & 0.025 & 0.031 & 0.876 & 0.058 & 0.392 & 0.685 & 0.060 & 0.090 & 0.169 & 0.849 & 0.050 & 0.060 & 0.479 & 0.767 & 0.055 \\
 LingbotMap$^{*}$-W & 1158 & 0.028 & 0.030 & 0.885 & 0.027 & 0.911 & 0.085 & 0.364 & 0.049 & 0.790 & 0.176 & 0.288 & 0.035 & 0.862 & 0.130 & 0.326 \\
 LingbotMap$^{*}$-S & 1158 & 0.028 & 0.030 & 0.885 & \textbf{0.022} & \textbf{0.929} & \textbf{0.055} & \textbf{0.524} & \cellcolor{secondorange}\textbf{0.021} & \cellcolor{bestred}\textbf{0.934} & \cellcolor{bestred}\textbf{0.055} & \cellcolor{secondorange}\textbf{0.494} & \textbf{0.024} & \cellcolor{thirdyellow}\textbf{0.916} & \cellcolor{secondorange}\textbf{0.055} & \textbf{0.509} \\
\midrule
\multicolumn{17}{c}{\cellcolor{catgray}\textbf{Chunk-wise}} \\
\midrule
 VGGT-Long & 1257 & 0.024 & 0.016 & 0.926 & 0.013 & 0.955 & 0.038 & 0.658 & 0.024 & 0.851 & 0.174 & \cellcolor{thirdyellow}0.481 & \cellcolor{secondorange}0.018 & 0.911 & 0.106 & \cellcolor{thirdyellow}0.570 \\
 $\pi^{3}$-Long & 958.7 & \textbf{0.021} & 0.017 & 0.930 & 0.027 & 0.952 & 0.032 & 0.568 & 0.143 & 0.850 & \cellcolor{thirdyellow}\textbf{0.147} & 0.160 & 0.062 & 0.911 & \textbf{0.090} & 0.364 \\
 DA3-Streaming & 1356 & 0.026 & \cellcolor{bestred}\textbf{0.009} & \cellcolor{bestred}\textbf{0.944} & \cellcolor{secondorange}\textbf{0.009} & \cellcolor{secondorange}\textbf{0.967} & \cellcolor{secondorange}\textbf{0.027} & \cellcolor{secondorange}\textbf{0.790} & \cellcolor{thirdyellow}\textbf{0.023} & \cellcolor{thirdyellow}\textbf{0.871} & 0.170 & \cellcolor{bestred}\textbf{0.551} & \cellcolor{bestred}\textbf{0.014} & \cellcolor{bestred}\textbf{0.927} & 0.098 & \cellcolor{bestred}\textbf{0.671} \\
\midrule
\multicolumn{17}{c}{\cellcolor{catgray}\textbf{SLAM-based}} \\
\midrule
 MASt3R-SLAM & 688.6 & 0.096 & 0.103 & 0.258 & 0.111 & 0.674 & 0.482 & 0.134 & 0.116 & 0.791 & 0.230 & 0.130 & 0.110 & 0.574 & 0.356 & 0.132 \\
 VGGT-SLAM & 1257 & \textbf{0.024} & \textbf{0.016} & \textbf{0.926} & \textbf{0.015} & \textbf{0.941} & \textbf{0.047} & \textbf{0.647} & \textbf{0.027} & \textbf{0.837} & \textbf{0.163} & \textbf{0.410} & \cellcolor{thirdyellow}\textbf{0.019} & \textbf{0.901} & \textbf{0.105} & \textbf{0.528} \\
\midrule
\multicolumn{17}{c}{\cellcolor{catgray}\textbf{Test-Time Training}} \\
\midrule
 TTT3R & 793.3 & 0.044 & 0.048 & 0.800 & 0.047 & 0.846 & 0.114 & 0.333 & 0.070 & 0.302 & 0.686 & 0.095 & 0.055 & 0.649 & 0.400 & 0.214 \\
 Scal3R & 1266 & 0.024 & 0.016 & 0.928 & 0.147 & 0.798 & \textbf{0.037} & \textbf{0.709} & 0.172 & 0.622 & \textbf{0.174} & \textbf{0.472} & 0.112 & 0.783 & \textbf{0.105} & \cellcolor{secondorange}\textbf{0.590} \\
 LoGeR & 1255 & 0.020 & 0.021 & 0.912 & 0.022 & 0.922 & 0.067 & 0.502 & 0.055 & 0.731 & 0.215 & 0.268 & 0.033 & 0.855 & 0.141 & 0.385 \\
 LoGeR$^*$ & 1255 & \cellcolor{secondorange}\textbf{0.015} & \cellcolor{thirdyellow}\textbf{0.013} & \textbf{0.935} & \textbf{0.017} & \textbf{0.938} & 0.056 & 0.593 & \textbf{0.052} & \textbf{0.828} & 0.187 & 0.289 & \textbf{0.028} & \textbf{0.900} & 0.121 & 0.441 \\
\bottomrule
\end{tabular}%
}
\end{table*}

\begin{table*}[t]
\centering
\setlength{\fboxsep}{0pt}
\caption{\textbf{Per-Dataset Results on OmniWorld}.
Performance across different input regimes: \textit{Single Frame}, \textit{Sparse}, \textit{Medium}, \textit{Dense}, and the \textit{Average}.
The best, second-best, and third-best results in each column are highlighted in
\colorbox{bestred}{deep blue}, \colorbox{secondorange}{medium blue}, and \colorbox{thirdyellow}{light blue}, respectively.
Out-of-memory (OOM) and Timeout (T.O) cells are shaded \colorbox{oomred}{light red};
\textit{Average} values for those rows are wrapped in parentheses and excluded from per-column ranking.
Within each sub-category, the \textbf{bold} value marks the in-group best.
Note that \ours (Ours) is excluded from the per-column rankings.
}
\label{tab:dataset_omniworld}
\resizebox{\textwidth}{!}{%
\renewcommand{\arraystretch}{1.15}
\setlength{\tabcolsep}{3pt}
\begin{tabular}{l c >{\columncolor{subcol}}c >{\columncolor{subcol}}c >{\columncolor{subcol}}c >{\columncolor{subcol}}c >{\columncolor{subcol}}c >{\columncolor{subcol}}c >{\columncolor{subcol}}c >{\columncolor{subcol}}c >{\columncolor{subcol}}c >{\columncolor{subcol}}c >{\columncolor{subcol}}c >{\columncolor{subcol}}c}
\toprule
\multirow{2.5}{*}{\textbf{Method}}
& \multirow{2.5}{*}{\makecell{\textbf{\#Params}\\\textbf{(M)}}}
& \multicolumn{1}{c}{\textbf{Single Frame}}
& \multicolumn{2}{c}{\textbf{Sparse}}
& \multicolumn{3}{c}{\textbf{Medium}}
& \multicolumn{3}{c}{\textbf{Dense}}
& \multicolumn{3}{c}{\textbf{Average}} \\
\cmidrule(lr){3-3} \cmidrule(lr){4-5} \cmidrule(lr){6-8} \cmidrule(lr){9-11} \cmidrule(lr){12-14}
 & &
 AbsRel$\downarrow$ &
 AbsRel$\downarrow$ & AUC@30$\uparrow$ &
 AbsRel$\downarrow$ & AUC@30$\uparrow$ & ATE$\downarrow$ &
 AbsRel$\downarrow$ & AUC@30$\uparrow$ & ATE$\downarrow$ &
 AbsRel$\downarrow$ & AUC@30$\uparrow$ & ATE$\downarrow$ \\
\midrule
\multicolumn{14}{c}{\cellcolor{catgray}\textbf{Optimization-based}} \\
\midrule
 DUSt3R & 571.2 & \textbf{0.211} & 0.305 & \textbf{0.601} & 0.371 & 0.550 & \textbf{2.484} & \cellcolor{oomred}OOM & \cellcolor{oomred}OOM & \cellcolor{oomred}OOM & (0.338) & (0.575) & (2.484) \\
 MASt3R & 688.6 & 0.216 & \textbf{0.247} & 0.598 & \textbf{0.248} & \textbf{0.595} & 2.923 & \cellcolor{oomred}OOM & \cellcolor{oomred}OOM & \cellcolor{oomred}OOM & (0.247) & (0.596) & (2.923) \\
\midrule
\multicolumn{14}{c}{\cellcolor{catgray}\textbf{End-to-End Feed-Forward}} \\
\midrule
 VGGT & 1257 & 0.147 & 0.170 & 0.774 & 0.176 & 0.752 & 3.763 & \cellcolor{oomred}OOM & \cellcolor{oomred}OOM & \cellcolor{oomred}OOM & (0.173) & (0.763) & (3.763) \\
 Fast3R & 647.5 & 0.246 & 0.352 & 0.449 & 0.348 & 0.337 & 7.692 & 0.438 & 0.165 & 12.5 & 0.380 & 0.317 & 10.1 \\
 FastVGGT & 1158 & 0.148 & 0.176 & 0.672 & 0.166 & 0.739 & 3.956 & 0.161 & 0.735 & 4.926 & 0.168 & 0.715 & 4.441 \\
 MUSt3R & 423.4 & 0.215 & 0.205 & 0.762 & 0.291 & 0.724 & 1.362 & \cellcolor{oomred}T.O & \cellcolor{oomred}T.O & \cellcolor{oomred}T.O & (0.248) & (0.743) & (1.362) \\
 MapAnything & 1228 & 0.144 & 0.122 & 0.824 & 0.125 & 0.802 & 2.256 & \cellcolor{oomred}OOM & \cellcolor{oomred}OOM & \cellcolor{oomred}OOM & (0.123) & (0.813) & (2.256) \\
 OmniVGGT & 1217 & 0.143 & 0.125 & 0.713 & 0.140 & 0.691 & 2.765 & \cellcolor{oomred}OOM & \cellcolor{oomred}OOM & \cellcolor{oomred}OOM & (0.132) & (0.702) & (2.765) \\
 $\pi^{3}$ & 958.7 & 0.123 & \cellcolor{thirdyellow}0.042 & 0.956 & \cellcolor{thirdyellow}0.037 & 0.962 & 0.270 & \cellcolor{bestred}\textbf{0.037} & \cellcolor{bestred}\textbf{0.959} & \cellcolor{bestred}\textbf{0.286} & \cellcolor{bestred}\textbf{0.039} & \cellcolor{secondorange}\textbf{0.959} & \cellcolor{secondorange}\textbf{0.278} \\
 $\pi^{3}$-X & 1360 & 0.125 & \cellcolor{secondorange}\textbf{0.041} & \cellcolor{thirdyellow}0.972 & \cellcolor{secondorange}\textbf{0.035} & 0.965 & \cellcolor{secondorange}0.217 & \cellcolor{oomred}OOM & \cellcolor{oomred}OOM & \cellcolor{oomred}OOM & (0.038) & (0.969) & (0.217) \\
 AMB3R & 1563 & 0.147 & 0.124 & 0.876 & 0.124 & 0.880 & 0.837 & \cellcolor{oomred}OOM & \cellcolor{oomred}OOM & \cellcolor{oomred}OOM & (0.124) & (0.878) & (0.837) \\
 DA3-Small & 34.3 & 0.223 & 0.211 & 0.531 & 0.227 & 0.476 & 5.359 & 0.261 & 0.462 & 7.659 & 0.233 & 0.490 & 6.509 \\
 DA3-Base & 135.4 & 0.195 & 0.155 & 0.613 & 0.170 & 0.640 & 4.013 & 0.183 & 0.583 & 4.623 & 0.169 & 0.612 & 4.318 \\
 DA3-Large & 410.9 & \cellcolor{thirdyellow}0.110 & 0.074 & 0.956 & 0.069 & 0.959 & 0.436 & \cellcolor{oomred}OOM & \cellcolor{oomred}OOM & \cellcolor{oomred}OOM & (0.071) & (0.958) & (0.436) \\
 DA3-Giant & 1356 & 0.114 & 0.050 & \cellcolor{secondorange}0.978 & 0.046 & \cellcolor{secondorange}0.982 & \cellcolor{thirdyellow}0.235 & \cellcolor{oomred}OOM & \cellcolor{oomred}OOM & \cellcolor{oomred}OOM & (0.048) & (0.980) & (0.235) \\
 DA3-Nested & 1690 & \cellcolor{secondorange}\textbf{0.106} & 0.050 & \cellcolor{bestred}\textbf{0.981} & 0.044 & \cellcolor{bestred}\textbf{0.985} & \cellcolor{bestred}\textbf{0.186} & \cellcolor{oomred}OOM & \cellcolor{oomred}OOM & \cellcolor{oomred}OOM & (0.047) & (0.983) & (0.186) \\
 WorldMirror & 1263 & 0.180 & 0.173 & 0.733 & 0.167 & 0.787 & 1.836 & \cellcolor{oomred}OOM & \cellcolor{oomred}OOM & \cellcolor{oomred}OOM & (0.170) & (0.760) & (1.836) \\
 VGGT-Omega & 1144 & \cellcolor{bestred}\textbf{0.061} & 0.050 & 0.969 & 0.047 & 0.960 & 0.299 & -- & -- & -- & (0.048) & (0.965) & (0.299) \\ \hdashline
 \ours (Ours) & 1304 & 0.135 & 0.057 & 0.973 & 0.049 & 0.974 & 0.278 & \cellcolor{oomred}OOM & \cellcolor{oomred}OOM & \cellcolor{oomred}OOM & (0.080) & (0.973) & (0.278) \\
\midrule
\multicolumn{14}{c}{\cellcolor{catgray}\textbf{Online}} \\
\midrule
 Spann3r$^{224}$ & 658.7 & 0.269 & 0.385 & 0.434 & 0.394 & 0.367 & 5.645 & 0.417 & 0.259 & 8.475 & 0.399 & 0.353 & 7.060 \\
 CUT3R & 793.3 & 0.172 & 0.216 & 0.726 & 0.312 & 0.608 & 2.828 & 0.389 & 0.373 & 9.160 & 0.306 & 0.569 & 5.994 \\
 MonST3R & 571.2 & 0.200 & 0.223 & 0.394 & 0.242 & 0.495 & 3.075 & \cellcolor{oomred}OOM & \cellcolor{oomred}OOM & \cellcolor{oomred}OOM & (0.233) & (0.445) & (3.075) \\
 Point3R & 828 & 0.200 & 0.246 & 0.260 & 0.262 & 0.261 & 7.981 & 0.339 & 0.092 & 9.878 & 0.282 & 0.205 & 8.929 \\
 Stream3R-S & 1191 & 0.160 & 0.146 & 0.637 & 0.184 & 0.643 & 6.196 & \cellcolor{oomred}OOM & \cellcolor{oomred}OOM & \cellcolor{oomred}OOM & (0.165) & (0.640) & (6.196) \\
 Stream3R-W & 1191 & 0.160 & 0.152 & 0.592 & 0.215 & 0.450 & 11.25 & \cellcolor{oomred}OOM & \cellcolor{oomred}OOM & \cellcolor{oomred}OOM & (0.183) & (0.521) & (11.25) \\
 StreamVGGT & 1257 & 0.158 & 0.298 & 0.622 & 0.382 & 0.578 & 6.522 & 0.376 & 0.383 & 9.389 & 0.352 & 0.528 & 7.956 \\
 Page4D & 1257 & 0.148 & 0.135 & 0.720 & 0.147 & 0.740 & 4.709 & \cellcolor{oomred}OOM & \cellcolor{oomred}OOM & \cellcolor{oomred}OOM & (0.141) & (0.730) & (4.709) \\
 InfiniteVGGT & 1257 & 0.160 & 0.299 & 0.623 & 0.381 & 0.586 & 6.464 & 0.379 & 0.387 & 9.373 & 0.353 & 0.532 & 7.919 \\
 Wint3R & 749.5 & 0.137 & \textbf{0.066} & 0.747 & \textbf{0.072} & 0.685 & 1.833 & 0.144 & 0.394 & 6.080 & 0.094 & 0.609 & 3.956 \\
 LongStream-B & 1191 & 0.153 & 0.185 & 0.420 & 0.247 & 0.780 & 1.356 & 0.224 & 0.650 & 2.156 & 0.219 & 0.617 & 1.756 \\
 LongStream-S & 1191 & 0.153 & 0.165 & 0.417 & 0.199 & 0.617 & 2.763 & 0.232 & 0.502 & 2.440 & 0.199 & 0.512 & 2.602 \\
 LingbotMap$^{*}$-W & 1158 & \textbf{0.131} & 0.080 & \textbf{0.948} & 0.094 & \textbf{0.936} & 0.402 & 0.117 & 0.906 & 0.881 & 0.097 & \textbf{0.930} & 0.642 \\
 LingbotMap$^{*}$-S & 1158 & \textbf{0.131} & 0.080 & \textbf{0.948} & 0.090 & 0.923 & \textbf{0.357} & \textbf{0.095} & \textbf{0.918} & \cellcolor{thirdyellow}\textbf{0.588} & \textbf{0.088} & 0.930 & \textbf{0.473} \\
\midrule
\multicolumn{14}{c}{\cellcolor{catgray}\textbf{Chunk-wise}} \\
\midrule
 VGGT-Long & 1257 & 0.147 & 0.170 & 0.774 & 0.170 & 0.814 & 2.701 & 0.349 & 0.613 & 5.409 & 0.230 & 0.734 & 4.055 \\
 $\pi^{3}$-Long & 958.7 & 0.123 & \cellcolor{thirdyellow}\textbf{0.042} & 0.956 & 0.082 & 0.957 & 0.288 & 0.337 & 0.909 & 0.739 & 0.154 & 0.940 & 0.513 \\
 DA3-Streaming & 1356 & \textbf{0.114} & 0.050 & \cellcolor{secondorange}\textbf{0.978} & \textbf{0.046} & \cellcolor{thirdyellow}\textbf{0.981} & \textbf{0.243} & \cellcolor{secondorange}\textbf{0.051} & \cellcolor{secondorange}\textbf{0.956} & \cellcolor{secondorange}\textbf{0.289} & \cellcolor{secondorange}\textbf{0.049} & \cellcolor{bestred}\textbf{0.972} & \cellcolor{bestred}\textbf{0.266} \\
\midrule
\multicolumn{14}{c}{\cellcolor{catgray}\textbf{SLAM-based}} \\
\midrule
 MASt3R-SLAM & 688.6 & 0.281 & 0.331 & 0.342 & 0.280 & 0.610 & \textbf{3.683} & \textbf{0.344} & \textbf{0.616} & \textbf{3.616} & 0.318 & 0.523 & \textbf{3.649} \\
 VGGT-SLAM & 1257 & \textbf{0.147} & \textbf{0.170} & \textbf{0.774} & \textbf{0.186} & \textbf{0.765} & 4.607 & 0.371 & 0.433 & 6.222 & \textbf{0.243} & \textbf{0.657} & 5.415 \\
\midrule
\multicolumn{14}{c}{\cellcolor{catgray}\textbf{Test-Time Training}} \\
\midrule
 TTT3R & 793.3 & 0.172 & 0.236 & 0.506 & 0.249 & 0.645 & 3.182 & 0.284 & 0.631 & 4.303 & 0.257 & 0.594 & 3.742 \\
 Scal3R & 1266 & 0.142 & 0.116 & 0.848 & 0.137 & 0.830 & 0.812 & 0.405 & 0.682 & 1.359 & 0.219 & 0.787 & 1.085 \\
 LoGeR & 1255 & 0.131 & \cellcolor{bestred}\textbf{0.039} & 0.936 & 0.048 & 0.960 & 0.311 & 0.087 & \cellcolor{thirdyellow}0.939 & 0.605 & 0.058 & 0.945 & 0.458 \\
 LoGeR$^*$ & 1255 & \textbf{0.130} & \cellcolor{secondorange}0.041 & \textbf{0.942} & \cellcolor{bestred}\textbf{0.033} & \textbf{0.971} & \textbf{0.238} & \cellcolor{thirdyellow}\textbf{0.074} & \cellcolor{thirdyellow}\textbf{0.939} & \textbf{0.602} & \cellcolor{thirdyellow}\textbf{0.050} & \cellcolor{thirdyellow}\textbf{0.951} & \cellcolor{thirdyellow}\textbf{0.420} \\
\bottomrule
\end{tabular}%
}
\end{table*}

\begin{table*}[t]
\centering
\setlength{\fboxsep}{0pt}
\caption{\textbf{Per-Dataset Results on RLBench}.
Performance across different input regimes: \textit{Single Frame}, \textit{Sparse}, \textit{Medium}, \textit{Dense}, and the \textit{Average}.
The best, second-best, and third-best results in each column are highlighted in
\colorbox{bestred}{deep blue}, \colorbox{secondorange}{medium blue}, and \colorbox{thirdyellow}{light blue}, respectively.
Out-of-memory (OOM) and Timeout (T.O) cells are shaded \colorbox{oomred}{light red};
\textit{Average} values for those rows are wrapped in parentheses and excluded from per-column ranking.
Within each sub-category, the \textbf{bold} value marks the in-group best.
Note that \ours (Ours) is excluded from the per-column rankings.
}
\label{tab:dataset_rlbench}
\resizebox{\textwidth}{!}{%
\renewcommand{\arraystretch}{1.15}
\setlength{\tabcolsep}{3pt}
\begin{tabular}{l c >{\columncolor{subcol}}c >{\columncolor{subcol}}c >{\columncolor{subcol}}c >{\columncolor{subcol}}c >{\columncolor{subcol}}c >{\columncolor{subcol}}c >{\columncolor{subcol}}c >{\columncolor{subcol}}c >{\columncolor{subcol}}c >{\columncolor{subcol}}c >{\columncolor{subcol}}c >{\columncolor{subcol}}c}
\toprule
\multirow{2.5}{*}{\textbf{Method}}
& \multirow{2.5}{*}{\makecell{\textbf{\#Params}\\\textbf{(M)}}}
& \multicolumn{1}{c}{\textbf{Single Frame}}
& \multicolumn{2}{c}{\textbf{Sparse}}
& \multicolumn{3}{c}{\textbf{Medium}}
& \multicolumn{3}{c}{\textbf{Dense}}
& \multicolumn{3}{c}{\textbf{Average}} \\
\cmidrule(lr){3-3} \cmidrule(lr){4-5} \cmidrule(lr){6-8} \cmidrule(lr){9-11} \cmidrule(lr){12-14}
 & &
 AbsRel$\downarrow$ &
 AbsRel$\downarrow$ & AUC@30$\uparrow$ &
 AbsRel$\downarrow$ & AUC@30$\uparrow$ & ATE$\downarrow$ &
 AbsRel$\downarrow$ & AUC@30$\uparrow$ & ATE$\downarrow$ &
 AbsRel$\downarrow$ & AUC@30$\uparrow$ & ATE$\downarrow$ \\
\midrule
\multicolumn{14}{c}{\cellcolor{catgray}\textbf{Optimization-based}} \\
\midrule
 DUSt3R & 571.2 & 0.437 & 0.830 & 0.181 & \textbf{0.746} & \textbf{0.148} & 0.154 & \cellcolor{oomred}OOM & \cellcolor{oomred}OOM & \cellcolor{oomred}OOM & (0.788) & (0.165) & (0.154) \\
 MASt3R & 688.6 & \textbf{0.308} & \textbf{0.419} & \textbf{0.216} & 0.790 & 0.133 & \textbf{0.150} & \cellcolor{oomred}OOM & \cellcolor{oomred}OOM & \cellcolor{oomred}OOM & (0.605) & (0.174) & (0.150) \\
\midrule
\multicolumn{14}{c}{\cellcolor{catgray}\textbf{End-to-End Feed-Forward}} \\
\midrule
 VGGT & 1257 & \cellcolor{bestred}\textbf{0.190} & 0.296 & 0.302 & 0.608 & 0.177 & 0.133 & \cellcolor{oomred}OOM & \cellcolor{oomred}OOM & \cellcolor{oomred}OOM & (0.452) & (0.239) & (0.133) \\
 Fast3R & 647.5 & 0.367 & 0.432 & 0.147 & 0.630 & 0.082 & 0.150 & 0.722 & 0.055 & 0.155 & 0.595 & 0.095 & 0.153 \\
 FastVGGT & 1158 & \cellcolor{secondorange}0.192 & 0.304 & 0.172 & 0.351 & 0.139 & 0.152 & 0.368 & 0.158 & 0.126 & 0.341 & 0.156 & 0.139 \\
 MUSt3R & 423.4 & 0.394 & 0.382 & 0.267 & 0.480 & 0.308 & 0.056 & \cellcolor{oomred}T.O & \cellcolor{oomred}T.O & \cellcolor{oomred}T.O & (0.431) & (0.287) & (0.056) \\
 MapAnything & 1228 & 0.281 & 0.328 & 0.284 & 0.443 & 0.176 & 0.101 & \cellcolor{oomred}OOM & \cellcolor{oomred}OOM & \cellcolor{oomred}OOM & (0.386) & (0.230) & (0.101) \\
 OmniVGGT & 1217 & 0.300 & 0.283 & 0.521 & 0.306 & 0.323 & 0.104 & \cellcolor{oomred}OOM & \cellcolor{oomred}OOM & \cellcolor{oomred}OOM & (0.295) & (0.422) & (0.104) \\
 $\pi^{3}$ & 958.7 & 0.201 & \textbf{0.205} & 0.559 & \cellcolor{thirdyellow}\textbf{0.204} & 0.433 & 0.061 & \cellcolor{bestred}\textbf{0.210} & \cellcolor{secondorange}\textbf{0.439} & \cellcolor{bestred}\textbf{0.062} & \cellcolor{bestred}\textbf{0.207} & \cellcolor{secondorange}\textbf{0.477} & \cellcolor{secondorange}\textbf{0.062} \\
 $\pi^{3}$-X & 1360 & 0.205 & 0.260 & 0.440 & 0.212 & 0.385 & 0.072 & \cellcolor{oomred}OOM & \cellcolor{oomred}OOM & \cellcolor{oomred}OOM & (0.236) & (0.413) & (0.072) \\
 AMB3R & 1563 & 0.204 & 0.229 & 0.525 & 0.256 & 0.389 & 0.086 & \cellcolor{oomred}OOM & \cellcolor{oomred}OOM & \cellcolor{oomred}OOM & (0.243) & (0.457) & (0.086) \\
 DA3-Small & 34.3 & 0.350 & 0.435 & 0.283 & 0.482 & 0.183 & 0.088 & 0.536 & 0.140 & 0.088 & 0.484 & 0.202 & 0.088 \\
 DA3-Base & 135.4 & 0.410 & 0.406 & 0.285 & 0.400 & 0.217 & 0.077 & 0.438 & 0.157 & 0.078 & 0.415 & 0.220 & 0.078 \\
 DA3-Large & 410.9 & 0.364 & 0.336 & 0.431 & 0.353 & 0.315 & 0.078 & \cellcolor{oomred}OOM & \cellcolor{oomred}OOM & \cellcolor{oomred}OOM & (0.344) & (0.373) & (0.078) \\
 DA3-Giant & 1356 & 0.355 & 0.270 & \cellcolor{thirdyellow}\textbf{0.565} & 0.250 & 0.471 & 0.058 & \cellcolor{oomred}OOM & \cellcolor{oomred}OOM & \cellcolor{oomred}OOM & (0.260) & (0.518) & (0.058) \\
 DA3-Nested & 1690 & 0.385 & 0.311 & 0.560 & 0.241 & \cellcolor{thirdyellow}\textbf{0.521} & \cellcolor{thirdyellow}\textbf{0.051} & \cellcolor{oomred}OOM & \cellcolor{oomred}OOM & \cellcolor{oomred}OOM & (0.276) & (0.541) & (0.051) \\
 WorldMirror & 1263 & 0.322 & 0.329 & 0.318 & 0.340 & 0.245 & 0.116 & \cellcolor{oomred}OOM & \cellcolor{oomred}OOM & \cellcolor{oomred}OOM & (0.335) & (0.282) & (0.116) \\
 VGGT-Omega & 1144 & \cellcolor{thirdyellow}0.193 & \cellcolor{secondorange}\textbf{0.194} & \cellcolor{bestred}\textbf{0.596} & \cellcolor{bestred}\textbf{0.182} & \cellcolor{bestred}\textbf{0.589} & \cellcolor{secondorange}\textbf{0.043} & -- & -- & -- & (0.188) & (0.592) & (0.043) \\ \hdashline
 \ours (Ours) & 1304 & 0.088 & 0.046 & 0.893 & 0.037 & 0.807 & 0.019 & \cellcolor{oomred}OOM & \cellcolor{oomred}OOM & \cellcolor{oomred}OOM & (0.057) & (0.850) & (0.019) \\
\midrule
\multicolumn{14}{c}{\cellcolor{catgray}\textbf{Online}} \\
\midrule
 Spann3r$^{224}$ & 658.7 & 0.444 & 0.434 & 0.131 & 0.618 & 0.129 & 0.108 & 0.813 & 0.045 & 0.167 & 0.622 & 0.102 & 0.138 \\
 CUT3R & 793.3 & 0.329 & 0.543 & 0.187 & 0.440 & 0.171 & 0.102 & 0.557 & 0.051 & 0.148 & 0.513 & 0.137 & 0.125 \\
 MonST3R & 571.2 & 0.393 & 0.553 & 0.080 & 0.476 & 0.147 & 0.125 & \cellcolor{oomred}OOM & \cellcolor{oomred}OOM & \cellcolor{oomred}OOM & (0.515) & (0.113) & (0.125) \\
 Point3R & 828 & 0.348 & 0.503 & 0.164 & 0.678 & 0.044 & 0.183 & 0.795 & 0.036 & 0.181 & 0.659 & 0.081 & 0.182 \\
 Stream3R-S & 1191 & 0.262 & 0.326 & 0.134 & 0.352 & 0.099 & 0.153 & \cellcolor{oomred}OOM & \cellcolor{oomred}OOM & \cellcolor{oomred}OOM & (0.339) & (0.116) & (0.153) \\
 Stream3R-W & 1191 & 0.262 & 0.326 & 0.134 & 0.484 & 0.064 & 0.198 & \cellcolor{oomred}OOM & \cellcolor{oomred}OOM & \cellcolor{oomred}OOM & (0.405) & (0.099) & (0.198) \\
 StreamVGGT & 1257 & 0.239 & 0.316 & 0.151 & 0.358 & 0.132 & 0.162 & 0.379 & 0.119 & 0.163 & 0.351 & 0.134 & 0.162 \\
 Page4D & 1257 & \textbf{0.219} & 0.338 & 0.330 & 0.327 & 0.211 & 0.113 & \cellcolor{oomred}OOM & \cellcolor{oomred}OOM & \cellcolor{oomred}OOM & (0.333) & (0.271) & (0.113) \\
 InfiniteVGGT & 1257 & 0.239 & \textbf{0.315} & 0.155 & 0.358 & 0.132 & 0.162 & 0.378 & 0.118 & 0.163 & 0.350 & 0.135 & 0.162 \\
 Wint3R & 749.5 & 0.364 & 0.383 & 0.201 & 0.398 & 0.142 & 0.107 & 0.584 & 0.076 & 0.154 & 0.455 & 0.140 & 0.131 \\
 LongStream-B & 1191 & 0.385 & 0.332 & 0.121 & 0.438 & 0.097 & 0.112 & 0.668 & 0.080 & 0.102 & 0.479 & 0.099 & 0.107 \\
 LongStream-S & 1191 & 0.385 & 0.331 & 0.121 & 0.363 & 0.081 & 0.110 & 0.750 & 0.063 & 0.113 & 0.482 & 0.088 & 0.112 \\
 LingbotMap$^{*}$-W & 1158 & 0.300 & 0.339 & \textbf{0.387} & \textbf{0.282} & \textbf{0.257} & 0.091 & 0.364 & \textbf{0.269} & \textbf{0.095} & 0.328 & \textbf{0.304} & \textbf{0.093} \\
 LingbotMap$^{*}$-S & 1158 & 0.300 & 0.339 & \textbf{0.387} & \textbf{0.282} & 0.257 & \textbf{0.091} & \cellcolor{secondorange}\textbf{0.324} & 0.183 & 0.114 & \textbf{0.315} & 0.276 & 0.103 \\
\midrule
\multicolumn{14}{c}{\cellcolor{catgray}\textbf{Chunk-wise}} \\
\midrule
 VGGT-Long & 1257 & \cellcolor{bestred}\textbf{0.190} & 0.296 & 0.302 & 0.609 & 0.178 & 0.136 & 0.599 & 0.177 & 0.113 & 0.501 & 0.219 & 0.124 \\
 $\pi^{3}$-Long & 958.7 & 0.201 & \textbf{0.205} & 0.559 & \textbf{0.206} & 0.433 & \textbf{0.061} & \textbf{0.350} & \textbf{0.375} & \cellcolor{thirdyellow}\textbf{0.069} & \cellcolor{thirdyellow}\textbf{0.254} & \cellcolor{thirdyellow}\textbf{0.456} & \cellcolor{thirdyellow}\textbf{0.065} \\
 DA3-Streaming & 1356 & 0.355 & 0.270 & \cellcolor{secondorange}\textbf{0.566} & 0.256 & \textbf{0.472} & 0.062 & 0.549 & 0.311 & 0.102 & 0.359 & 0.450 & 0.082 \\
\midrule
\multicolumn{14}{c}{\cellcolor{catgray}\textbf{SLAM-based}} \\
\midrule
 MASt3R-SLAM & 688.6 & 0.410 & 0.594 & 0.106 & 0.773 & 0.148 & 0.139 & 0.859 & \textbf{0.201} & \textbf{0.107} & 0.742 & 0.152 & 0.123 \\
 VGGT-SLAM & 1257 & \cellcolor{bestred}\textbf{0.190} & \textbf{0.296} & \textbf{0.302} & \textbf{0.432} & \textbf{0.182} & \textbf{0.135} & \textbf{0.350} & 0.120 & 0.108 & \textbf{0.359} & \textbf{0.201} & \textbf{0.121} \\
\midrule
\multicolumn{14}{c}{\cellcolor{catgray}\textbf{Test-Time Training}} \\
\midrule
 TTT3R & 793.3 & 0.329 & 0.559 & 0.156 & 0.493 & 0.158 & 0.106 & 0.506 & 0.112 & 0.113 & 0.519 & 0.142 & 0.110 \\
 Scal3R & 1266 & 0.238 & 0.233 & \textbf{0.524} & 0.282 & 0.347 & 0.089 & 0.386 & 0.248 & 0.073 & 0.300 & 0.373 & 0.081 \\
 LoGeR & 1255 & 0.235 & \cellcolor{thirdyellow}0.202 & 0.428 & 0.321 & 0.452 & 0.052 & 0.433 & \cellcolor{thirdyellow}0.435 & \cellcolor{secondorange}\textbf{0.066} & 0.319 & 0.438 & \cellcolor{bestred}0.059 \\
 LoGeR$^*$ & 1255 & \textbf{0.210} & \cellcolor{bestred}\textbf{0.171} & 0.523 & \cellcolor{secondorange}\textbf{0.202} & \cellcolor{secondorange}\textbf{0.525} & \cellcolor{bestred}\textbf{0.040} & \cellcolor{thirdyellow}\textbf{0.349} & \cellcolor{bestred}\textbf{0.465} & 0.077 & \cellcolor{secondorange}\textbf{0.241} & \cellcolor{bestred}\textbf{0.505} & \cellcolor{bestred}\textbf{0.059} \\
\bottomrule
\end{tabular}%
}
\end{table*}

\begin{table*}[t]
\centering
\setlength{\fboxsep}{0pt}
\caption{\textbf{Per-Dataset Results on Robolab}.
Performance across different input regimes: \textit{Single Frame}, \textit{Sparse}, \textit{Medium}, \textit{Dense}, and the \textit{Average}.
The best, second-best, and third-best results in each column are highlighted in
\colorbox{bestred}{deep blue}, \colorbox{secondorange}{medium blue}, and \colorbox{thirdyellow}{light blue}, respectively.
Out-of-memory (OOM) and Timeout (T.O) cells are shaded \colorbox{oomred}{light red};
\textit{Average} values for those rows are wrapped in parentheses and excluded from per-column ranking.
Within each sub-category, the \textbf{bold} value marks the in-group best.
Note that \ours (Ours) is excluded from the per-column rankings.
}
\label{tab:dataset_robolab}
\resizebox{\textwidth}{!}{%
\renewcommand{\arraystretch}{1.15}
\setlength{\tabcolsep}{3pt}
\begin{tabular}{l c >{\columncolor{subcol}}c >{\columncolor{subcol}}c >{\columncolor{subcol}}c >{\columncolor{subcol}}c >{\columncolor{subcol}}c >{\columncolor{subcol}}c >{\columncolor{subcol}}c >{\columncolor{subcol}}c >{\columncolor{subcol}}c >{\columncolor{subcol}}c >{\columncolor{subcol}}c >{\columncolor{subcol}}c}
\toprule
\multirow{2.5}{*}{\textbf{Method}}
& \multirow{2.5}{*}{\makecell{\textbf{\#Params}\\\textbf{(M)}}}
& \multicolumn{1}{c}{\textbf{Single Frame}}
& \multicolumn{2}{c}{\textbf{Sparse}}
& \multicolumn{3}{c}{\textbf{Medium}}
& \multicolumn{3}{c}{\textbf{Dense}}
& \multicolumn{3}{c}{\textbf{Average}} \\
\cmidrule(lr){3-3} \cmidrule(lr){4-5} \cmidrule(lr){6-8} \cmidrule(lr){9-11} \cmidrule(lr){12-14}
 & &
 AbsRel$\downarrow$ &
 AbsRel$\downarrow$ & AUC@30$\uparrow$ &
 AbsRel$\downarrow$ & AUC@30$\uparrow$ & ATE$\downarrow$ &
 AbsRel$\downarrow$ & AUC@30$\uparrow$ & ATE$\downarrow$ &
 AbsRel$\downarrow$ & AUC@30$\uparrow$ & ATE$\downarrow$ \\
\midrule
\multicolumn{14}{c}{\cellcolor{catgray}\textbf{Optimization-based}} \\
\midrule
 DUSt3R & 571.2 & \textbf{0.371} & \textbf{0.495} & 0.274 & 0.512 & 0.132 & 0.117 & \cellcolor{oomred}OOM & \cellcolor{oomred}OOM & \cellcolor{oomred}OOM & (0.504) & (0.203) & (0.117) \\
 MASt3R & 688.6 & 0.386 & 0.515 & \textbf{0.429} & \textbf{0.463} & \textbf{0.242} & \textbf{0.103} & \cellcolor{oomred}OOM & \cellcolor{oomred}OOM & \cellcolor{oomred}OOM & (0.489) & (0.336) & (0.103) \\
\midrule
\multicolumn{14}{c}{\cellcolor{catgray}\textbf{End-to-End Feed-Forward}} \\
\midrule
 VGGT & 1257 & 0.174 & 0.201 & 0.566 & 0.177 & 0.504 & 0.035 & \cellcolor{oomred}OOM & \cellcolor{oomred}OOM & \cellcolor{oomred}OOM & (0.189) & (0.535) & (0.035) \\
 Fast3R & 647.5 & 0.340 & 0.412 & 0.174 & 0.403 & 0.176 & 0.085 & 0.480 & 0.099 & 0.103 & 0.432 & 0.150 & 0.094 \\
 FastVGGT & 1158 & 0.171 & 0.215 & 0.453 & 0.175 & 0.483 & 0.038 & \cellcolor{bestred}\textbf{0.180} & \cellcolor{bestred}\textbf{0.468} & \cellcolor{bestred}\textbf{0.041} & \cellcolor{secondorange}0.190 & 0.468 & \cellcolor{secondorange}0.040 \\
 MUSt3R & 423.4 & 0.491 & 0.457 & 0.376 & 0.262 & 0.505 & 0.036 & \cellcolor{oomred}T.O & \cellcolor{oomred}T.O & \cellcolor{oomred}T.O & (0.359) & (0.440) & (0.036) \\
 MapAnything & 1228 & 0.194 & 0.381 & 0.221 & 0.321 & 0.179 & 0.069 & \cellcolor{oomred}OOM & \cellcolor{oomred}OOM & \cellcolor{oomred}OOM & (0.351) & (0.200) & (0.069) \\
 OmniVGGT & 1217 & 0.157 & 0.185 & 0.601 & 0.162 & 0.562 & 0.030 & \cellcolor{oomred}OOM & \cellcolor{oomred}OOM & \cellcolor{oomred}OOM & (0.173) & (0.582) & (0.030) \\
 $\pi^{3}$ & 958.7 & 0.168 & 0.171 & 0.663 & \cellcolor{thirdyellow}0.150 & \cellcolor{secondorange}\textbf{0.742} & \cellcolor{secondorange}0.018 & \cellcolor{secondorange}0.183 & \cellcolor{secondorange}0.458 & \cellcolor{thirdyellow}0.049 & \cellcolor{bestred}\textbf{0.168} & \cellcolor{bestred}\textbf{0.621} & \cellcolor{bestred}\textbf{0.033} \\
 $\pi^{3}$-X & 1360 & 0.157 & \cellcolor{thirdyellow}\textbf{0.155} & 0.635 & \cellcolor{bestred}\textbf{0.137} & 0.682 & \cellcolor{thirdyellow}0.024 & \cellcolor{oomred}OOM & \cellcolor{oomred}OOM & \cellcolor{oomred}OOM & (0.146) & (0.658) & (0.024) \\
 AMB3R & 1563 & 0.213 & 0.239 & 0.639 & 0.188 & 0.582 & 0.032 & \cellcolor{oomred}OOM & \cellcolor{oomred}OOM & \cellcolor{oomred}OOM & (0.213) & (0.611) & (0.032) \\
 DA3-Small & 34.3 & 0.338 & 0.490 & 0.198 & 0.389 & 0.134 & 0.082 & 0.402 & 0.100 & 0.093 & 0.427 & 0.144 & 0.087 \\
 DA3-Base & 135.4 & 0.278 & 0.326 & 0.352 & 0.273 & 0.225 & 0.067 & 0.276 & 0.177 & 0.073 & 0.292 & 0.251 & 0.070 \\
 DA3-Large & 410.9 & 0.243 & 0.237 & 0.447 & 0.184 & 0.469 & 0.036 & \cellcolor{oomred}OOM & \cellcolor{oomred}OOM & \cellcolor{oomred}OOM & (0.210) & (0.458) & (0.036) \\
 DA3-Giant & 1356 & \cellcolor{bestred}\textbf{0.132} & 0.174 & \cellcolor{thirdyellow}\textbf{0.719} & 0.170 & 0.671 & 0.027 & \cellcolor{oomred}OOM & \cellcolor{oomred}OOM & \cellcolor{oomred}OOM & (0.172) & (0.695) & (0.027) \\
 DA3-Nested & 1690 & \cellcolor{thirdyellow}0.148 & 0.194 & 0.716 & \cellcolor{bestred}0.137 & \cellcolor{thirdyellow}0.721 & \cellcolor{secondorange}\textbf{0.018} & \cellcolor{oomred}OOM & \cellcolor{oomred}OOM & \cellcolor{oomred}OOM & (0.166) & (0.718) & (0.018) \\
 WorldMirror & 1263 & 0.271 & 0.311 & 0.409 & 0.275 & 0.405 & 0.050 & \cellcolor{oomred}OOM & \cellcolor{oomred}OOM & \cellcolor{oomred}OOM & (0.293) & (0.407) & (0.050) \\
 VGGT-Omega & 1144 & 0.207 & \cellcolor{secondorange}\textbf{0.147} & \cellcolor{bestred}\textbf{0.742} & \cellcolor{secondorange}0.149 & \cellcolor{bestred}\textbf{0.811} & \cellcolor{bestred}\textbf{0.012} & -- & -- & -- & (0.148) & (0.777) & (0.012) \\ \hdashline
 \ours (Ours) & 1304 & 0.025 & 0.030 & 0.802 & 0.018 & 0.916 & 0.006 & \cellcolor{oomred}OOM & \cellcolor{oomred}OOM & \cellcolor{oomred}OOM & (0.024) & (0.859) & (0.006) \\
\midrule
\multicolumn{14}{c}{\cellcolor{catgray}\textbf{Online}} \\
\midrule
 Spann3r$^{224}$ & 658.7 & 0.380 & 0.485 & 0.041 & 0.399 & 0.154 & 0.079 & 0.434 & 0.076 & 0.108 & 0.439 & 0.090 & 0.094 \\
 CUT3R & 793.3 & 0.274 & 0.451 & 0.287 & 0.404 & 0.122 & 0.090 & 0.449 & 0.029 & 0.123 & 0.434 & 0.146 & 0.107 \\
 MonST3R & 571.2 & 0.381 & 0.501 & 0.152 & 0.416 & 0.124 & 0.114 & \cellcolor{oomred}OOM & \cellcolor{oomred}OOM & \cellcolor{oomred}OOM & (0.458) & (0.138) & (0.114) \\
 Point3R & 828 & 0.424 & 0.469 & 0.139 & 0.416 & 0.100 & 0.114 & 0.434 & 0.053 & 0.129 & 0.440 & 0.097 & 0.121 \\
 Stream3R-S & 1191 & 0.196 & 0.270 & 0.424 & 0.244 & 0.260 & 0.092 & \cellcolor{oomred}OOM & \cellcolor{oomred}OOM & \cellcolor{oomred}OOM & (0.257) & (0.342) & (0.092) \\
 Stream3R-W & 1191 & 0.196 & 0.271 & 0.421 & 0.348 & 0.117 & 0.137 & \cellcolor{oomred}OOM & \cellcolor{oomred}OOM & \cellcolor{oomred}OOM & (0.310) & (0.269) & (0.137) \\
 StreamVGGT & 1257 & 0.170 & 0.266 & 0.392 & 0.240 & 0.335 & 0.060 & 0.254 & 0.175 & 0.092 & 0.253 & 0.301 & 0.076 \\
 Page4D & 1257 & 0.189 & \textbf{0.259} & \textbf{0.494} & 0.222 & 0.389 & 0.051 & \cellcolor{oomred}OOM & \cellcolor{oomred}OOM & \cellcolor{oomred}OOM & (0.241) & (0.441) & (0.051) \\
 InfiniteVGGT & 1257 & \textbf{0.169} & 0.265 & 0.390 & 0.239 & 0.339 & 0.060 & 0.252 & 0.175 & 0.090 & \textbf{0.252} & 0.301 & 0.075 \\
 Wint3R & 749.5 & 0.319 & 0.333 & 0.292 & 0.268 & 0.169 & 0.077 & 0.364 & 0.020 & 0.119 & 0.322 & 0.161 & 0.098 \\
 LongStream-B & 1191 & 0.216 & 0.270 & 0.412 & 0.360 & 0.165 & 0.087 & 0.358 & 0.062 & 0.100 & 0.329 & 0.213 & 0.093 \\
 LongStream-S & 1191 & 0.216 & 0.270 & 0.411 & 0.326 & 0.129 & 0.095 & 0.358 & 0.038 & 0.103 & 0.318 & 0.193 & 0.099 \\
 LingbotMap$^{*}$-W & 1158 & 0.202 & 0.327 & 0.394 & \textbf{0.201} & 0.360 & 0.044 & 0.323 & 0.200 & 0.077 & 0.284 & 0.318 & 0.060 \\
 LingbotMap$^{*}$-S & 1158 & 0.202 & 0.327 & 0.394 & 0.208 & \textbf{0.412} & \textbf{0.043} & \textbf{0.230} & \textbf{0.359} & \cellcolor{secondorange}\textbf{0.047} & 0.255 & \textbf{0.388} & \cellcolor{thirdyellow}\textbf{0.045} \\
\midrule
\multicolumn{14}{c}{\cellcolor{catgray}\textbf{Chunk-wise}} \\
\midrule
 VGGT-Long & 1257 & 0.174 & 0.201 & 0.566 & 0.192 & 0.506 & 0.049 & 0.345 & 0.278 & 0.083 & 0.246 & 0.450 & 0.066 \\
 $\pi^{3}$-Long & 958.7 & 0.168 & \textbf{0.171} & 0.663 & 0.185 & \textbf{0.639} & 0.042 & 0.301 & \cellcolor{thirdyellow}\textbf{0.378} & 0.075 & 0.219 & \cellcolor{secondorange}\textbf{0.560} & 0.058 \\
 DA3-Streaming & 1356 & \cellcolor{bestred}\textbf{0.132} & 0.174 & \cellcolor{secondorange}\textbf{0.720} & \textbf{0.183} & 0.578 & \textbf{0.037} & \cellcolor{thirdyellow}\textbf{0.223} & 0.325 & \textbf{0.074} & \cellcolor{thirdyellow}\textbf{0.193} & \cellcolor{thirdyellow}0.541 & \textbf{0.056} \\
\midrule
\multicolumn{14}{c}{\cellcolor{catgray}\textbf{SLAM-based}} \\
\midrule
 MASt3R-SLAM & 688.6 & 0.473 & 0.547 & 0.118 & 0.471 & 0.147 & 0.110 & 0.459 & 0.181 & 0.094 & 0.492 & 0.149 & 0.102 \\
 VGGT-SLAM & 1257 & \textbf{0.174} & \textbf{0.201} & \textbf{0.566} & \textbf{0.209} & \textbf{0.436} & \textbf{0.062} & \textbf{0.320} & \textbf{0.239} & \textbf{0.081} & \textbf{0.244} & \textbf{0.414} & \textbf{0.072} \\
\midrule
\multicolumn{14}{c}{\cellcolor{catgray}\textbf{Test-Time Training}} \\
\midrule
 TTT3R & 793.3 & 0.274 & 0.464 & 0.278 & 0.367 & 0.212 & 0.082 & 0.412 & 0.075 & 0.109 & 0.415 & 0.189 & 0.096 \\
 Scal3R & 1266 & 0.189 & 0.216 & \textbf{0.711} & 0.308 & 0.413 & \textbf{0.052} & 0.329 & 0.188 & \textbf{0.079} & 0.284 & 0.437 & \textbf{0.066} \\
 LoGeR & 1255 & 0.190 & 0.220 & 0.586 & 0.301 & 0.448 & 0.066 & 0.452 & 0.223 & 0.100 & 0.324 & 0.419 & 0.083 \\
 LoGeR$^*$ & 1255 & \cellcolor{secondorange}\textbf{0.142} & \cellcolor{bestred}\textbf{0.130} & 0.617 & \textbf{0.206} & \textbf{0.490} & 0.057 & \textbf{0.325} & \textbf{0.240} & 0.096 & \textbf{0.220} & \textbf{0.449} & 0.077 \\
\bottomrule
\end{tabular}%
}
\end{table*}

\begin{table*}[t]
\centering
\setlength{\fboxsep}{0pt}
\caption{\textbf{Per-Dataset Results on RoboTwin}.
Performance across different input regimes: \textit{Single Frame}, \textit{Sparse}, \textit{Medium}, \textit{Dense}, and the \textit{Average}.
The best, second-best, and third-best results in each column are highlighted in
\colorbox{bestred}{deep blue}, \colorbox{secondorange}{medium blue}, and \colorbox{thirdyellow}{light blue}, respectively.
Out-of-memory (OOM) and Timeout (T.O) cells are shaded \colorbox{oomred}{light red};
\textit{Average} values for those rows are wrapped in parentheses and excluded from per-column ranking.
Within each sub-category, the \textbf{bold} value marks the in-group best.
Note that \ours (Ours) is excluded from the per-column rankings.
}
\label{tab:dataset_robotwin}
\resizebox{\textwidth}{!}{%
\renewcommand{\arraystretch}{1.15}
\setlength{\tabcolsep}{3pt}
\begin{tabular}{l c >{\columncolor{subcol}}c >{\columncolor{subcol}}c >{\columncolor{subcol}}c >{\columncolor{subcol}}c >{\columncolor{subcol}}c >{\columncolor{subcol}}c >{\columncolor{subcol}}c >{\columncolor{subcol}}c >{\columncolor{subcol}}c >{\columncolor{subcol}}c >{\columncolor{subcol}}c >{\columncolor{subcol}}c}
\toprule
\multirow{2.5}{*}{\textbf{Method}}
& \multirow{2.5}{*}{\makecell{\textbf{\#Params}\\\textbf{(M)}}}
& \multicolumn{1}{c}{\textbf{Single Frame}}
& \multicolumn{2}{c}{\textbf{Sparse}}
& \multicolumn{3}{c}{\textbf{Medium}}
& \multicolumn{3}{c}{\textbf{Dense}}
& \multicolumn{3}{c}{\textbf{Average}} \\
\cmidrule(lr){3-3} \cmidrule(lr){4-5} \cmidrule(lr){6-8} \cmidrule(lr){9-11} \cmidrule(lr){12-14}
 & &
 AbsRel$\downarrow$ &
 AbsRel$\downarrow$ & AUC@30$\uparrow$ &
 AbsRel$\downarrow$ & AUC@30$\uparrow$ & ATE$\downarrow$ &
 AbsRel$\downarrow$ & AUC@30$\uparrow$ & ATE$\downarrow$ &
 AbsRel$\downarrow$ & AUC@30$\uparrow$ & ATE$\downarrow$ \\
\midrule
\multicolumn{14}{c}{\cellcolor{catgray}\textbf{Optimization-based}} \\
\midrule
 DUSt3R & 571.2 & \textbf{0.342} & \textbf{0.728} & 0.061 & 0.928 & 0.067 & 0.120 & \cellcolor{oomred}OOM & \cellcolor{oomred}OOM & \cellcolor{oomred}OOM & (0.828) & (0.064) & (0.120) \\
 MASt3R & 688.6 & 0.388 & 0.756 & \textbf{0.190} & \textbf{0.695} & \textbf{0.132} & \textbf{0.107} & \cellcolor{oomred}OOM & \cellcolor{oomred}OOM & \cellcolor{oomred}OOM & (0.725) & (0.161) & (0.107) \\
\midrule
\multicolumn{14}{c}{\cellcolor{catgray}\textbf{End-to-End Feed-Forward}} \\
\midrule
 VGGT & 1257 & 0.218 & 0.369 & 0.253 & 0.323 & 0.105 & 0.093 & \cellcolor{oomred}OOM & \cellcolor{oomred}OOM & \cellcolor{oomred}OOM & (0.346) & (0.179) & (0.093) \\
 Fast3R & 647.5 & 0.351 & 0.698 & 0.079 & 0.580 & 0.030 & 0.109 & 0.624 & 0.045 & 0.110 & 0.634 & 0.051 & 0.110 \\
 FastVGGT & 1158 & 0.218 & 0.416 & 0.185 & 0.317 & 0.100 & 0.085 & \cellcolor{bestred}\textbf{0.300} & 0.088 & 0.090 & \cellcolor{thirdyellow}0.345 & 0.124 & 0.088 \\
 MUSt3R & 423.4 & 0.366 & 0.507 & 0.126 & 0.473 & 0.137 & 0.093 & \cellcolor{oomred}T.O & \cellcolor{oomred}T.O & \cellcolor{oomred}T.O & (0.490) & (0.131) & (0.093) \\
 MapAnything & 1228 & 0.232 & 0.415 & 0.142 & 0.410 & 0.124 & 0.096 & \cellcolor{oomred}OOM & \cellcolor{oomred}OOM & \cellcolor{oomred}OOM & (0.412) & (0.133) & (0.096) \\
 OmniVGGT & 1217 & 0.212 & 0.333 & 0.090 & 0.495 & 0.070 & 0.106 & \cellcolor{oomred}OOM & \cellcolor{oomred}OOM & \cellcolor{oomred}OOM & (0.414) & (0.080) & (0.106) \\
 $\pi^{3}$ & 958.7 & \cellcolor{secondorange}0.181 & 0.309 & \cellcolor{secondorange}\textbf{0.482} & 0.317 & 0.250 & \cellcolor{secondorange}\textbf{0.064} & \cellcolor{secondorange}0.313 & \cellcolor{bestred}\textbf{0.263} & \cellcolor{bestred}\textbf{0.066} & \cellcolor{bestred}\textbf{0.313} & \cellcolor{bestred}\textbf{0.332} & \cellcolor{bestred}\textbf{0.065} \\
 $\pi^{3}$-X & 1360 & \cellcolor{bestred}\textbf{0.170} & 0.271 & 0.345 & 0.310 & 0.090 & 0.084 & \cellcolor{oomred}OOM & \cellcolor{oomred}OOM & \cellcolor{oomred}OOM & (0.291) & (0.217) & (0.084) \\
 AMB3R & 1563 & 0.184 & \cellcolor{secondorange}\textbf{0.252} & \cellcolor{thirdyellow}0.454 & \cellcolor{thirdyellow}\textbf{0.273} & 0.223 & \cellcolor{thirdyellow}0.068 & \cellcolor{oomred}OOM & \cellcolor{oomred}OOM & \cellcolor{oomred}OOM & (0.263) & (0.339) & (0.068) \\
 DA3-Small & 34.3 & 0.310 & 0.575 & 0.060 & 0.485 & 0.106 & 0.087 & 0.494 & 0.108 & 0.091 & 0.518 & 0.091 & 0.089 \\
 DA3-Base & 135.4 & 0.285 & 0.565 & 0.095 & 0.476 & 0.133 & 0.084 & 0.484 & 0.144 & 0.091 & 0.509 & 0.124 & 0.088 \\
 DA3-Large & 410.9 & 0.255 & 0.426 & 0.285 & 0.348 & 0.210 & 0.070 & \cellcolor{oomred}OOM & \cellcolor{oomred}OOM & \cellcolor{oomred}OOM & (0.387) & (0.248) & (0.070) \\
 DA3-Giant & 1356 & 0.287 & 0.320 & 0.441 & 0.315 & 0.253 & 0.071 & \cellcolor{oomred}OOM & \cellcolor{oomred}OOM & \cellcolor{oomred}OOM & (0.318) & (0.347) & (0.071) \\
 DA3-Nested & 1690 & 0.259 & 0.317 & 0.385 & 0.280 & \cellcolor{secondorange}\textbf{0.263} & 0.069 & \cellcolor{oomred}OOM & \cellcolor{oomred}OOM & \cellcolor{oomred}OOM & (0.298) & (0.324) & (0.069) \\
 WorldMirror & 1263 & 0.205 & 0.363 & 0.179 & 0.344 & 0.121 & 0.095 & \cellcolor{oomred}OOM & \cellcolor{oomred}OOM & \cellcolor{oomred}OOM & (0.353) & (0.150) & (0.095) \\
 VGGT-Omega & 1144 & \cellcolor{thirdyellow}0.182 & \cellcolor{bestred}\textbf{0.217} & \cellcolor{bestred}\textbf{0.529} & \cellcolor{bestred}\textbf{0.189} & \cellcolor{bestred}\textbf{0.372} & \cellcolor{bestred}\textbf{0.058} & -- & -- & -- & (0.203) & (0.450) & (0.058) \\ \hdashline
 \ours (Ours) & 1304 & 0.072 & 0.068 & 0.765 & 0.045 & 0.541 & 0.036 & \cellcolor{oomred}OOM & \cellcolor{oomred}OOM & \cellcolor{oomred}OOM & (0.062) & (0.653) & (0.036) \\
\midrule
\multicolumn{14}{c}{\cellcolor{catgray}\textbf{Online}} \\
\midrule
 Spann3r$^{224}$ & 658.7 & 0.307 & 0.612 & 0.085 & 0.421 & 0.067 & 0.099 & 0.460 & 0.072 & 0.103 & 0.498 & 0.075 & 0.101 \\
 CUT3R & 793.3 & 0.320 & 0.617 & 0.043 & 0.501 & 0.064 & 0.101 & 0.539 & 0.032 & 0.104 & 0.552 & 0.046 & 0.102 \\
 MonST3R & 571.2 & 0.361 & 0.627 & 0.027 & 0.427 & 0.061 & 0.127 & \cellcolor{oomred}OOM & \cellcolor{oomred}OOM & \cellcolor{oomred}OOM & (0.527) & (0.044) & (0.127) \\
 Point3R & 828 & 0.295 & 0.583 & 0.017 & 0.591 & 0.045 & 0.109 & 0.613 & 0.053 & 0.114 & 0.596 & 0.038 & 0.112 \\
 Stream3R-S & 1191 & 0.216 & 0.354 & \textbf{0.289} & 0.449 & 0.106 & 0.091 & \cellcolor{oomred}OOM & \cellcolor{oomred}OOM & \cellcolor{oomred}OOM & (0.401) & (0.198) & (0.091) \\
 Stream3R-W & 1191 & 0.216 & 0.354 & \textbf{0.289} & 0.484 & 0.079 & 0.138 & \cellcolor{oomred}OOM & \cellcolor{oomred}OOM & \cellcolor{oomred}OOM & (0.419) & (0.184) & (0.138) \\
 StreamVGGT & 1257 & 0.203 & 0.372 & 0.155 & 0.391 & 0.128 & 0.085 & 0.369 & 0.116 & 0.081 & 0.377 & 0.133 & 0.083 \\
 Page4D & 1257 & 0.224 & \textbf{0.326} & 0.122 & 0.347 & 0.082 & 0.092 & \cellcolor{oomred}OOM & \cellcolor{oomred}OOM & \cellcolor{oomred}OOM & (0.337) & (0.102) & (0.092) \\
 InfiniteVGGT & 1257 & \textbf{0.203} & 0.369 & 0.152 & 0.392 & 0.131 & 0.085 & 0.369 & 0.121 & \textbf{0.080} & \textbf{0.377} & 0.135 & 0.082 \\
 Wint3R & 749.5 & 0.245 & 0.569 & 0.106 & 0.438 & 0.111 & 0.085 & 0.517 & 0.078 & 0.095 & 0.508 & 0.099 & 0.090 \\
 LongStream-B & 1191 & 0.243 & 0.462 & 0.199 & 0.512 & 0.098 & 0.098 & 0.444 & 0.070 & 0.092 & 0.473 & 0.122 & 0.095 \\
 LongStream-S & 1191 & 0.243 & 0.462 & 0.199 & 0.420 & 0.085 & 0.097 & 0.448 & 0.106 & 0.083 & 0.443 & 0.130 & 0.090 \\
 LingbotMap$^{*}$-W & 1158 & 0.303 & 0.509 & 0.262 & \textbf{0.336} & 0.201 & 0.082 & \cellcolor{thirdyellow}\textbf{0.354} & 0.191 & 0.088 & 0.400 & 0.218 & 0.085 \\
 LingbotMap$^{*}$-S & 1158 & 0.303 & 0.509 & 0.262 & 0.361 & \textbf{0.202} & \textbf{0.081} & 0.429 & \textbf{0.199} & 0.084 & 0.433 & \textbf{0.221} & \textbf{0.082} \\
\midrule
\multicolumn{14}{c}{\cellcolor{catgray}\textbf{Chunk-wise}} \\
\midrule
 VGGT-Long & 1257 & 0.218 & 0.369 & 0.253 & 0.334 & 0.088 & 0.093 & \textbf{0.415} & 0.100 & 0.096 & 0.373 & 0.147 & 0.095 \\
 $\pi^{3}$-Long & 958.7 & \cellcolor{secondorange}\textbf{0.181} & \textbf{0.309} & \cellcolor{secondorange}\textbf{0.482} & \textbf{0.309} & 0.238 & \cellcolor{thirdyellow}\textbf{0.068} & 0.439 & \cellcolor{secondorange}\textbf{0.243} & \cellcolor{secondorange}\textbf{0.076} & \textbf{0.353} & \cellcolor{secondorange}\textbf{0.321} & \cellcolor{secondorange}\textbf{0.072} \\
 DA3-Streaming & 1356 & 0.287 & 0.320 & 0.442 & 0.310 & \cellcolor{thirdyellow}\textbf{0.257} & 0.074 & 0.461 & 0.189 & 0.098 & 0.364 & \cellcolor{thirdyellow}0.296 & 0.086 \\
\midrule
\multicolumn{14}{c}{\cellcolor{catgray}\textbf{SLAM-based}} \\
\midrule
 MASt3R-SLAM & 688.6 & 0.504 & 0.937 & 0.109 & 0.806 & 0.094 & 0.100 & 0.854 & \textbf{0.151} & 0.101 & 0.866 & 0.118 & 0.101 \\
 VGGT-SLAM & 1257 & \textbf{0.218} & \textbf{0.369} & \textbf{0.253} & \textbf{0.423} & \textbf{0.110} & \textbf{0.085} & \textbf{0.426} & 0.110 & \textbf{0.099} & \textbf{0.406} & \textbf{0.158} & \textbf{0.092} \\
\midrule
\multicolumn{14}{c}{\cellcolor{catgray}\textbf{Test-Time Training}} \\
\midrule
 TTT3R & 793.3 & 0.320 & 0.606 & 0.051 & 0.480 & 0.070 & 0.103 & 0.502 & 0.062 & 0.097 & 0.529 & 0.061 & 0.100 \\
 Scal3R & 1266 & 0.193 & 0.350 & \textbf{0.280} & 0.309 & 0.147 & 0.083 & \textbf{0.391} & 0.132 & 0.085 & 0.350 & 0.186 & 0.084 \\
 LoGeR & 1255 & 0.223 & 0.350 & 0.195 & 0.354 & 0.160 & \textbf{0.070} & 0.466 & 0.181 & 0.083 & 0.390 & 0.178 & 0.076 \\
 LoGeR$^*$ & 1255 & \textbf{0.191} & \cellcolor{thirdyellow}\textbf{0.259} & 0.233 & \cellcolor{secondorange}\textbf{0.271} & \textbf{0.173} & 0.070 & 0.430 & \cellcolor{thirdyellow}\textbf{0.203} & \cellcolor{thirdyellow}\textbf{0.077} & \cellcolor{secondorange}\textbf{0.320} & \textbf{0.203} & \cellcolor{thirdyellow}\textbf{0.074} \\
\bottomrule
\end{tabular}%
}
\end{table*}

\begin{table*}[t]
\centering
\setlength{\fboxsep}{0pt}
\caption{\textbf{Per-Dataset Results on Xperience}.
Performance across different input regimes: \textit{Single Frame}, \textit{Sparse}, \textit{Medium}, \textit{Dense}, and the \textit{Average}.
The best, second-best, and third-best results in each column are highlighted in
\colorbox{bestred}{deep blue}, \colorbox{secondorange}{medium blue}, and \colorbox{thirdyellow}{light blue}, respectively.
Out-of-memory (OOM) and Timeout (T.O) cells are shaded \colorbox{oomred}{light red};
\textit{Average} values for those rows are wrapped in parentheses and excluded from per-column ranking.
Within each sub-category, the \textbf{bold} value marks the in-group best.
Note that \ours (Ours) is excluded from the per-column rankings.
}
\label{tab:dataset_ropedia}
\resizebox{\textwidth}{!}{%
\renewcommand{\arraystretch}{1.15}
\setlength{\tabcolsep}{3pt}
\begin{tabular}{l c >{\columncolor{subcol}}c >{\columncolor{subcol}}c >{\columncolor{subcol}}c >{\columncolor{subcol}}c >{\columncolor{subcol}}c >{\columncolor{subcol}}c >{\columncolor{subcol}}c >{\columncolor{subcol}}c >{\columncolor{subcol}}c >{\columncolor{subcol}}c >{\columncolor{subcol}}c >{\columncolor{subcol}}c}
\toprule
\multirow{2.5}{*}{\textbf{Method}}
& \multirow{2.5}{*}{\makecell{\textbf{\#Params}\\\textbf{(M)}}}
& \multicolumn{1}{c}{\textbf{Single Frame}}
& \multicolumn{2}{c}{\textbf{Sparse}}
& \multicolumn{3}{c}{\textbf{Medium}}
& \multicolumn{3}{c}{\textbf{Dense}}
& \multicolumn{3}{c}{\textbf{Average}} \\
\cmidrule(lr){3-3} \cmidrule(lr){4-5} \cmidrule(lr){6-8} \cmidrule(lr){9-11} \cmidrule(lr){12-14}
 & &
 AbsRel$\downarrow$ &
 AbsRel$\downarrow$ & AUC@30$\uparrow$ &
 AbsRel$\downarrow$ & AUC@30$\uparrow$ & ATE$\downarrow$ &
 AbsRel$\downarrow$ & AUC@30$\uparrow$ & ATE$\downarrow$ &
 AbsRel$\downarrow$ & AUC@30$\uparrow$ & ATE$\downarrow$ \\
\midrule
\multicolumn{14}{c}{\cellcolor{catgray}\textbf{Optimization-based}} \\
\midrule
 DUSt3R & 571.2 & \cellcolor{secondorange}\textbf{0.054} & 0.092 & 0.471 & \textbf{0.240} & 0.079 & \cellcolor{thirdyellow}\textbf{4.312} & \cellcolor{oomred}OOM & \cellcolor{oomred}OOM & \cellcolor{oomred}OOM & (0.166) & (0.275) & (4.312) \\
 MASt3R & 688.6 & 0.087 & \textbf{0.068} & \cellcolor{bestred}\textbf{0.694} & 0.261 & \textbf{0.202} & 4.373 & \cellcolor{oomred}OOM & \cellcolor{oomred}OOM & \cellcolor{oomred}OOM & (0.164) & (0.448) & (4.373) \\
\midrule
\multicolumn{14}{c}{\cellcolor{catgray}\textbf{End-to-End Feed-Forward}} \\
\midrule
 VGGT & 1257 & 0.126 & 0.051 & 0.070 & 0.078 & 0.047 & 7.239 & \cellcolor{oomred}OOM & \cellcolor{oomred}OOM & \cellcolor{oomred}OOM & (0.065) & (0.059) & (7.239) \\
 Fast3R & 647.5 & 0.085 & 0.359 & 0.005 & 0.468 & 0.001 & 7.571 & -- & -- & -- & 0.413 & 0.003 & 7.571 \\
 FastVGGT & 1158 & 0.128 & 0.048 & 0.073 & 0.076 & 0.036 & 7.024 & 0.113 & 0.062 & 7.943 & \cellcolor{thirdyellow}0.079 & 0.057 & 7.483 \\
 MUSt3R & 423.4 & 0.058 & 0.059 & 0.496 & 0.111 & \cellcolor{secondorange}\textbf{0.282} & 5.645 & \cellcolor{oomred}T.O & \cellcolor{oomred}T.O & \cellcolor{oomred}T.O & (0.085) & (0.389) & (5.645) \\
 MapAnything & 1228 & 0.062 & 0.040 & \cellcolor{secondorange}\textbf{0.617} & 0.098 & 0.138 & 5.422 & \cellcolor{oomred}OOM & \cellcolor{oomred}OOM & \cellcolor{oomred}OOM & (0.069) & (0.377) & (5.422) \\
 OmniVGGT & 1217 & \cellcolor{thirdyellow}\textbf{0.056} & \cellcolor{thirdyellow}0.033 & 0.042 & 0.072 & 0.042 & 6.971 & \cellcolor{oomred}OOM & \cellcolor{oomred}OOM & \cellcolor{oomred}OOM & (0.052) & (0.042) & (6.971) \\
 $\pi^{3}$ & 958.7 & 0.133 & \cellcolor{secondorange}0.031 & 0.127 & \cellcolor{secondorange}0.050 & 0.130 & 7.219 & 0.265 & 0.009 & 7.460 & 0.115 & \textbf{0.089} & 7.340 \\
 $\pi^{3}$-X & 1360 & 0.089 & \cellcolor{bestred}\textbf{0.029} & 0.421 & \cellcolor{bestred}\textbf{0.046} & 0.022 & 6.461 & \cellcolor{oomred}OOM & \cellcolor{oomred}OOM & \cellcolor{oomred}OOM & (0.038) & (0.221) & (6.461) \\
 AMB3R & 1563 & 0.104 & 0.041 & 0.022 & 0.069 & 0.045 & 7.102 & \cellcolor{oomred}OOM & \cellcolor{oomred}OOM & \cellcolor{oomred}OOM & (0.055) & (0.034) & (7.102) \\
 DA3-Small & 34.3 & 0.112 & 0.061 & 0.011 & 0.094 & 0.025 & 6.493 & 0.104 & 0.041 & 6.741 & 0.086 & 0.026 & 6.617 \\
 DA3-Base & 135.4 & 0.103 & 0.053 & 0.071 & 0.078 & 0.053 & 5.507 & \cellcolor{bestred}\textbf{0.088} & \textbf{0.081} & \textbf{5.624} & \cellcolor{secondorange}\textbf{0.073} & 0.068 & \textbf{5.565} \\
 DA3-Large & 410.9 & 0.099 & 0.064 & 0.103 & 0.073 & 0.116 & 5.989 & \cellcolor{oomred}OOM & \cellcolor{oomred}OOM & \cellcolor{oomred}OOM & (0.068) & (0.109) & (5.989) \\
 DA3-Giant & 1356 & 0.162 & 0.049 & 0.291 & 0.057 & 0.193 & \textbf{4.895} & \cellcolor{oomred}OOM & \cellcolor{oomred}OOM & \cellcolor{oomred}OOM & (0.053) & (0.242) & (4.895) \\
 DA3-Nested & 1690 & 0.180 & 0.044 & 0.201 & \cellcolor{thirdyellow}0.051 & 0.185 & 6.236 & \cellcolor{oomred}OOM & \cellcolor{oomred}OOM & \cellcolor{oomred}OOM & (0.047) & (0.193) & (6.236) \\
 WorldMirror & 1263 & 0.076 & 0.048 & 0.197 & 0.091 & 0.040 & 7.172 & \cellcolor{oomred}OOM & \cellcolor{oomred}OOM & \cellcolor{oomred}OOM & (0.069) & (0.118) & (7.172) \\
 VGGT-Omega & 1144 & 0.070 & 0.038 & \cellcolor{thirdyellow}0.528 & 0.069 & \cellcolor{thirdyellow}0.275 & 5.317 & -- & -- & -- & (0.054) & (0.401) & (5.317) \\ \hdashline
 \ours (Ours) & 1304 & 0.064 & 0.037 & 0.703 & 0.046 & 0.385 & 4.508 & \cellcolor{oomred}OOM & \cellcolor{oomred}OOM & \cellcolor{oomred}OOM & (0.041) & (0.544) & (4.508) \\
\midrule
\multicolumn{14}{c}{\cellcolor{catgray}\textbf{Online}} \\
\midrule
 Spann3r$^{224}$ & 658.7 & 0.156 & 0.292 & 0.066 & 0.221 & 0.012 & 5.472 & 0.171 & 0.216 & 4.852 & 0.228 & 0.098 & 5.162 \\
 CUT3R & 793.3 & 0.148 & 0.074 & 0.124 & 0.114 & 0.027 & 4.663 & 0.316 & 0.029 & 7.244 & 0.168 & 0.060 & 5.954 \\
 MonST3R & 571.2 & 0.191 & 0.116 & 0.021 & 0.283 & 0.007 & 5.104 & \cellcolor{oomred}OOM & \cellcolor{oomred}OOM & \cellcolor{oomred}OOM & (0.200) & (0.014) & (5.104) \\
 Point3R & 828 & 0.114 & 0.074 & 0.031 & 0.156 & 0.017 & 6.804 & 0.234 & 0.029 & 7.762 & 0.154 & 0.025 & 7.283 \\
 Stream3R-S & 1191 & 0.083 & 0.055 & 0.046 & 0.419 & 0.005 & 7.469 & \cellcolor{oomred}OOM & \cellcolor{oomred}OOM & \cellcolor{oomred}OOM & (0.237) & (0.025) & (7.469) \\
 Stream3R-W & 1191 & 0.083 & 0.056 & 0.047 & 0.407 & 0.006 & 7.484 & \cellcolor{oomred}OOM & \cellcolor{oomred}OOM & \cellcolor{oomred}OOM & (0.231) & (0.027) & (7.484) \\
 StreamVGGT & 1257 & 0.108 & 0.131 & 0.054 & 0.159 & 0.033 & 7.306 & 0.185 & 0.053 & 8.103 & 0.158 & 0.047 & 7.705 \\
 Page4D & 1257 & 0.134 & \textbf{0.049} & 0.075 & \textbf{0.073} & 0.032 & 7.165 & \cellcolor{oomred}OOM & \cellcolor{oomred}OOM & \cellcolor{oomred}OOM & (0.061) & (0.054) & (7.165) \\
 InfiniteVGGT & 1257 & 0.109 & 0.131 & 0.051 & 0.159 & 0.033 & 7.317 & 0.185 & 0.054 & 8.087 & 0.159 & 0.046 & 7.702 \\
 Wint3R & 749.5 & \cellcolor{bestred}\textbf{0.053} & 0.054 & 0.039 & 0.074 & 0.065 & 6.512 & 0.290 & 0.059 & 7.698 & 0.139 & 0.054 & 7.105 \\
 LongStream-B & 1191 & 0.142 & 0.079 & 0.029 & 0.118 & 0.004 & 6.245 & 0.259 & 0.027 & 2.884 & 0.152 & 0.020 & 4.565 \\
 LongStream-S & 1191 & 0.142 & 0.075 & 0.028 & 0.137 & 0.004 & 6.278 & 0.223 & 0.020 & 6.397 & 0.145 & 0.017 & 6.337 \\
 LingbotMap$^{*}$-W & 1158 & 0.060 & 0.050 & \textbf{0.197} & 0.086 & \cellcolor{bestred}\textbf{0.299} & \cellcolor{bestred}\textbf{2.164} & 0.157 & 0.193 & \textbf{1.458} & 0.098 & 0.230 & \cellcolor{bestred}\textbf{1.811} \\
 LingbotMap$^{*}$-S & 1158 & 0.060 & 0.050 & \textbf{0.197} & 0.097 & 0.270 & \cellcolor{secondorange}2.445 & \cellcolor{thirdyellow}\textbf{0.100} & \cellcolor{thirdyellow}\textbf{0.451} & 3.616 & \textbf{0.082} & \cellcolor{secondorange}\textbf{0.306} & \cellcolor{thirdyellow}3.031 \\
\midrule
\multicolumn{14}{c}{\cellcolor{catgray}\textbf{Chunk-wise}} \\
\midrule
 VGGT-Long & 1257 & \textbf{0.126} & 0.051 & 0.070 & 0.244 & 0.067 & 7.389 & 1.390 & 0.075 & 5.495 & 0.562 & 0.071 & 6.442 \\
 $\pi^{3}$-Long & 958.7 & 0.133 & \cellcolor{secondorange}\textbf{0.031} & 0.127 & \textbf{0.091} & 0.134 & 6.415 & \textbf{0.465} & \cellcolor{bestred}\textbf{0.809} & \cellcolor{bestred}\textbf{0.215} & \textbf{0.196} & \cellcolor{bestred}\textbf{0.357} & \textbf{3.315} \\
 DA3-Streaming & 1356 & 0.162 & 0.049 & \textbf{0.285} & 0.175 & \textbf{0.173} & \textbf{5.291} & 0.474 & 0.319 & 2.817 & 0.233 & \cellcolor{thirdyellow}0.259 & 4.054 \\
\midrule
\multicolumn{14}{c}{\cellcolor{catgray}\textbf{SLAM-based}} \\
\midrule
 MASt3R-SLAM & 688.6 & 0.206 & 0.144 & 0.000 & \textbf{0.201} & 0.001 & \textbf{6.986} & \textbf{0.279} & \textbf{0.423} & \textbf{4.841} & \textbf{0.208} & \textbf{0.141} & \textbf{5.913} \\
 VGGT-SLAM & 1257 & \textbf{0.126} & \textbf{0.051} & \textbf{0.070} & 0.302 & \textbf{0.044} & 7.735 & 0.973 & 0.051 & 5.342 & 0.442 & 0.055 & 6.538 \\
\midrule
\multicolumn{14}{c}{\cellcolor{catgray}\textbf{Test-Time Training}} \\
\midrule
 TTT3R & 793.3 & 0.148 & 0.064 & \textbf{0.107} & 0.102 & 0.061 & 7.159 & 0.143 & 0.119 & 5.582 & 0.103 & 0.096 & 6.371 \\
 Scal3R & 1266 & 0.256 & 0.045 & 0.056 & 0.192 & 0.010 & 7.411 & 0.462 & 0.195 & 3.739 & 0.233 & 0.087 & 5.575 \\
 LoGeR & 1255 & 0.122 & 0.040 & 0.036 & 0.076 & 0.043 & \textbf{4.821} & 0.158 & 0.400 & \cellcolor{thirdyellow}1.088 & 0.092 & 0.160 & \cellcolor{secondorange}\textbf{2.954} \\
 LoGeR$^*$ & 1255 & \textbf{0.113} & \textbf{0.034} & 0.071 & \textbf{0.066} & \textbf{0.093} & 7.053 & \cellcolor{secondorange}\textbf{0.090} & \cellcolor{secondorange}\textbf{0.583} & \cellcolor{secondorange}\textbf{0.608} & \cellcolor{bestred}\textbf{0.064} & \textbf{0.249} & 3.830 \\
\bottomrule
\end{tabular}%
}
\end{table*}

\begin{table*}[t]
\centering
\setlength{\fboxsep}{0pt}
\caption{\textbf{Per-Dataset Results on Scannet++}.
Performance across different input regimes: \textit{Single Frame}, \textit{Sparse}, \textit{Medium}, \textit{Dense}, and the \textit{Average}.
The best, second-best, and third-best results in each column are highlighted in
\colorbox{bestred}{deep blue}, \colorbox{secondorange}{medium blue}, and \colorbox{thirdyellow}{light blue}, respectively.
Out-of-memory (OOM) and Timeout (T.O) cells are shaded \colorbox{oomred}{light red};
\textit{Average} values for those rows are wrapped in parentheses and excluded from per-column ranking.
Within each sub-category, the \textbf{bold} value marks the in-group best.
Note that \ours (Ours) is excluded from the per-column rankings.
}
\label{tab:dataset_scannetpp}
\resizebox{\textwidth}{!}{%
\renewcommand{\arraystretch}{1.15}
\setlength{\tabcolsep}{3pt}
\begin{tabular}{l c >{\columncolor{subcol}}c >{\columncolor{subcol}}c >{\columncolor{subcol}}c >{\columncolor{subcol}}c >{\columncolor{subcol}}c >{\columncolor{subcol}}c >{\columncolor{subcol}}c >{\columncolor{subcol}}c >{\columncolor{subcol}}c >{\columncolor{subcol}}c >{\columncolor{subcol}}c >{\columncolor{subcol}}c >{\columncolor{subcol}}c >{\columncolor{subcol}}c >{\columncolor{subcol}}c}
\toprule
\multirow{2.5}{*}{\textbf{Method}}
& \multirow{2.5}{*}{\makecell{\textbf{\#Params}\\\textbf{(M)}}}
& \multicolumn{1}{c}{\textbf{Single Frame}}
& \multicolumn{2}{c}{\textbf{Sparse}}
& \multicolumn{4}{c}{\textbf{Medium}}
& \multicolumn{4}{c}{\textbf{Dense}}
& \multicolumn{4}{c}{\textbf{Average}} \\
\cmidrule(lr){3-3} \cmidrule(lr){4-5} \cmidrule(lr){6-9} \cmidrule(lr){10-13} \cmidrule(lr){14-17}
 & &
 AbsRel$\downarrow$ &
 AbsRel$\downarrow$ & AUC@30$\uparrow$ &
 AbsRel$\downarrow$ & AUC@30$\uparrow$ & ATE$\downarrow$ & F-Score$\uparrow$ &
 AbsRel$\downarrow$ & AUC@30$\uparrow$ & ATE$\downarrow$ & F-Score$\uparrow$ &
 AbsRel$\downarrow$ & AUC@30$\uparrow$ & ATE$\downarrow$ & F-Score$\uparrow$ \\
\midrule
\multicolumn{17}{c}{\cellcolor{catgray}\textbf{Optimization-based}} \\
\midrule
 DUSt3R & 571.2 & \textbf{0.038} & \textbf{0.064} & 0.762 & \textbf{0.057} & 0.713 & 0.392 & 0.374 & \cellcolor{oomred}OOM & \cellcolor{oomred}OOM & \cellcolor{oomred}OOM & \cellcolor{oomred}OOM & (0.060) & (0.738) & (0.392) & (0.374) \\
 MASt3R & 688.6 & 0.061 & 0.069 & \textbf{0.858} & 0.073 & \textbf{0.757} & \textbf{0.184} & \textbf{0.431} & \cellcolor{oomred}OOM & \cellcolor{oomred}OOM & \cellcolor{oomred}OOM & \cellcolor{oomred}OOM & (0.071) & (0.807) & (0.184) & (0.431) \\
\midrule
\multicolumn{17}{c}{\cellcolor{catgray}\textbf{End-to-End Feed-Forward}} \\
\midrule
 VGGT & 1257 & 0.038 & 0.049 & 0.901 & 0.043 & 0.942 & 0.067 & 0.657 & \cellcolor{oomred}OOM & \cellcolor{oomred}OOM & \cellcolor{oomred}OOM & \cellcolor{oomred}OOM & (0.046) & (0.921) & (0.067) & (0.657) \\
 Fast3R & 647.5 & 0.046 & 0.102 & 0.567 & 0.095 & 0.619 & 0.555 & 0.325 & 0.112 & 0.599 & 0.639 & 0.308 & 0.103 & 0.595 & 0.597 & 0.316 \\
 FastVGGT & 1158 & 0.038 & 0.052 & 0.838 & 0.043 & 0.922 & 0.079 & 0.524 & \cellcolor{thirdyellow}0.040 & \cellcolor{thirdyellow}0.915 & 0.104 & 0.536 & 0.045 & 0.892 & 0.092 & 0.530 \\
 MUSt3R & 423.4 & 0.039 & 0.044 & 0.881 & 0.042 & 0.921 & 0.077 & 0.506 & \cellcolor{oomred}T.O & \cellcolor{oomred}T.O & \cellcolor{oomred}T.O & \cellcolor{oomred}T.O & (0.043) & (0.901) & (0.077) & (0.506) \\
 MapAnything & 1228 & 0.041 & 0.039 & 0.854 & 0.039 & 0.897 & 0.100 & 0.498 & \cellcolor{oomred}OOM & \cellcolor{oomred}OOM & \cellcolor{oomred}OOM & \cellcolor{oomred}OOM & (0.039) & (0.875) & (0.100) & (0.498) \\
 OmniVGGT & 1217 & 0.047 & 0.033 & 0.728 & 0.039 & 0.875 & 0.162 & 0.526 & \cellcolor{oomred}OOM & \cellcolor{oomred}OOM & \cellcolor{oomred}OOM & \cellcolor{oomred}OOM & (0.036) & (0.801) & (0.162) & (0.526) \\
 $\pi^{3}$ & 958.7 & \cellcolor{secondorange}0.035 & \cellcolor{secondorange}\textbf{0.031} & 0.901 & \cellcolor{bestred}\textbf{0.031} & 0.952 & 0.044 & 0.709 & \cellcolor{bestred}\textbf{0.030} & \cellcolor{bestred}\textbf{0.953} & \cellcolor{bestred}\textbf{0.058} & \cellcolor{bestred}\textbf{0.723} & \cellcolor{bestred}\textbf{0.031} & \cellcolor{bestred}\textbf{0.935} & \cellcolor{bestred}\textbf{0.051} & \cellcolor{bestred}\textbf{0.716} \\
 $\pi^{3}$-X & 1360 & \cellcolor{thirdyellow}0.036 & \cellcolor{secondorange}0.031 & 0.903 & \cellcolor{bestred}0.031 & 0.951 & \cellcolor{thirdyellow}0.039 & \cellcolor{thirdyellow}0.728 & \cellcolor{oomred}OOM & \cellcolor{oomred}OOM & \cellcolor{oomred}OOM & \cellcolor{oomred}OOM & (0.031) & (0.927) & (0.039) & (0.728) \\
 AMB3R & 1563 & \cellcolor{secondorange}\textbf{0.035} & 0.036 & 0.871 & 0.035 & 0.929 & 0.084 & 0.557 & \cellcolor{oomred}OOM & \cellcolor{oomred}OOM & \cellcolor{oomred}OOM & \cellcolor{oomred}OOM & (0.036) & (0.900) & (0.084) & (0.557) \\
 DA3-Small & 34.3 & 0.073 & 0.075 & 0.677 & 0.071 & 0.614 & 0.616 & 0.262 & 0.073 & 0.533 & 0.787 & 0.216 & 0.073 & 0.608 & 0.701 & 0.239 \\
 DA3-Base & 135.4 & 0.058 & 0.056 & 0.795 & 0.048 & 0.799 & 0.215 & 0.388 & 0.048 & 0.759 & 0.288 & 0.346 & 0.050 & 0.785 & 0.251 & 0.367 \\
 DA3-Large & 410.9 & \cellcolor{thirdyellow}0.036 & 0.039 & 0.899 & 0.035 & 0.941 & 0.062 & 0.609 & \cellcolor{oomred}OOM & \cellcolor{oomred}OOM & \cellcolor{oomred}OOM & \cellcolor{oomred}OOM & (0.037) & (0.920) & (0.062) & (0.609) \\
 DA3-Giant & 1356 & \cellcolor{secondorange}0.035 & \cellcolor{thirdyellow}0.032 & \cellcolor{bestred}\textbf{0.959} & \cellcolor{bestred}0.031 & \cellcolor{bestred}0.983 & \cellcolor{bestred}\textbf{0.020} & \cellcolor{bestred}\textbf{0.761} & \cellcolor{oomred}OOM & \cellcolor{oomred}OOM & \cellcolor{oomred}OOM & \cellcolor{oomred}OOM & (0.031) & (0.971) & (0.020) & (0.761) \\
 DA3-Nested & 1690 & \cellcolor{secondorange}\textbf{0.035} & \cellcolor{secondorange}0.031 & \cellcolor{secondorange}0.958 & \cellcolor{bestred}0.031 & \cellcolor{bestred}\textbf{0.983} & \cellcolor{bestred}0.020 & \cellcolor{secondorange}0.758 & \cellcolor{oomred}OOM & \cellcolor{oomred}OOM & \cellcolor{oomred}OOM & \cellcolor{oomred}OOM & (0.031) & (0.970) & (0.020) & (0.758) \\
 WorldMirror & 1263 & 0.037 & 0.037 & 0.889 & \cellcolor{thirdyellow}0.033 & 0.932 & 0.070 & 0.686 & \cellcolor{oomred}OOM & \cellcolor{oomred}OOM & \cellcolor{oomred}OOM & \cellcolor{oomred}OOM & (0.035) & (0.911) & (0.070) & (0.686) \\
 VGGT-Omega & 1144 & \cellcolor{secondorange}\textbf{0.035} & 0.042 & \cellcolor{thirdyellow}0.946 & 0.041 & \cellcolor{secondorange}0.969 & \cellcolor{secondorange}0.036 & 0.709 & -- & -- & -- & -- & (0.042) & (0.958) & (0.036) & (0.709) \\ \hdashline
 \ours (Ours) & 1304 & 0.039 & 0.029 & 0.949 & 0.026 & 0.976 & 0.025 & 0.805 & \cellcolor{oomred}OOM & \cellcolor{oomred}OOM & \cellcolor{oomred}OOM & \cellcolor{oomred}OOM & (0.032) & (0.962) & (0.025) & (0.805) \\
\midrule
\multicolumn{17}{c}{\cellcolor{catgray}\textbf{Online}} \\
\midrule
 Spann3r$^{224}$ & 658.7 & 0.072 & 0.175 & 0.337 & 0.123 & 0.431 & 0.731 & 0.161 & 0.130 & 0.361 & 0.925 & 0.154 & 0.143 & 0.376 & 0.828 & 0.158 \\
 CUT3R & 793.3 & 0.042 & 0.069 & 0.717 & 0.070 & 0.673 & 0.506 & 0.264 & 0.083 & 0.275 & 1.132 & 0.169 & 0.074 & 0.555 & 0.819 & 0.217 \\
 MonST3R & 571.2 & 0.049 & 0.110 & 0.232 & 0.157 & 0.022 & 1.189 & 0.084 & \cellcolor{oomred}OOM & \cellcolor{oomred}OOM & \cellcolor{oomred}OOM & \cellcolor{oomred}OOM & (0.133) & (0.127) & (1.189) & (0.084) \\
 Point3R & 828 & 0.040 & 0.061 & 0.445 & 0.062 & 0.459 & 0.528 & 0.189 & 0.067 & 0.412 & 0.597 & 0.159 & 0.063 & 0.438 & 0.562 & 0.174 \\
 Stream3R-S & 1191 & \cellcolor{bestred}\textbf{0.032} & 0.045 & 0.836 & 0.226 & 0.514 & 1.088 & 0.291 & \cellcolor{oomred}OOM & \cellcolor{oomred}OOM & \cellcolor{oomred}OOM & \cellcolor{oomred}OOM & (0.136) & (0.675) & (1.088) & (0.291) \\
 Stream3R-W & 1191 & \cellcolor{bestred}\textbf{0.032} & 0.046 & 0.828 & 0.217 & 0.437 & 1.299 & 0.236 & \cellcolor{oomred}OOM & \cellcolor{oomred}OOM & \cellcolor{oomred}OOM & \cellcolor{oomred}OOM & (0.131) & (0.633) & (1.299) & (0.236) \\
 StreamVGGT & 1257 & 0.038 & 0.077 & 0.837 & 0.088 & 0.740 & 0.500 & 0.302 & 0.087 & 0.670 & 0.622 & 0.263 & 0.084 & 0.749 & 0.561 & 0.283 \\
 Page4D & 1257 & 0.044 & 0.054 & 0.812 & 0.050 & 0.867 & 0.130 & 0.385 & \cellcolor{oomred}OOM & \cellcolor{oomred}OOM & \cellcolor{oomred}OOM & \cellcolor{oomred}OOM & (0.052) & (0.840) & (0.130) & (0.385) \\
 InfiniteVGGT & 1257 & \cellcolor{secondorange}0.035 & 0.078 & 0.811 & 0.087 & 0.759 & 0.477 & 0.313 & 0.085 & 0.676 & 0.602 & 0.267 & 0.083 & 0.749 & 0.540 & 0.290 \\
 Wint3R & 749.5 & 0.049 & 0.057 & 0.565 & 0.058 & 0.442 & 0.920 & 0.213 & 0.067 & 0.344 & 0.982 & 0.169 & 0.061 & 0.450 & 0.951 & 0.191 \\
 LongStream-B & 1191 & 0.047 & 0.062 & 0.574 & 0.110 & 0.309 & 1.044 & 0.123 & 0.125 & 0.280 & 1.018 & 0.119 & 0.099 & 0.388 & 1.031 & 0.121 \\
 LongStream-S & 1191 & 0.047 & 0.062 & 0.574 & 0.089 & 0.158 & 1.313 & 0.104 & 0.118 & 0.055 & 1.351 & 0.133 & 0.090 & 0.262 & 1.332 & 0.119 \\
 LingbotMap$^{*}$-W & 1158 & 0.042 & \textbf{0.042} & \textbf{0.906} & 0.046 & 0.869 & 0.165 & 0.494 & 0.044 & 0.875 & 0.124 & 0.481 & 0.044 & 0.884 & 0.145 & 0.488 \\
 LingbotMap$^{*}$-S & 1158 & 0.042 & \textbf{0.042} & \textbf{0.906} & \textbf{0.040} & \textbf{0.900} & \textbf{0.082} & \textbf{0.556} & \cellcolor{secondorange}\textbf{0.038} & \textbf{0.912} & \cellcolor{thirdyellow}\textbf{0.088} & \textbf{0.565} & \textbf{0.040} & \textbf{0.906} & \textbf{0.085} & \textbf{0.561} \\
\midrule
\multicolumn{17}{c}{\cellcolor{catgray}\textbf{Chunk-wise}} \\
\midrule
 VGGT-Long & 1257 & 0.038 & 0.049 & 0.901 & 0.046 & 0.911 & 0.126 & 0.597 & \textbf{0.045} & 0.893 & 0.111 & 0.543 & 0.047 & 0.902 & 0.119 & 0.570 \\
 $\pi^{3}$-Long & 958.7 & \cellcolor{secondorange}0.035 & \cellcolor{secondorange}\textbf{0.031} & 0.901 & 0.043 & 0.946 & \textbf{0.050} & 0.643 & 0.102 & \cellcolor{secondorange}\textbf{0.929} & \cellcolor{secondorange}\textbf{0.085} & 0.342 & 0.059 & \cellcolor{secondorange}\textbf{0.925} & \cellcolor{secondorange}\textbf{0.067} & 0.493 \\
 DA3-Streaming & 1356 & \cellcolor{secondorange}\textbf{0.035} & \cellcolor{thirdyellow}0.032 & \cellcolor{secondorange}\textbf{0.958} & \cellcolor{secondorange}\textbf{0.032} & \cellcolor{thirdyellow}\textbf{0.956} & 0.084 & \textbf{0.707} & 0.054 & 0.817 & 0.381 & \cellcolor{thirdyellow}\textbf{0.606} & \cellcolor{thirdyellow}\textbf{0.039} & \cellcolor{thirdyellow}0.910 & 0.233 & \cellcolor{thirdyellow}\textbf{0.656} \\
\midrule
\multicolumn{17}{c}{\cellcolor{catgray}\textbf{SLAM-based}} \\
\midrule
 MASt3R-SLAM & 688.6 & 0.165 & 0.205 & 0.096 & 0.199 & 0.093 & 1.638 & 0.043 & 0.188 & 0.164 & 1.500 & 0.074 & 0.197 & 0.118 & 1.569 & 0.059 \\
 VGGT-SLAM & 1257 & \textbf{0.038} & \textbf{0.049} & \textbf{0.901} & \textbf{0.048} & \textbf{0.884} & \textbf{0.134} & \textbf{0.538} & \textbf{0.053} & \textbf{0.759} & \textbf{0.275} & \textbf{0.437} & \textbf{0.050} & \textbf{0.848} & \textbf{0.204} & \textbf{0.487} \\
\midrule
\multicolumn{17}{c}{\cellcolor{catgray}\textbf{Test-Time Training}} \\
\midrule
 TTT3R & 793.3 & 0.042 & 0.071 & 0.634 & 0.067 & 0.637 & 0.662 & 0.261 & 0.071 & 0.679 & 0.484 & 0.244 & 0.069 & 0.650 & 0.573 & 0.253 \\
 Scal3R & 1266 & 0.065 & 0.043 & 0.891 & 0.097 & 0.843 & \textbf{0.056} & \textbf{0.707} & 0.144 & 0.728 & \textbf{0.092} & \cellcolor{secondorange}\textbf{0.651} & 0.095 & 0.821 & \cellcolor{thirdyellow}\textbf{0.074} & \cellcolor{secondorange}\textbf{0.679} \\
 LoGeR & 1255 & 0.038 & 0.033 & \textbf{0.908} & 0.040 & 0.908 & 0.077 & 0.574 & 0.058 & 0.826 & 0.182 & 0.447 & 0.043 & 0.881 & 0.129 & 0.510 \\
 LoGeR$^*$ & 1255 & \cellcolor{secondorange}\textbf{0.035} & \cellcolor{bestred}\textbf{0.030} & 0.900 & \textbf{0.034} & \textbf{0.923} & 0.073 & 0.632 & \cellcolor{thirdyellow}\textbf{0.040} & \textbf{0.865} & 0.143 & 0.548 & \cellcolor{secondorange}\textbf{0.035} & \textbf{0.896} & 0.108 & 0.590 \\
\bottomrule
\end{tabular}%
}
\end{table*}

\begin{table*}[t]
\centering
\setlength{\fboxsep}{0pt}
\caption{\textbf{Per-Dataset Results on Tanks and Temples}.
Performance across different input regimes: \textit{Single Frame}, \textit{Sparse}, \textit{Medium}, \textit{Dense}, and the \textit{Average}.
The best, second-best, and third-best results in each column are highlighted in
\colorbox{bestred}{deep blue}, \colorbox{secondorange}{medium blue}, and \colorbox{thirdyellow}{light blue}, respectively.
Out-of-memory (OOM) and Timeout (T.O) cells are shaded \colorbox{oomred}{light red};
\textit{Average} values for those rows are wrapped in parentheses and excluded from per-column ranking.
Within each sub-category, the \textbf{bold} value marks the in-group best.
Note that \ours (Ours) is excluded from the per-column rankings.
}
\label{tab:dataset_tanksandtemples}
\resizebox{\textwidth}{!}{%
\renewcommand{\arraystretch}{1.15}
\setlength{\tabcolsep}{3pt}
\begin{tabular}{l c >{\columncolor{subcol}}c >{\columncolor{subcol}}c >{\columncolor{subcol}}c >{\columncolor{subcol}}c >{\columncolor{subcol}}c >{\columncolor{subcol}}c >{\columncolor{subcol}}c >{\columncolor{subcol}}c >{\columncolor{subcol}}c >{\columncolor{subcol}}c >{\columncolor{subcol}}c >{\columncolor{subcol}}c}
\toprule
\multirow{2.5}{*}{\textbf{Method}}
& \multirow{2.5}{*}{\makecell{\textbf{\#Params}\\\textbf{(M)}}}
& \multicolumn{1}{c}{\textbf{Single Frame}}
& \multicolumn{2}{c}{\textbf{Sparse}}
& \multicolumn{3}{c}{\textbf{Medium}}
& \multicolumn{3}{c}{\textbf{Dense}}
& \multicolumn{3}{c}{\textbf{Average}} \\
\cmidrule(lr){3-3} \cmidrule(lr){4-5} \cmidrule(lr){6-8} \cmidrule(lr){9-11} \cmidrule(lr){12-14}
 & &
 AbsRel$\downarrow$ &
 AbsRel$\downarrow$ & AUC@30$\uparrow$ &
 AbsRel$\downarrow$ & AUC@30$\uparrow$ & ATE$\downarrow$ &
 AbsRel$\downarrow$ & AUC@30$\uparrow$ & ATE$\downarrow$ &
 AbsRel$\downarrow$ & AUC@30$\uparrow$ & ATE$\downarrow$ \\
\midrule
\multicolumn{14}{c}{\cellcolor{catgray}\textbf{Optimization-based}} \\
\midrule
 DUSt3R & 571.2 & 0.045 & 0.097 & 0.368 & 0.132 & 0.333 & 11.93 & \cellcolor{oomred}OOM & \cellcolor{oomred}OOM & \cellcolor{oomred}OOM & (0.114) & (0.351) & (11.93) \\
 MASt3R & 688.6 & \textbf{0.043} & \textbf{0.074} & \textbf{0.548} & \textbf{0.113} & \textbf{0.578} & \textbf{7.868} & \cellcolor{oomred}OOM & \cellcolor{oomred}OOM & \cellcolor{oomred}OOM & (0.094) & (0.563) & (7.868) \\
\midrule
\multicolumn{14}{c}{\cellcolor{catgray}\textbf{End-to-End Feed-Forward}} \\
\midrule
 VGGT & 1257 & \cellcolor{secondorange}0.022 & \cellcolor{secondorange}\textbf{0.039} & \cellcolor{bestred}\textbf{0.712} & \cellcolor{thirdyellow}0.038 & \cellcolor{bestred}\textbf{0.780} & \cellcolor{thirdyellow}4.238 & \cellcolor{oomred}OOM & \cellcolor{oomred}OOM & \cellcolor{oomred}OOM & (0.038) & (0.746) & (4.238) \\
 Fast3R & 647.5 & 0.045 & 0.182 & 0.303 & 0.181 & 0.342 & 11.37 & 0.172 & 0.354 & 14.09 & 0.178 & 0.333 & 12.73 \\
 FastVGGT & 1158 & \cellcolor{secondorange}0.022 & 0.043 & 0.507 & 0.039 & 0.748 & 4.854 & \cellcolor{bestred}\textbf{0.027} & \cellcolor{bestred}\textbf{0.832} & \cellcolor{bestred}\textbf{3.614} & \cellcolor{bestred}\textbf{0.036} & \cellcolor{secondorange}0.695 & \cellcolor{bestred}\textbf{4.234} \\
 MUSt3R & 423.4 & 0.054 & 0.070 & \cellcolor{thirdyellow}0.670 & 0.064 & 0.729 & 5.223 & \cellcolor{oomred}T.O & \cellcolor{oomred}T.O & \cellcolor{oomred}T.O & (0.067) & (0.699) & (5.223) \\
 MapAnything & 1228 & 0.027 & 0.054 & 0.604 & 0.045 & 0.696 & 4.929 & \cellcolor{oomred}OOM & \cellcolor{oomred}OOM & \cellcolor{oomred}OOM & (0.050) & (0.650) & (4.929) \\
 OmniVGGT & 1217 & \cellcolor{thirdyellow}0.023 & 0.052 & 0.419 & 0.051 & 0.498 & 11.06 & \cellcolor{oomred}OOM & \cellcolor{oomred}OOM & \cellcolor{oomred}OOM & (0.051) & (0.458) & (11.06) \\
 $\pi^{3}$ & 958.7 & 0.026 & 0.053 & 0.626 & 0.043 & 0.732 & 4.345 & \cellcolor{secondorange}0.030 & \cellcolor{secondorange}0.813 & 7.929 & \cellcolor{thirdyellow}0.042 & \cellcolor{bestred}\textbf{0.724} & 6.137 \\
 $\pi^{3}$-X & 1360 & \cellcolor{thirdyellow}0.023 & 0.048 & 0.479 & \cellcolor{thirdyellow}0.038 & 0.730 & \cellcolor{bestred}\textbf{3.709} & \cellcolor{oomred}OOM & \cellcolor{oomred}OOM & \cellcolor{oomred}OOM & (0.043) & (0.605) & (3.709) \\
 AMB3R & 1563 & \cellcolor{bestred}\textbf{0.021} & \cellcolor{thirdyellow}0.040 & 0.561 & 0.039 & \cellcolor{secondorange}0.779 & 7.555 & \cellcolor{oomred}OOM & \cellcolor{oomred}OOM & \cellcolor{oomred}OOM & (0.039) & (0.670) & (7.555) \\
 DA3-Small & 34.3 & 0.093 & 0.075 & 0.373 & 0.069 & 0.457 & 9.981 & 0.072 & 0.499 & 10.56 & 0.072 & 0.443 & 10.27 \\
 DA3-Base & 135.4 & 0.066 & 0.073 & 0.385 & 0.060 & 0.613 & 8.527 & 0.060 & 0.624 & 11.09 & 0.064 & 0.541 & 9.807 \\
 DA3-Large & 410.9 & 0.034 & 0.076 & 0.468 & 0.063 & 0.713 & 6.316 & \cellcolor{oomred}OOM & \cellcolor{oomred}OOM & \cellcolor{oomred}OOM & (0.069) & (0.590) & (6.316) \\
 DA3-Giant & 1356 & \cellcolor{thirdyellow}0.023 & 0.041 & 0.624 & \cellcolor{bestred}\textbf{0.034} & \cellcolor{thirdyellow}0.751 & 6.912 & \cellcolor{oomred}OOM & \cellcolor{oomred}OOM & \cellcolor{oomred}OOM & (0.037) & (0.687) & (6.912) \\
 DA3-Nested & 1690 & \cellcolor{secondorange}0.022 & 0.043 & 0.625 & 0.042 & 0.688 & 7.066 & \cellcolor{oomred}OOM & \cellcolor{oomred}OOM & \cellcolor{oomred}OOM & (0.043) & (0.657) & (7.066) \\
 WorldMirror & 1263 & 0.031 & 0.061 & 0.389 & 0.055 & 0.682 & 7.159 & \cellcolor{oomred}OOM & \cellcolor{oomred}OOM & \cellcolor{oomred}OOM & (0.058) & (0.535) & (7.159) \\
 VGGT-Omega & 1144 & 0.028 & 0.041 & 0.593 & \cellcolor{secondorange}0.037 & 0.629 & 10.21 & -- & -- & -- & (0.039) & (0.611) & (10.21) \\ \hdashline
 \ours (Ours) & 1304 & 0.030 & 0.040 & 0.444 & 0.035 & 0.813 & 7.456 & \cellcolor{oomred}OOM & \cellcolor{oomred}OOM & \cellcolor{oomred}OOM & (0.035) & (0.628) & (7.456) \\
\midrule
\multicolumn{14}{c}{\cellcolor{catgray}\textbf{Online}} \\
\midrule
 Spann3r$^{224}$ & 658.7 & 0.142 & 0.677 & 0.260 & 0.517 & 0.316 & 11.5 & 0.410 & 0.261 & 13.84 & 0.535 & 0.279 & 12.67 \\
 CUT3R & 793.3 & 0.040 & 0.105 & 0.335 & 0.150 & 0.535 & 9.351 & 0.153 & 0.385 & 10.4 & 0.136 & 0.418 & 9.877 \\
 MonST3R & 571.2 & 0.058 & 0.108 & 0.088 & 0.179 & 0.029 & 10.42 & \cellcolor{oomred}OOM & \cellcolor{oomred}OOM & \cellcolor{oomred}OOM & (0.143) & (0.059) & (10.42) \\
 Point3R & 828 & 0.038 & 0.284 & 0.244 & 0.218 & 0.226 & 13.01 & 0.593 & 0.229 & 14 & 0.365 & 0.233 & 13.51 \\
 Stream3R-S & 1191 & \cellcolor{secondorange}\textbf{0.022} & 0.050 & 0.366 & 0.081 & 0.488 & 10.99 & \cellcolor{oomred}OOM & \cellcolor{oomred}OOM & \cellcolor{oomred}OOM & (0.065) & (0.427) & (10.99) \\
 Stream3R-W & 1191 & \cellcolor{secondorange}\textbf{0.022} & 0.065 & 0.373 & 0.090 & 0.411 & 13.03 & \cellcolor{oomred}OOM & \cellcolor{oomred}OOM & \cellcolor{oomred}OOM & (0.078) & (0.392) & (13.03) \\
 StreamVGGT & 1257 & \cellcolor{thirdyellow}0.023 & 0.100 & 0.378 & 0.085 & 0.510 & 11.51 & 0.085 & 0.590 & 8.610 & 0.090 & 0.493 & 10.06 \\
 Page4D & 1257 & 0.029 & \cellcolor{bestred}\textbf{0.036} & \cellcolor{secondorange}\textbf{0.695} & \cellcolor{thirdyellow}\textbf{0.038} & \textbf{0.730} & \textbf{4.257} & \cellcolor{oomred}OOM & \cellcolor{oomred}OOM & \cellcolor{oomred}OOM & (0.037) & (0.712) & (4.257) \\
 InfiniteVGGT & 1257 & 0.026 & 0.099 & 0.412 & 0.087 & 0.507 & 11.88 & 0.089 & 0.615 & 9.897 & 0.092 & 0.511 & 10.89 \\
 Wint3R & 749.5 & 0.037 & 0.094 & 0.344 & 0.104 & 0.448 & 10.57 & 0.092 & 0.361 & 14.37 & 0.097 & 0.384 & 12.47 \\
 LongStream-B & 1191 & 0.034 & 0.097 & 0.374 & 0.171 & 0.327 & 11.43 & 0.191 & 0.284 & 12.92 & 0.153 & 0.329 & 12.18 \\
 LongStream-S & 1191 & 0.034 & 0.068 & 0.368 & 0.141 & 0.320 & 11.48 & 0.166 & 0.108 & 12 & 0.125 & 0.265 & 11.74 \\
 LingbotMap$^{*}$-W & 1158 & 0.031 & 0.060 & 0.532 & 0.058 & 0.647 & 7.800 & 0.060 & 0.636 & 7.732 & 0.059 & 0.605 & 7.766 \\
 LingbotMap$^{*}$-S & 1158 & 0.031 & 0.060 & 0.532 & 0.057 & 0.667 & 7.531 & \textbf{0.054} & \cellcolor{thirdyellow}\textbf{0.740} & \textbf{7.282} & \textbf{0.057} & \textbf{0.646} & \textbf{7.406} \\
\midrule
\multicolumn{14}{c}{\cellcolor{catgray}\textbf{Chunk-wise}} \\
\midrule
 VGGT-Long & 1257 & \cellcolor{secondorange}\textbf{0.022} & \cellcolor{secondorange}\textbf{0.039} & \cellcolor{bestred}\textbf{0.712} & \cellcolor{thirdyellow}0.038 & \cellcolor{secondorange}\textbf{0.779} & \textbf{4.242} & \cellcolor{thirdyellow}\textbf{0.044} & 0.586 & \cellcolor{secondorange}\textbf{6.455} & \cellcolor{secondorange}\textbf{0.040} & \cellcolor{thirdyellow}\textbf{0.692} & \cellcolor{secondorange}\textbf{5.349} \\
 $\pi^{3}$-Long & 958.7 & 0.026 & 0.053 & 0.626 & 0.047 & 0.732 & 4.331 & 0.134 & \textbf{0.698} & 6.858 & 0.078 & 0.685 & \cellcolor{thirdyellow}5.594 \\
 DA3-Streaming & 1356 & \cellcolor{thirdyellow}0.023 & 0.041 & 0.624 & \cellcolor{bestred}\textbf{0.034} & 0.750 & 6.913 & 2.078 & 0.537 & 9.225 & 0.718 & 0.637 & 8.069 \\
\midrule
\multicolumn{14}{c}{\cellcolor{catgray}\textbf{SLAM-based}} \\
\midrule
 MASt3R-SLAM & 688.6 & 0.074 & 0.173 & 0.052 & 0.178 & 0.049 & 12.21 & 0.196 & 0.172 & 15.25 & 0.182 & 0.091 & 13.73 \\
 VGGT-SLAM & 1257 & \cellcolor{secondorange}\textbf{0.022} & \cellcolor{secondorange}\textbf{0.039} & \cellcolor{bestred}\textbf{0.712} & \textbf{0.051} & \textbf{0.698} & \cellcolor{secondorange}\textbf{4.195} & \textbf{0.139} & \textbf{0.491} & \textbf{8.844} & \textbf{0.076} & \textbf{0.633} & \textbf{6.520} \\
\midrule
\multicolumn{14}{c}{\cellcolor{catgray}\textbf{Test-Time Training}} \\
\midrule
 TTT3R & 793.3 & 0.040 & 0.109 & 0.373 & 0.119 & 0.413 & 11.95 & 0.180 & \textbf{0.567} & 11.93 & 0.136 & 0.451 & 11.94 \\
 Scal3R & 1266 & \textbf{0.027} & \textbf{0.041} & \textbf{0.531} & \textbf{0.056} & 0.629 & \textbf{4.730} & 0.234 & 0.504 & \cellcolor{thirdyellow}\textbf{6.711} & 0.110 & 0.555 & \textbf{5.720} \\
 LoGeR & 1255 & 0.031 & 0.044 & 0.413 & 0.066 & 0.650 & 8.799 & 0.135 & 0.478 & 10.81 & 0.082 & 0.513 & 9.803 \\
 LoGeR$^*$ & 1255 & 0.034 & 0.053 & 0.482 & 0.061 & \textbf{0.671} & 8.614 & \textbf{0.087} & 0.536 & 9.533 & \textbf{0.067} & \textbf{0.563} & 9.073 \\
\bottomrule
\end{tabular}%
}
\end{table*}

\begin{table*}[t]
\centering
\setlength{\fboxsep}{0pt}
\caption{\textbf{Per-Dataset Results on TUM}.
Performance across different input regimes: \textit{Single Frame}, \textit{Sparse}, \textit{Medium}, \textit{Dense}, and the \textit{Average}.
The best, second-best, and third-best results in each column are highlighted in
\colorbox{bestred}{deep blue}, \colorbox{secondorange}{medium blue}, and \colorbox{thirdyellow}{light blue}, respectively.
Out-of-memory (OOM) and Timeout (T.O) cells are shaded \colorbox{oomred}{light red};
\textit{Average} values for those rows are wrapped in parentheses and excluded from per-column ranking.
Within each sub-category, the \textbf{bold} value marks the in-group best.
Note that \ours (Ours) is excluded from the per-column rankings.
}
\label{tab:dataset_tum}
\resizebox{\textwidth}{!}{%
\renewcommand{\arraystretch}{1.15}
\setlength{\tabcolsep}{3pt}
\begin{tabular}{l c >{\columncolor{subcol}}c >{\columncolor{subcol}}c >{\columncolor{subcol}}c >{\columncolor{subcol}}c >{\columncolor{subcol}}c >{\columncolor{subcol}}c >{\columncolor{subcol}}c >{\columncolor{subcol}}c >{\columncolor{subcol}}c >{\columncolor{subcol}}c >{\columncolor{subcol}}c >{\columncolor{subcol}}c}
\toprule
\multirow{2.5}{*}{\textbf{Method}}
& \multirow{2.5}{*}{\makecell{\textbf{\#Params}\\\textbf{(M)}}}
& \multicolumn{1}{c}{\textbf{Single Frame}}
& \multicolumn{2}{c}{\textbf{Sparse}}
& \multicolumn{3}{c}{\textbf{Medium}}
& \multicolumn{3}{c}{\textbf{Dense}}
& \multicolumn{3}{c}{\textbf{Average}} \\
\cmidrule(lr){3-3} \cmidrule(lr){4-5} \cmidrule(lr){6-8} \cmidrule(lr){9-11} \cmidrule(lr){12-14}
 & &
 AbsRel$\downarrow$ &
 AbsRel$\downarrow$ & AUC@30$\uparrow$ &
 AbsRel$\downarrow$ & AUC@30$\uparrow$ & ATE$\downarrow$ &
 AbsRel$\downarrow$ & AUC@30$\uparrow$ & ATE$\downarrow$ &
 AbsRel$\downarrow$ & AUC@30$\uparrow$ & ATE$\downarrow$ \\
\midrule
\multicolumn{14}{c}{\cellcolor{catgray}\textbf{Optimization-based}} \\
\midrule
 DUSt3R & 571.2 & 0.235 & \textbf{0.123} & 0.418 & 0.237 & 0.251 & 0.208 & \cellcolor{oomred}OOM & \cellcolor{oomred}OOM & \cellcolor{oomred}OOM & (0.180) & (0.335) & (0.208) \\
 MASt3R & 688.6 & \textbf{0.223} & 0.125 & \textbf{0.472} & \textbf{0.201} & \textbf{0.376} & \textbf{0.168} & \cellcolor{oomred}OOM & \cellcolor{oomred}OOM & \cellcolor{oomred}OOM & (0.163) & (0.424) & (0.168) \\
\midrule
\multicolumn{14}{c}{\cellcolor{catgray}\textbf{End-to-End Feed-Forward}} \\
\midrule
 VGGT & 1257 & 0.120 & 0.070 & 0.728 & 0.075 & 0.836 & 0.017 & \cellcolor{oomred}OOM & \cellcolor{oomred}OOM & \cellcolor{oomred}OOM & (0.072) & (0.782) & (0.017) \\
 Fast3R & 647.5 & 0.350 & 0.258 & 0.323 & 0.237 & 0.507 & 0.176 & 0.258 & 0.097 & 0.127 & 0.251 & 0.309 & 0.152 \\
 FastVGGT & 1158 & 0.119 & 0.075 & 0.703 & 0.074 & 0.812 & 0.023 & \cellcolor{secondorange}0.070 & \cellcolor{bestred}\textbf{0.821} & \cellcolor{bestred}\textbf{0.022} & 0.073 & \cellcolor{bestred}\textbf{0.778} & \cellcolor{bestred}\textbf{0.022} \\
 MUSt3R & 423.4 & 0.237 & 0.124 & 0.627 & 0.194 & 0.714 & 0.032 & \cellcolor{oomred}T.O & \cellcolor{oomred}T.O & \cellcolor{oomred}T.O & (0.159) & (0.670) & (0.032) \\
 MapAnything & 1228 & 0.175 & 0.089 & 0.542 & 0.092 & 0.656 & 0.062 & \cellcolor{oomred}OOM & \cellcolor{oomred}OOM & \cellcolor{oomred}OOM & (0.090) & (0.599) & (0.062) \\
 OmniVGGT & 1217 & 0.140 & 0.063 & 0.727 & 0.072 & 0.786 & 0.057 & \cellcolor{oomred}OOM & \cellcolor{oomred}OOM & \cellcolor{oomred}OOM & (0.068) & (0.756) & (0.057) \\
 $\pi^{3}$ & 958.7 & 0.089 & \cellcolor{thirdyellow}\textbf{0.046} & 0.700 & \cellcolor{secondorange}\textbf{0.045} & 0.806 & 0.023 & \cellcolor{bestred}\textbf{0.057} & 0.161 & 0.454 & \cellcolor{bestred}\textbf{0.049} & 0.556 & 0.238 \\
 $\pi^{3}$-X & 1360 & \cellcolor{thirdyellow}\textbf{0.084} & \cellcolor{thirdyellow}0.046 & 0.730 & \cellcolor{secondorange}0.045 & 0.833 & 0.018 & \cellcolor{oomred}OOM & \cellcolor{oomred}OOM & \cellcolor{oomred}OOM & (0.046) & (0.781) & (0.018) \\
 AMB3R & 1563 & 0.087 & 0.056 & 0.690 & 0.065 & 0.734 & 0.033 & \cellcolor{oomred}OOM & \cellcolor{oomred}OOM & \cellcolor{oomred}OOM & (0.061) & (0.712) & (0.033) \\
 DA3-Small & 34.3 & 0.180 & 0.134 & 0.546 & 0.116 & 0.632 & 0.092 & 0.105 & 0.616 & 0.151 & 0.118 & 0.598 & 0.121 \\
 DA3-Base & 135.4 & 0.181 & 0.112 & 0.561 & 0.109 & 0.703 & 0.075 & 0.094 & \cellcolor{secondorange}0.697 & 0.106 & 0.105 & 0.654 & 0.091 \\
 DA3-Large & 410.9 & 0.160 & 0.089 & 0.643 & 0.081 & 0.811 & 0.038 & \cellcolor{oomred}OOM & \cellcolor{oomred}OOM & \cellcolor{oomred}OOM & (0.085) & (0.727) & (0.038) \\
 DA3-Giant & 1356 & 0.189 & 0.088 & \cellcolor{thirdyellow}0.747 & 0.079 & \cellcolor{secondorange}\textbf{0.865} & \cellcolor{secondorange}\textbf{0.014} & \cellcolor{oomred}OOM & \cellcolor{oomred}OOM & \cellcolor{oomred}OOM & (0.084) & (0.806) & (0.014) \\
 DA3-Nested & 1690 & 0.180 & 0.088 & \cellcolor{secondorange}\textbf{0.769} & 0.080 & \cellcolor{thirdyellow}0.854 & \cellcolor{thirdyellow}0.015 & \cellcolor{oomred}OOM & \cellcolor{oomred}OOM & \cellcolor{oomred}OOM & (0.084) & (0.812) & (0.015) \\
 WorldMirror & 1263 & 0.121 & 0.082 & 0.694 & 0.090 & 0.795 & 0.023 & \cellcolor{oomred}OOM & \cellcolor{oomred}OOM & \cellcolor{oomred}OOM & (0.086) & (0.744) & (0.023) \\
 VGGT-Omega & 1144 & \cellcolor{bestred}\textbf{0.066} & \cellcolor{bestred}\textbf{0.042} & \cellcolor{bestred}\textbf{0.821} & \cellcolor{bestred}\textbf{0.040} & \cellcolor{bestred}\textbf{0.899} & \cellcolor{bestred}\textbf{0.013} & -- & -- & -- & (0.041) & (0.860) & (0.013) \\ \hdashline
 \ours (Ours) & 1304 & 0.151 & 0.057 & 0.697 & 0.048 & 0.813 & 0.016 & \cellcolor{oomred}OOM & \cellcolor{oomred}OOM & \cellcolor{oomred}OOM & (0.085) & (0.755) & (0.016) \\
\midrule
\multicolumn{14}{c}{\cellcolor{catgray}\textbf{Online}} \\
\midrule
 Spann3r$^{224}$ & 658.7 & 0.195 & 0.220 & 0.239 & 0.133 & 0.391 & 0.241 & 0.130 & 0.250 & 0.432 & 0.161 & 0.293 & 0.336 \\
 CUT3R & 793.3 & 0.154 & 0.102 & 0.634 & 0.088 & 0.551 & 0.088 & 0.105 & 0.115 & 0.478 & 0.099 & 0.433 & 0.283 \\
 MonST3R & 571.2 & 0.237 & 0.127 & 0.342 & 0.235 & 0.141 & 0.303 & \cellcolor{oomred}OOM & \cellcolor{oomred}OOM & \cellcolor{oomred}OOM & (0.181) & (0.241) & (0.303) \\
 Point3R & 828 & 0.153 & 0.134 & 0.410 & 0.110 & 0.421 & 0.145 & 0.114 & 0.218 & 0.292 & 0.119 & 0.350 & 0.218 \\
 Stream3R-S & 1191 & 0.106 & 0.069 & 0.591 & 0.262 & 0.208 & 0.519 & \cellcolor{oomred}OOM & \cellcolor{oomred}OOM & \cellcolor{oomred}OOM & (0.166) & (0.400) & (0.519) \\
 Stream3R-W & 1191 & 0.106 & 0.069 & 0.591 & 0.221 & 0.221 & 0.553 & \cellcolor{oomred}OOM & \cellcolor{oomred}OOM & \cellcolor{oomred}OOM & (0.145) & (0.406) & (0.553) \\
 StreamVGGT & 1257 & 0.106 & 0.087 & 0.621 & 0.094 & 0.699 & 0.052 & 0.108 & 0.553 & 0.115 & 0.097 & 0.624 & 0.084 \\
 Page4D & 1257 & \textbf{0.098} & \textbf{0.053} & 0.634 & \cellcolor{secondorange}\textbf{0.045} & \textbf{0.726} & \textbf{0.033} & \cellcolor{oomred}OOM & \cellcolor{oomred}OOM & \cellcolor{oomred}OOM & (0.049) & (0.680) & (0.033) \\
 InfiniteVGGT & 1257 & 0.106 & 0.087 & 0.601 & 0.093 & 0.701 & 0.052 & 0.108 & 0.554 & 0.117 & 0.096 & 0.619 & 0.084 \\
 Wint3R & 749.5 & 0.147 & 0.102 & 0.588 & 0.108 & 0.489 & 0.165 & 0.126 & 0.163 & 0.301 & 0.112 & 0.413 & 0.233 \\
 LongStream-B & 1191 & 0.108 & 0.064 & 0.573 & 0.145 & 0.576 & 0.104 & 0.183 & 0.188 & 0.262 & 0.131 & 0.446 & 0.183 \\
 LongStream-S & 1191 & 0.108 & 0.064 & 0.573 & 0.108 & 0.448 & 0.179 & 0.181 & 0.146 & 0.271 & 0.118 & 0.389 & 0.225 \\
 LingbotMap$^{*}$-W & 1158 & 0.167 & 0.066 & \textbf{0.655} & 0.116 & 0.671 & 0.048 & 0.207 & 0.378 & 0.265 & 0.130 & 0.568 & 0.157 \\
 LingbotMap$^{*}$-S & 1158 & 0.167 & 0.066 & \textbf{0.655} & 0.104 & 0.677 & 0.048 & \textbf{0.095} & \cellcolor{thirdyellow}\textbf{0.658} & \cellcolor{secondorange}\textbf{0.051} & \textbf{0.088} & \textbf{0.663} & \cellcolor{thirdyellow}\textbf{0.049} \\
\midrule
\multicolumn{14}{c}{\cellcolor{catgray}\textbf{Chunk-wise}} \\
\midrule
 VGGT-Long & 1257 & 0.120 & 0.070 & 0.728 & 0.080 & 0.757 & 0.036 & 0.227 & 0.266 & 0.311 & 0.126 & 0.584 & 0.174 \\
 $\pi^{3}$-Long & 958.7 & \textbf{0.089} & \cellcolor{thirdyellow}\textbf{0.046} & 0.700 & \textbf{0.057} & 0.784 & 0.025 & 0.172 & \textbf{0.441} & 0.146 & \textbf{0.091} & 0.642 & 0.086 \\
 DA3-Streaming & 1356 & 0.189 & 0.088 & \cellcolor{thirdyellow}\textbf{0.747} & 0.081 & \textbf{0.822} & \textbf{0.019} & \textbf{0.107} & 0.429 & \textbf{0.144} & 0.092 & \cellcolor{thirdyellow}\textbf{0.666} & \textbf{0.081} \\
\midrule
\multicolumn{14}{c}{\cellcolor{catgray}\textbf{SLAM-based}} \\
\midrule
 MASt3R-SLAM & 688.6 & 0.230 & 0.232 & 0.129 & 0.337 & 0.306 & 0.361 & 0.323 & \textbf{0.426} & \textbf{0.125} & 0.297 & 0.287 & 0.243 \\
 VGGT-SLAM & 1257 & \textbf{0.120} & \textbf{0.070} & \textbf{0.728} & \textbf{0.085} & \textbf{0.653} & \textbf{0.056} & \textbf{0.303} & 0.196 & 0.352 & \textbf{0.152} & \textbf{0.526} & \textbf{0.204} \\
\midrule
\multicolumn{14}{c}{\cellcolor{catgray}\textbf{Test-Time Training}} \\
\midrule
 TTT3R & 793.3 & 0.154 & 0.097 & 0.613 & 0.084 & 0.651 & 0.058 & 0.090 & 0.257 & 0.325 & 0.090 & 0.507 & 0.192 \\
 Scal3R & 1266 & 0.186 & 0.070 & 0.668 & 0.126 & 0.680 & 0.054 & 0.217 & 0.197 & 0.260 & 0.138 & 0.515 & 0.157 \\
 LoGeR & 1255 & 0.085 & \cellcolor{secondorange}\textbf{0.045} & \textbf{0.746} & 0.068 & 0.690 & 0.064 & 0.103 & 0.411 & 0.142 & \cellcolor{thirdyellow}0.072 & 0.616 & 0.103 \\
 LoGeR$^*$ & 1255 & \cellcolor{secondorange}\textbf{0.079} & \cellcolor{secondorange}0.045 & 0.717 & \cellcolor{thirdyellow}\textbf{0.053} & \textbf{0.761} & \textbf{0.035} & \cellcolor{thirdyellow}\textbf{0.074} & \textbf{0.657} & \cellcolor{thirdyellow}\textbf{0.061} & \cellcolor{secondorange}\textbf{0.057} & \cellcolor{secondorange}\textbf{0.712} & \cellcolor{secondorange}\textbf{0.048} \\
\bottomrule
\end{tabular}%
}
\end{table*}

\begin{table*}[t]
\centering
\setlength{\fboxsep}{0pt}
\caption{\textbf{Per-Dataset Results on Vkitti}.
Performance across different input regimes: \textit{Single Frame}, \textit{Sparse}, \textit{Medium}, \textit{Dense}, and the \textit{Average}.
The best, second-best, and third-best results in each column are highlighted in
\colorbox{bestred}{deep blue}, \colorbox{secondorange}{medium blue}, and \colorbox{thirdyellow}{light blue}, respectively.
Out-of-memory (OOM) and Timeout (T.O) cells are shaded \colorbox{oomred}{light red};
\textit{Average} values for those rows are wrapped in parentheses and excluded from per-column ranking.
Within each sub-category, the \textbf{bold} value marks the in-group best.
Note that \ours (Ours) is excluded from the per-column rankings.
}
\label{tab:dataset_vkitti}
\resizebox{\textwidth}{!}{%
\renewcommand{\arraystretch}{1.15}
\setlength{\tabcolsep}{3pt}
\begin{tabular}{l c >{\columncolor{subcol}}c >{\columncolor{subcol}}c >{\columncolor{subcol}}c >{\columncolor{subcol}}c >{\columncolor{subcol}}c >{\columncolor{subcol}}c >{\columncolor{subcol}}c >{\columncolor{subcol}}c >{\columncolor{subcol}}c >{\columncolor{subcol}}c >{\columncolor{subcol}}c >{\columncolor{subcol}}c}
\toprule
\multirow{2.5}{*}{\textbf{Method}}
& \multirow{2.5}{*}{\makecell{\textbf{\#Params}\\\textbf{(M)}}}
& \multicolumn{1}{c}{\textbf{Single Frame}}
& \multicolumn{2}{c}{\textbf{Sparse}}
& \multicolumn{3}{c}{\textbf{Medium}}
& \multicolumn{3}{c}{\textbf{Dense}}
& \multicolumn{3}{c}{\textbf{Average}} \\
\cmidrule(lr){3-3} \cmidrule(lr){4-5} \cmidrule(lr){6-8} \cmidrule(lr){9-11} \cmidrule(lr){12-14}
 & &
 AbsRel$\downarrow$ &
 AbsRel$\downarrow$ & AUC@30$\uparrow$ &
 AbsRel$\downarrow$ & AUC@30$\uparrow$ & ATE$\downarrow$ &
 AbsRel$\downarrow$ & AUC@30$\uparrow$ & ATE$\downarrow$ &
 AbsRel$\downarrow$ & AUC@30$\uparrow$ & ATE$\downarrow$ \\
\midrule
\multicolumn{14}{c}{\cellcolor{catgray}\textbf{Optimization-based}} \\
\midrule
 DUSt3R & 571.2 & 0.215 & 0.286 & \textbf{0.668} & \textbf{0.326} & \textbf{0.702} & \textbf{16.67} & \cellcolor{oomred}OOM & \cellcolor{oomred}OOM & \cellcolor{oomred}OOM & (0.306) & (0.685) & (16.67) \\
 MASt3R & 688.6 & \textbf{0.114} & \textbf{0.210} & 0.386 & 0.504 & 0.677 & 21.24 & \cellcolor{oomred}OOM & \cellcolor{oomred}OOM & \cellcolor{oomred}OOM & (0.357) & (0.532) & (21.24) \\
\midrule
\multicolumn{14}{c}{\cellcolor{catgray}\textbf{End-to-End Feed-Forward}} \\
\midrule
 VGGT & 1257 & \cellcolor{secondorange}\textbf{0.037} & \cellcolor{secondorange}\textbf{0.055} & \cellcolor{thirdyellow}0.922 & \cellcolor{secondorange}\textbf{0.046} & 0.921 & 5.778 & \cellcolor{oomred}OOM & \cellcolor{oomred}OOM & \cellcolor{oomred}OOM & (0.051) & (0.922) & (5.778) \\
 Fast3R & 647.5 & 0.228 & 0.424 & 0.115 & 0.303 & 0.135 & 72.45 & 0.324 & 0.146 & 75.05 & 0.350 & 0.132 & 73.75 \\
 FastVGGT & 1158 & \cellcolor{thirdyellow}0.038 & 0.062 & 0.918 & \cellcolor{thirdyellow}0.048 & 0.918 & 4.304 & \cellcolor{thirdyellow}\textbf{0.047} & 0.784 & 6.673 & \cellcolor{thirdyellow}\textbf{0.052} & \textbf{0.873} & 5.488 \\
 MUSt3R & 423.4 & 0.117 & 0.108 & 0.634 & 0.094 & 0.583 & 47.54 & \cellcolor{oomred}T.O & \cellcolor{oomred}T.O & \cellcolor{oomred}T.O & (0.101) & (0.608) & (47.54) \\
 MapAnything & 1228 & 0.120 & 0.113 & 0.690 & 0.106 & 0.681 & 28.69 & \cellcolor{oomred}OOM & \cellcolor{oomred}OOM & \cellcolor{oomred}OOM & (0.109) & (0.685) & (28.69) \\
 OmniVGGT & 1217 & 0.045 & \cellcolor{thirdyellow}0.058 & 0.816 & 0.070 & 0.795 & 12.71 & \cellcolor{oomred}OOM & \cellcolor{oomred}OOM & \cellcolor{oomred}OOM & (0.064) & (0.806) & (12.71) \\
 $\pi^{3}$ & 958.7 & 0.112 & 0.082 & 0.843 & 0.076 & 0.898 & 5.051 & 0.074 & \textbf{0.804} & \textbf{5.849} & 0.077 & 0.848 & \textbf{5.450} \\
 $\pi^{3}$-X & 1360 & 0.062 & 0.064 & 0.882 & 0.056 & 0.903 & \textbf{1.783} & \cellcolor{oomred}OOM & \cellcolor{oomred}OOM & \cellcolor{oomred}OOM & (0.060) & (0.892) & (1.783) \\
 AMB3R & 1563 & 0.045 & 0.061 & \cellcolor{bestred}\textbf{0.950} & 0.050 & \textbf{0.921} & 4.534 & \cellcolor{oomred}OOM & \cellcolor{oomred}OOM & \cellcolor{oomred}OOM & (0.056) & (0.936) & (4.534) \\
 DA3-Small & 34.3 & 0.119 & 0.122 & 0.448 & 0.120 & 0.458 & 55.2 & 0.123 & 0.397 & 59.85 & 0.122 & 0.434 & 57.52 \\
 DA3-Base & 135.4 & 0.142 & 0.105 & 0.550 & 0.105 & 0.485 & 45.83 & 0.103 & 0.440 & 56.79 & 0.104 & 0.492 & 51.31 \\
 DA3-Large & 410.9 & 0.085 & 0.079 & 0.655 & 0.075 & 0.649 & 44.85 & \cellcolor{oomred}OOM & \cellcolor{oomred}OOM & \cellcolor{oomred}OOM & (0.077) & (0.652) & (44.85) \\
 DA3-Giant & 1356 & 0.072 & 0.064 & 0.787 & 0.061 & 0.749 & 20.83 & \cellcolor{oomred}OOM & \cellcolor{oomred}OOM & \cellcolor{oomred}OOM & (0.063) & (0.768) & (20.83) \\
 DA3-Nested & 1690 & 0.071 & 0.072 & 0.767 & 0.072 & 0.655 & 37.2 & \cellcolor{oomred}OOM & \cellcolor{oomred}OOM & \cellcolor{oomred}OOM & (0.072) & (0.711) & (37.2) \\
 WorldMirror & 1263 & 0.081 & 0.104 & 0.734 & 0.099 & 0.773 & 13.65 & \cellcolor{oomred}OOM & \cellcolor{oomred}OOM & \cellcolor{oomred}OOM & (0.101) & (0.754) & (13.65) \\
 VGGT-Omega & 1144 & 0.110 & 0.100 & 0.793 & 0.094 & 0.787 & 4.123 & -- & -- & -- & (0.097) & (0.790) & (4.123) \\ \hdashline
 \ours (Ours) & 1304 & 0.083 & 0.073 & 0.742 & 0.073 & 0.691 & 26.178 & \cellcolor{oomred}OOM & \cellcolor{oomred}OOM & \cellcolor{oomred}OOM & (0.077) & (0.717) & (26.178) \\
\midrule
\multicolumn{14}{c}{\cellcolor{catgray}\textbf{Online}} \\
\midrule
 Spann3r$^{224}$ & 658.7 & 0.327 & 0.453 & 0.366 & 0.443 & 0.346 & 38.77 & 0.497 & 0.278 & 33.67 & 0.464 & 0.330 & 36.22 \\
 CUT3R & 793.3 & 0.070 & 0.075 & 0.650 & 0.069 & 0.588 & 41.24 & 0.063 & 0.349 & 58.95 & 0.069 & 0.529 & 50.1 \\
 MonST3R & 571.2 & 0.254 & 0.216 & 0.485 & 0.391 & 0.478 & 17.98 & \cellcolor{oomred}OOM & \cellcolor{oomred}OOM & \cellcolor{oomred}OOM & (0.304) & (0.482) & (17.98) \\
 Point3R & 828 & 0.064 & 0.073 & 0.348 & 0.067 & 0.244 & 70.09 & 0.062 & 0.249 & 60.33 & 0.067 & 0.280 & 65.21 \\
 Stream3R-S & 1191 & 0.053 & 0.092 & 0.708 & 0.128 & 0.584 & 51.36 & \cellcolor{oomred}OOM & \cellcolor{oomred}OOM & \cellcolor{oomred}OOM & (0.110) & (0.646) & (51.36) \\
 Stream3R-W & 1191 & 0.053 & 0.126 & 0.666 & 0.133 & 0.479 & 63.01 & \cellcolor{oomred}OOM & \cellcolor{oomred}OOM & \cellcolor{oomred}OOM & (0.130) & (0.572) & (63.01) \\
 StreamVGGT & 1257 & 0.054 & 0.322 & 0.808 & 0.280 & 0.602 & 58.14 & 0.227 & 0.417 & 65.29 & 0.276 & 0.609 & 61.71 \\
 Page4D & 1257 & \textbf{0.043} & \textbf{0.061} & 0.823 & \textbf{0.053} & 0.837 & 7.484 & \cellcolor{oomred}OOM & \cellcolor{oomred}OOM & \cellcolor{oomred}OOM & (0.057) & (0.830) & (7.484) \\
 InfiniteVGGT & 1257 & 0.054 & 0.322 & 0.809 & 0.281 & 0.588 & 59.06 & 0.223 & 0.415 & 65.02 & 0.275 & 0.604 & 62.04 \\
 Wint3R & 749.5 & 0.126 & 0.142 & 0.688 & 0.110 & 0.505 & 48.24 & 0.146 & 0.317 & 74.98 & 0.133 & 0.503 & 61.61 \\
 LongStream-B & 1191 & 0.055 & 0.073 & 0.822 & 0.069 & 0.875 & 2.261 & 0.069 & 0.585 & 2.175 & 0.070 & 0.760 & 2.218 \\
 LongStream-S & 1191 & 0.055 & 0.073 & 0.787 & 0.074 & 0.805 & 3.625 & 0.072 & 0.557 & 2.488 & 0.073 & 0.716 & 3.057 \\
 LingbotMap$^{*}$-W & 1158 & 0.075 & 0.079 & \textbf{0.884} & 0.060 & \textbf{0.927} & \cellcolor{thirdyellow}\textbf{1.310} & 0.063 & 0.815 & 1.637 & 0.067 & 0.875 & 1.473 \\
 LingbotMap$^{*}$-S & 1158 & 0.075 & 0.079 & \textbf{0.884} & 0.059 & 0.923 & 1.464 & \textbf{0.055} & \cellcolor{thirdyellow}\textbf{0.827} & \textbf{1.396} & \textbf{0.064} & \textbf{0.878} & \cellcolor{thirdyellow}\textbf{1.430} \\
\midrule
\multicolumn{14}{c}{\cellcolor{catgray}\textbf{Chunk-wise}} \\
\midrule
 VGGT-Long & 1257 & \cellcolor{secondorange}\textbf{0.037} & \cellcolor{secondorange}\textbf{0.055} & \cellcolor{thirdyellow}\textbf{0.922} & \cellcolor{bestred}\textbf{0.041} & \cellcolor{bestred}\textbf{0.959} & \cellcolor{secondorange}\textbf{0.719} & \cellcolor{bestred}\textbf{0.044} & \cellcolor{bestred}\textbf{0.853} & \cellcolor{thirdyellow}\textbf{1.324} & \cellcolor{bestred}\textbf{0.047} & \cellcolor{bestred}\textbf{0.911} & \cellcolor{secondorange}\textbf{1.021} \\
 $\pi^{3}$-Long & 958.7 & 0.112 & 0.082 & 0.843 & 0.095 & \cellcolor{thirdyellow}0.930 & 2.012 & 0.159 & 0.722 & 6.218 & 0.112 & 0.831 & 4.115 \\
 DA3-Streaming & 1356 & 0.072 & 0.064 & 0.787 & 0.116 & 0.897 & 5.066 & 0.052 & 0.779 & 1.386 & 0.077 & 0.821 & 3.226 \\
\midrule
\multicolumn{14}{c}{\cellcolor{catgray}\textbf{SLAM-based}} \\
\midrule
 MASt3R-SLAM & 688.6 & 0.253 & 0.459 & 0.165 & 0.492 & 0.152 & 77.07 & 0.566 & 0.139 & 68.34 & 0.506 & 0.152 & 72.71 \\
 VGGT-SLAM & 1257 & \cellcolor{secondorange}\textbf{0.037} & \cellcolor{secondorange}\textbf{0.055} & \cellcolor{thirdyellow}\textbf{0.922} & \textbf{0.054} & \cellcolor{secondorange}\textbf{0.956} & \textbf{2.172} & \cellcolor{secondorange}\textbf{0.045} & \cellcolor{secondorange}\textbf{0.847} & \cellcolor{secondorange}\textbf{1.076} & \cellcolor{secondorange}\textbf{0.051} & \cellcolor{secondorange}\textbf{0.909} & \textbf{1.624} \\
\midrule
\multicolumn{14}{c}{\cellcolor{catgray}\textbf{Test-Time Training}} \\
\midrule
 TTT3R & 793.3 & 0.070 & 0.073 & 0.541 & 0.069 & 0.655 & 32.52 & 0.064 & 0.405 & 33 & 0.068 & 0.534 & 32.76 \\
 Scal3R & 1266 & \cellcolor{bestred}\textbf{0.036} & \cellcolor{bestred}\textbf{0.048} & \cellcolor{secondorange}\textbf{0.928} & 0.087 & \textbf{0.905} & \cellcolor{bestred}\textbf{0.437} & 0.089 & \textbf{0.804} & \cellcolor{bestred}\textbf{0.650} & 0.075 & \cellcolor{thirdyellow}\textbf{0.879} & \cellcolor{bestred}\textbf{0.543} \\
 LoGeR & 1255 & 0.054 & 0.065 & 0.751 & 0.061 & 0.819 & 3.494 & 0.058 & 0.707 & 2.756 & 0.061 & 0.759 & 3.125 \\
 LoGeR$^*$ & 1255 & 0.054 & 0.068 & 0.715 & \textbf{0.056} & 0.846 & 2.877 & \textbf{0.052} & 0.771 & 2.949 & \textbf{0.059} & 0.778 & 2.913 \\
\bottomrule
\end{tabular}%
}
\end{table*}

\begin{table*}[t]
\centering
\setlength{\fboxsep}{0pt}
\caption{\textbf{Per-Dataset Results on Waymo}.
Performance across different input regimes: \textit{Single Frame}, \textit{Sparse}, \textit{Medium}, \textit{Dense}, and the \textit{Average}.
The best, second-best, and third-best results in each column are highlighted in
\colorbox{bestred}{deep blue}, \colorbox{secondorange}{medium blue}, and \colorbox{thirdyellow}{light blue}, respectively.
Out-of-memory (OOM) and Timeout (T.O) cells are shaded \colorbox{oomred}{light red};
\textit{Average} values for those rows are wrapped in parentheses and excluded from per-column ranking.
Within each sub-category, the \textbf{bold} value marks the in-group best.
Note that \ours (Ours) is excluded from the per-column rankings.
}
\label{tab:dataset_waymo}
\resizebox{\textwidth}{!}{%
\renewcommand{\arraystretch}{1.15}
\setlength{\tabcolsep}{3pt}
\begin{tabular}{l c >{\columncolor{subcol}}c >{\columncolor{subcol}}c >{\columncolor{subcol}}c >{\columncolor{subcol}}c >{\columncolor{subcol}}c >{\columncolor{subcol}}c >{\columncolor{subcol}}c >{\columncolor{subcol}}c >{\columncolor{subcol}}c >{\columncolor{subcol}}c >{\columncolor{subcol}}c >{\columncolor{subcol}}c}
\toprule
\multirow{2.5}{*}{\textbf{Method}}
& \multirow{2.5}{*}{\makecell{\textbf{\#Params}\\\textbf{(M)}}}
& \multicolumn{1}{c}{\textbf{Single Frame}}
& \multicolumn{2}{c}{\textbf{Sparse}}
& \multicolumn{3}{c}{\textbf{Medium}}
& \multicolumn{3}{c}{\textbf{Dense}}
& \multicolumn{3}{c}{\textbf{Average}} \\
\cmidrule(lr){3-3} \cmidrule(lr){4-5} \cmidrule(lr){6-8} \cmidrule(lr){9-11} \cmidrule(lr){12-14}
 & &
 AbsRel$\downarrow$ &
 AbsRel$\downarrow$ & AUC@30$\uparrow$ &
 AbsRel$\downarrow$ & AUC@30$\uparrow$ & ATE$\downarrow$ &
 AbsRel$\downarrow$ & AUC@30$\uparrow$ & ATE$\downarrow$ &
 AbsRel$\downarrow$ & AUC@30$\uparrow$ & ATE$\downarrow$ \\
\midrule
\multicolumn{14}{c}{\cellcolor{catgray}\textbf{Optimization-based}} \\
\midrule
 DUSt3R & 571.2 & 0.125 & 0.270 & 0.725 & \textbf{0.256} & 0.601 & \textbf{5.117} & \cellcolor{oomred}OOM & \cellcolor{oomred}OOM & \cellcolor{oomred}OOM & (0.263) & (0.663) & (5.117) \\
 MASt3R & 688.6 & \textbf{0.119} & \textbf{0.224} & \textbf{0.751} & 0.334 & \textbf{0.682} & 8.506 & \cellcolor{oomred}OOM & \cellcolor{oomred}OOM & \cellcolor{oomred}OOM & (0.279) & (0.716) & (8.506) \\
\midrule
\multicolumn{14}{c}{\cellcolor{catgray}\textbf{End-to-End Feed-Forward}} \\
\midrule
 VGGT & 1257 & \cellcolor{secondorange}\textbf{0.053} & \cellcolor{secondorange}0.048 & \cellcolor{secondorange}0.975 & \cellcolor{bestred}\textbf{0.035} & 0.958 & 0.683 & \cellcolor{oomred}OOM & \cellcolor{oomred}OOM & \cellcolor{oomred}OOM & (0.042) & (0.966) & (0.683) \\
 Fast3R & 647.5 & 0.159 & 0.226 & 0.508 & 0.285 & 0.230 & 43.43 & 0.274 & 0.248 & 47.63 & 0.262 & 0.329 & 45.53 \\
 FastVGGT & 1158 & \cellcolor{secondorange}0.053 & \cellcolor{thirdyellow}0.049 & 0.962 & \cellcolor{secondorange}0.037 & 0.946 & 0.944 & \cellcolor{bestred}\textbf{0.036} & \textbf{0.946} & \textbf{0.776} & \cellcolor{bestred}\textbf{0.041} & \textbf{0.951} & \textbf{0.860} \\
 MUSt3R & 423.4 & 0.125 & 0.128 & 0.792 & 0.147 & 0.648 & 14.2 & \cellcolor{oomred}T.O & \cellcolor{oomred}T.O & \cellcolor{oomred}T.O & (0.138) & (0.720) & (14.2) \\
 MapAnything & 1228 & 0.110 & 0.293 & 0.905 & 0.137 & 0.863 & 3.648 & \cellcolor{oomred}OOM & \cellcolor{oomred}OOM & \cellcolor{oomred}OOM & (0.215) & (0.884) & (3.648) \\
 OmniVGGT & 1217 & 0.055 & 0.054 & 0.951 & 0.044 & 0.923 & 5.228 & \cellcolor{oomred}OOM & \cellcolor{oomred}OOM & \cellcolor{oomred}OOM & (0.049) & (0.937) & (5.228) \\
 $\pi^{3}$ & 958.7 & 0.101 & 0.086 & 0.887 & 0.069 & 0.883 & 1.200 & 0.068 & 0.866 & 1.207 & 0.074 & 0.879 & 1.203 \\
 $\pi^{3}$-X & 1360 & 0.060 & 0.058 & \cellcolor{thirdyellow}0.972 & 0.044 & 0.968 & 0.601 & \cellcolor{oomred}OOM & \cellcolor{oomred}OOM & \cellcolor{oomred}OOM & (0.051) & (0.970) & (0.601) \\
 AMB3R & 1563 & \cellcolor{thirdyellow}0.054 & \cellcolor{bestred}\textbf{0.047} & \cellcolor{secondorange}0.975 & \cellcolor{secondorange}0.037 & 0.956 & 0.666 & \cellcolor{oomred}OOM & \cellcolor{oomred}OOM & \cellcolor{oomred}OOM & (0.042) & (0.965) & (0.666) \\
 DA3-Small & 34.3 & 0.144 & 0.179 & 0.671 & 0.145 & 0.537 & 30.89 & 0.141 & 0.532 & 30.19 & 0.155 & 0.580 & 30.54 \\
 DA3-Base & 135.4 & 0.112 & 0.147 & 0.754 & 0.114 & 0.580 & 23.83 & 0.112 & 0.543 & 22.18 & 0.124 & 0.626 & 23.01 \\
 DA3-Large & 410.9 & 0.115 & 0.164 & 0.862 & 0.117 & 0.851 & 9.856 & \cellcolor{oomred}OOM & \cellcolor{oomred}OOM & \cellcolor{oomred}OOM & (0.140) & (0.856) & (9.856) \\
 DA3-Giant & 1356 & 0.083 & 0.078 & \cellcolor{bestred}\textbf{0.985} & 0.067 & \cellcolor{bestred}\textbf{0.986} & \cellcolor{secondorange}\textbf{0.303} & \cellcolor{oomred}OOM & \cellcolor{oomred}OOM & \cellcolor{oomred}OOM & (0.072) & (0.985) & (0.303) \\
 DA3-Nested & 1690 & 0.099 & 0.125 & 0.905 & 0.094 & 0.903 & 2.907 & \cellcolor{oomred}OOM & \cellcolor{oomred}OOM & \cellcolor{oomred}OOM & (0.110) & (0.904) & (2.907) \\
 WorldMirror & 1263 & 0.068 & 0.104 & 0.902 & 0.075 & 0.875 & 4.761 & \cellcolor{oomred}OOM & \cellcolor{oomred}OOM & \cellcolor{oomred}OOM & (0.090) & (0.888) & (4.761) \\
 VGGT-Omega & 1144 & 0.088 & 0.088 & \cellcolor{thirdyellow}0.972 & 0.073 & 0.949 & 0.975 & -- & -- & -- & (0.081) & (0.960) & (0.975) \\ \hdashline
 \ours (Ours) & 1304 & 0.064 & 0.059 & 0.943 & 0.052 & 0.960 & 0.884 & \cellcolor{oomred}OOM & \cellcolor{oomred}OOM & \cellcolor{oomred}OOM & (0.058) & (0.952) & (0.884) \\
\midrule
\multicolumn{14}{c}{\cellcolor{catgray}\textbf{Online}} \\
\midrule
 Spann3r$^{224}$ & 658.7 & 0.178 & 0.335 & 0.526 & 0.288 & 0.431 & 29.38 & 0.294 & 0.360 & 29.13 & 0.306 & 0.439 & 29.25 \\
 CUT3R & 793.3 & 0.057 & 0.072 & 0.865 & 0.062 & 0.709 & 6.451 & 0.061 & 0.573 & 10.89 & 0.065 & 0.716 & 8.670 \\
 MonST3R & 571.2 & 0.092 & 0.232 & 0.730 & 0.296 & 0.726 & 9.918 & \cellcolor{oomred}OOM & \cellcolor{oomred}OOM & \cellcolor{oomred}OOM & (0.264) & (0.728) & (9.918) \\
 Point3R & 828 & 0.062 & 0.074 & 0.577 & 0.062 & 0.331 & 43.64 & 0.062 & 0.282 & 44.75 & 0.066 & 0.396 & 44.19 \\
 Stream3R-S & 1191 & 0.059 & 0.069 & 0.879 & 0.240 & 0.573 & 42.25 & \cellcolor{oomred}OOM & \cellcolor{oomred}OOM & \cellcolor{oomred}OOM & (0.155) & (0.726) & (42.25) \\
 Stream3R-W & 1191 & 0.059 & 0.071 & 0.850 & 0.205 & 0.514 & 47.43 & \cellcolor{oomred}OOM & \cellcolor{oomred}OOM & \cellcolor{oomred}OOM & (0.138) & (0.682) & (47.43) \\
 StreamVGGT & 1257 & \cellcolor{bestred}0.050 & 0.205 & 0.856 & 0.297 & 0.674 & 28.85 & 0.308 & 0.600 & 29.34 & 0.270 & 0.710 & 29.09 \\
 Page4D & 1257 & \cellcolor{bestred}0.050 & \cellcolor{secondorange}\textbf{0.048} & \textbf{0.954} & \textbf{0.041} & 0.947 & 0.696 & \cellcolor{oomred}OOM & \cellcolor{oomred}OOM & \cellcolor{oomred}OOM & (0.045) & (0.950) & (0.696) \\
 InfiniteVGGT & 1257 & \cellcolor{bestred}\textbf{0.050} & 0.204 & 0.856 & 0.293 & 0.676 & 28.54 & 0.305 & 0.603 & 29.63 & 0.268 & 0.712 & 29.09 \\
 Wint3R & 749.5 & 0.130 & 0.124 & 0.765 & 0.125 & 0.597 & 21.38 & 0.131 & 0.458 & 25.96 & 0.127 & 0.607 & 23.67 \\
 LongStream-B & 1191 & 0.057 & 0.065 & 0.895 & 0.057 & 0.906 & 0.786 & 0.058 & 0.840 & 0.896 & 0.060 & 0.880 & 0.841 \\
 LongStream-S & 1191 & 0.057 & 0.064 & 0.839 & 0.056 & 0.855 & 2.367 & \textbf{0.053} & 0.806 & 1.323 & \textbf{0.058} & 0.833 & 1.845 \\
 LingbotMap$^{*}$-W & 1158 & 0.073 & 0.086 & 0.950 & 0.060 & \textbf{0.967} & \textbf{0.509} & 0.063 & \cellcolor{thirdyellow}\textbf{0.954} & \textbf{0.644} & 0.069 & \cellcolor{thirdyellow}\textbf{0.957} & \textbf{0.576} \\
 LingbotMap$^{*}$-S & 1158 & 0.073 & 0.086 & 0.950 & 0.060 & 0.965 & 0.645 & 0.058 & 0.945 & 0.659 & 0.068 & 0.954 & 0.652 \\
\midrule
\multicolumn{14}{c}{\cellcolor{catgray}\textbf{Chunk-wise}} \\
\midrule
 VGGT-Long & 1257 & \cellcolor{secondorange}\textbf{0.053} & \cellcolor{secondorange}\textbf{0.048} & \cellcolor{secondorange}0.975 & \textbf{0.047} & 0.930 & 0.889 & \textbf{0.059} & 0.878 & 1.408 & \textbf{0.051} & 0.927 & 1.148 \\
 $\pi^{3}$-Long & 958.7 & 0.101 & 0.086 & 0.887 & 0.117 & 0.865 & 1.642 & 0.111 & 0.812 & 2.409 & 0.105 & 0.855 & 2.026 \\
 DA3-Streaming & 1356 & 0.083 & 0.078 & \cellcolor{bestred}\textbf{0.985} & 0.065 & \cellcolor{secondorange}\textbf{0.981} & \cellcolor{bestred}\textbf{0.286} & 0.062 & \cellcolor{bestred}\textbf{0.967} & \cellcolor{bestred}\textbf{0.233} & 0.068 & \cellcolor{bestred}\textbf{0.977} & \cellcolor{bestred}\textbf{0.260} \\
\midrule
\multicolumn{14}{c}{\cellcolor{catgray}\textbf{SLAM-based}} \\
\midrule
 MASt3R-SLAM & 688.6 & 0.216 & 0.494 & 0.478 & 0.477 & 0.392 & 32.29 & 0.468 & 0.422 & 23.64 & 0.480 & 0.430 & 27.96 \\
 VGGT-SLAM & 1257 & \cellcolor{secondorange}\textbf{0.053} & \cellcolor{secondorange}\textbf{0.048} & \cellcolor{secondorange}\textbf{0.975} & \textbf{0.061} & \textbf{0.841} & \textbf{1.431} & \textbf{0.095} & \textbf{0.797} & \textbf{2.519} & \textbf{0.068} & \textbf{0.871} & \textbf{1.975} \\
\midrule
\multicolumn{14}{c}{\cellcolor{catgray}\textbf{Test-Time Training}} \\
\midrule
 TTT3R & 793.3 & 0.057 & 0.072 & 0.815 & 0.061 & 0.847 & 3.565 & 0.061 & 0.813 & 4.001 & 0.065 & 0.825 & 3.783 \\
 Scal3R & 1266 & 0.059 & \cellcolor{thirdyellow}\textbf{0.049} & \textbf{0.968} & 0.180 & 0.870 & 0.673 & 0.088 & 0.856 & 1.382 & 0.106 & 0.898 & 1.027 \\
 LoGeR & 1255 & 0.058 & 0.057 & 0.944 & 0.051 & 0.966 & 0.681 & \cellcolor{thirdyellow}0.043 & 0.953 & \cellcolor{thirdyellow}0.428 & \cellcolor{thirdyellow}0.050 & 0.954 & \cellcolor{thirdyellow}0.555 \\
 LoGeR$^*$ & 1255 & \cellcolor{thirdyellow}\textbf{0.054} & 0.053 & 0.949 & \cellcolor{thirdyellow}\textbf{0.039} & \cellcolor{thirdyellow}\textbf{0.976} & \cellcolor{thirdyellow}\textbf{0.391} & \cellcolor{secondorange}\textbf{0.037} & \cellcolor{secondorange}\textbf{0.965} & \cellcolor{secondorange}\textbf{0.275} & \cellcolor{secondorange}\textbf{0.043} & \cellcolor{secondorange}\textbf{0.963} & \cellcolor{secondorange}\textbf{0.333} \\
\bottomrule
\end{tabular}%
}
\end{table*}

\clearpage

\begin{table}[!ht]
\centering
\setlength{\fboxsep}{0pt}
\caption{\textbf{Metric-Depth Comparison on \benchmark.}
We compare the native metric-depth predictions (no median/scale alignment) of
all 6 methods on \benchmark{} that emit metric-scale depth, across the
\textit{Single Frame}, \textit{Sparse}, and \textit{Medium} settings,
together with the \textit{Average} across all three settings.
The best, second-best, and third-best results in each column are highlighted in
\colorbox{bestred}{deep blue}, \colorbox{secondorange}{medium blue}, and
\colorbox{thirdyellow}{light blue}, respectively.}
\label{tab:metric_depth}
\resizebox{\textwidth}{!}{%
\renewcommand{\arraystretch}{1.15}
\setlength{\tabcolsep}{3pt}
\begin{tabular}{l c >{\columncolor{subcol}}c >{\columncolor{subcol}}c >{\columncolor{subcol}}c >{\columncolor{subcol}}c >{\columncolor{subcol}}c >{\columncolor{subcol}}c >{\columncolor{subcol}}c >{\columncolor{subcol}}c >{\columncolor{subcol}}c >{\columncolor{subcol}}c >{\columncolor{subcol}}c >{\columncolor{subcol}}c >{\columncolor{subcol}}c >{\columncolor{subcol}}c >{\columncolor{subcol}}c >{\columncolor{subcol}}c}
\toprule
\multirow{2.5}{*}{\textbf{Method}}
 & \multirow{2.5}{*}{\makecell{\textbf{\#Params}\\\textbf{(M)}}}
 & \multicolumn{4}{c}{\textbf{Single Frame}}
 & \multicolumn{4}{c}{\textbf{Sparse}}
 & \multicolumn{4}{c}{\textbf{Medium}}
 & \multicolumn{4}{c}{\textbf{Average}} \\
\cmidrule(lr){3-6} \cmidrule(lr){7-10} \cmidrule(lr){11-14} \cmidrule(lr){15-18}
 &  & AbsRel$\downarrow$ & SqRel$\downarrow$ & RMSE$\downarrow$ & $\delta_{1.25}\uparrow$ & AbsRel$\downarrow$ & SqRel$\downarrow$ & RMSE$\downarrow$ & $\delta_{1.25}\uparrow$ & AbsRel$\downarrow$ & SqRel$\downarrow$ & RMSE$\downarrow$ & $\delta_{1.25}\uparrow$ & AbsRel$\downarrow$ & SqRel$\downarrow$ & RMSE$\downarrow$ & $\delta_{1.25}\uparrow$ \\
\midrule
DA3-Nested & 1689.85 & \cellcolor{secondorange}1.599 & 16.57 & \cellcolor{secondorange}1.477 & \cellcolor{bestred}\textbf{0.507} & \cellcolor{secondorange}0.870 & 5.295 & \cellcolor{secondorange}1.569 & \cellcolor{bestred}\textbf{0.617} & \cellcolor{secondorange}1.188 & 19.64 & \cellcolor{thirdyellow}2.000 & \cellcolor{bestred}\textbf{0.552} & \cellcolor{secondorange}1.219 & 13.83 & \cellcolor{secondorange}1.682 & \cellcolor{bestred}\textbf{0.559} \\
MapAnything & 1228.49 & 2.476 & 31.96 & 2.440 & 0.365 & 2.368 & 23.54 & 2.592 & 0.437 & 3.103 & 27.87 & 2.374 & 0.407 & 2.649 & 27.79 & 2.469 & 0.403 \\
AMB3R & 1563.12 & \cellcolor{thirdyellow}2.235 & 48.59 & \cellcolor{bestred}\textbf{1.473} & \cellcolor{thirdyellow}0.388 & \cellcolor{thirdyellow}1.133 & \cellcolor{thirdyellow}4.934 & \cellcolor{bestred}\textbf{1.360} & \cellcolor{secondorange}0.519 & \cellcolor{thirdyellow}1.407 & \cellcolor{secondorange}3.990 & \cellcolor{bestred}\textbf{1.242} & \cellcolor{secondorange}0.521 & \cellcolor{thirdyellow}1.592 & 19.17 & \cellcolor{bestred}\textbf{1.358} & \cellcolor{secondorange}0.476 \\
$\pi^{3}$-X & 1360.03 & 3.187 & \cellcolor{thirdyellow}13.87 & \cellcolor{thirdyellow}1.842 & \cellcolor{secondorange}0.400 & 1.771 & 9.361 & \cellcolor{thirdyellow}1.779 & \cellcolor{thirdyellow}0.461 & 1.957 & 6.269 & \cellcolor{secondorange}1.705 & \cellcolor{thirdyellow}0.449 & 2.305 & \cellcolor{thirdyellow}9.832 & \cellcolor{thirdyellow}1.775 & \cellcolor{thirdyellow}0.436 \\
DANext & 1303.76 & \cellcolor{bestred}\textbf{1.484} & \cellcolor{bestred}\textbf{3.369} & 3.595 & 0.070 & \cellcolor{bestred}\textbf{0.713} & \cellcolor{bestred}\textbf{3.280} & 4.235 & 0.117 & \cellcolor{bestred}\textbf{0.713} & \cellcolor{bestred}\textbf{3.350} & 4.261 & 0.116 & \cellcolor{bestred}\textbf{0.970} & \cellcolor{bestred}\textbf{3.333} & 4.030 & 0.101 \\
MASt3R-SLAM & 688.64 & 2.672 & \cellcolor{secondorange}5.479 & 2.962 & 0.231 & 1.570 & \cellcolor{secondorange}4.043 & 3.468 & 0.191 & 2.264 & \cellcolor{thirdyellow}4.992 & 3.555 & 0.209 & 2.169 & \cellcolor{secondorange}4.838 & 3.329 & 0.210 \\
\bottomrule
\end{tabular}%
}
\vspace{-10pt}
\end{table}

\section{Limitations}
\label{appendix:limit}

We report the limitations of \benchmark in this section.

\noindent\textbf{Evaluation Cost.} 
Evaluating 41 models across 100+ scenes per model under the dense regime is time-consuming. This can be mitigated by distributing evaluation across multiple GPUs in parallel.

\noindent\textbf{Memory Constraints.}
All evaluations are conducted on H200 GPUs with 141\,GB VRAM. 
We have not tested models under larger memory configurations such as B100 or B200, and performance or behavior may differ on such hardware.

\noindent\textbf{Hyperparameter Selection.}
We acknowledge that some evaluated methods may require task- or scene-specific hyperparameter tuning to achieve optimal performance. However, such tuning falls outside the scope of this benchmark. We follow the recommended configurations provided in each method's official codebase to ensure a fair and consistent comparison across all methods.

\noindent\textbf{Expanding Method Coverage.}
As of the submission deadline, a number of newly released models continue to be open-sourced. 
Therefore, we cannot guarantee complete coverage of all existing methods.
We commit to continuously integrating and evaluating new methods as they become available.


\end{document}